\documentclass[conference]{IEEEtran}
\IEEEoverridecommandlockouts

\usepackage{amsmath,amsfonts,bm}

\def\1{\bm{1}}

\DeclareMathAlphabet{\mathsfit}{\encodingdefault}{\sfdefault}{m}{sl}
\SetMathAlphabet{\mathsfit}{bold}{\encodingdefault}{\sfdefault}{bx}{n}

\DeclareMathOperator*{\argmax}{arg\,max}

\usepackage{caption} 
\usepackage{hyperref}
\usepackage{url}
\usepackage{comment}
\usepackage[utf8]{inputenc} 
\usepackage[T1]{fontenc}    
\usepackage{hyperref}       
\usepackage{url}            
\usepackage{booktabs}       
\usepackage{amsfonts}       
\usepackage{nicefrac}       
\usepackage{microtype}      
\usepackage{xcolor}         
\usepackage{amsfonts}
\usepackage{amsmath}
\usepackage{algorithm,algpseudocode}
\usepackage{booktabs}
\usepackage{multirow} 
\usepackage{graphicx}
\usepackage{bbm}
\usepackage{longtable}
\usepackage{tabularx}
\usepackage{tabularray}
\usepackage{subfigure}
\usepackage{enumitem}
\usepackage{xspace}
\newcommand{\IDNet}{\textsc{IDNet}\xspace}
\newcommand{\IDSpace}{\textsc{IDSpace}\xspace}

\def\BibTeX{{\rm B\kern-.05em{\sc i\kern-.025em b}\kern-.08em
    T\kern-.1667em\lower.7ex\hbox{E}\kern-.125emX}}

\makeatletter
\def\linebreakand{%
  \end{@IEEEauthorhalign}
  \hfill\mbox{}\par
  \begingroup
  \@IEEEauthorhalign
}
\makeatother

\begin{document}

\title{\IDSpace: A Novel Document Generator for Reliable Evaluation of Digital Identity Verification Systems [Extended Technical Report]}

\author{
\centering
\IEEEauthorblockN{ Lulu Xie, Yancheng Wang}
\IEEEauthorblockA{
\textit{Arizona State University}
}
\and
\IEEEauthorblockN{Kanchan Chowdhury}
\IEEEauthorblockA{
\textit{Marquette University}
}
\and
\IEEEauthorblockN{ Rolando Garcia, Yingzhen Yang, Jia Zou}
\IEEEauthorblockA{
\textit{Arizona State University}
}
}

\maketitle

\begin{abstract}
As services move online, trust institutions such as banks, lenders, and governments must verify the identity of remote users. 
Fraud detection tools are widely available, but evaluating and fine-tuning them remains difficult because identity documents are sensitive and therefore scarce. 
Synthetic data generation offers a path forward, and demand is clear: our prior work in this area has been downloaded over $11{,}000$ times (aggregated from eight parts). 
We introduce \IDSpace, extending this line of research in three directions. 
First, we propose model-guided Bayesian optimization, which tunes generation parameters to maximize both visual similarity and prediction consistency with target-domain models given only a few samples from a target domain. Second, we decouple user-specified metadata (demographics, fraud patterns, capture device) from automatically tuned control parameters (font styles, noise levels, image quality), allowing users to configure evaluations without low-level expertise. Third, we expand beyond template images to support scanned and mobile-captured documents. Experiments show \IDSpace improves evaluation consistency by $15-45\%$ over baselines including CycleGAN, diffusion inpainting, and non-guided optimization, using only a few real samples, while improving training accuracy by up to $9\%$ and SSIM similarity with the target domain by $10\%$.
We also released a new dataset consisting of $359{,}240$ high-quality synthetic documents across ten European ID types.

\end{abstract}

 
\section{Introduction}
\label{sec:intro}
The surge in digital platforms offering remote identity verification has raised concerns about forged identity documents, such as passports, driver's licenses, and ID cards. In fiscal year 2023, the Financial Crimes Enforcement Network received approximately 4.6 million Suspicious Activity Reports, with around 1.75 million related to identity fraud~\cite{FinCEN}. Accurate detection of fraudulent identity documents is crucial for reducing authentication risks across various sectors, including finance, healthcare, travel, retail, government, and gambling~\cite{onfido}. Therefore, it is important to evaluate the accuracy of fraud detection capabilities before deploying digital identity verification services in these application scenarios. However, due to the sensitivity of personal information in these documents, a significant \textbf{data scarcity} challenge arises in assembling comprehensive real-world datasets for flexible and reliable evaluation of identity verification systems. 

\vspace{3pt}
\noindent
\textbf{Motivating Example.}
Between August 2023 and April 2024, the US General Services Administration conducted a large-scale study evaluating the fairness of commercial remote identity verification services across demographic groups~\cite{fatima2024large}. They recruited 3,991 participants from five racial and ethnic groups, who captured and uploaded images of their identity documents and selfies using mobile devices. 
This study illustrates the fundamental challenges of evaluating identity verification systems with real data: recruitment costs scale linearly with coverage requirements, making it expensive to achieve statistical significance across demographic subgroups and document types; image quality variance is uncontrolled, as participants use diverse devices under varying conditions; and demographic balance is difficult to achieve, with minority groups, precisely the populations where fairness audits matter most, often underrepresented. 

\vspace{3pt}
\noindent
\textbf{Synthetic data generation offers a path forward.} Existing synthetic identity document generation falls into three main categories, and they all suffer from significant limitations: \textbf{(Category 1) Manual generation} of identity document datasets, such as MIDV-500~\cite{arlazarov2019midv} and MIDV-2020~\cite{bulatovich2022midv}, is expensive and labor-intensive, leading to limited scale. For example, MIDV-2020~\cite{bulatovich2022midv} contains only $1{,}000$ distinct template images across ten document categories. \textbf{(Category 2) Training-based approaches}~\cite{benalcazar2023synthetic}, including those using GANs~\cite{goodfellow2020generative}, differential privacy~\cite{dwork2006differential}, generative AI~\cite{vinogradov2025can}, and diffusion models~\cite{saifullah2025dp}, can still require large numbers of labeled real documents that are difficult to obtain due to regulatory constraints, and hard to generalize to different countries and regions. \textbf{(Category 3) Few-shot approaches}~\cite{bothra2023synthetic, lerouge2024docxpand, xie2024idnet} aim to minimize data requirements by  
using inpainting techniques to fill in synthetic personal data and face images into a template. Our prior work, \IDNet~\cite{xie2024idnet}, a representative of this category, has been downloaded more than $\textbf{11{,}000}$ times (aggregated from eight parts) on the Zenodo platform since 2024, underscoring both the unmet demand and the practical impact of few-shot approaches on generating large-scale synthetic documents with only one or a few real-world document examples. However, despite of the cost-effectiveness of the few-shot approaches, they often produce content that diverges from the target domain distribution. This domain shift can mislead fraud detection models and result in unreliable evaluation outcomes.  

Furthermore, existing methods lack flexible, declarative specificity over what documents to generate, yet comprehensive benchmarking requires testing under a wide range of conditions, such as documents from specific demographic groups, or scanned documents at particular rotation angles.

\begin{figure*}[t]
    \centering
    \includegraphics[width=0.98\linewidth]{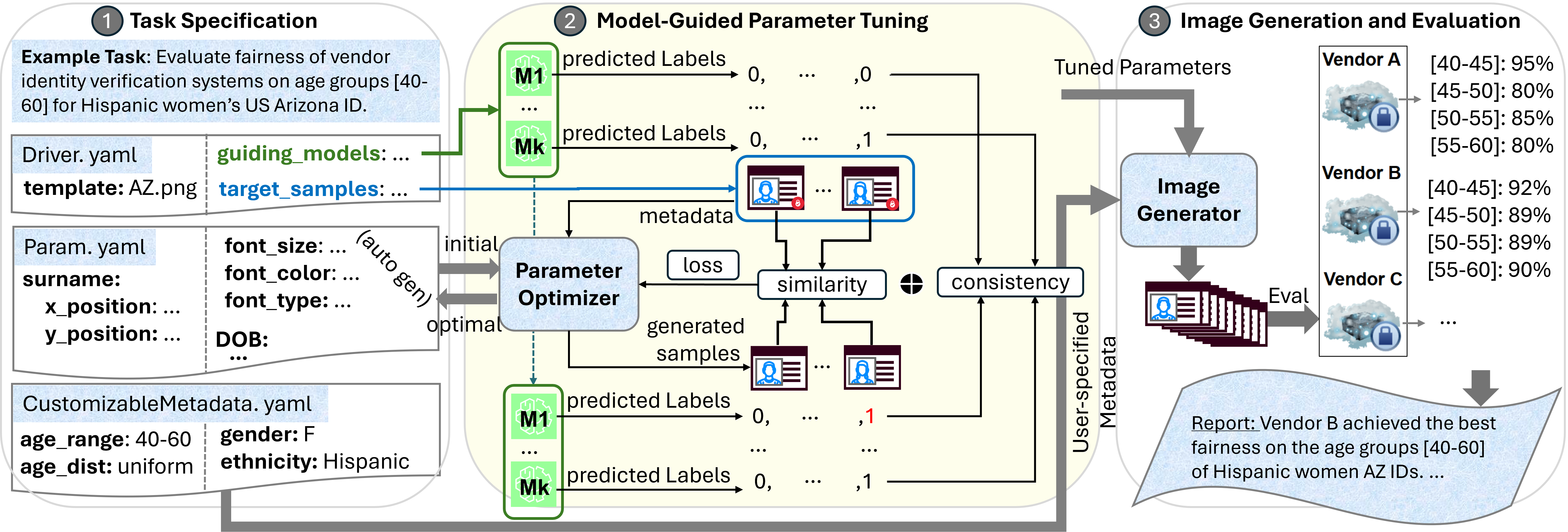}
    \caption{\small Overview of the \IDSpace framework. The pipeline consists of three stages: (1) Task specification, where users specify high-level metadata constraints, and provide a few real samples and models from the target-domain ; (2) Model-guided parameter tuning, which optimizes rendering parameters to maximize structural similarity and prediction consistency; and (3) Image generation and evaluation, which generates a synthetic benchmark dataset to evaluate remote identity verification systems and report results.}
    \label{fig:overview}
    \vspace{-10pt}
\end{figure*}

\begin{figure}[t]
    \centering
    \includegraphics[width=0.8\linewidth]{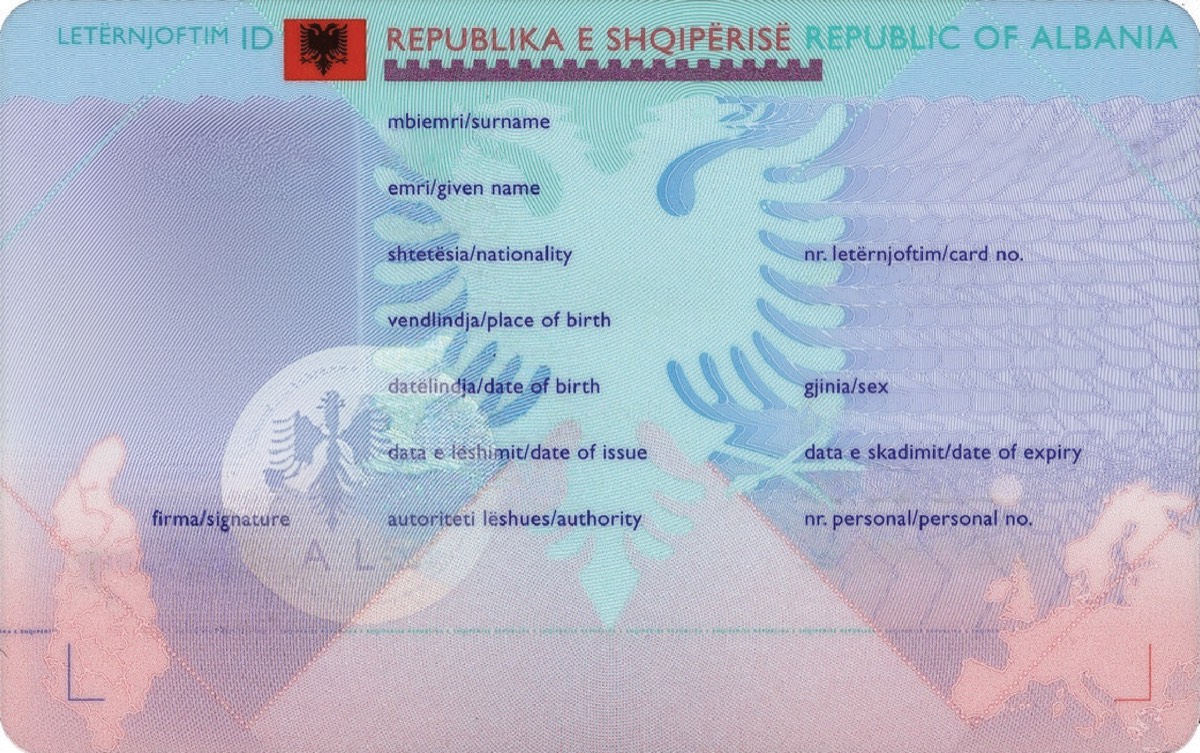}
    \caption{\small Template of Albania ID card.}
    \label{fig:albania-template}
    \vspace{-10pt}
\end{figure}

\vspace{3pt}
\noindent
\textbf{This paper introduces \IDSpace}, a 
 model-guided few-shot framework for generating high-quality synthetic identity documents with a flexible declarative task-specification interface, while enabling reliable and reproducible evaluation and training of identity verification systems, as illustrated in Fig.~\ref{fig:overview}. 
First, \IDSpace introduces model-guided Bayesian optimization, which automatically tunes the generation parameters that control the filling of \emph{user-specified metadata}, such as age, gender, ethnicity, and fraud patterns, into a user-specified template image as illustrated in Fig.~\ref{fig:albania-template} to maximize both visual similarity and prediction consistency with the target domain, using only a few real samples. 
Second, it decouples \emph{user-specified metadata} from \emph{automatically tuned control parameters} that govern low-level rendering details, such as font styles, font sizes, font colors, noise levels, brightness, and sharpness. This separation allows users to configure evaluations without low-level expertise. 
Third, \IDSpace expands beyond template images to support scanned and mobile-captured documents, enabling more realistic and diverse benchmarking settings. 

Onfido's survey of recent ID frauds~\cite{onfido} showed that an attacker can produce a convincing forged \emph{physical} document, printed or digitally composited, and then capture it through a legitimate app on a real device. The resulting image has clean metadata and passes pipeline integrity checks, yet the underlying document may be fraudulent. 
\IDSpace is designed to support evaluation under such document-level forgery scenarios: detecting manipulated text fields and photos, inconsistent fonts, inconsistent fields, and other artifacts that distinguish forged documents from genuine ones, regardless of how the image was captured. 

Concretely, we make the following contributions:

\begin{itemize}[leftmargin=*, itemsep=2pt, topsep=2pt]

\item \textbf{Reliable evaluation of identity verification systems under data scarcity.} \IDSpace reframes synthetic identity document generation from a data augmentation problem to a \emph{benchmarking and evaluation} problem under severe data scarcity. While it builds on \IDNet~\cite{xie2024idnet}, which focuses on visual realism for training, \IDSpace is designed to ensure \emph{evaluation validity} by preserving model behavior between real and synthetic documents. Rather than optimizing solely for perceptual similarity, \IDSpace introduces \emph{prediction consistency} metric and \emph{user-provided target models} that are trained on the target domain of documents, and it uses model-guided Bayesian optimization to auto-tune rendering (or ``control'') parameters that preserve prediction consistency between the documents in the target domain and the generated synthetic documents. Therefore, it can reliably serve as a proxy for real data in downstream evaluation, a capability not supported by any prior synthetic datasets.

\item \textbf{Specification-driven generation with decoupled user intent and rendering control.} 
Different from existing document generation frameworks, \IDSpace separates user-specified domain parameters (or ``metadata''; the \emph{what} to evaluate) from automatically tuned rendering parameters (the \emph{how} to generate). Users specify metadata declaratively via YAML/JSON/CSV, while rendering parameters are optimized internally to support reproducible and controlled benchmarking. Table~\ref{tab:parameters} formalizes this separation across template, scanned, and mobile formats.

\item \textbf{A large-scale, multi-modal dataset for benchmarking identity verification systems.} 
We release a dataset of $359{,}240$ synthetic documents across ten European ID types, spanning templated, scanned, and mobile-captured modalities, publicly available under CC-BY 4.0\footnote {\footnotesize\url{https://huggingface.co/datasets/cactuslab/IDSpace}\label{footnote1}} with source code available under Apache 2.0 license\footnote{\footnotesize\url{https://github.com/asu-cactus/IDSpace}\label{footnote2}}.

Experiments show that with as few as two real samples, \IDSpace improves evaluation consistency by 15--45\% over CycleGAN~\cite{CycleGAN2017}, diffusion inpainting~\cite{rombach2022high}, and non-guided optimization~\cite{xie2024idnet}, enabling reliable evaluation and benchmarks. The evaluation results also showed that the documents generated by \IDSpace have consistently better SSIM similarity with the documents in the target domain than \IDNet. In addition, the accuracy of the models trained on \IDSpace outperformed \IDNet by up to $9\%$.

\end{itemize}



\begin{figure*}[t]
    \centering
    \includegraphics[width=0.99\linewidth]{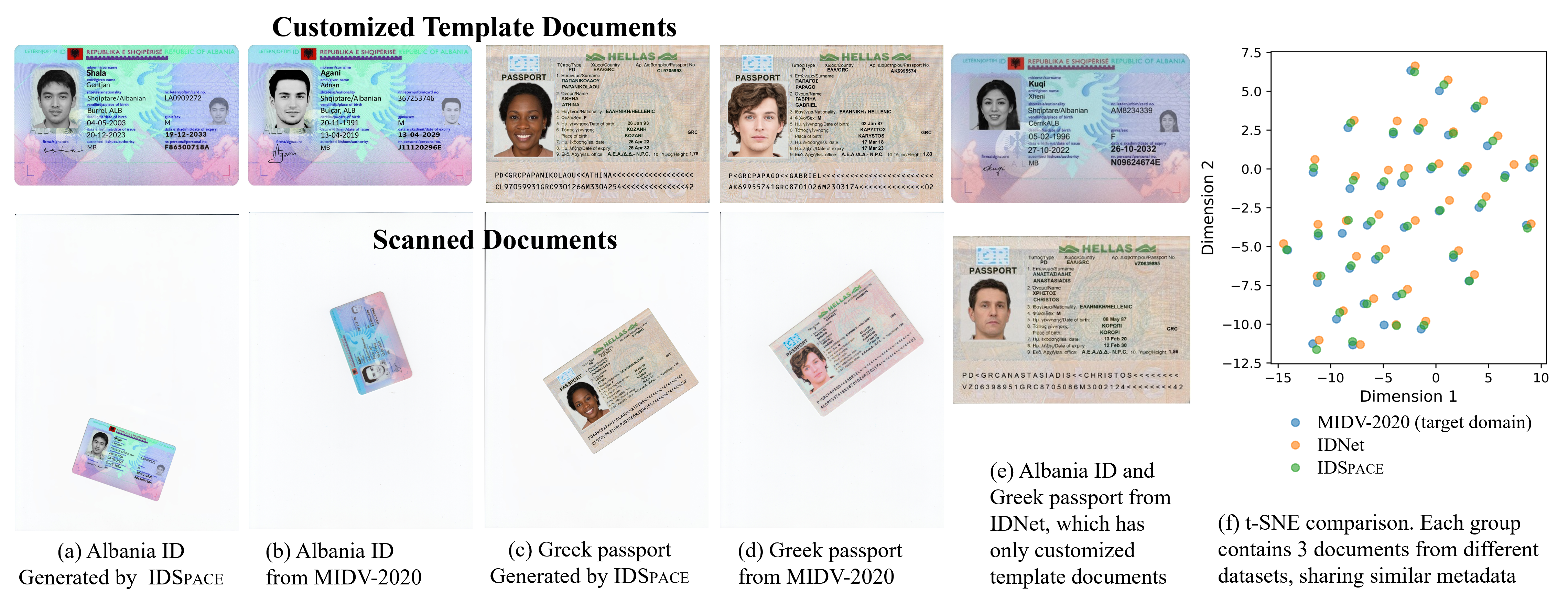}
    \caption{\small Examples and t-SNE comparison for non-fraud documents generated by \IDSpace, \IDNet~\cite{xie2024idnet}, and MIDV-2020 (the target domain).}
    \label{fig:dataset}
    \vspace{-10pt}
\end{figure*}

\begin{figure}[t]
    \centering
    \subfigure[User Specified Background]{
        \includegraphics[width=0.19\textwidth]{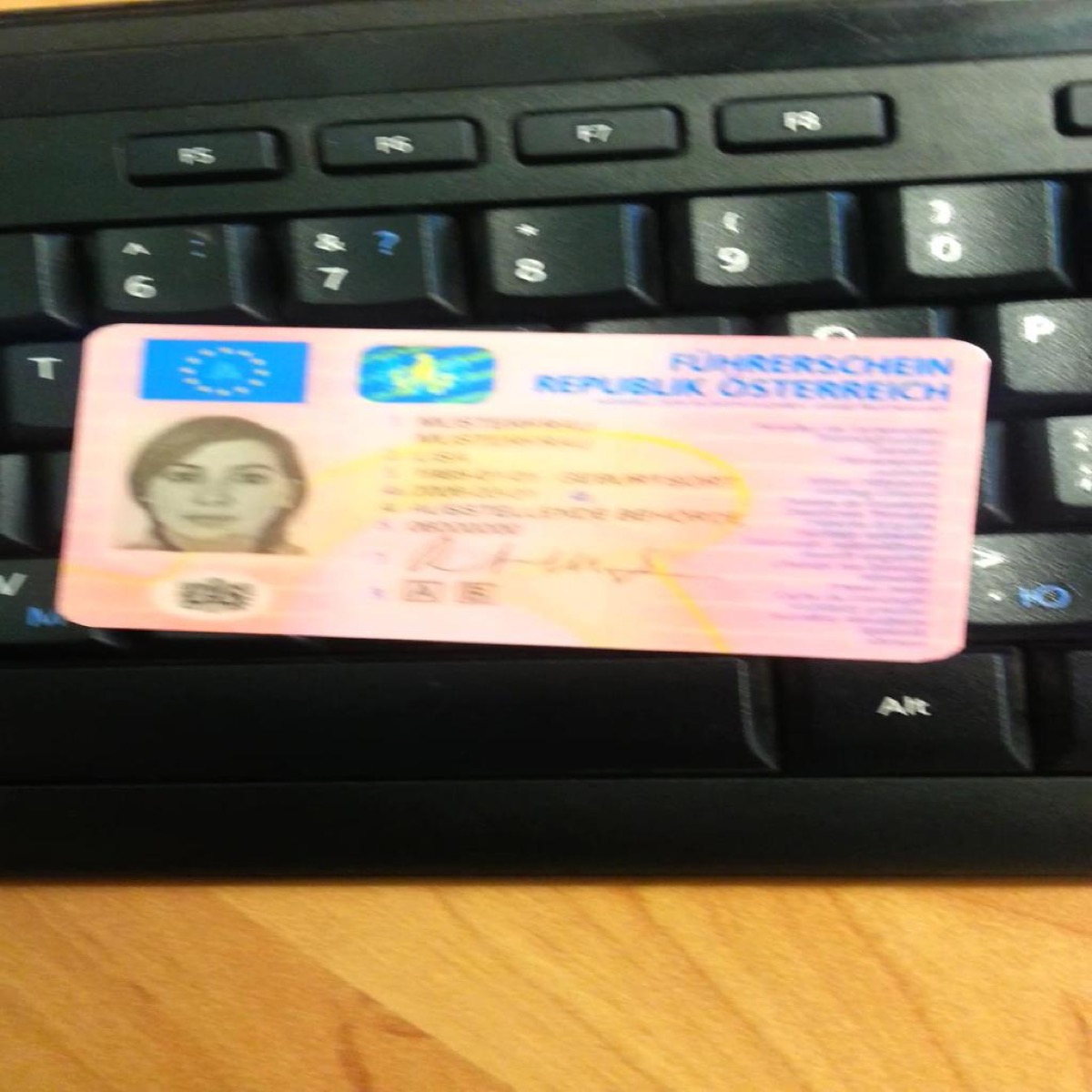}
    }
    \subfigure[Generated Mobile ID]{
        \includegraphics[width=0.19\textwidth]{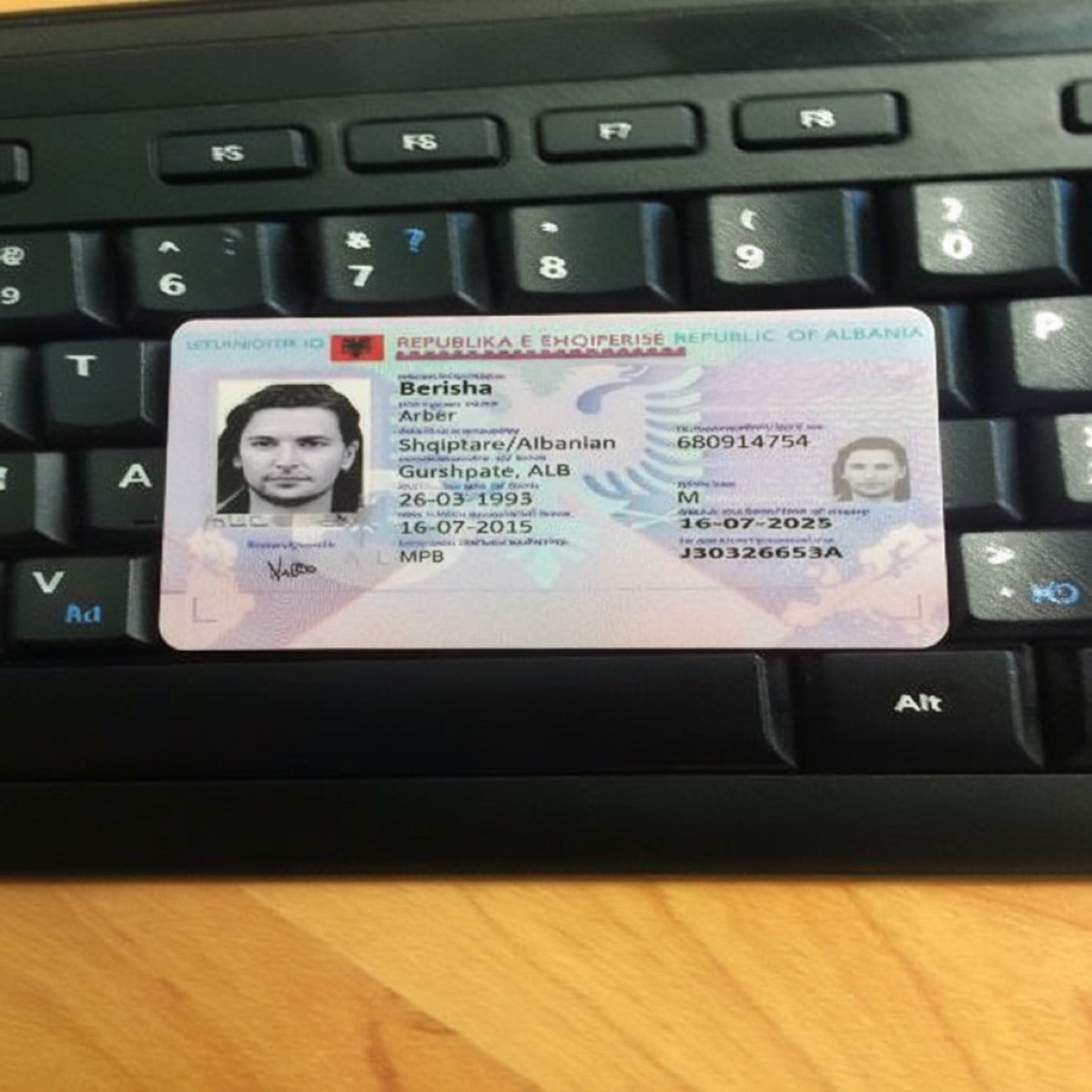}
    }
    \caption{\small Example Mobile document images generated.}
    \label{fig:mobile_docs}
\end{figure}

\vspace{-5pt}
\section{Related Work}
\label{sec:relatedworks}
\noindent
\textbf{Public Synthetic Datasets and Synthetic Data Generation Methods for Identity Documents.}
As mentioned, manually crafted datasets such as MIDV-500~\cite{arlazarov2019midv}, MIDV-2020~\cite{bulatovich2022midv}, MIDV-UP~\cite{chernyshova2025midv}, and \textcolor{black}{KID34K~\cite{park2023kid34k}} suffer from the document generation cost, limited to around $100$ templated documents for one country. SIDTD~\cite{boned2024synthetic} and FMIDV~\cite{al2023guilloche} rely on MIDV-2020 to provide fraud patterns using inpainting or crop-and-move techniques. Overall, these datasets lack diversity and flexibility in performing comprehensive and customizable model evaluation tasks.
Other large-scale datasets suffer from different quality issues, which fall into the following categories. \textbf{(1) Redaction-based document generation.} For example,   \textbf{BID}~\cite{de2020bid} redacts sensitive information such as portrait photos from real-world documents, which reduces their utility for many portrait-based fraud detection applications, e.g., face morphing detection. \textbf{(2) Training-based generation~\cite{benalcazar2023synthetic}} Generative-Adversarial Network (GAN) is widely used for synthetic document generation. For example, StyleGAN2~\cite{karras2020analyzing} was used to generate ID images building on its strong capability in synthesizing highly realistic human faces. 
However, it struggled with alphanumeric characters, and it also needed thousands of real IDs for model training, which is impractical in our target scenarios where only a small number of documents from the target domain of models to be evaluated is available.  DP-DocLDM~\cite{saifullah2025dp} uses private documents to finetune a public Diffusion model with differential privacy guarantees to generate documents, such as images of resumes, news articles, memos, scientific reports, etc. However, they do not focus on identity documents. In addition, their finetuning process used RVL-CDIP and Tobacco3482 datasets. The former consists of $400{,}000$ document images, while the latter consists of more than $3{,}000$ documents.
CycleGAN~\cite{xie2020self} has been used to adapt low-fidelity identity documents to the target domain. However, this approach usually requires a large number of labeled samples of the target domain, which is not practical in our target scenarios. 
Ben-David et al.~\cite{david2010impossibility} proposed impossibility theorems, indicating that 
  even with small distributional divergence or a universally good classifier, domain adaptation can fail without labeled target data.
\textbf{(3) Few-shot Approaches.} DocXPand-25k~\cite{lerouge2024docxpand} applies inpainting to fill metadata into a self-designed document template, and it can hardly generalize to real-world document types. 
Our prior work \IDNet~\cite{guan2024idnet, xie2024idnet} is a low-cost framework that tunes a couple of parameters to maximize the similarity between the generated documents and target domain documents. While it is instrumental in augmenting training data and its parts have been downloaded for more than $11{,}000$ times in total on Zenodo, its quality issues have been noticed~\cite{chernyshova2025midv}. We further identified a prediction inconsistency using multiple fraud-detection models between the target domain and the \IDSpace-generated documents, due to an observed discrepancy between the generated dataset and the target domain, as illustrated in Fig.~\ref{fig:dataset}(f). In this work, we addressed this limitation by introducing a novel model-guided data generation technique. In addition, \IDNet only involved templated document images (i.e., images with information filled in the fields of templates), while this work has expanded the modality to documents captured by scanners and mobile phones. Our experiments in Sec.~\ref{sec:experiment} showed that compared to \IDNet, \IDSpace significantly improved the prediction consistency, training accuracy, and SSIM similarity with the documents from the target domain.

\noindent
\textbf{Synthetic Data Generation for Model Evaluation.} \cite{nikolenko2021synthetic} discusses adaptation of the synthetic data to the real domain using GANs. Van Breugel et al.~\cite{van2023can} proposed to use a deep generative model on the test dataset to create synthetic data for evaluating model performance on underrepresented subgroups and under distributional shifts. However, their approach relies on the availability of test data, which is often scarce in practice in our targeting scenarios. 


\begin{table*}
  \caption{\small Overview of Examplar Synthetic ID Generation Parameters 
  }
  \label{tab:parameters}
  \scriptsize
  \centering
  \begin{tabular}{p{1.5cm}|p{7cm}|p{7cm}}
  \hline
         & Control Parameters (AutoTuned)    & User-Customizable Metadata  \\
    \hline
    Customized Template Documents &
    \textbf{Font:} Font style and size of each text field.
    \newline
    \textbf{Text color:} Color of texts that overlap with the background. \newline 
    \textbf{Positions:} Positions of each field in the template, and character spacing within each field.\newline
    \textbf{Quality:} JPEG image quality settings (e.g., quality factor in [50--95]).
   &      \textbf{ID Template:} File path of the template image used for ID generation.\newline \textbf{Fraud Pattern:} Whether to simulate fraud and which type---e.g., \emph{crop-and-move}, \emph{inpaint-and-rewrite}, following SIDTD~\cite{boned2024synthetic}.\newline
    \textbf{Entity Information:} First name, last name, ID number, DOB, ethnicity group, gender, height, weight, issue date, expiration date, etc.
\\\hline

    Scanned/Mobile Documents & 
    \textbf{Noise Level:} Amount of Gaussian noise added---e.g., standard deviation in range [0, 25].\newline
    \textbf{Subtle Blurring:} Gaussian blur with sigma in range [0.5, 2.0] to simulate out-of-focus scans.\newline
    \textbf{Brightness:} Adjustment factor for image brightness---e.g., randomly sampled from [0.8, 1.2].\newline
    \textbf{Contrast:} Adjustment factor for image contrast---e.g., randomly sampled from [0.8, 1.2].\newline
    \textbf{Sharpness:} Strength of sharpening filter---e.g., factor in [0.5, 2.0].
    &
    \textbf{ID Template Image:} File path of the ID image (a customized template document) to be scanned.\newline
    \textbf{Resolution:} DPI value, e.g., 200, 300, or 400.\newline
    \textbf{Color Mode:} One of \{"color", "grayscale", "black and white"\}.\newline 
\textbf{Position/Orientation (for scanned images):} Placement of the ID---e.g., rotated by \ensuremath{\pm}5\ensuremath{^\circ} or shifted \ensuremath{\pm}10px.\newline
    \textbf{Background image (for mobile images):} An image to serve as the background of the ID placement. 
    \\
    %
    %
    %
    \bottomrule
  \end{tabular}
\end{table*}

\section{A Novel Problem Abstraction}
\label{sec:problem}

In this section, we present a few-shot synthetic data generation methodology designed to balance cost and quality in evaluating identity verification systems. Our approach is to combine the decoupling of the user-specified metadata and the control parameters finetuned by a model-guiding framework. 

Table~\ref{tab:parameters} illustrates how we distinguish metadata that should be explicitly specified by users from parameters that should be automatically adjusted in this work.  Parameters that can be reliably inferred using external tools are excluded. For instance, the background color of a portrait photo can often be extracted using standard color analysis tools and does not require automated tuning. In contrast, detecting text color when it overlaps with complex background images is significantly more difficult, and identifying the original font styles used in official identity documents is often infeasible due to their proprietary nature. In such cases, we rely on auto-tuning.
Importantly, advanced users can redefine the boundary between user-specified metadata and automatically controlled parameters, depending on their available tools, expertise, and resources. 
Based on this flexible and modular design, we are the first to formalize the synthetic identity document generation problem as follows.

\noindent
\textbf{Portrait Photos.} By default, portrait photos are selected from an open academic dataset published by generated.photos \cite{generated.photos} that consists of $10{,}000$ portrait photos, with $5{,}979$ photos suitable for identity documents. For each document, the selection of the portrait photo must be consistent with the specified metadata, which means the user-specified age (DOB), gender, ethnicity group, and weight, must match the face of the portrait photo. This is not difficult given that each photo in the generated.photos dataset has all annotated information specifying age, gender, and ethnicity groups. Advanced users can also replace the generated.photos dataset with their own portrait photo database.

\noindent
\textbf{Fraud Patterns.} As shown in Tab.~\ref{tab:parameters}, users can specify the numbers of non-fraud and fraud documents, as well as the distribution of fraud patterns among the fraud documents. Our current implementation supports two representative fraud patterns identified in Onfido/Entrust's \textit{Identity Fraud Report 2024}~\cite{onfido}: (1) \textit{Inpaint-and-Rewrite}. A text field is randomly selected from all available ID fields, excluding the portrait photo and signature. A realistic mask is applied to the selected field region, and the replacement text is rendered using a randomly selected font size and style. (2) \textit{Crop-and-Move}. A field, such as last name, date of birth, or address, is cropped from one ID and replaced with the corresponding field from another ID. In both patterns, fields are selected randomly, with a $95\%$ probability of selecting the same field across the two IDs and a $5\%$ probability of selecting different PII fields, following the standard implementations of SIDTD~\cite{boned2024synthetic} and \IDNet~\cite{xie2024idnet}. These are also the only two fraud patterns supported by both SIDTD and \IDNet.

\IDNet supports three additional patterns: face morphing, photo replacement, and a trivial mixture mode. Since these patterns are confined to the portrait photo field and do not affect other ID fields, incorporating them into our framework is straightforward. For example, users can provide a set of portrait photos with morphed faces in the metadata to replace the default generated.photos dataset~\cite{generated.photos}. We exclude these photo-based fraud patterns from our evaluation because mature benchmarks already exist for face morphing detection, such as FRLL, FERET, and FRGC, and inserting morphed or replaced portraits into synthetic documents does not introduce additional document-level generation challenges.

\noindent
\textbf{Problem Definition.} Given a template of a type of ID document (e.g., a template of the Albania ID card, as illustrated in Fig.~\ref{fig:albania-template}), denoted as $T$, generating the $i$-th ID image has two steps: (1) obtaining or generating the metadata information $\boldsymbol{x^i_{meta}}$ following user-specification, which includes fraud patterns, capturing device (e.g., scanner), capturing environments (e.g., rotation and position of the document, color mode, and resolution), and various personal information (e.g., first name, last name, date of birth, ID card number, portrait photo, eye color, height, weight, card issue date, expiration date), listed as user-customizable metadata in Tab.~\ref{tab:parameters}. (2) filling in the metadata information into the template to generate the final image $x_i$, denoted as $x_i=G_{\theta}(\boldsymbol{x^i_{meta}}, T)$. Here, $\theta$ represents the parameters that control filling the metadata into the template, such as those control parameters listed in Tab.~\ref{tab:parameters}. While $G_{\theta}(\cdot)$ represents the process of transforming the metadata $\boldsymbol{x^i_{meta}}$ and the given template $T$ into a synthetic ID image $x_i$ using the control parameters $\theta$. 

Given an existing machine learning model $f$ trained for fraud detection for a target domain consisting of ID documents sharing the same template $T$, denoted as $\mathcal{D}_{\text{real}} = \{x_i\}$, with $f$'s training dataset $\mathcal{D}_{\text{training}}\subset\mathcal{D}_{\text{real}}$. Assuming each sample $x_i\in \mathcal{D}_{\text{real}}$ having metadata $meta(x^i)$, given a small number of samples from $\mathcal{D}_{\text{real}}$, we would like to learn $\theta$ so that $\forall x_i \in \mathcal{D}_{\text{real}}$, we have $x_i=G_{\theta}(meta(x_i), T)$, and thus $f(x_i)=f(G_{\theta}(meta(x_i), T))$.

\section{Model Guided Control Parameter Tuning} 
\label{sec:parameter-tuning}
To address the problem described in Sec.~\ref{sec:problem}, we propose combining Bayesian optimization~\cite{frazier2018tutorial} with our custom optimization objective that introduces $l$ guiding models $f_1,\dots,f_l$, which are trained in the target domain. \textit{Importantly, these $k$ models are not necessarily the target models to be evaluated.} The objective is not only to maximize the overall similarity (e.g., measured using structural similarity index measure (SSIM)) between each input document  $x_i$ and the corresponding generated document $G_{\theta}(meta(x_i), T)$ for $i=1,..., m$ (See Eq.~\ref{eq:similarity}), but also to improve evaluation consistency between  $f_k(x_i)$ and $f_k(G_{\theta}(meta(x_i), T))$ for $i=1,...,m$ and $k=1,...,l$ (See Eq.~\ref{eq:consistency}). Here, $m$ denotes the number of samples from the target domain, which is assumed to be small, given the scarcity of ID data. %
Our custom optimization objective is formalized in Eq.~\ref{eq:objective}, which is a weighted sum of the similarity metric (Eq.~\ref{eq:similarity}) and the evaluation consistency metric (Eq.~\ref{eq:consistency}). 

\begin{equation}
\label{eq:objective}
    \vspace{-10pt}
  \small
    \theta^* = \argmax_{\theta \in \Theta} (\lambda_0\cdot\mu_{\text{similarity}}(\theta)  + \dots + \lambda_l \cdot \mu^l_{\text{consistency}}(\theta))
\end{equation}

\smallskip
\begin{equation}
\label{eq:similarity}
\small
    \mu_{\text{similarity}}(\theta) = \frac{1}{m} \sum_{xi \in D_{real}} similarity(x_i, G_{\theta}(meta(x_i), T))
        \vspace{-10pt}
\end{equation}

\smallskip
\begin{equation}
\label{eq:consistency}
\small
    \mu^k_{\text{consistency}}(\theta) = \frac{1}{m} \sum_{xi \in D_{real}} \mathbbm{1}(f_k(x_i) = f(G_{\theta}(meta(x_i), T)))
        \vspace{-5pt}
\end{equation}


Our proposed Bayesian optimization algorithm formalized in Alg.~\ref{alg:parameter-tuning} first trains a surrogate model to learn the relationship between $\theta$ (the control parameters of the data generation process) and an objective metric as formalized in Eq.~\ref{eq:objective}. It then iteratively selects $\theta$ guided by the surrogate model to optimize the objective. In each iteration, it evaluates the effectiveness of the selected $\theta$ using the objective function, and use the measured results to update the surrogate model.  The Bayesian optimization strategy can be replaced with other search methods, such as Hyperband~\cite{li2018hyperband}, which are evaluated and compared to our approach in Sec.~\ref{sec:tuning}.

\begin{algorithm}[t]
\caption{\small Model-Guided Bayesian Optimization of Control Parameters (For simplicity, it only used one guiding model $f$, and the extension to multiple guiding models as illustrated in Eq.~\ref{eq:objective} is trivial)
}
\label{alg:parameter-tuning}
\begin{algorithmic}[1]
\scriptsize
\State \textbf{Input:} A small dataset from $\mathcal{D}_{\text{real}}$: $\mathcal{D}_{\text{sample}} = \{x_1, x_2, ..., x_m\}$, a template $T$
\State synthetic data generator $G$, the guiding model $f$, parameter space $\Theta$, weight $\lambda_0$, $\lambda_1$
\State \textbf{Output:} Optimized parameter set $\theta^*$

\State Initialize Bayesian Optimization over $\Theta$ \Comment{Setup BO framework}

\For{iteration $t = 1$ to $T$} 
    \State $\theta_t \gets \text{BO acquisition function}$ \Comment{Sample candidate parameters}
    \State $\mathcal{D}_{\text{syn}}  \gets \{{G_{\theta_t}}(meta(x_i), T)| x_i \in \mathcal{D}_{sample}\}$ \Comment{Generate synthetic data}
    
    
    \State $\mu_{\text{similarity}}^t \gets \frac{1}{m} \sum_{i=1}^m similarity(x_i, D_{\text{syn}}^{(i)})$ \Comment{Compute similarity, e.g. SSIM (Eq.~\ref{eq:similarity})}
    
    \State $\mathbf{p}_{\text{real}} \gets f(\mathcal{D}_{\text{sample}})$ \Comment{Get model prediction results}
    \State $\mathbf{p}_{\text{syn}} \gets f(\mathcal{D}_{\text{synth}})$
    \State $\mu_{\text{consistency}}^t \gets \frac{1}{m} \sum_{i=1}^m \mathbbm{1}(p_{\text{real}}^{(i)} = p_{\text{syn}}^{(i)})$ \Comment{Compute consistency (Eq.~\ref{eq:consistency})}
    
    \State $J(\theta_t) \gets \lambda_0 \mu_{\text{similarity}}^t + \lambda_1 \mu_{\text{consistency}}^t $ \Comment{Compute Objective score (Eq.~\ref{eq:objective})}
    \State Update BO model with $(\theta_t, J(\theta_t))$
\EndFor

\State \Return $\theta^* = \arg\max_{\theta \in \{\theta_1, \ldots, \theta_T\}} J(\theta)$ \Comment{Select best parameters}
\end{algorithmic}
\end{algorithm}

\vspace{5pt}
\section{Document Generation}
\label{sec:data-generation}

After optimizing the control parameters using Alg.~\ref{alg:parameter-tuning}, we apply a synthetic data generation process formalized in Alg.~\ref{alg:data_generation}. It consists of two steps:

\textbf{Step 1. Generating metadata following user specification} (line 3 in Alg.~\ref{alg:data_generation}). By customizing the metadata, users can control which aspects of the model performance to focus on during the evaluation process. 
For example, if a user needs to evaluate the model's fraud detection fairness on different gender, age, or ethnicity groups, s/he may request to generate documents with uniform distributions across all groups, including minority groups. 
Users can specify metadata by either uploading a CSV file that lists metadata for each document or a YAML file that defines the probability distribution of each type of metadata, used as $\theta_{user}$ in Alg.~\ref{alg:data_generation}.

\begin{algorithm}[t]
\caption{\small Synthetic Data Generation
}
\label{alg:data_generation}
\begin{algorithmic}[1]
\scriptsize
\State \textbf{Input:}  User-specified metadata: $\theta_{user}$; control parameters tuned using Alg.~\ref{alg:parameter-tuning}: $\theta^*; $ID template: $T$; Synthetic data generator: $G$; Metadata generator: $F$
\State \textbf{Output:} Generated document $x_{syn}$

\State $meta$ $\gets$ $F_{\theta_{user}}()$\Comment{Generate metadata based on user specification}
\State \Return $x_{syn} = G_{\theta^*}(meta, T)$\Comment{Generate data using tuned control parameters}
\end{algorithmic}
\end{algorithm}

\textbf{Step 2. Generating the documents} (line 4 in Alg.~\ref{alg:data_generation}). Customized template documents, as shown in Fig.~\ref{fig:overview} and Fig.~\ref{fig:dataset}, emphasize geometric precision in text alignment and compliance with governmental design specifications. Such quality requirements are satisfied via the control parameters ($\theta^*$) tuned by our model-guided Bayesian optimization algorithm and the function ($G_{\theta^*}(\cdot)$) that applies those parameters to drive varying-scale synthesis by filling in the template following user-specified information.
In addition to the generated document, each output includes annotations such as typographic specifications, positional coordinates, semantic segment masks, and bounding boxes in JSON format, facilitating downstream tasks, e.g., constructing scanned or mobile documents. 

Taking scanned documents as an example, the pipeline first loads the user-specified customized template document. Then, it applies the tuned control parameters to fill user-specified metadata into the template image. 
After that, the function generates the scanner background image and combines the transformed customized template document and the background following the user-specified position and rotation parameters. 

The generation of mobile documents is similar. A user can flexibly specify to use the background of a certain existing mobile ID image (e.g., a picture of A's driver's license (DL) placed on top of the keyboard of a computer), and a template image, e.g., B's DL, to be blended by replacing A's DL in the image with B's DL. We used the MIDV's collection of mobile documents,  users can specify to generate a new mobile dataset with $x\%$ of indoor backgrounds and $1-x\%$ of outdoor backgrounds. Users can also specify the distribution of age, gender, ethnicity groups, and fraud patterns of the entities involved in the new template images to be blended with existing documents. 
To generate the mobile and scan documents, we employed a combination of advanced computer vision models and image processing techniques. In particular, we use Grounding DINO~\cite{liu2023grounding}, a state-of-the-art model that integrates object detection and language grounding, to detect and localize the document in the original background image. Once localized, the Segment Anything Model (SAM)~\cite{kirillov2023segment} is applied to obtain an accurate segmentation mask of the original document. The prompt-based interface of SAM allows for precise and flexible segmentation, which is critical for accurate geometric alignment.

For the blending stage, we adopt the Deep Image Blending (DIB) framework~\cite{Zhang2019DeepIB}, which synthesizes high-quality images by optimizing a combination of loss functions including Poisson gradient loss, content loss, style loss, histogram loss, and total variation loss. To enhance structural fidelity, we extend the DIB loss with an additional differentiable Structural Similarity Index (SSIM) loss. This augmentation improves both local and global consistency between the blended image and the background.

\section{\IDSpace Dataset}
\label{sec:dataset}

We first extracted $5{,}979$ adult entities from the generated.photos~\cite{generated.photos} dataset. Each entity has metadata information, such as gender, ethnicity group, and age, and a photo. Among these entities, we have $3{,}009$ females, and $2{,}970$ males. Among females, there are $825$ entities associated with the Asian ethnicity group, $925$ with Latino, $467$ with black, and $752$ with white. Among the males, there are $612$ with Asian, $787$ with Latino, $746$ with black, and $825$ with white. We used this information as user-specified metadata, of which the distribution is consistent with the \IDNet\cite{idnet-data}  dataset.

We used ten templates of identity document types from ten European countries, from our prior \IDNet work, including the ID cards and passports Albania, Azerbaijan, Estonia, Finland, Greece, Latvia, Russia, Serbia, Slovakia, and Spain. 

For each of these $5{,}979$ entities, using our \IDSpace data generation methodology detailed in Sec.~\ref{sec:problem} to Sec.~\ref{sec:data-generation}, we generated one non-fraud document, two fraud documents with the \textit{crop-and-move} fraud pattern and the \textit{inpaint-and-replacement} fraud pattern, respectively, for each of the ten European country identity document types. In the process, we used the ResNet50 model detailed in Sec.~\ref{sec:models} as the guiding models  We used the \textit{tuning set} with $40$ documents with balanced labels from the target domain to tune the control parameters. 

Then, similarly, for each of these generated documents, we further create one scanned document using randomly selected positions and rotations. We also generated $50$ mobile documents for each of the ten European identity document types. For each mobile document, we randomly sample one mobile ID document from the MIDV dataset as the background, and one templated document we generated for the given document type. 

In total, our new dataset has $359{,}240$ images, including $179{,}370$ templated documents, $179{,}370$ scanned documents, and a small set of $500$ mobile documents. We published our dataset on HuggingFace (See footnote~\ref{footnote1}) for public access. 

\section{Evaluation of \IDSpace Data Generator}
\label{sec:experiment}
We conducted a comprehensive empirical study to investigate the following research questions. R1. Will models trained on the target domain achieve consistent evaluation results on our generated document and the corresponding document from the target domain with the same metadata, and will the evaluation (prediction) consistency of our generated documents outperform alternatives? \textcolor{black}{R2. Does the dataset generated by \IDSpace also benefit the learning process?} R3. How is our proposed model-guided parameter tuning approach compared to alternatives?

\noindent
\textcolor{black}{\textbf{System Environment.} All experiments were conducted on an Ubuntu Linux server equipped with 48 CPU cores (Intel Xeon Silver 4310, 2.10 GHz), 125 GB of RAM, and two NVIDIA A10 GPUs (24GB VRAM each). The server is installed with 256GB NVMe SSD and 1TB HDD drive.}

\noindent
\textbf{Experiment Setup.} We used the MIDV~\cite{midv-dataset} and SIDTD~\cite{sidtd-dataset} datasets, which are under the CC BY-SA 2.5 and CC-BY-4.0 licenses, as the target domain. They include ten types of European country identity documents, with $100$ templated documents and around {$100$} fraud templated documents in each type. We sampled $50\%$ of documents to form a \textit{training set} for learning the fraud detection models, which are used as the guiding models for parameter tuning and the target models to be evaluated. $30\%$ of samples form the \textit{testing set} to evaluate the consistency of the prediction of the model of all the baseline data generation methods. The additional $20\%$ of the documents compose a \textit{tuning set} to adjust the control parameters following Alg.~\ref{alg:parameter-tuning}.
 Due to space limitations, our evaluation focuses on customized template images, which \textcolor{black}{is fundamental to identity fraud detection~\cite{boned2024synthetic, midv-dataset, guan2024idnet}}. We put the corresponding evaluation results for scanned documents in the Appendix. 
 In our implementation of Alg.~\ref{alg:parameter-tuning}, we used SSIM as our similarity metric. 
We set $\lambda_0$ to $1$ and $\lambda_i$ to $1/l$ ($l$ is the number of guiding models, $i=1,\dots,l$) to balance similarity and consistency. The tuning of $\lambda_i$ is discussed in the Appendix.
%
%

\subsection{R1. Prediction Consistency Comparison}
\label{sec:r1}

 \subsubsection{Using Models Independently Finetuned on the Target Domain} \label{sec:models} To answer R1, we first selected five fraud detection models with model architectures, including ViT-large, ResNet50, Inception-v3, VGG16, and DenseNet, and finetuned them independently using documents from the target domain. We choose those models because they are widely adopted in academia and industry for fraud detection on identity documents~\cite{boned2024synthetic, park2023kid34k, onfido, bayer2025authentication, khare2024predictive, mahadevan2023generalized, bruveris2020reducing, gietema2024method}, 

 We then use each of the following baselines with the \textit{tuning set} to generate a dataset, using the same metadata as the samples from the \textit{testing set} that is also from the target domain. We further compare the model prediction consistency, defined in Eq.~\ref{eq:consistency}, between the dataset generated by each baseline and the \textit{testing set} for each of the five models.

$\bullet$ \textbf{BO w/ SSIM-only objective}: This approach represents \IDNet, which did not adopt any model guidance, i.e., using Eq.~\ref{eq:similarity} as the objective function of the Bayesian Optimization (BO) search.

$\bullet$ \textbf{CycleGAN}: This is a widely used domain adaptation approach with pre-trained models publicly available~\cite{CycleGAN2017, cyclegan-code} under the BSD license. We chose the CycleGAN model pretrained on the Flickr dataset~\cite{flickr}, finetuned it using the \textit{tuning set}, and used the finetuned model to adapt an ID dataset generated by vanilla BO (i.e., BO w/ SSIM-only objective) to the target domain. 

$\bullet$ \textbf{Diffusion-based inpainting}: We include a diffusion-based model as a baseline using image-to-image generation with inpainting. Specifically, we use the pretrained Stable Diffusion v1.5 inpainting model (stable-diffusion-v1-5/stable-diffusion-inpainting)~\cite{rombach2022high}. Given an input identity document image, a text prompt specifying the target segment, and a mask indicating the editable region, the model generates an output image in which the segment is updated while the remaining content is preserved. This baseline enables comparison between diffusion-based inpainting and other approaches for controlled identity document editing. 

$\bullet$ \textbf{\IDSpace}: This is our approach using different combinations of guiding models for BO search. 




\vspace{3pt}
\noindent
\textbf{Prediction Consistency}
The comparison results are presented in Tab.~\ref{tab:cross_model}. We observed significant improvement of consistency scores using our proposed model-guided Bayesian optimization methodology, ranging from ${15.62}{\%}$ to ${30.55}{\%}$ compared to BO w/ SSIM-only objective, \textcolor{black}{${27.40}{\%}$ to ${42.33}{\%}$ compared to CycleGAN}, and \textcolor{black}{${30.60}{\%}$ to ${45.51}{\%}$ compared to Diffusion-based inpainting}. We also found that using our proposed approach, the consistency is not only improved for the guiding models used in the BO objective function, but also improved for other models trained on the target domain. In addition, incorporating more guiding models further improved consistency in the majority of the cases. The algorithm demonstrated robustness across different architectures (mean consistency = ${0.9354}{\pm 0.003}{\%}$). Using our approach, small models such as Inception-v3 and DenseNet are easier to achieve better consistency than other larger models.

\begin{table*}[h]
\centering
\caption{\small Model prediction consistency evaluation on different models with best results highlighted in \textcolor{blue}{blue}}
\scriptsize
\label{tab:cross_model}
\resizebox{\textwidth}{!}{%
\begin{tabular}{lcccccc}
\toprule
\multirow{2}{*}& \multicolumn{5}{c}{\textbf{Model Prediction Consistency} with Target Test Data (Mean $\pm$ Std)} &\multirow{2}{*}{Average(row)}\\
\cmidrule(lr){2-6}
 & ViT-Large & ResNet50 & Inception-v3 & VGG16 & DenseNet  \\
\midrule
CycleGAN & $0.5648 \pm 0.004$ & $0.5324 \pm 0.000$ & $0.5093 \pm 0.000$ & $0.5648 \pm 0.000$ & $0.5000 \pm 0.000$ & $0.5343 \pm 0.001 $ \\

Diffusion-based inpainting & $0.5000 \pm 0.000$ & $0.4958 \pm 0.000$ & $0.5000 \pm 0.000$ & $0.5167 \pm 0.000$ & $0.5000 \pm 0.000$ & $0.5025 \pm 0.000$ \\
BO w/ SSIM-only objective (\IDNet)   & $0.7257 \pm 0.018$ & $0.5382 \pm 0.006$ & $0.8646 \pm 0.012$ & $0.6146 \pm 0.006$ & $0.5174 \pm 0.012$ & $0.6521 \pm 0.017$ \\
\hline\hline
Guiding models&\multicolumn{6}{c}{\texttt{\IDSpace}} \\ 
\hline
DenseNet & $0.8819 \pm 0.093$ & $0.8646 \pm 0.018$ & $0.9965 \pm 0.006$ & $0.8715 \pm 0.040$ & $\textcolor{blue}{1.0000 \pm 0.000}$ & $0.9229 \pm 0.005$ \\
DenseNet + Inception-v3 & $0.8021 \pm 0.136$ & $0.8750 \pm 0.017$ & $0.9931 \pm 0.007$ & $0.8715 \pm 0.060$ & $\textcolor{blue}{1.0000 \pm 0.000}$ & $0.9083 \pm 0.007$ \\
DenseNet + Inception-v3 + ResNet50 & $0.8194 \pm 0.091$ & $0.9306 \pm 0.017$ & $\textcolor{blue}{1.0000 \pm 0.000}$ & $0.8889 \pm 0.028$ & $\textcolor{blue}{1.0000 \pm 0.000}$ & $0.9278 \pm 0.006$ \\
DenseNet + Inception-v3 + VGG16 & $0.8924 \pm 0.015$ & $0.8646 \pm 0.036$ & $\textcolor{blue}{1.0000 \pm 0.000}$ & $0.9410 \pm 0.025$ & $\textcolor{blue}{1.0000 \pm 0.000}$ & $0.9396 \pm 0.004$ \\
DenseNet + Inception-v3 + ViT-Large & $0.9306 \pm 0.020$ & $0.8889 \pm 0.045$ & $0.9965 \pm 0.006$ & $0.9097 \pm 0.036$ & $\textcolor{blue}{1.0000 \pm 0.000}$ & $0.9451 \pm 0.003$ \\
DenseNet + ResNet50 & $0.9340 \pm 0.015$ & $0.9236 \pm 0.016$ & $\textcolor{blue}{1.0000 \pm 0.000}$ & $0.8819 \pm 0.035$ & $\textcolor{blue}{1.0000 \pm 0.000}$ & $0.9479 \pm 0.003$ \\
DenseNet + ResNet50 + VGG16 & $0.9236 \pm 0.029$ & $0.8924 \pm 0.006$ & $1.0000 \pm 0.000$ & $0.9375 \pm 0.016$ & $\textcolor{blue}{1.0000 \pm 0.000}$ & $0.9507 \pm 0.002$ \\
DenseNet + Resnet50 + ViT-Large & $0.9201 \pm 0.033$ & $0.9062 \pm 0.041$ & $0.9792 \pm 0.036$ & $0.8403 \pm 0.094$ & $0.9965 \pm 0.006$ & $0.9285 \pm 0.004$ \\
DenseNet + VGG16 & $0.9132 \pm 0.012$ & $0.8819 \pm 0.029$ & $\textcolor{blue}{1.0000 \pm 0.000}$ & $0.9306 \pm 0.034$ & $\textcolor{blue}{1.0000 \pm 0.000}$ & $0.9451 \pm 0.003$ \\
DenseNet + VGG16 + ViT-Large & $0.9306 \pm 0.017$ & $0.8542 \pm 0.030$ & $\textcolor{blue}{1.0000 \pm 0.000}$ & $0.9479 \pm 0.032$ & $\textcolor{blue}{1.0000 \pm 0.000}$ & $0.9465 \pm 0.004$ \\
DenseNet + ViT-Large & $0.9306 \pm 0.039$ & $0.8646 \pm 0.060$ & $\textcolor{blue}{1.0000 \pm 0.000}$ & $0.8958 \pm 0.029$ & $\textcolor{blue}{1.0000 \pm 0.000}$ & $0.9382 \pm 0.004$ \\
Inception-v3 & $0.6146 \pm 0.108$ & $0.7778 \pm 0.057$ & $0.9931 \pm 0.007$ & $0.7118 \pm 0.129$ & $0.9444 \pm 0.056$ & $0.8083 \pm 0.025$ \\
Inception-v3 + ResNet50 & $0.7708 \pm 0.068$ & $0.9340 \pm 0.030$ & $0.9965 \pm 0.006$ & $0.8681 \pm 0.012$ & $0.9965 \pm 0.006$ & $0.9132 \pm 0.009$ \\
Inception-v3 + ResNet50 + VGG16 & $0.9375 \pm 0.023$ & $0.9097 \pm 0.025$ & $0.9965 \pm 0.006$ & $0.9479 \pm 0.027$ & $0.9965 \pm 0.006$ & $\textcolor{blue}{0.9576 \pm 0.001}$ \\
Inception-v3 + ResNet50 + ViT-Large & $0.9375 \pm 0.007$ & $0.9271 \pm 0.027$ & $0.9965 \pm 0.006$ & $0.8785 \pm 0.027$ & $\textcolor{blue}{1.0000 \pm 0.000}$ & $0.9479 \pm 0.003$ \\
Inception-v3 + VGG16 & $0.9375 \pm 0.030$ & $0.8750 \pm 0.026$ & $\textcolor{blue}{1.0000 \pm 0.000}$ & $\textcolor{blue}{0.9549 \pm 0.025}$ & $\textcolor{blue}{1.0000 \pm 0.000}$ & $0.9535 \pm 0.003$ \\
Inception-v3 + VGG16 + ViT-Large & $0.9410 \pm 0.030$ & $0.8194 \pm 0.033$ & $\textcolor{blue}{1.0000 \pm 0.000}$ & $0.9410 \pm 0.025$ & $\textcolor{blue}{1.0000 \pm 0.000}$ & $0.9403 \pm 0.005$ \\
Inception-v3 + ViT-Large & $0.9340 \pm 0.023$ & $0.8576 \pm 0.021$ & $0.9965 \pm 0.006$ & $0.9062 \pm 0.032$ & $\textcolor{blue}{1.0000 \pm 0.000}$ & $0.9389 \pm 0.004$ \\
ResNet50 & $0.9271 \pm 0.015$ & $\textcolor{blue}{0.9514 \pm 0.012}$ & $\textcolor{blue}{1.0000 \pm 0.000}$ & $0.8993 \pm 0.023$ & $\textcolor{blue}{1.0000 \pm 0.000}$ & $0.9556 \pm 0.002$ \\
ResNet50 + VGG16 & $0.9097 \pm 0.016$ & $0.9097 \pm 0.029$ & $\textcolor{blue}{1.0000 \pm 0.000}$ & $0.9306 \pm 0.014$ & $\textcolor{blue}{1.0000 \pm 0.000}$ & $0.9500 \pm 0.002$ \\
ResNet50 + VGG16 + ViT-Large & $0.9375 \pm 0.012$ & $0.9167 \pm 0.026$ & $\textcolor{blue}{1.0000 \pm 0.000}$ & $0.9132 \pm 0.006$ & $\textcolor{blue}{1.0000 \pm 0.000}$ & $0.9535 \pm 0.002$ \\
ResNet50 + ViT-Large & $0.9132 \pm 0.023$ & $0.9340 \pm 0.050$ & $0.9965 \pm 0.006$ & $0.8785 \pm 0.032$ & $1.0000 \pm 0.000$ & $0.9444 \pm 0.003$ \\
VGG16 & $0.9271 \pm 0.040$ & $0.8368 \pm 0.054$ & $0.9965 \pm 0.006$ & $0.9306 \pm 0.052$ & $\textcolor{blue}{1.0000 \pm 0.000}$ & $0.9382 \pm 0.004$ \\
VGG16 + ViT-Large & $\textcolor{blue}{0.9514 \pm 0.021}$ & $0.8542 \pm 0.025$ & $\textcolor{blue}{1.0000 \pm 0.000}$ & $0.9444 \pm 0.017$ & $\textcolor{blue}{1.0000 \pm 0.000}$ & $0.9500 \pm 0.004$ \\
ViT-Large & $0.9375 \pm 0.016$ & $0.8646 \pm 0.065$ & $0.9861 \pm 0.010$ & $0.8750 \pm 0.024$ & $0.9965 \pm 0.006$ & $0.9319 \pm 0.004$ \\
Average(Column) & $0.8982 \pm 0.006$ & $0.8846 \pm 0.002$ & $0.9969 \pm 0.000$ & $0.8999 \pm 0.003$ & $0.9972 \pm 0.000$ & \\ 
\bottomrule
\end{tabular}%
}
\end{table*}

\begin{table}[t]
\centering
\caption{\small Comparison of consistency using different numbers of target samples (the first column) with the best results in \textcolor{blue}{blue}.
}
\scriptsize
\label{tab:results1}
\begin{tabular}{ccccc}
\toprule
 &{CycleGAN} &{Diffusion} &{\IDNet} & {\IDSpace} \\
\midrule
2  &$0.546 \pm 0.000$& $0.500 \pm 0.000$ & $0.413 \pm 0.000$  & $\textcolor{blue}{0.806 \pm 0.008}$ \\
20 &$0.542 \pm 0.000$& $0.500 \pm 0.000$ & $0.170 \pm 0.000$ & $\textcolor{blue}{0.944 \pm 0.000}$ \\
40 &$0.532 \pm 0.000$ & $0.496 \pm 0.000$ & $0.538 \pm 0.006$ & $\textcolor{blue}{0.951 \pm 0.012}$\\
\bottomrule
\end{tabular}
\end{table}

To investigate how the model prediction consistency changes with the number of samples available from the target domain for different baselines, 
%
%
we applied $2$, $20$, and $40$ samples (with balanced fraud and non-fraud labels) from the \textit{tuning set} to \textcolor{black}{finetune} the CycleGAN model, and to tune the control parameters for the baseline using BO w/ SSIM-only objective and our model-guided \IDSpace approach. We used the \textit{testing set} that is disjoint with the \textit{tuning set} to measure the prediction consistency of the ResNet50 model used in Tab.~\ref{tab:cross_model} between the documents from the \textit{testing set} and the generated documents, both of which share the same metadata (i.e., field values and photos). Our \IDSpace approach used the target ResNet50 model as the guiding model.

The results in Tab.~\ref{tab:results1} showed that our approach consistently and significantly improved prediction consistency even when fewer samples are available from the target domain. This result demonstrated the benefits of incorporating guiding models into the search objective for a few-shot approach. Furthermore, the overall objective function improved as the number of samples from the target domain increased. Notably, even with just two samples, our algorithm demonstrated substantial improvement compared to baselines, proving the effectiveness of our proposed model-guided framework. 

\begin{figure}[h]
    \centering    \includegraphics[width=0.85\linewidth]{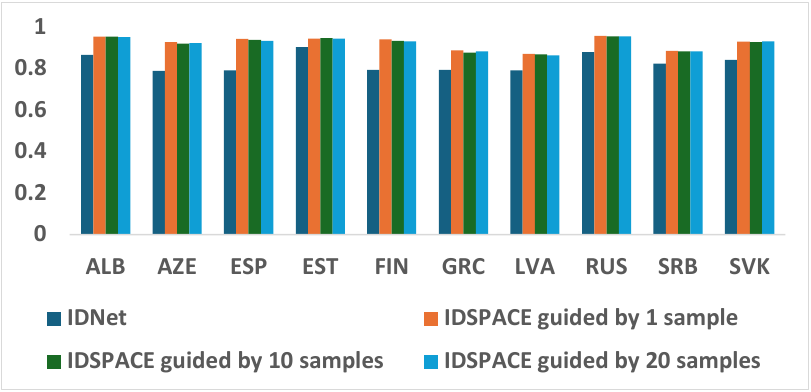}
    \caption{\small Structural Similarity (SSIM) comparison between \IDNet and MIDV vs. \IDSpace and MIDV, guided by different numbers of target-domain samples across multiple regions.}
    \label{fig:ssim_comparison}
\end{figure}

\vspace{3pt}
\noindent
\textbf{Fidelity.} We further evaluate the SSIM similarity between MIDV documents and the documents generated by \IDNet and \IDSpace across multiple target regions over multiple test samples, as shown in Figure~\ref{fig:ssim_comparison}. \IDSpace consistently outperforms \IDNet across all regions. Notably, substantial gains are achieved even with a single guidance sample, and increasing the number of guidance samples from 1 to 10 or 20 does not consistently improve SSIM and can slightly degrade performance in some regions. This behavior suggests that a small number of guidance samples induces a strong, coherent structural prior that generalizes well across the test set, whereas incorporating additional guidance samples increases structural variability, leading to a more flexible but less tightly constrained mapping. Since SSIM is particularly sensitive to fine-grained structural alignment, this increased variability can lead to slightly lower average SSIM scores, while still maintaining clear improvements over \IDNet.

\begin{figure*}[h]
\scriptsize
  \begin{minipage}[b]{0.7\textwidth}
    \centering
    \begin{tabular}{p{1.2cm}cccc}

\toprule
\multirow{2}{*}{} & \multicolumn{4}{c}{Target Models} \\ \cmidrule{2-5} 
                        &  ResNet50  &Inception-v3& 
                        DenseNet  &EfficientNet  
                        \\ \midrule
\IDNet                & $0.9156 \pm 0.016$    & $0.9536 \pm 0.006$& 
$0.8734 \pm 0.012$ & $0.9873 \pm 0.000$ 
\\
\IDSpace                & $ \textcolor{blue}{0.9866 \pm 0.000}$  &$\textcolor{blue}{0.9958 \pm 0.000}$ & 
$\textcolor{blue}{0.9536 \pm 0.006}$& $\textcolor{blue}{0.9958 \pm 0.000}$ 
\\
 \bottomrule
\end{tabular}
    \captionof{table}{\small \textcolor{black}{Utility (detection accuracy of the copy-and-move frauds) of the models trained on \IDNet and \IDSpace, w/ best values highlighted in \textcolor{blue}{blue}.}}
    \label{tab:training-utility-comparison}
  \end{minipage}
  \hfill
  \begin{minipage}[b]{0.28\textwidth}
    \centering
    \includegraphics[width=\linewidth]{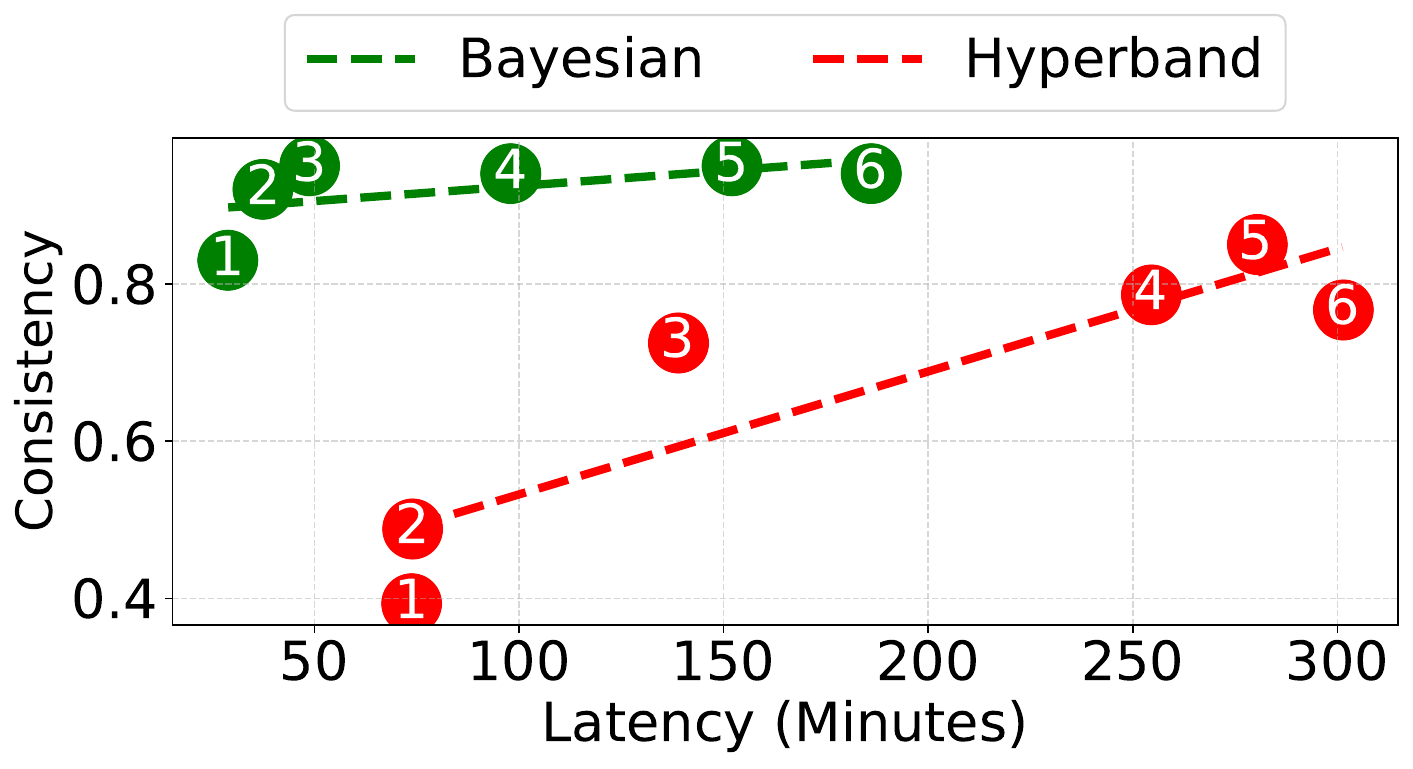} 
    \caption{\small{Bayesian search vs. Hyperband search}}
    \label{fig:cmp-hyperband}
  \end{minipage}
\end{figure*}

\noindent

\vspace{3pt}
\noindent
\textbf{Control Parameter Value Variation during the Model-Guided Optimization Process.} Models optimized jointly usually (e.g., two- or three-model combinations)
exhibit less variance than their single-model counterparts. 
This indicates that multi-model BO
acts as a regularizer: because the joint objective must satisfy multiple
architectures simultaneously, the search converges toward parameter
regions that are robust across models, resulting in more stable and
predictable behavior. For most of the models, parameters such as \texttt{xc}, \texttt{yc}, and \texttt{zc}, which encode RGB color channels, \texttt{font\_style} and \texttt{font\_size}, and  \texttt{stroke\_w}, representing stroke width, exhibit 
high variance, indicating their significant impact on the effectiveness of generated examples.

\vspace{3pt}
\noindent
\subsubsection{Limitations of Generative AI Models}
We evaluated representative generative approaches---including StyleGAN~\cite{karras2020analyzing}, diffusion-based image generation~\cite{rombach2022high} (i.e., text-to-image, different from the Diffusion-based inpainting), and recent large generative models such as GPT-4o and GPT-image-1 from OpenAI~\cite{chatgpt}---under few-shot and limited-data regimes consistent with real-world identity verification settings. StyleGAN and diffusion models trained with limited target-domain data (e.g., tens of images) fail to preserve the fine-grained structural properties of identity documents, exhibiting distortions in layout, typography, and field alignment. While large generative models demonstrate stronger global coherence in few-shot prompting scenarios, they similarly struggle to maintain precise document structure and layout fidelity required for reliable benchmarking. As a result, documents generated by these approaches struggle to maintain the precise document structure and layout fidelity, limiting their suitability for evaluation under data scarcity, as elaborated in the Appendix.

\subsubsection{Can LLM detect \IDSpace documents as generated or synthetic?}
\label{sec:detect}
We conducted experiments to test whether existing LLM models such as GPT-4o could detect that our generated documents are generated. As shown in the table below, we tested with a zero-shot method and the few-shot method, providing $2$, $4$, and $6$ examples in the generated and real categories, respectively, but the performance of GPT-4o remained poor as shown in Tab.~\ref{tab:gpt-4o-experiment}. Our synthetic data generation method is stealthy to GPT-4o.

\begin{table*}
  \caption{\small Results of using GPT-4o to detect whether our generated documents are real or generated.
  }
  \label{tab:gpt-4o-experiment}
  \scriptsize
  \centering
  \begin{tabular}{ccccc}
    \toprule
      Methods   &  ACC	& PRC	& Recall	& F1\\
    \midrule
    Zero-shot & 0.4667 \ensuremath{\pm} 0.058	& 0.4167 \ensuremath{\pm} 0.144	& 0.1167 \ensuremath{\pm} 0.020  &	 0.1801 \ensuremath{\pm} 0.04\\\hline
Few-shot (2 demonstrations for real and generated respectively)	& 0.5167 \ensuremath{\pm} 0.058 &	 0.5170 \ensuremath{\pm} 0.058	& 0.5167 \ensuremath{\pm} 0.076 &	 0.5159 \ensuremath{\pm} 0.061\\\hline 
Few-shot (4 demonstrations for real and generated respectively)	& 0.5917 \ensuremath{\pm} 0.076 &	 0.5796 \ensuremath{\pm} 0.060	& 0.6667 \ensuremath{\pm} 0.126 &	 0.6177 \ensuremath{\pm} 0.081\\\hline 
Few-shot (6 demonstrations for real and generated respectively)	& 0.5750 \ensuremath{\pm} 0.090 & 0.5693 \ensuremath{\pm} 0.077	& 0.5833 \ensuremath{\pm} 0.144	& 0.5747 \ensuremath{\pm} 0.107
    \\
    \bottomrule
  \end{tabular}
\end{table*}

\subsection{R2.  Does the dataset generated by \IDSpace also benefit the learning process?}
As demonstrated in Sec.~\ref{sec:r1}, our proposed approach significantly improved the evaluation consistency of the generated documents with the \textit{testing set}. Next, we will evaluate whether the fraud detection models trained on our generated dataset could also generalize well to the documents in the target domain. 
{\color{black}
To compare the utility of the \IDSpace-generated dataset described in Sec.~\ref{sec:dataset} to the \IDNet\cite{idnet-data} dataset, we trained different fraud detection models on each dataset and evaluated the utility of these models using the evaluation dataset from MIDV~\cite{bulatovich2022midv} and SIDTD~\cite{benalcazar2023synthetic} (i.e., the target domain). The results are shown in Tab.~\ref{tab:training-utility-comparison}, which demonstrates the excellent utility of our \IDSpace framework in learning fraud detection tasks in the target domain, outperforming \IDNet by up to $9\%$.

}

\subsection{R3. Tuning Algorithm Comparison}
\label{sec:tuning}

In this section, we compare our model-guided Bayesian optimization (Bayesian) method with Hyperband, which accelerates the search for optimal configurations by adaptively allocating resources to promising candidates using early-stopping and successive halving~\cite{li2018hyperband}. For both approaches, we used $40$ (i.e., $20\%$) samples from the \textit{tuning set}, and ResNet50 served as the guiding and the target model. In our approach, hyperparameters \textit{init\_point} and \textit{n\_iter} control the accuracy vs latency tradeoff. Similarly, in Hyperband search, \textit{max\_resources} controls the maximum amount of resources that can be allocated to a single configuration, and $\eta$ controls the proportion of configurations discarded in each round of successive halving. Let us denote each instance of $(init\_point, n\_iter)$ pairs and $(max\_resources, \eta)$ pairs as $b_i$ and $h_i$, respectively. In Fig.~\ref{fig:cmp-hyperband}, the evaluated $\{b_1, b_2, ..., b_6\}$ (green points) for our approach are $\{(50, 100), (50, 150), (50, 200), (100, 400), (100, 600), (100, $ $800)\}$, and $\{h_1, h_2, ..., h_6\}$ (red points) for Hyperband search include $\{(500, 3), (700, 3), $ $(500, 2), (900, 2), (1000, 2), (1050, 2)\}$. As shown in the figure, our approach outperforms Hyperband search in terms of both tuning latency and consistency score. We observe that our approach achieves the peak consistency score of $0.95$ for $b_3$, which took $48.9$ minutes, while the Hyperband search reached its maximum consistency score of $0.85$ for $h_5$, taking $280.15$ minutes. Therefore, our approach outperforms the Hyperband by $5.7\times$ in terms of tuning latency while the peak consistency score achieved by Bayesian optimization is $11.76\%$ better than the best consistency score of Hyperband search.

\section{Our \IDNet Dataset and Conclusions}\label{sec:conclusion}

The work is motivated by real-world requirements for a flexible and parameterized synthetic identity document generation framework from US General Services Administration and US Department of Homeland Security, where privacy regulations cause a lack of data for evaluating vendors' software. We proposed \IDSpace, a novel model-guided synthetic data generation approach, to address the shortage of accessible real data for trustworthy, reliable, comprehensive, and flexible evaluation of existing fraud detection models. Furthermore, our model-guided optimization-based strategy ensures that synthetic data can be aligned with the target domain using only a small set of real samples, significantly reducing the costs and dependency on sensitive real documents for both evaluation and training. Empirical evidence shows that \IDSpace improves evaluation consistency by $15-45\%$ over baselines including CycleGAN, diffusion inpainting, and non-guided optimization, such as \IDNet, using only $2$ to $40$ real samples, while improving training accuracy by up to $9\%$ and SSIM similarity with the target domain by $10\%$, compared to \IDNet.

\noindent
\textbf{Ethics.} Synthetic data plays a critical role in reducing privacy risks, yet we recognize the potential for dual-use. Malicious actors might attempt to misuse our framework for producing counterfeit documents. To mitigate this, we deliberately restrict the realism of generated outputs, ensuring that synthetic IDs do not contain functionally valid elements such as scannable barcodes. In addition, all portrait photos and ID entity information used in this study are collected from publicly available generated.photos dataset, which is $100\%$ synthetically generated. All portrait photos are $100\%$ collected from \textcolor{black}{a public synthetic dataset for academic research~\cite{generated.photos}}.

\bibliographystyle{ieeetr}
\bibliography{ref}

\appendix

\section{Technical Appendices and Supplementary Material}

\subsection{More Document Examples.}
\label{sec:examples}

\begin{figure*}[h!]
    \centering
    \subfigure[\tiny Templated sample from MIDV]{
        \includegraphics[width=0.3\textwidth]{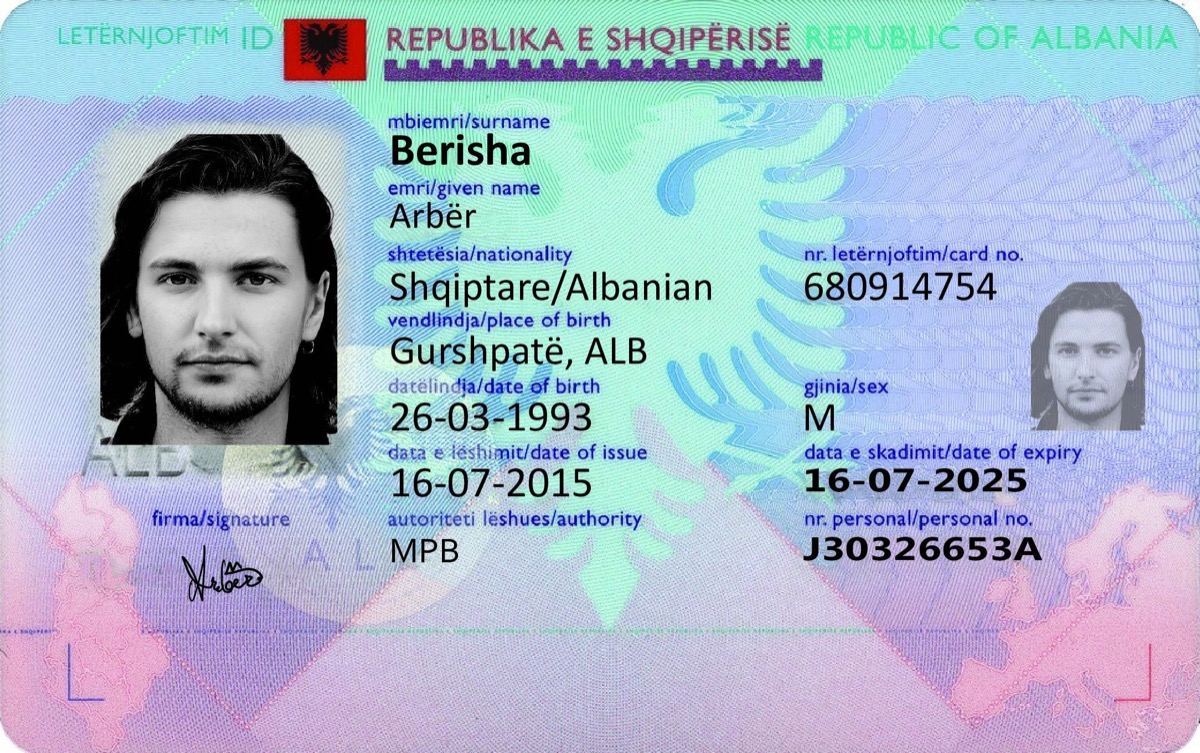}
    }
    \subfigure[\tiny Inpaint\&Rewrite fraud sample from SIDTD]{
        \includegraphics[width=0.3\textwidth]{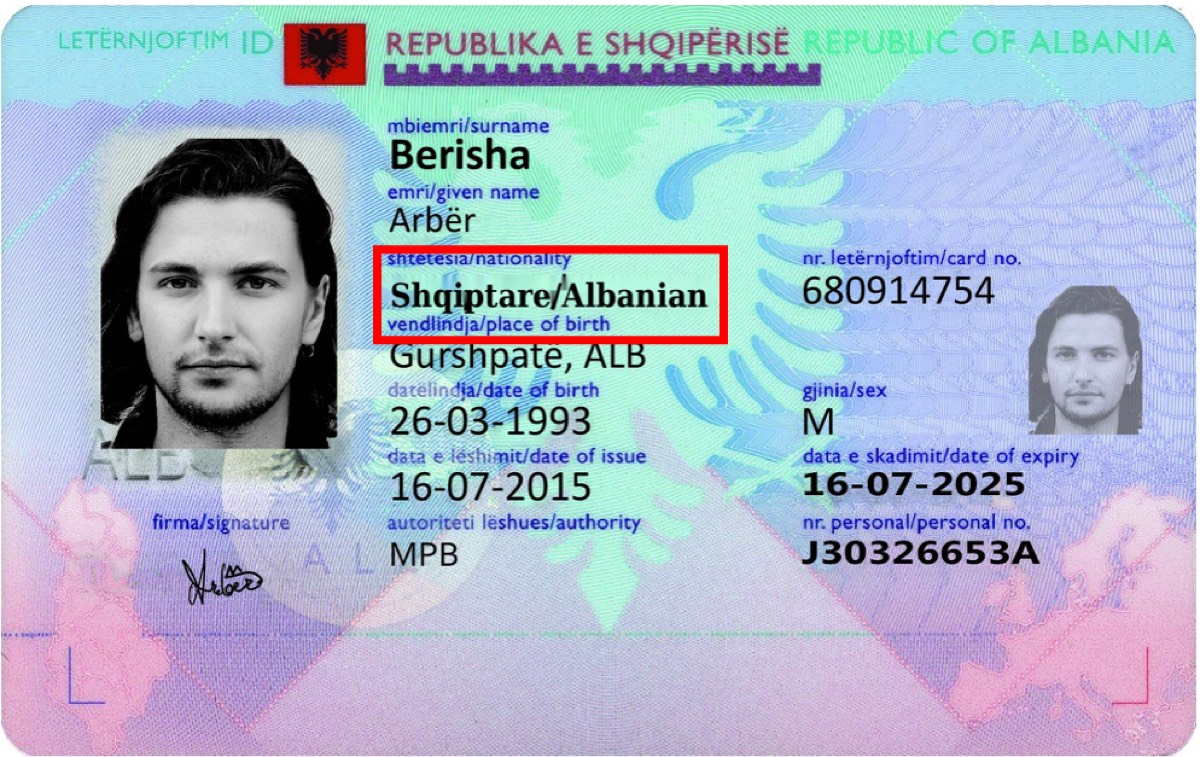}
    }
    \subfigure[\tiny Crop\&Replace fraud sample from SIDTD]{
        \includegraphics[width=0.3\textwidth]{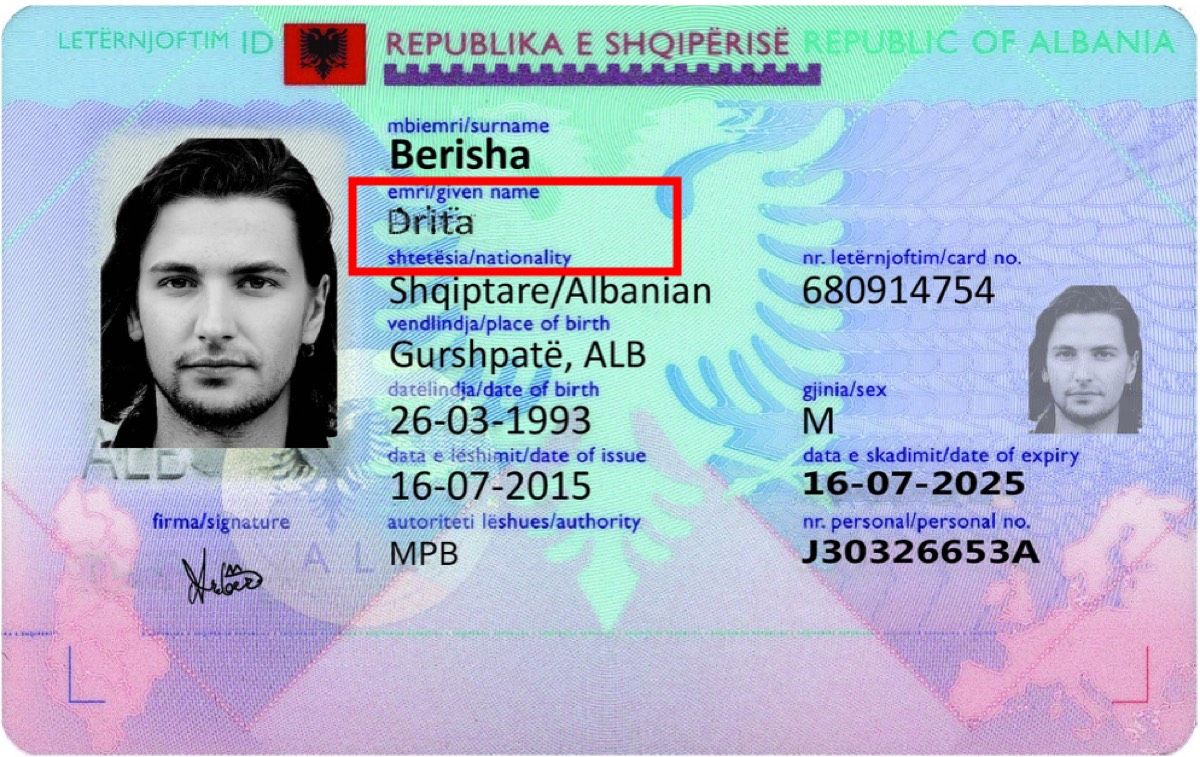}
    }
        \subfigure[\tiny Templated sample from \IDSpace]{
        \includegraphics[width=0.3\textwidth]{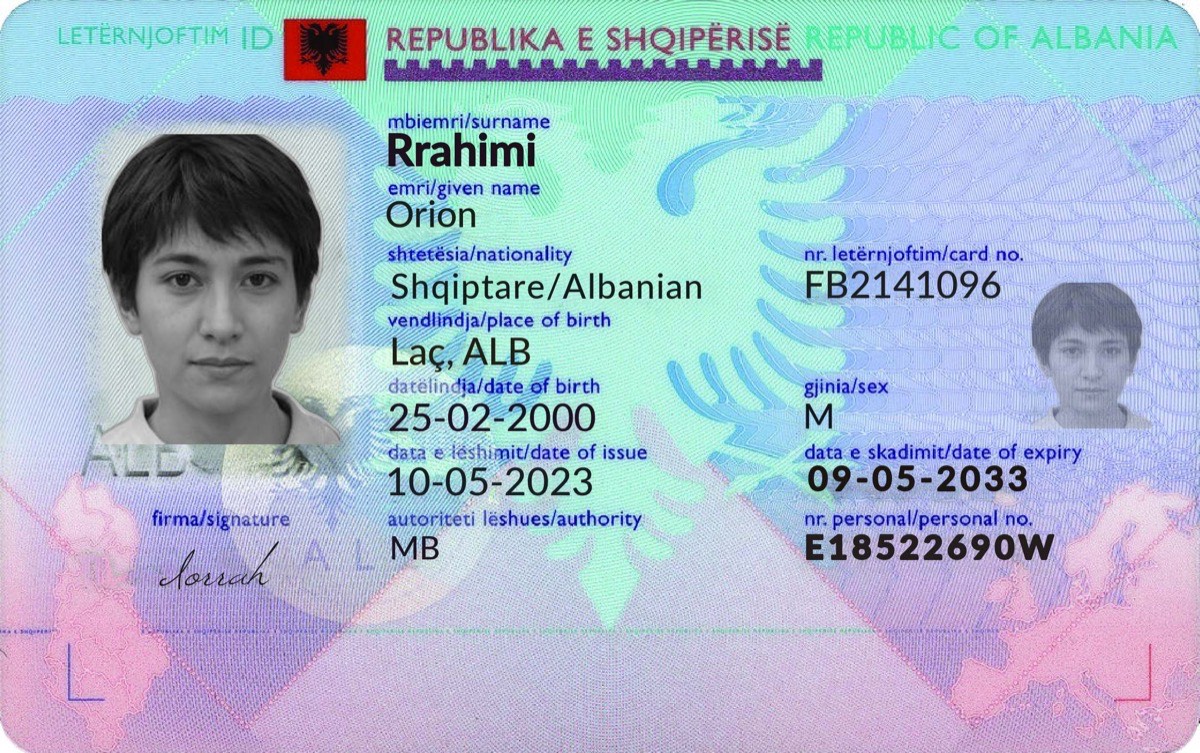}
    }
    \subfigure[\tiny Inpaint\&Rewrite fraud sample from \IDSpace]{
        \includegraphics[width=0.3\textwidth]{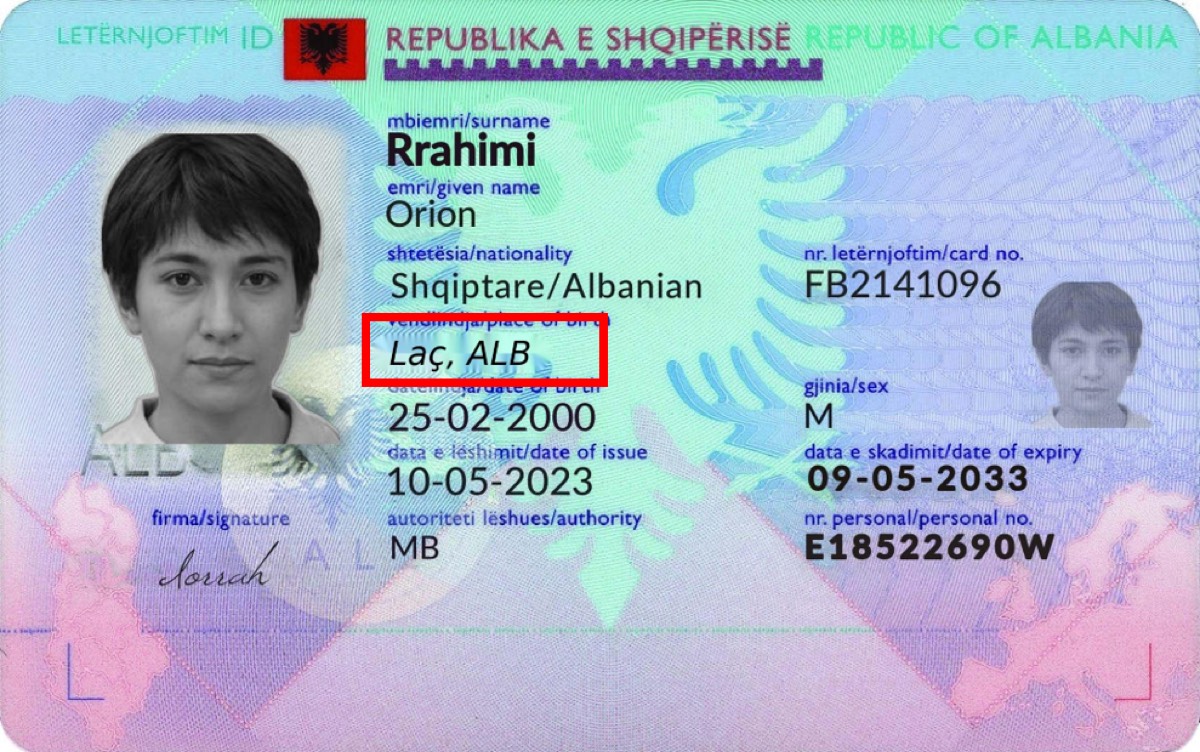}
    }
    \subfigure[\tiny Crop\&Replace fraud sample from \IDSpace]{
        \includegraphics[width=0.3\textwidth]{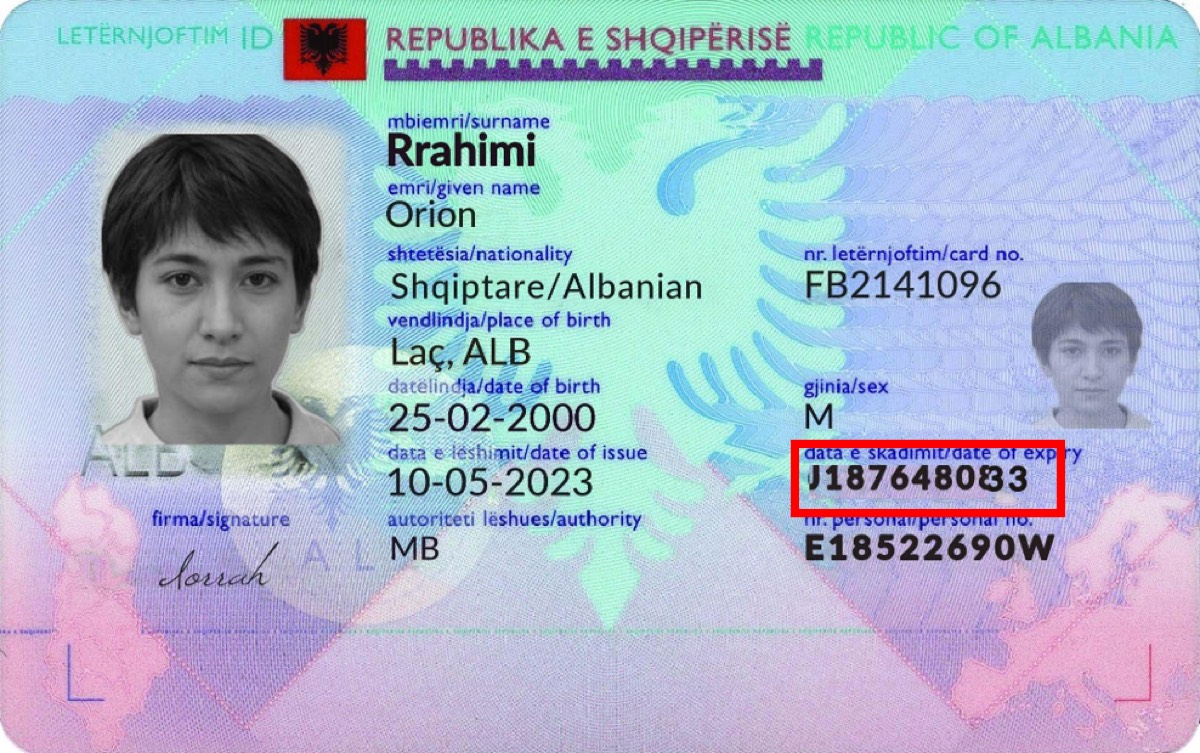}
    }

        \subfigure[\tiny Scanned sample from MIDV]{
        \includegraphics[width=0.22\textwidth]{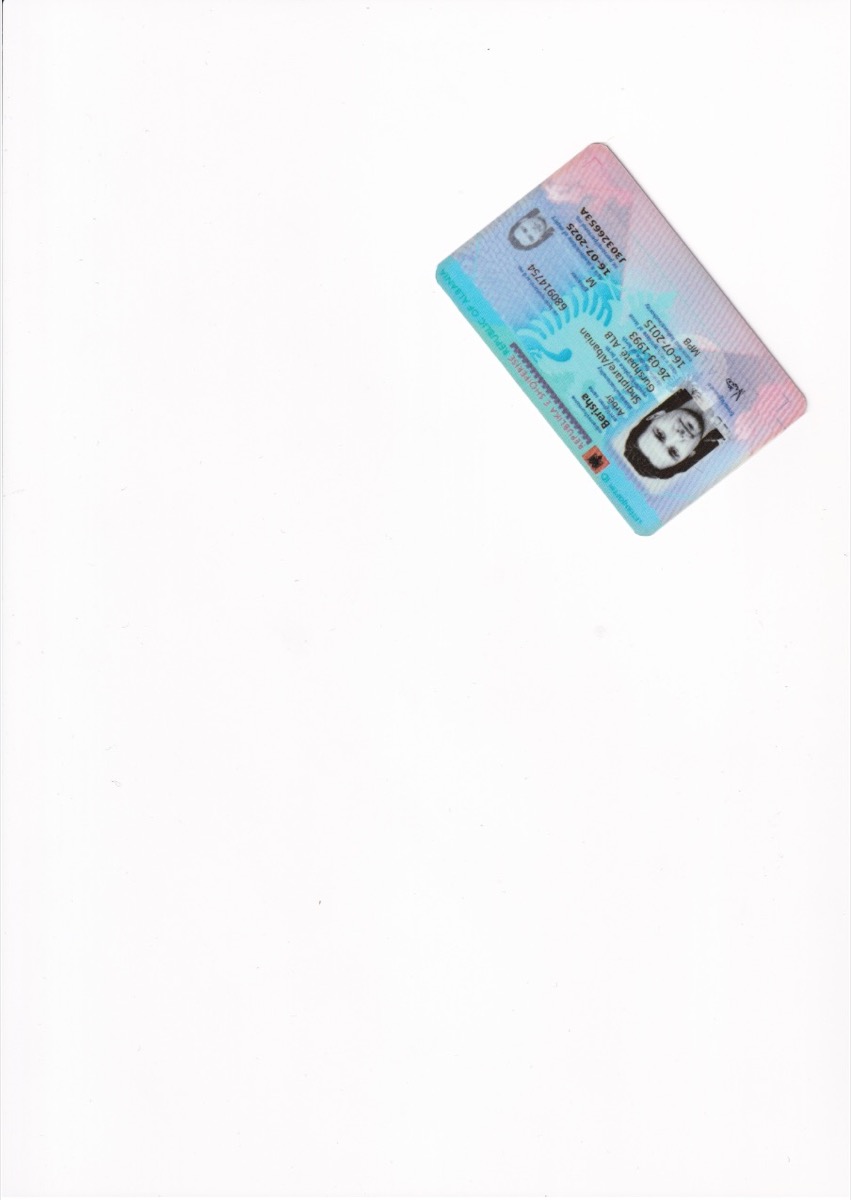}
    }
    \subfigure[\tiny Scanned sample from \IDSpace]{
        \includegraphics[width=0.22\textwidth]{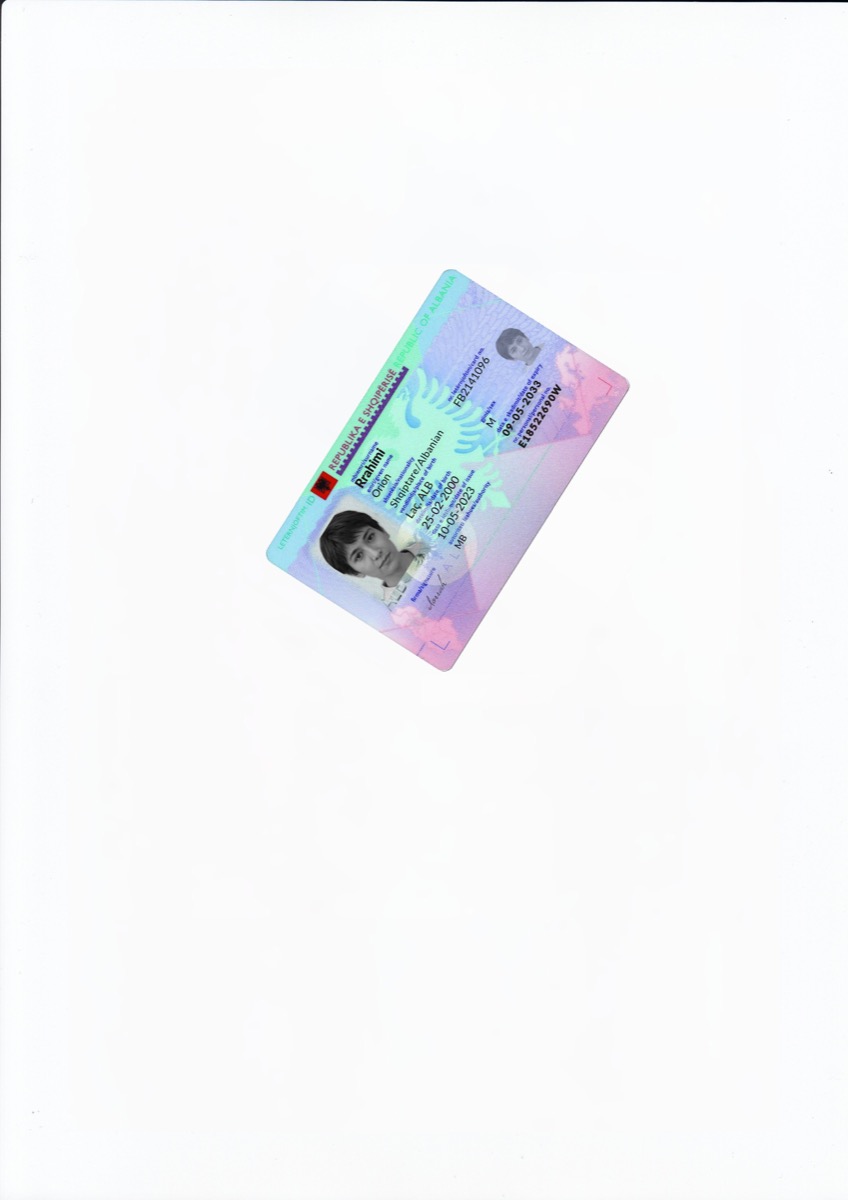}
    }
    \subfigure[\tiny Scanned sample with Inpaint\&Rewrite fraud from \IDSpace]{
        \includegraphics[width=0.22\textwidth]{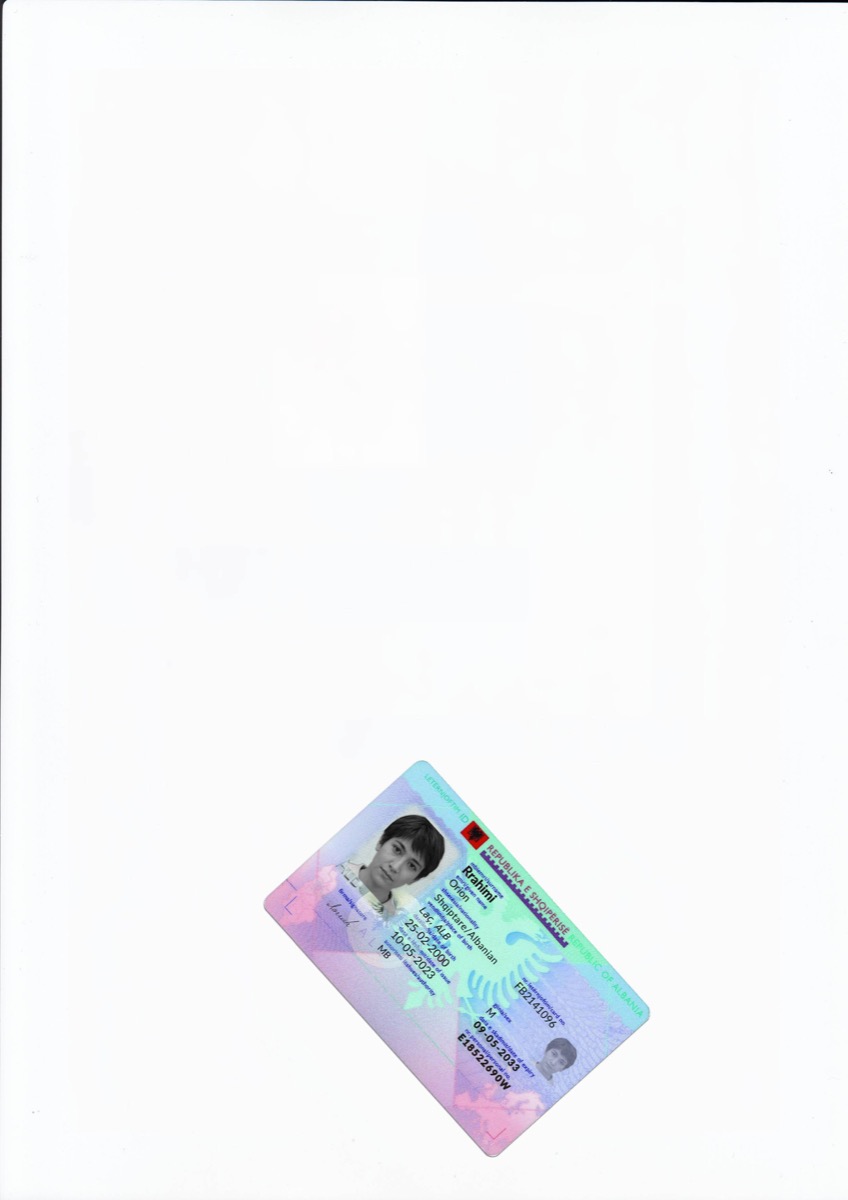}
    }
    \subfigure[\tiny Scanned sample with Crop\&Replace fraud from \IDSpace]{
        \includegraphics[width=0.22\textwidth]{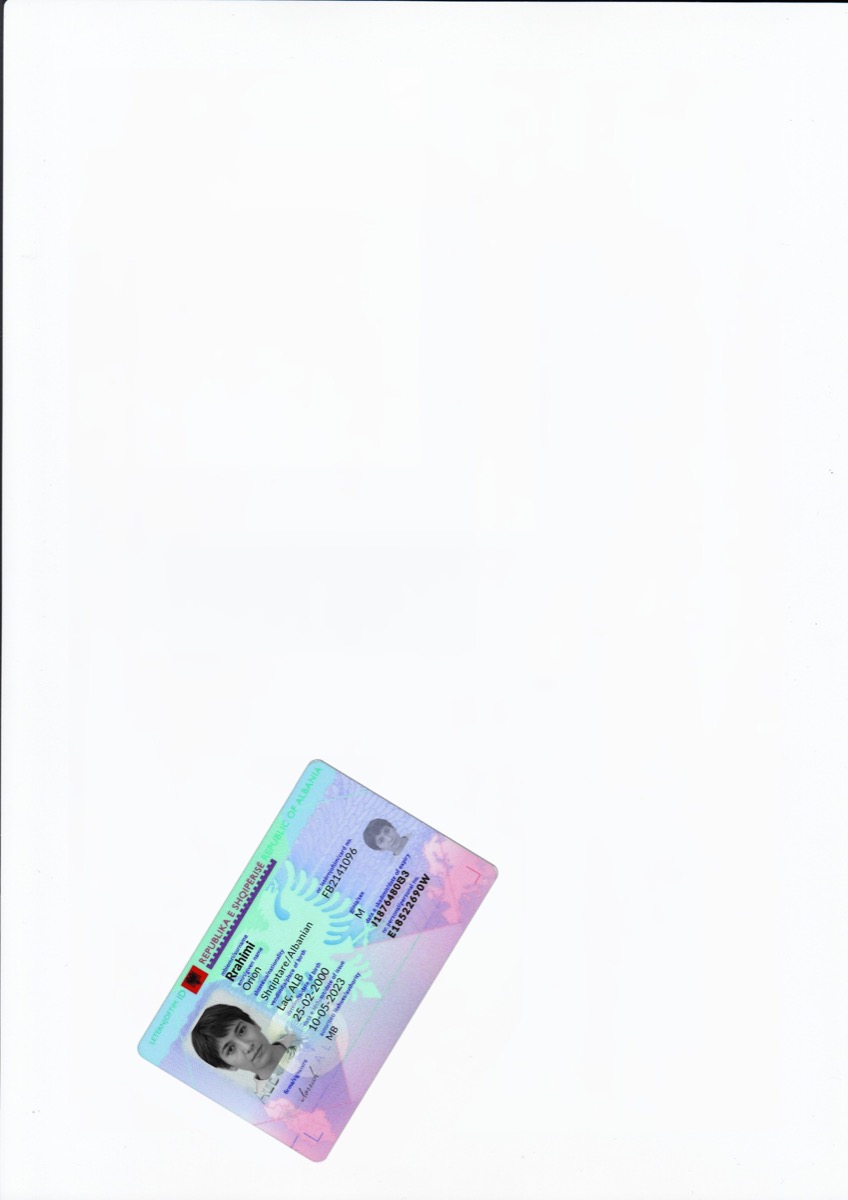}
    }
    \caption{Examples from multiple datasets containing Albanian ID card images.}
    \label{fig:alb_samples}
\end{figure*}

\begin{figure*}[h!]
    \centering
    \subfigure[\tiny Templated sample from MIDV]{
        \includegraphics[width=0.3\textwidth]{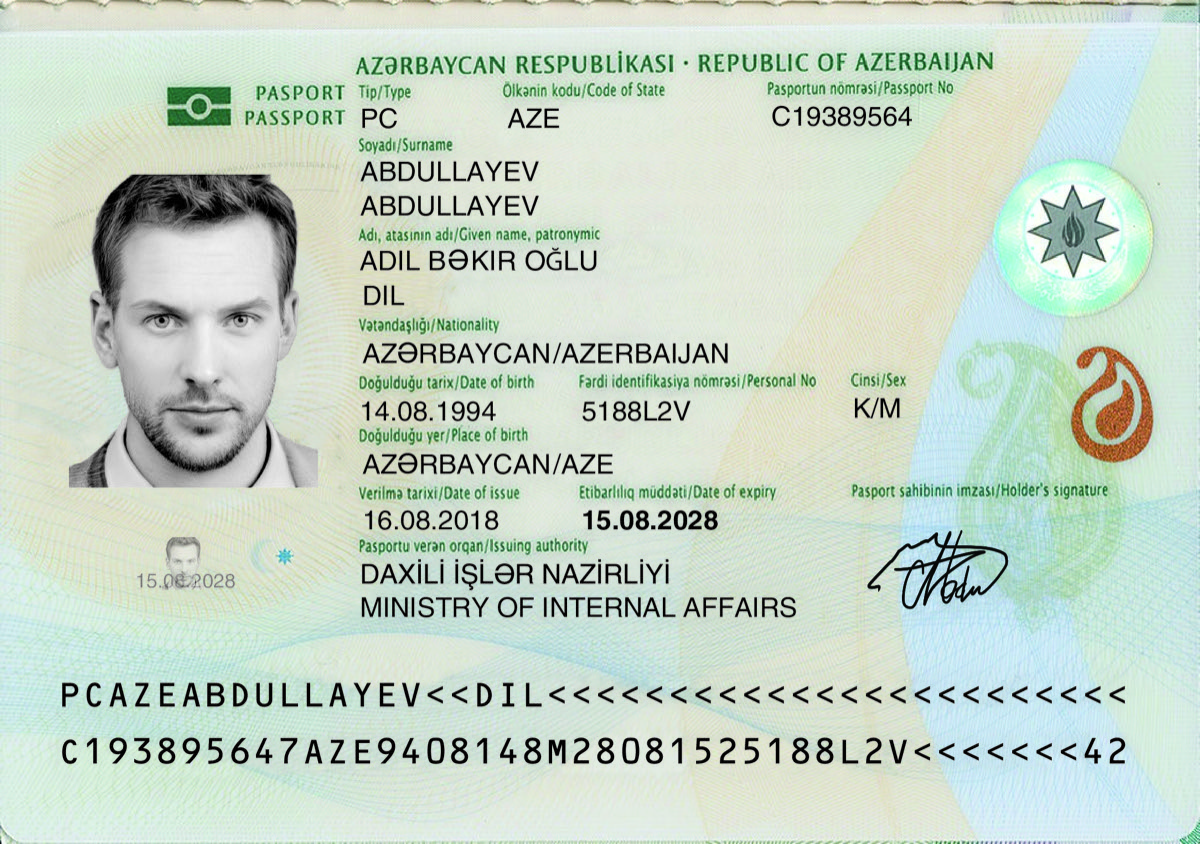}
    }
    \subfigure[\tiny Inpaint\&Rewrite fraud sample from SIDTD]{
        \includegraphics[width=0.3\textwidth]{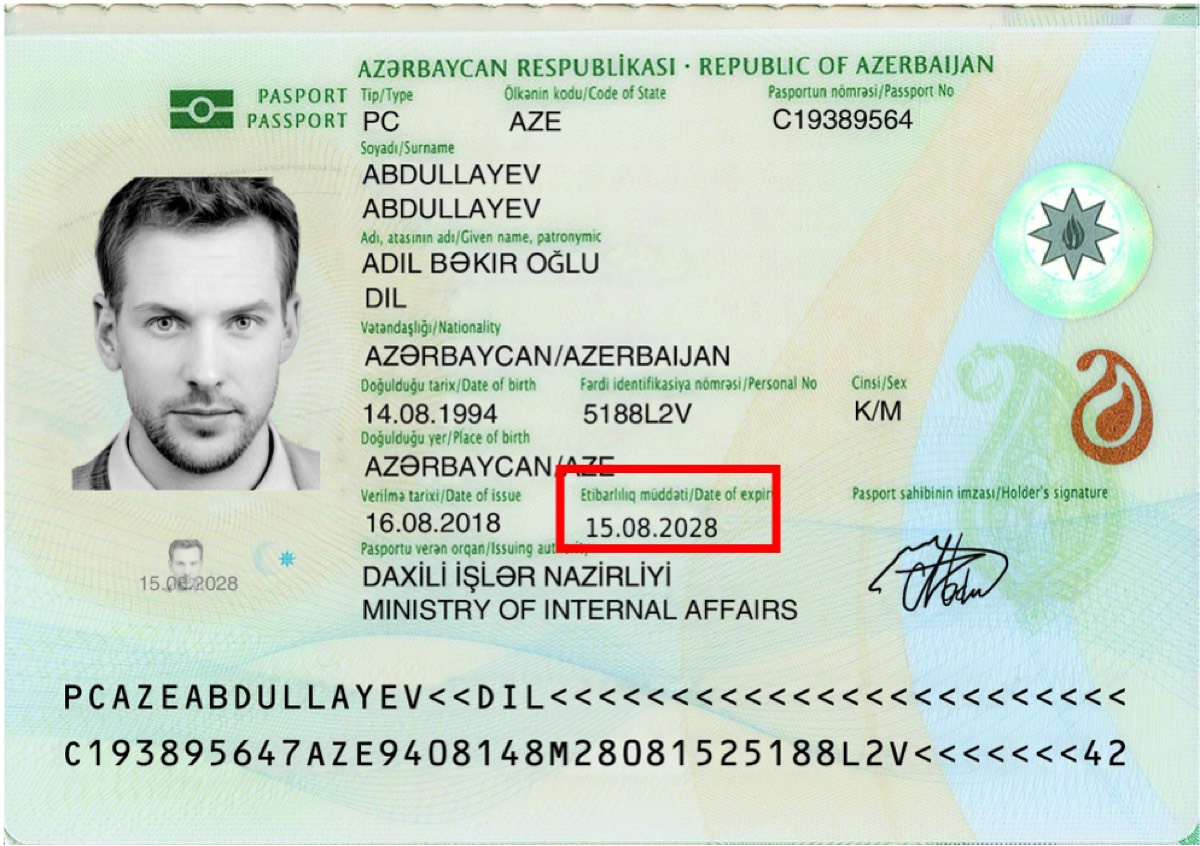}
    }
    \subfigure[\tiny Crop\&Replace fraud sample from SIDTD]{
        \includegraphics[width=0.3\textwidth]{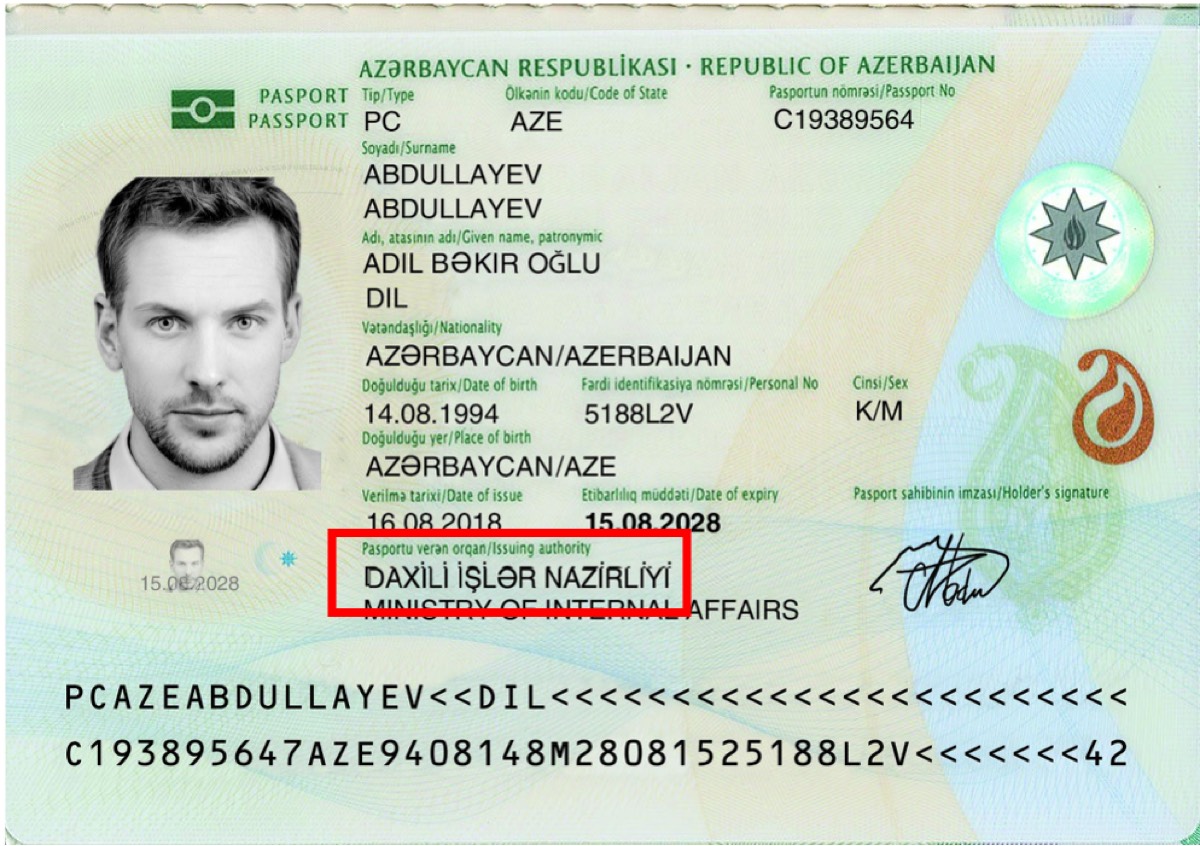}
    }
        \subfigure[\tiny Templated sample from \IDSpace]{
        \includegraphics[width=0.3\textwidth]{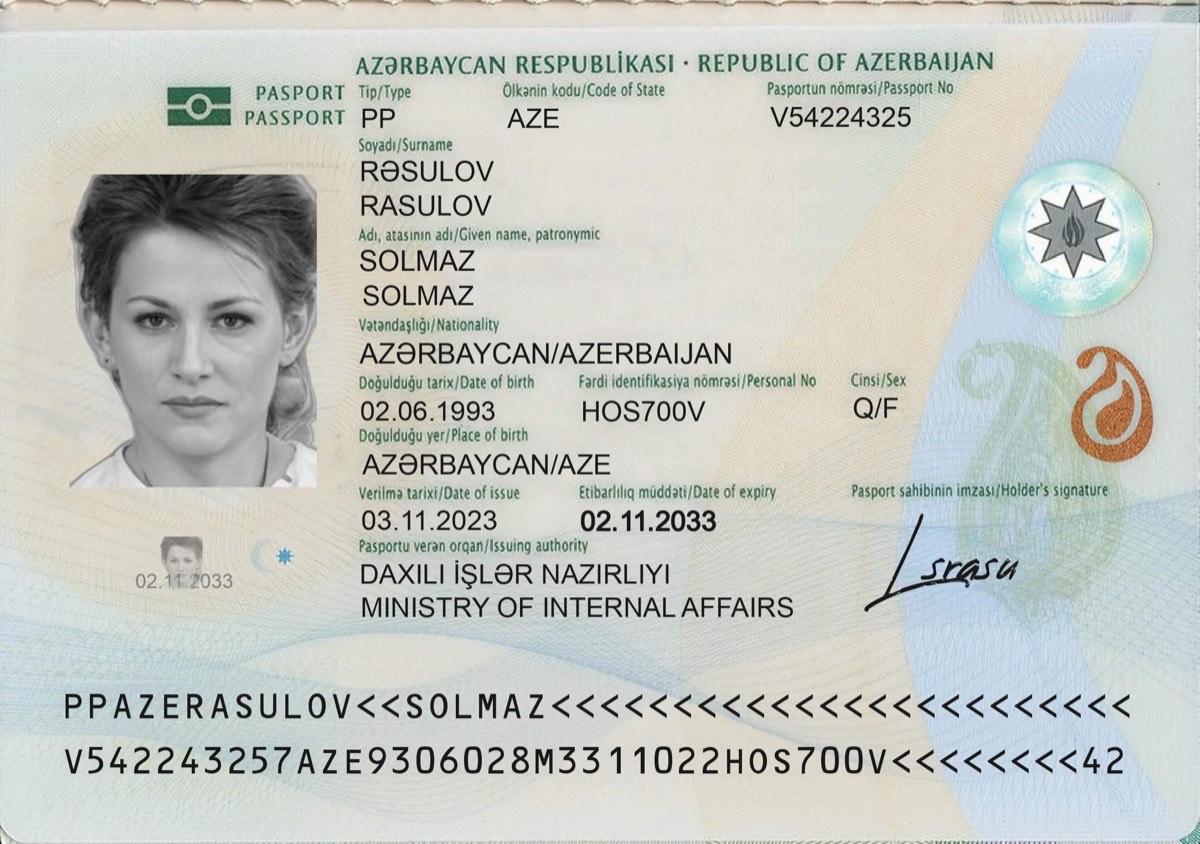}
    }
    \subfigure[\tiny Inpaint\&Rewrite fraud sample from \IDSpace]{
        \includegraphics[width=0.3\textwidth]{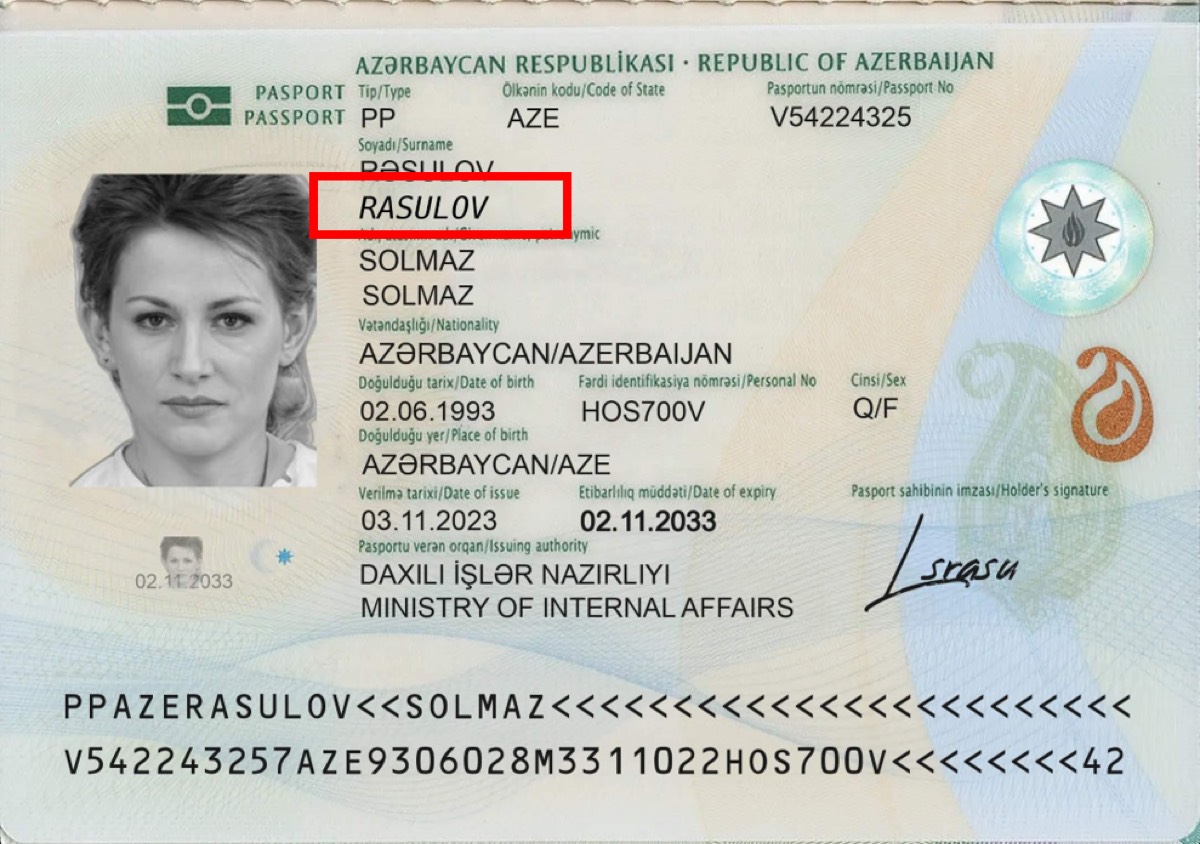}
    }
    \subfigure[\tiny Crop\&Replace fraud sample from \IDSpace]{
        \includegraphics[width=0.3\textwidth]{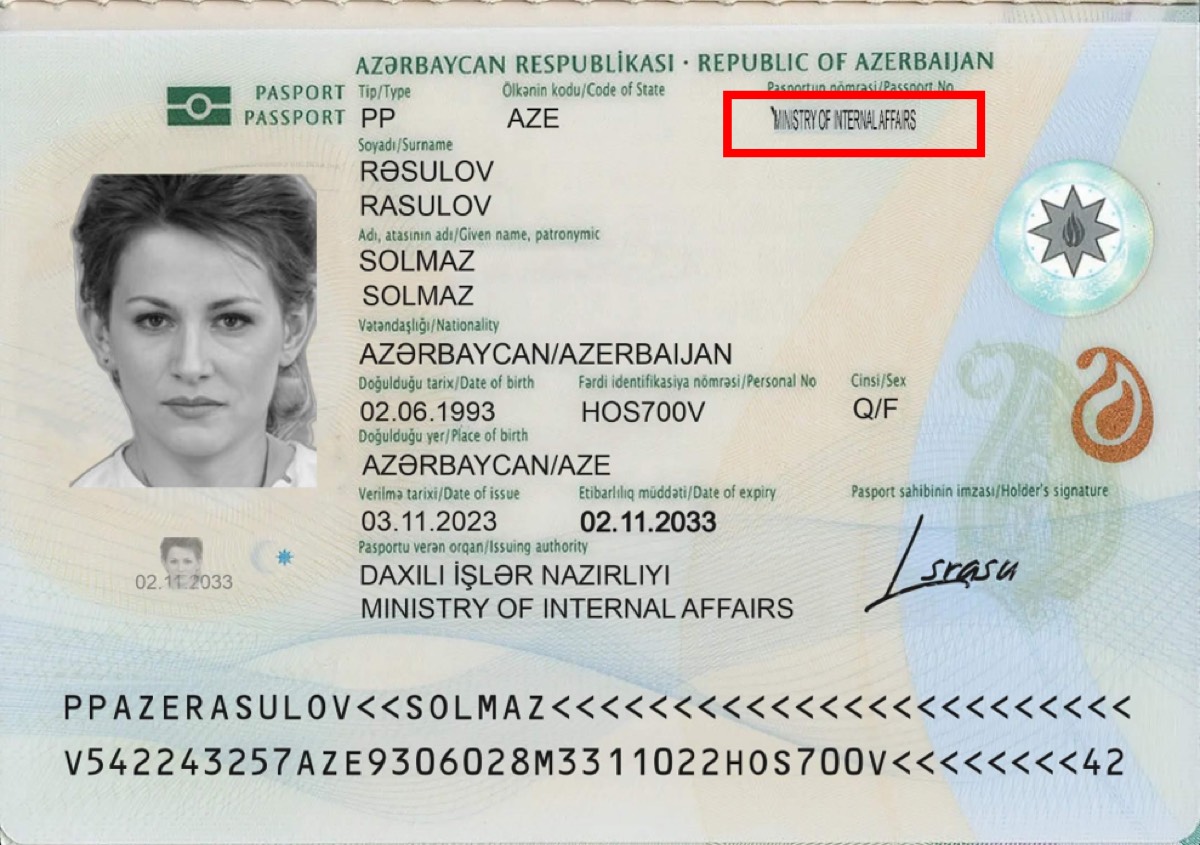}
    }

        \subfigure[\tiny Scanned sample from MIDV]{
        \includegraphics[width=0.22\textwidth]{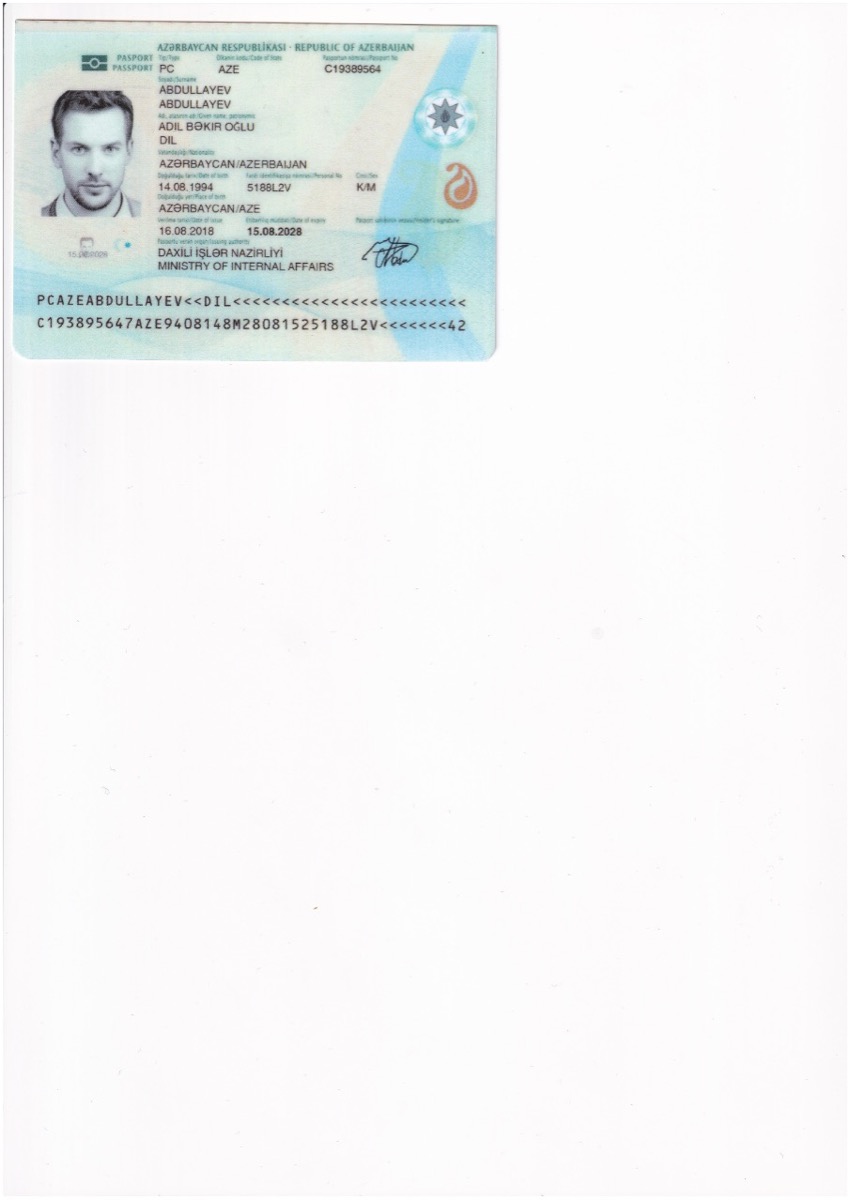}
    }
    \subfigure[\tiny Scanned sample from \IDSpace]{
        \includegraphics[width=0.22\textwidth]{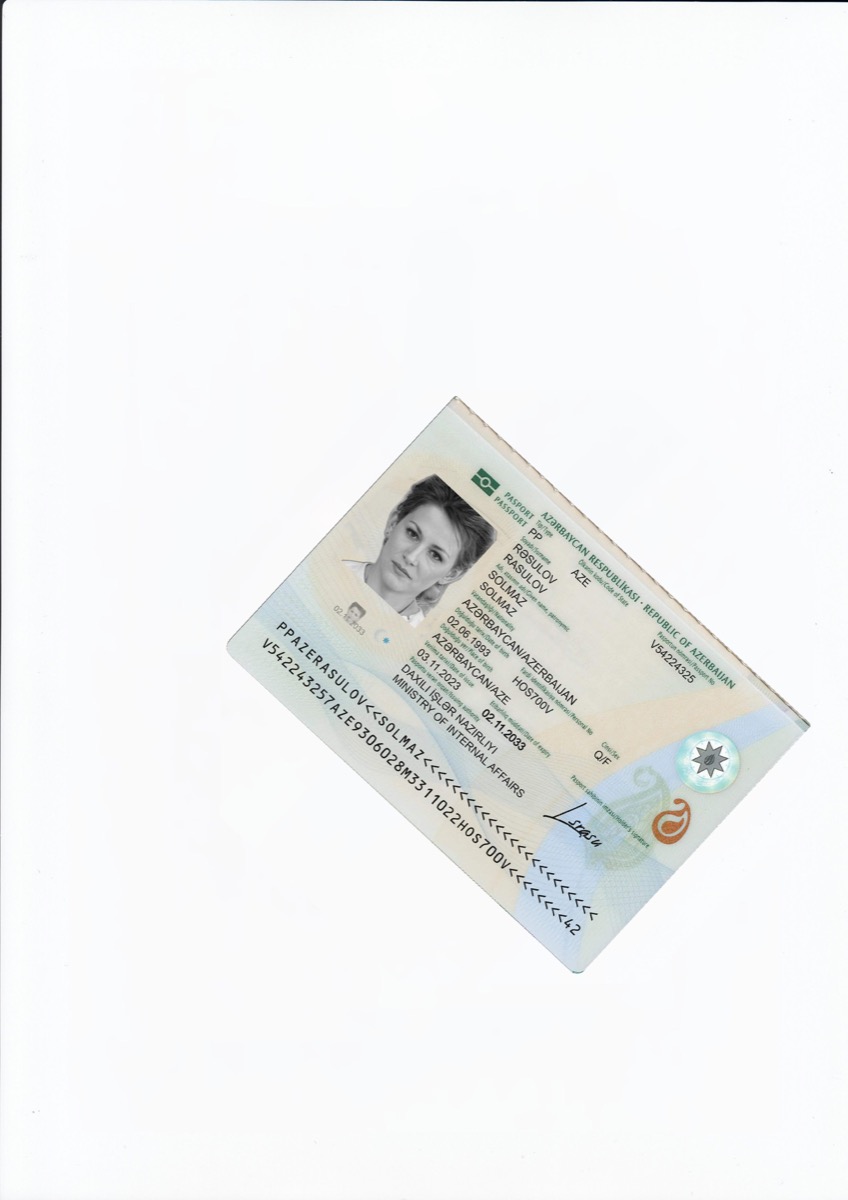}
    }
    \subfigure[\tiny Scanned sample with Inpaint\&Rewrite fraud from \IDSpace]{
        \includegraphics[width=0.22\textwidth]{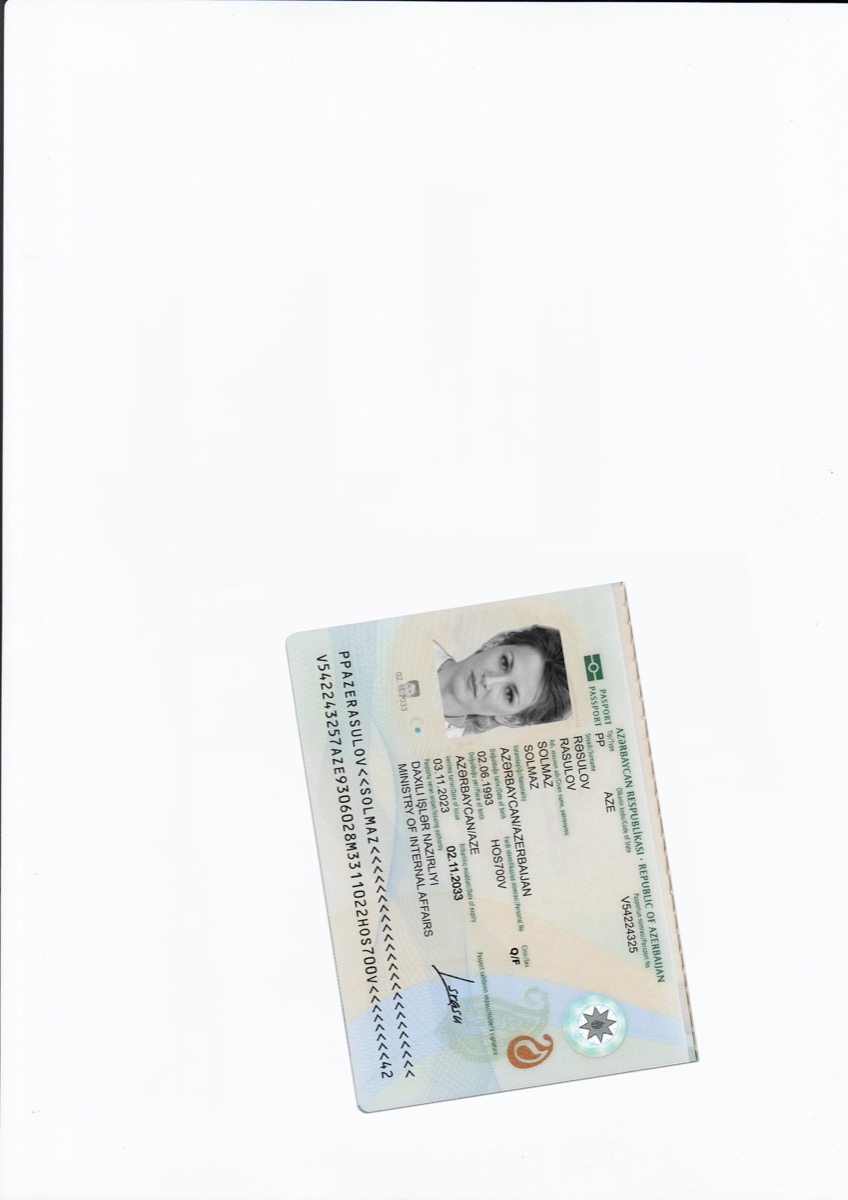}
    }
    \subfigure[\tiny Scanned sample with Crop\&Replace fraud from \IDSpace]{
        \includegraphics[width=0.22\textwidth]{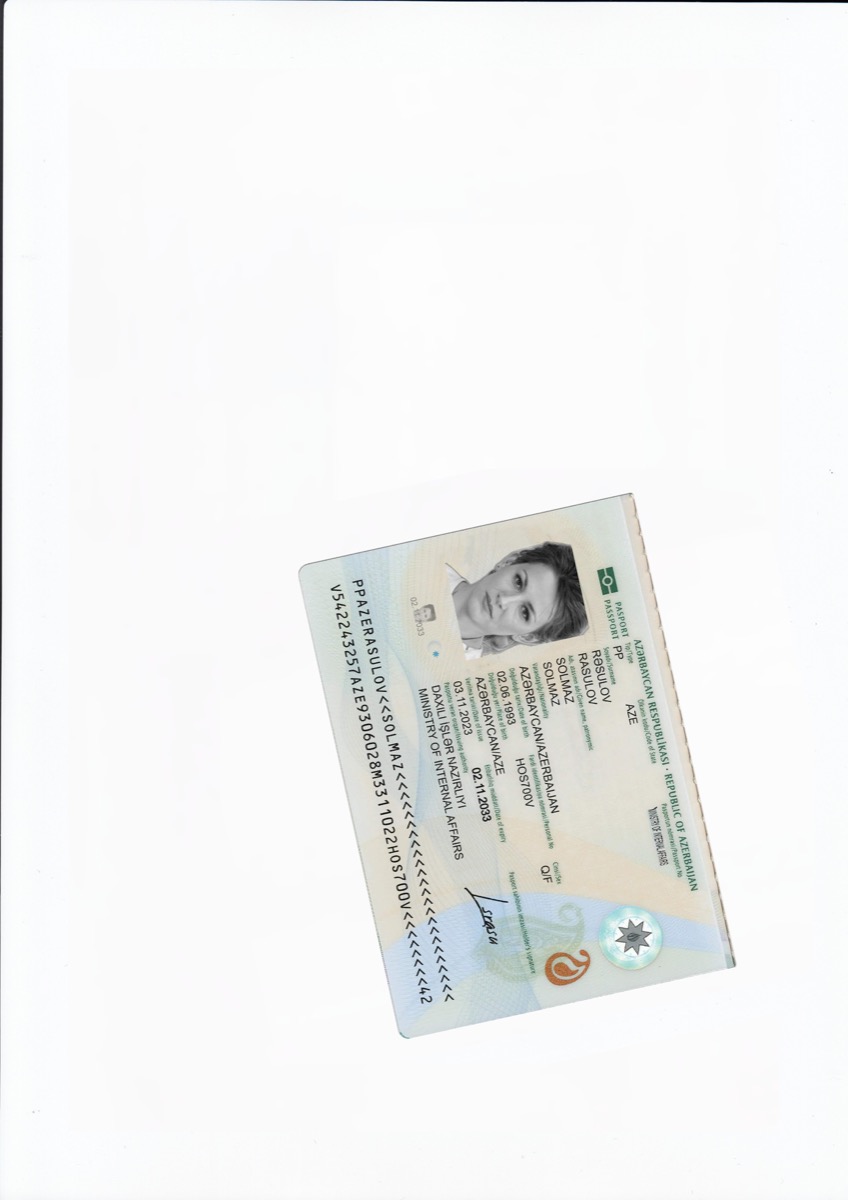}
    }
    \caption{Examples from multiple datasets containing Azerbaijani passport images.}
    \label{fig:aze_samples}
\end{figure*}

\begin{figure*}[h!]
    \centering
    \subfigure[\tiny Templated sample from MIDV]{
        \includegraphics[width=0.3\textwidth]{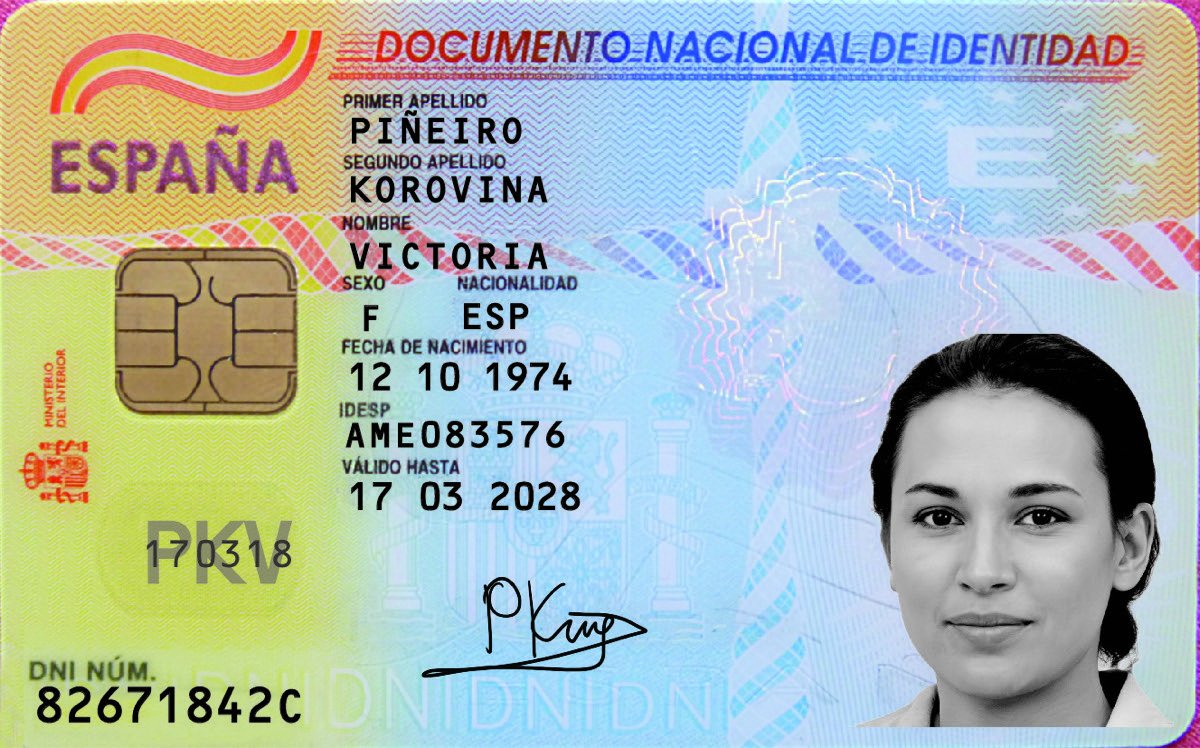}
    }
    \subfigure[\tiny Inpaint\&Rewrite fraud sample from SIDTD]{
        \includegraphics[width=0.3\textwidth]{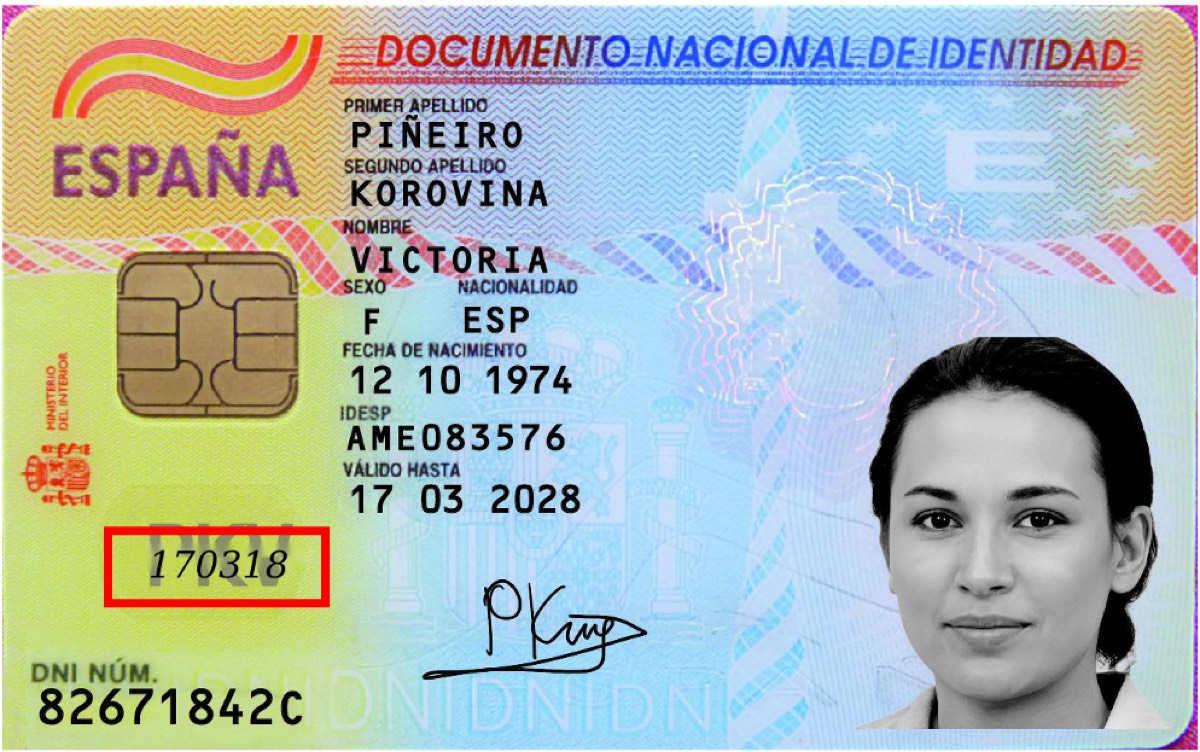}
    }
    \subfigure[\tiny Crop\&Replace fraud sample from SIDTD]{
        \includegraphics[width=0.3\textwidth]{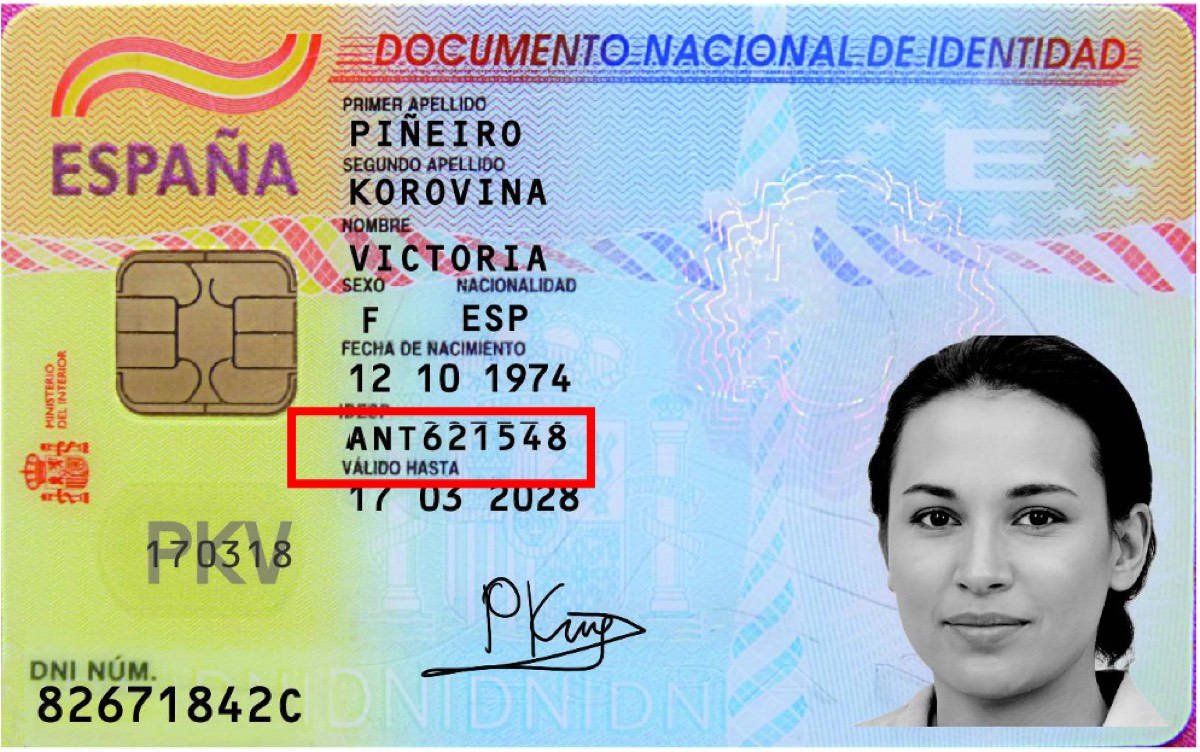}
    }
        \subfigure[\tiny Templated sample from \IDSpace]{
        \includegraphics[width=0.3\textwidth]{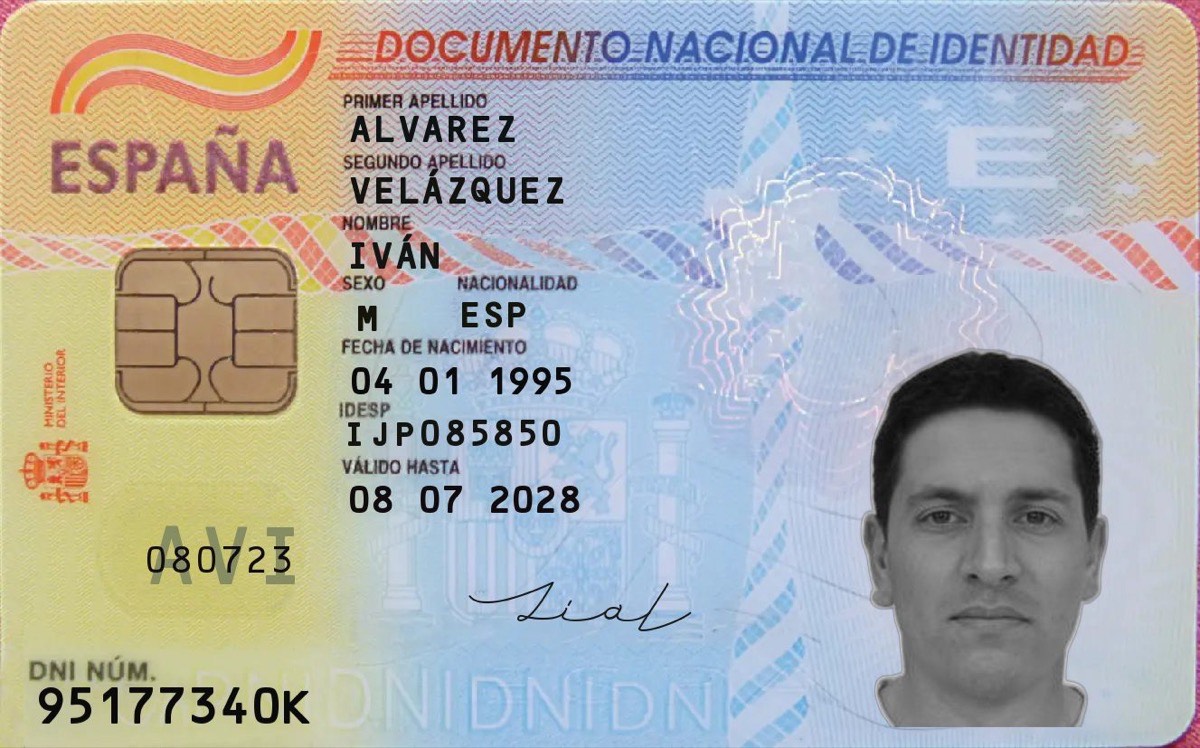}
    }
    \subfigure[\tiny Inpaint\&Rewrite fraud sample from \IDSpace]{
        \includegraphics[width=0.3\textwidth]{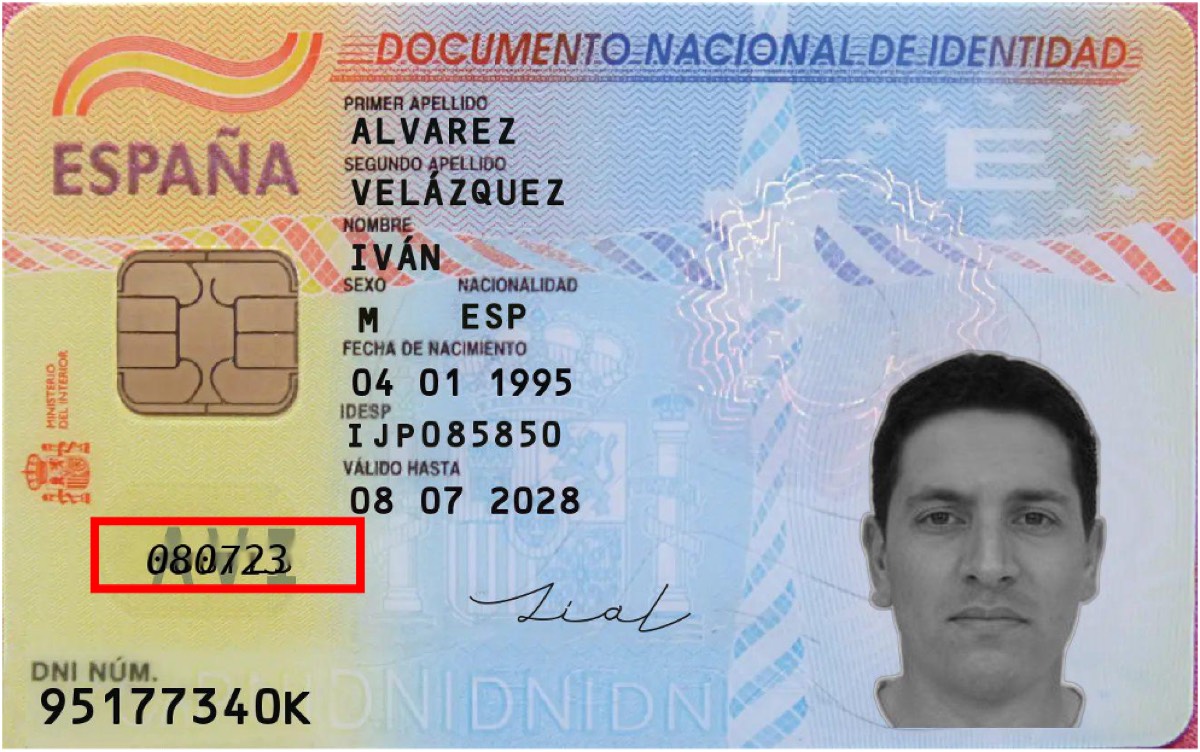}
    }
    \subfigure[\tiny Crop\&Replace fraud sample from \IDSpace]{
        \includegraphics[width=0.3\textwidth]{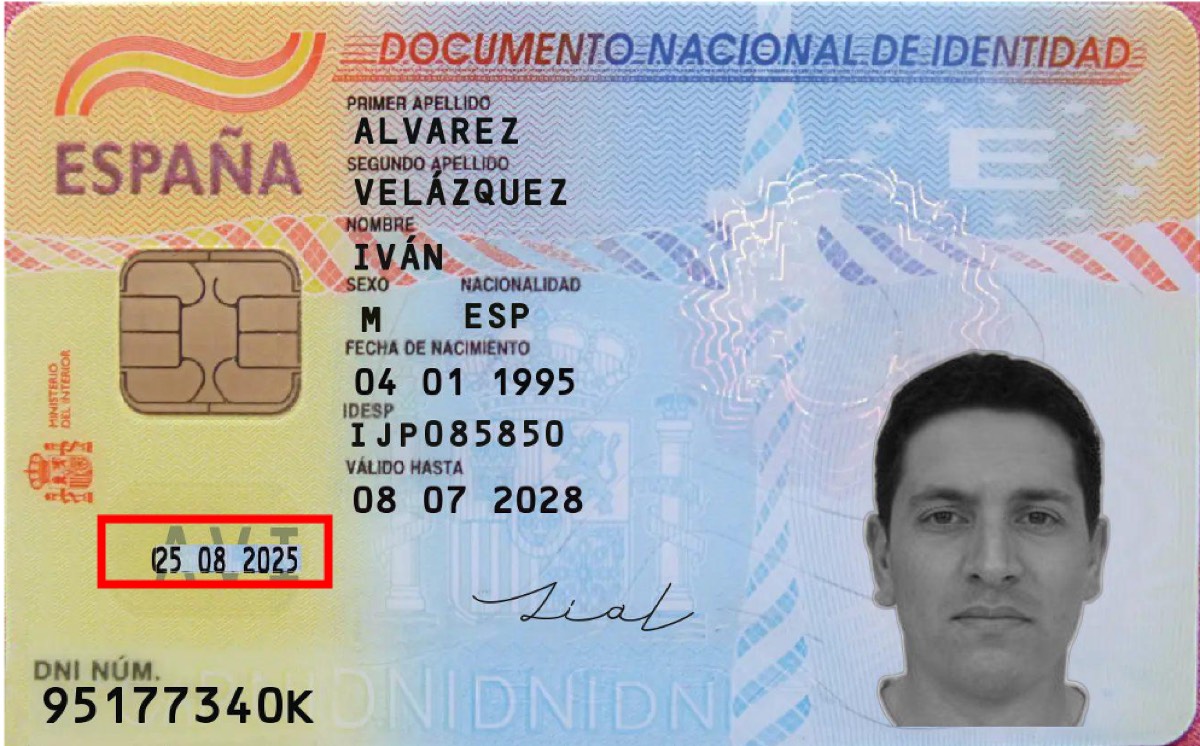}
    }

        \subfigure[\tiny Scanned sample from MIDV]{
        \includegraphics[width=0.22\textwidth]{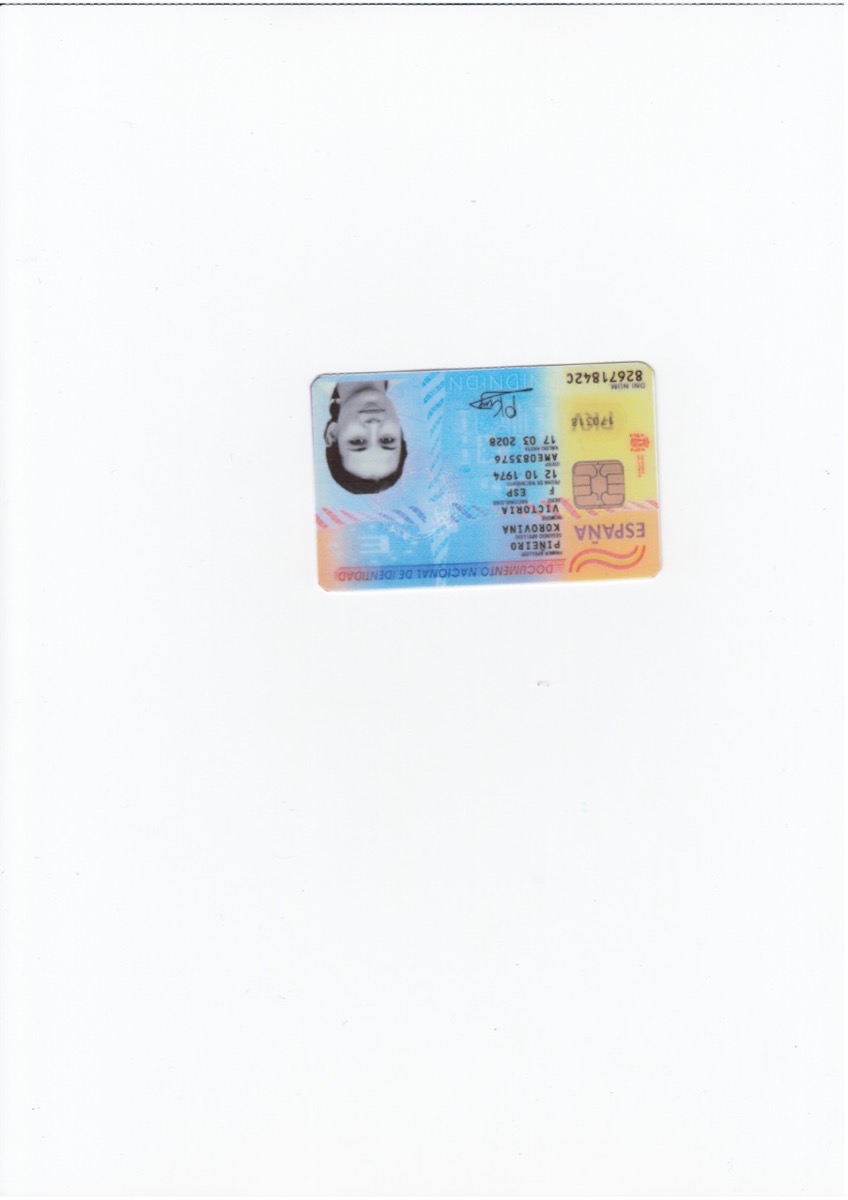}
    }
    \subfigure[\tiny Scanned sample from \IDSpace]{
        \includegraphics[width=0.22\textwidth]{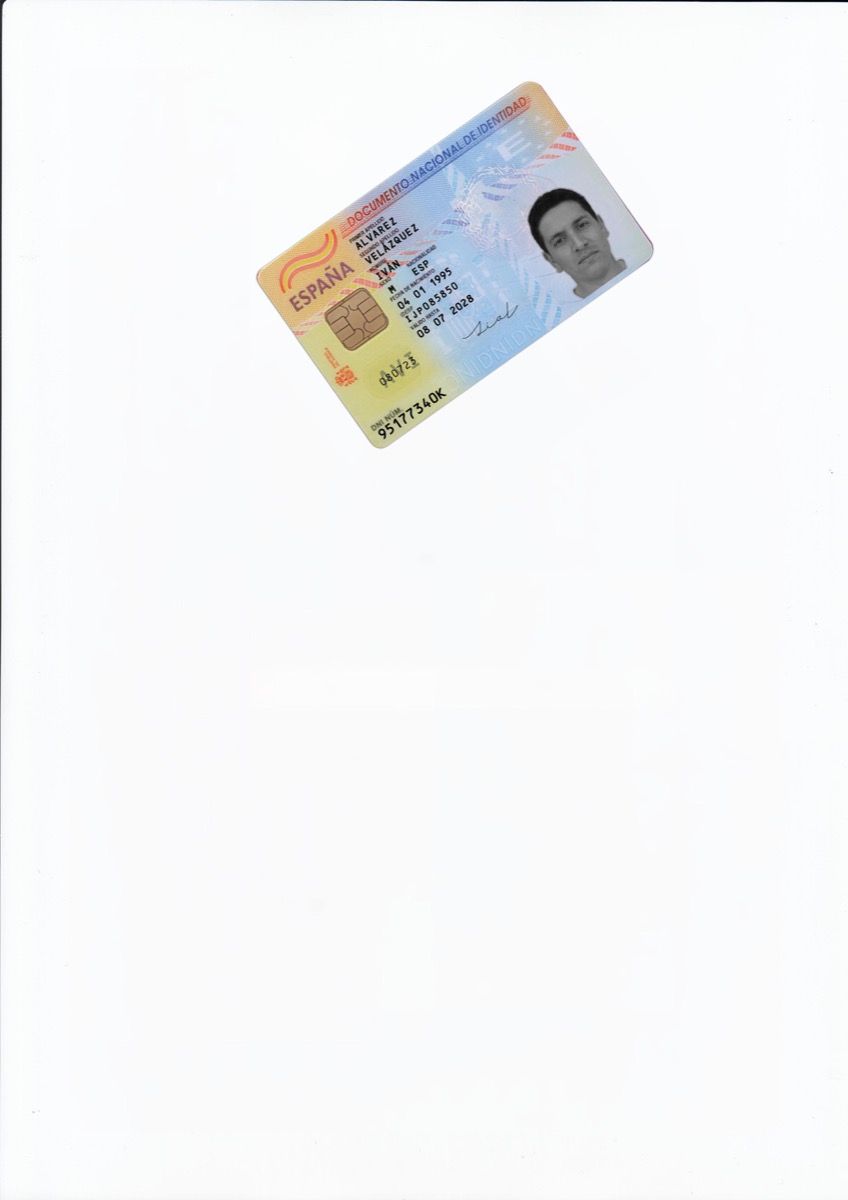}
    }
    \subfigure[\tiny Scanned sample with Inpaint\&Rewrite fraud from \IDSpace]{
        \includegraphics[width=0.22\textwidth]{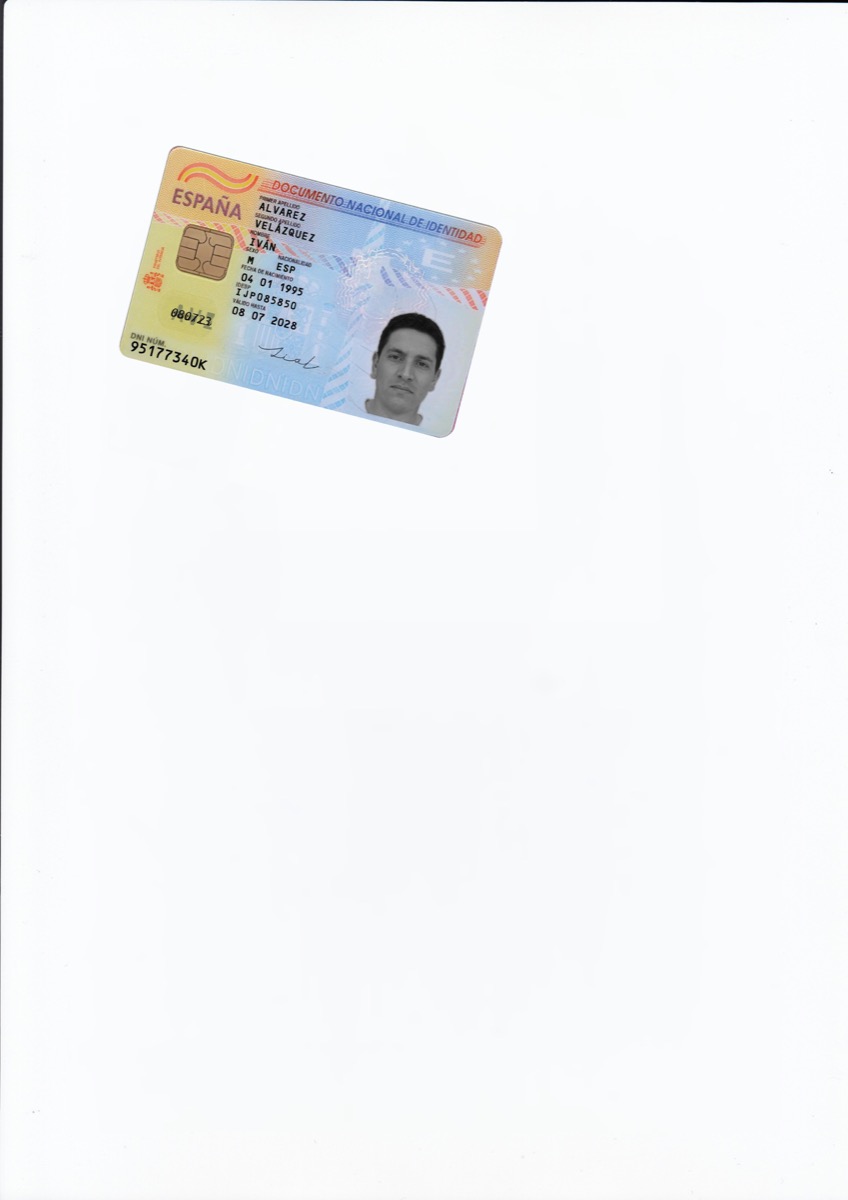}
    }
    \subfigure[\tiny Scanned sample with Crop\&Replace fraud from \IDSpace]{
        \includegraphics[width=0.22\textwidth]{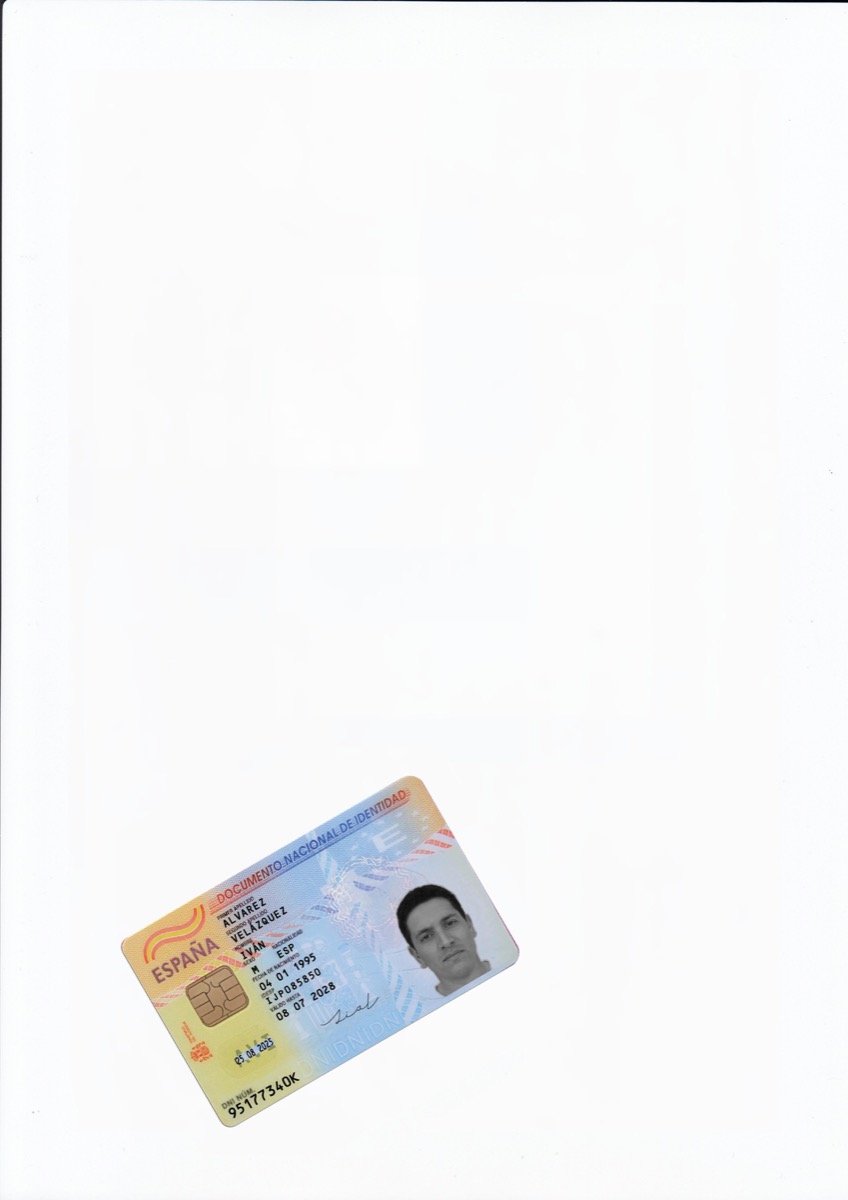}
    }
    \caption{Examples from multiple datasets containing Spanish ID card images.}
    \label{fig:esp_samples}
\end{figure*}

\begin{figure*}[h!]
    \centering
    \subfigure[\tiny Templated sample from MIDV]{
        \includegraphics[width=0.3\textwidth]{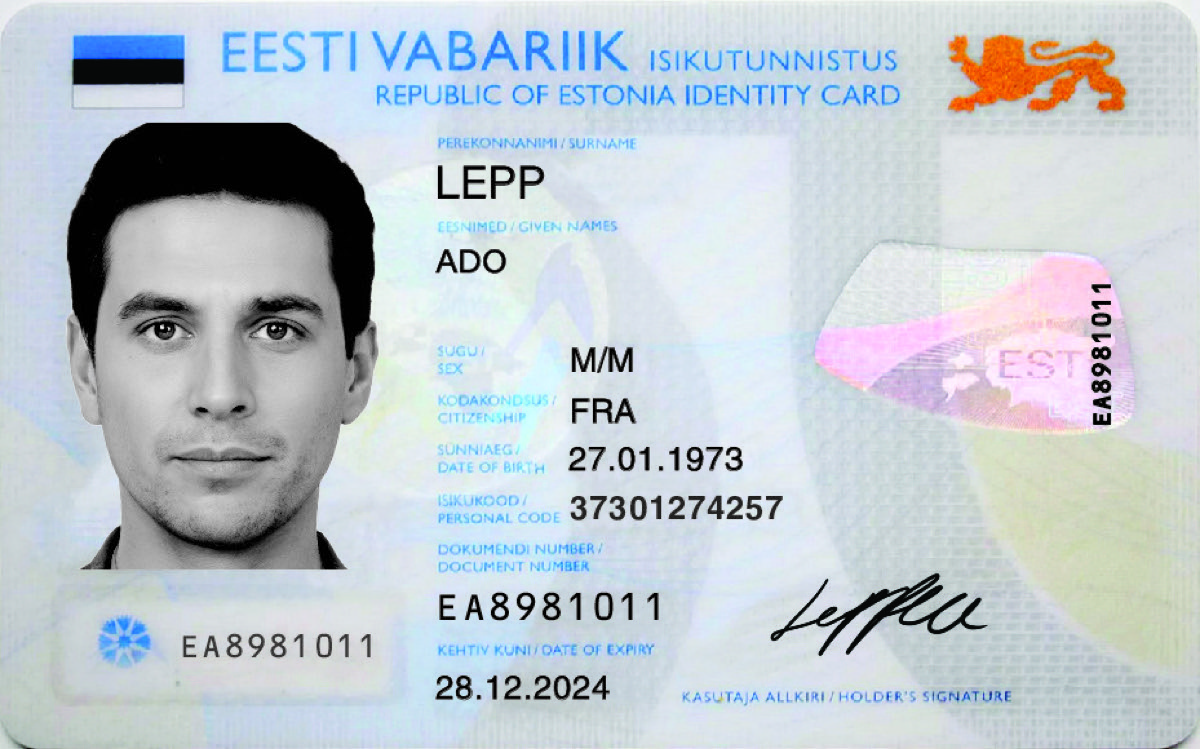}
    }
    \subfigure[\tiny Inpaint\&Rewrite fraud sample from SIDTD]{
        \includegraphics[width=0.3\textwidth]{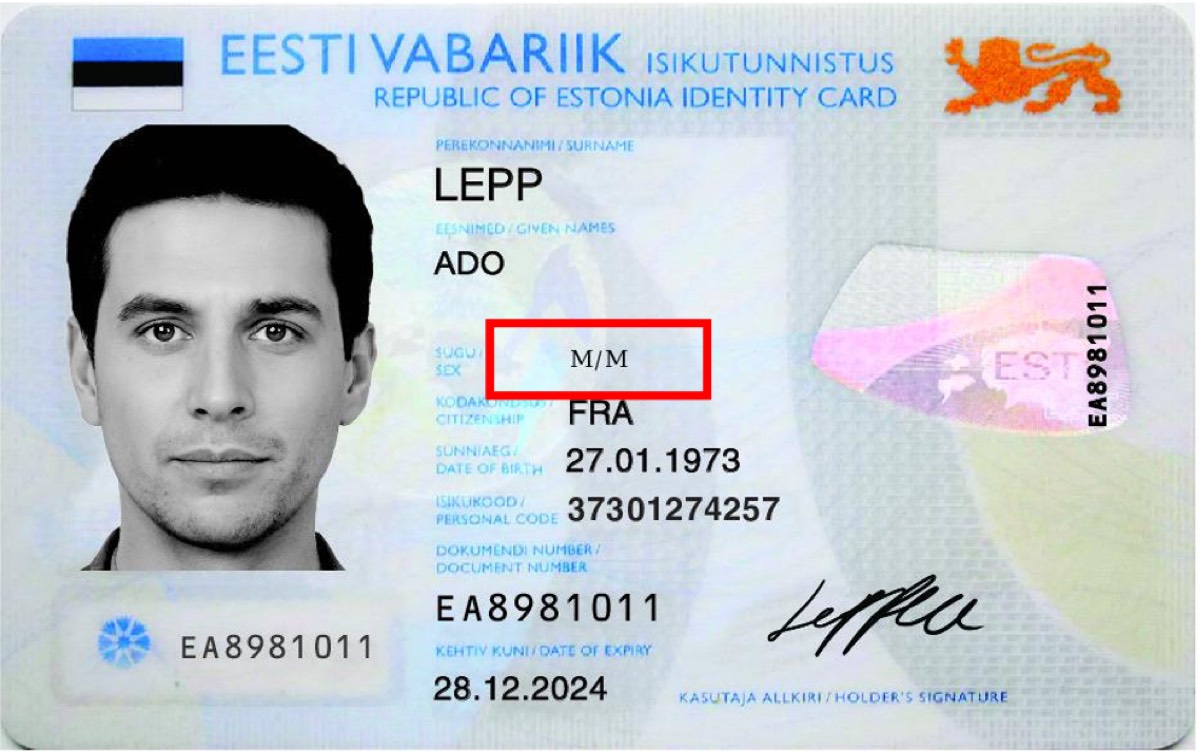}
    }
    \subfigure[\tiny Crop\&Replace fraud sample from SIDTD]{
        \includegraphics[width=0.3\textwidth]{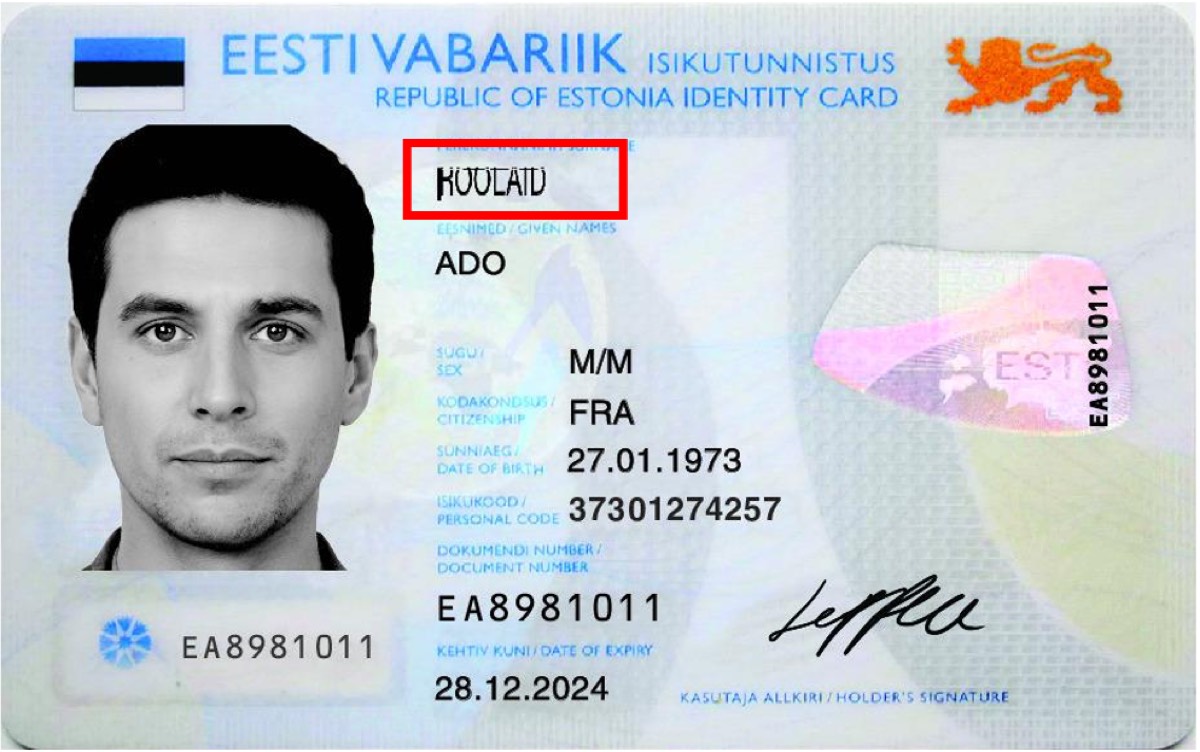}
    }
        \subfigure[\tiny Templated sample from \IDSpace]{
        \includegraphics[width=0.3\textwidth]{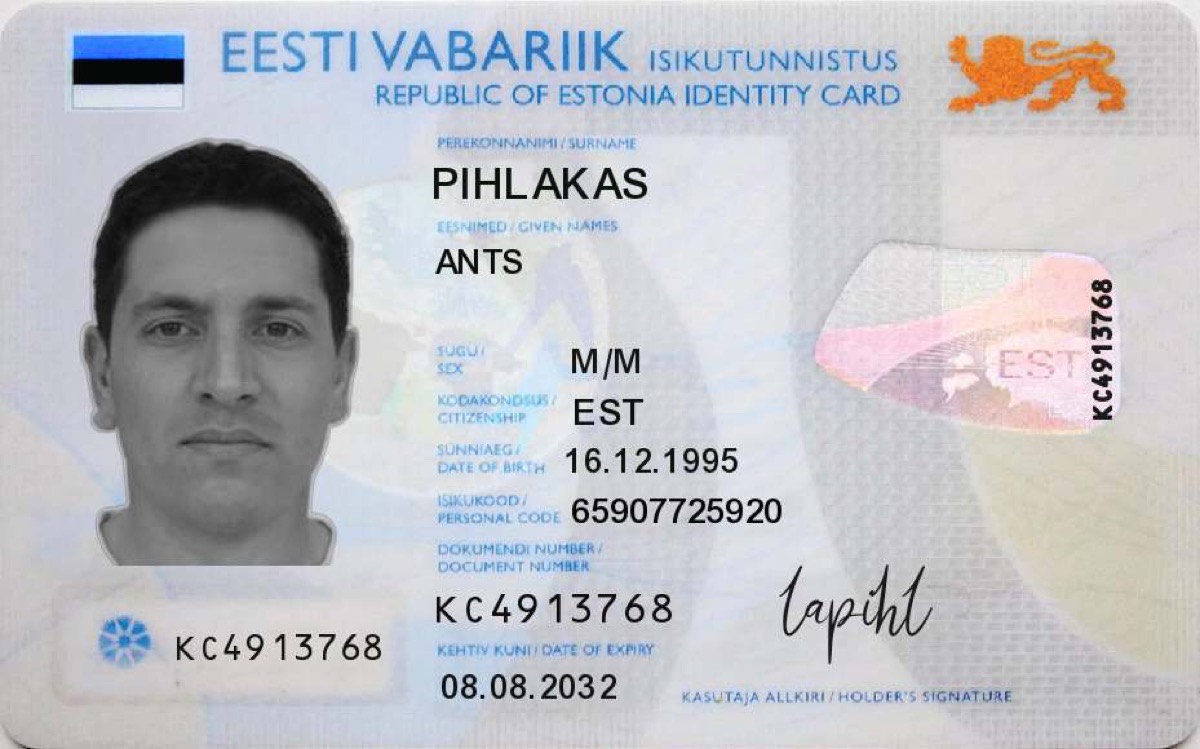}
    }
    \subfigure[\tiny Inpaint\&Rewrite fraud sample from \IDSpace]{
        \includegraphics[width=0.3\textwidth]{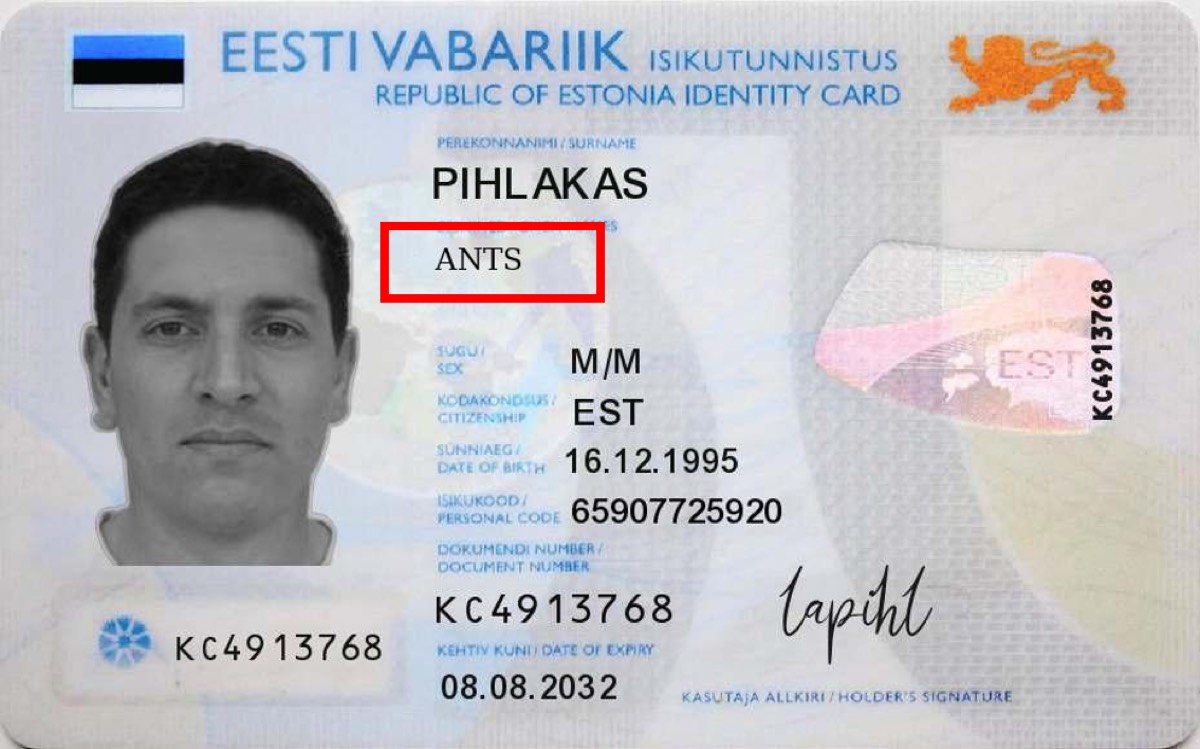}
    }
    \subfigure[\tiny Crop\&Replace fraud sample from \IDSpace]{
        \includegraphics[width=0.3\textwidth]{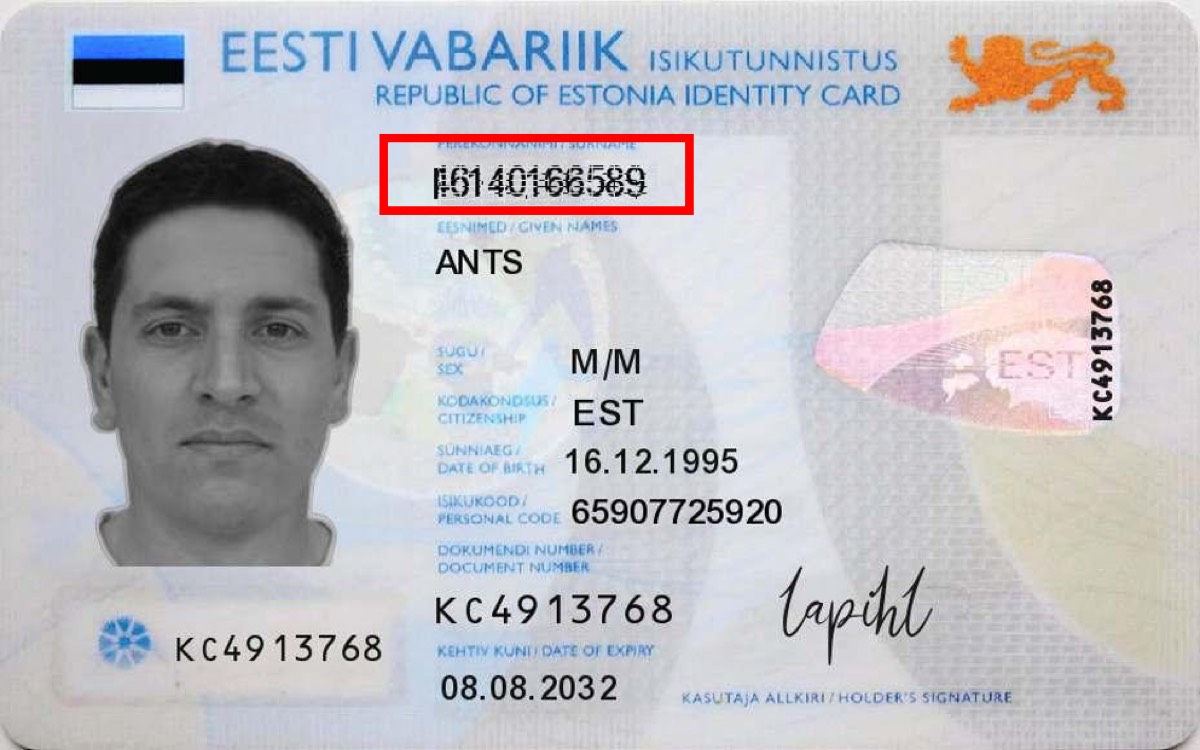}
    }

        \subfigure[\tiny Scanned sample from MIDV]{
        \includegraphics[width=0.22\textwidth]{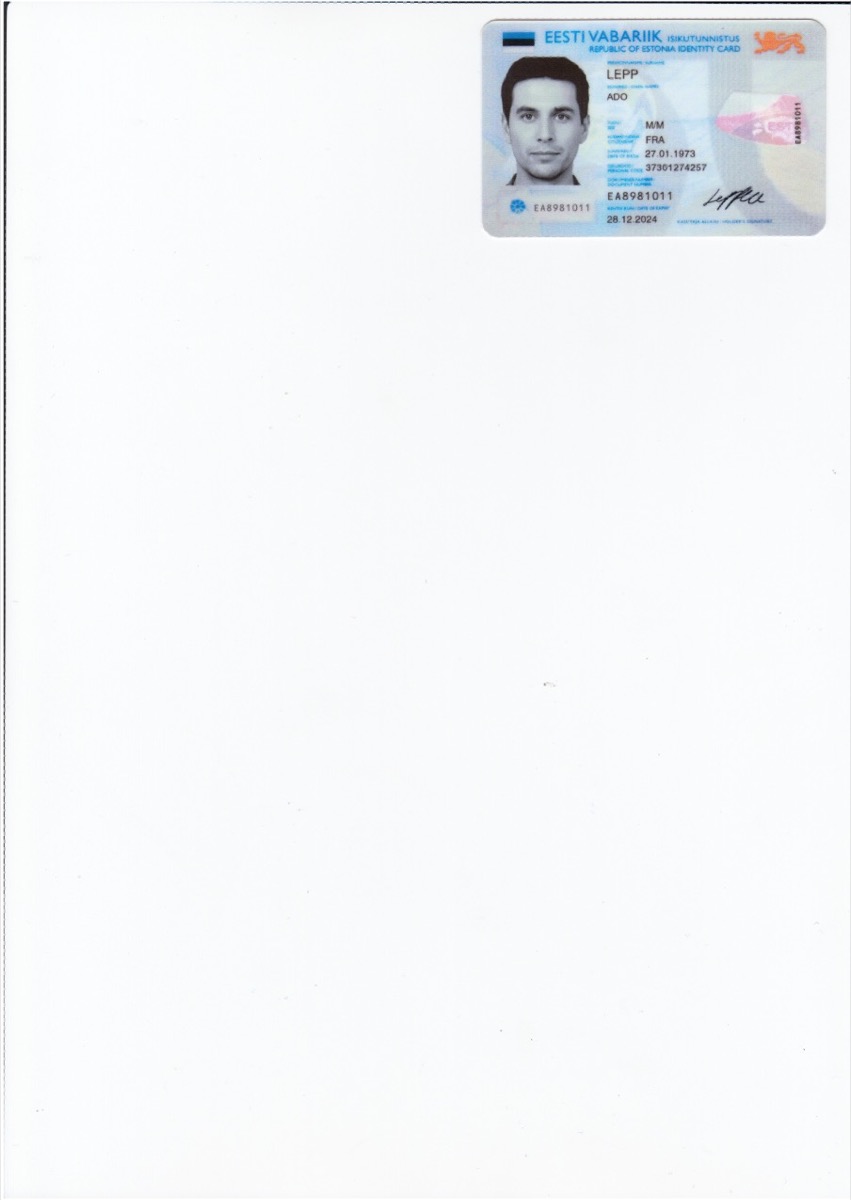}
    }
    \subfigure[\tiny Scanned sample from \IDSpace]{
        \includegraphics[width=0.22\textwidth]{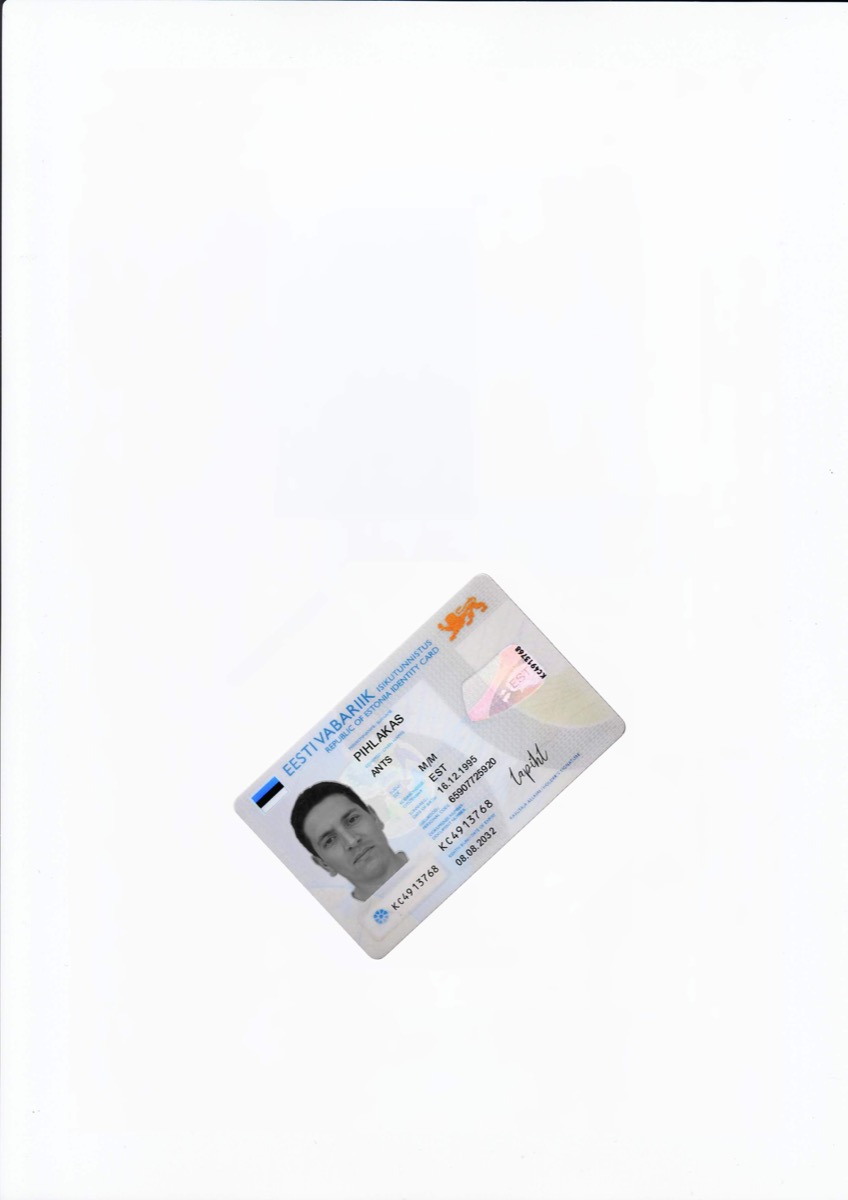}
    }
    \subfigure[\tiny Scanned sample with Inpaint\&Rewrite fraud from \IDSpace]{
        \includegraphics[width=0.22\textwidth]{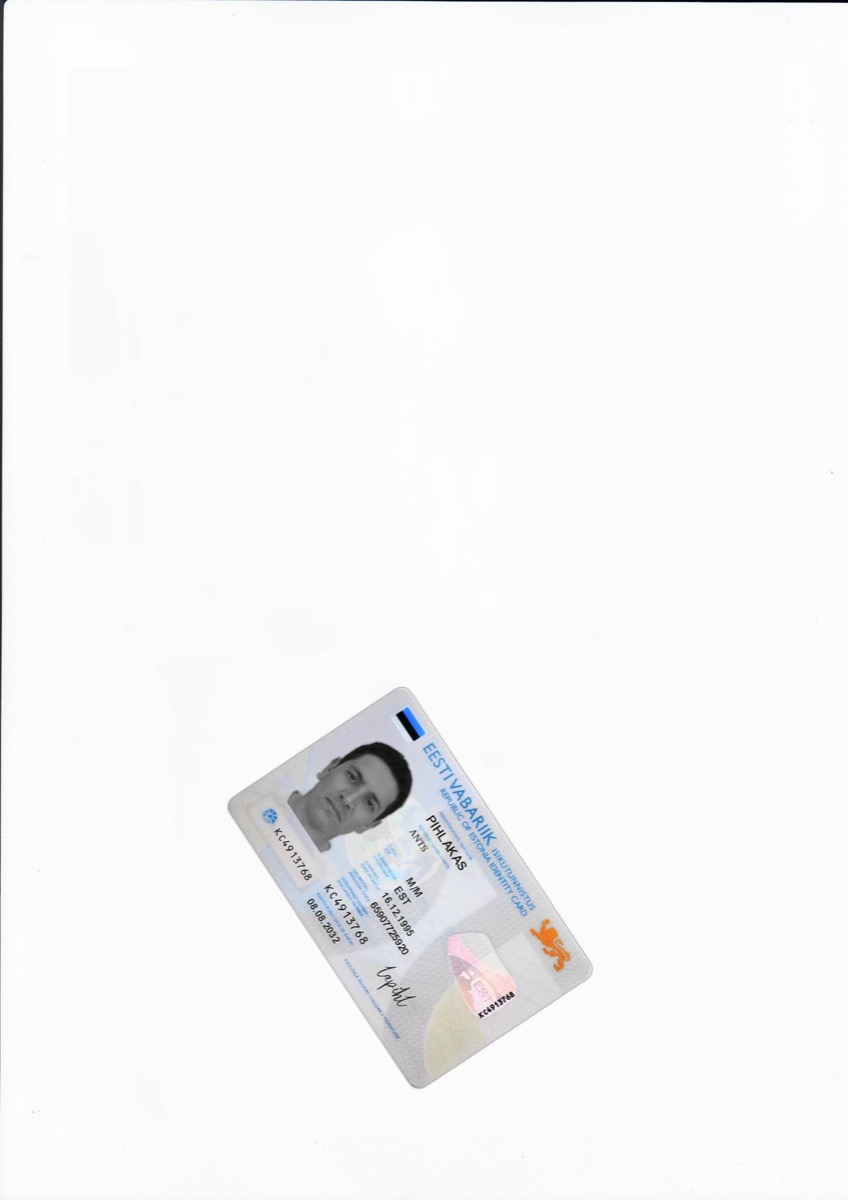}
    }
    \subfigure[\tiny Scanned sample with Crop\&Replace fraud from \IDSpace]{
        \includegraphics[width=0.22\textwidth]{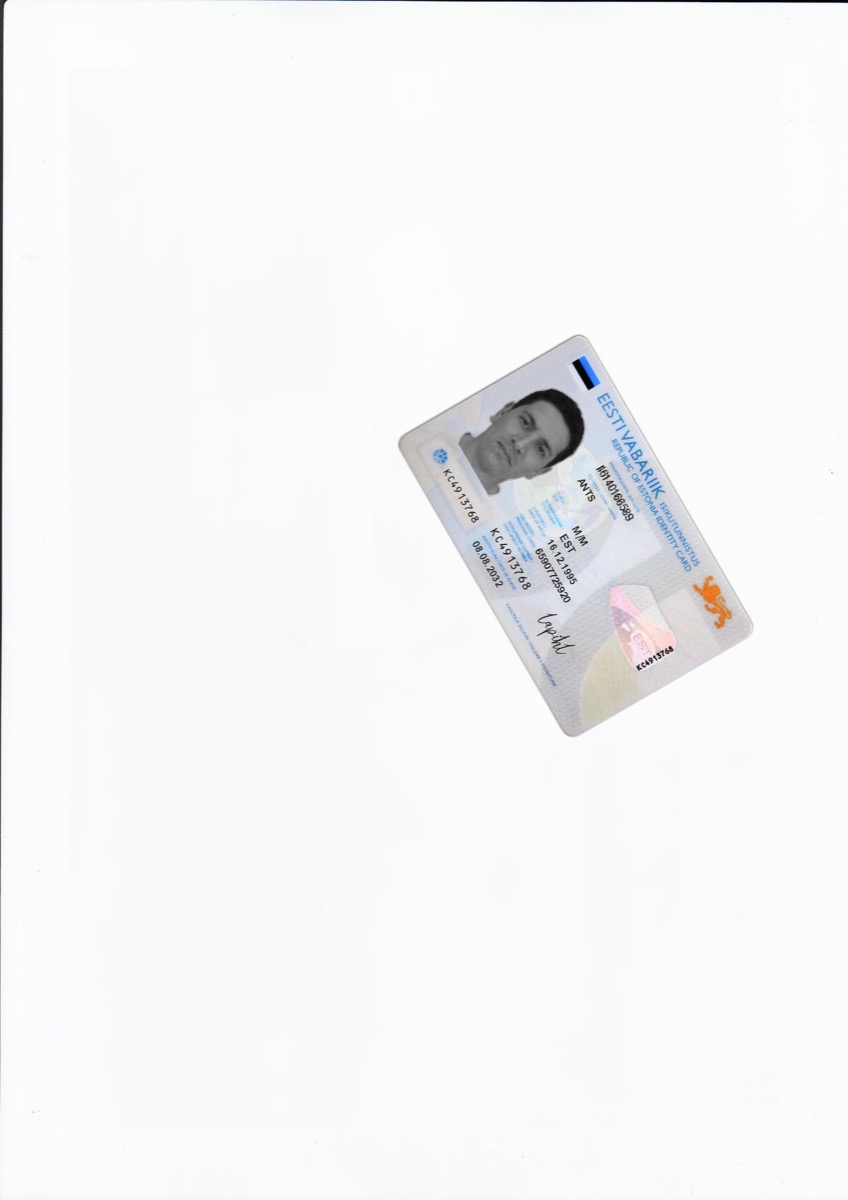}
    }
    \caption{Examples from multiple datasets containing Estonian ID card images.}
    \label{fig:est_samples}
\end{figure*}

\begin{figure*}[h!]
    \centering
    \subfigure[\tiny Templated sample from MIDV]{
        \includegraphics[width=0.3\textwidth]{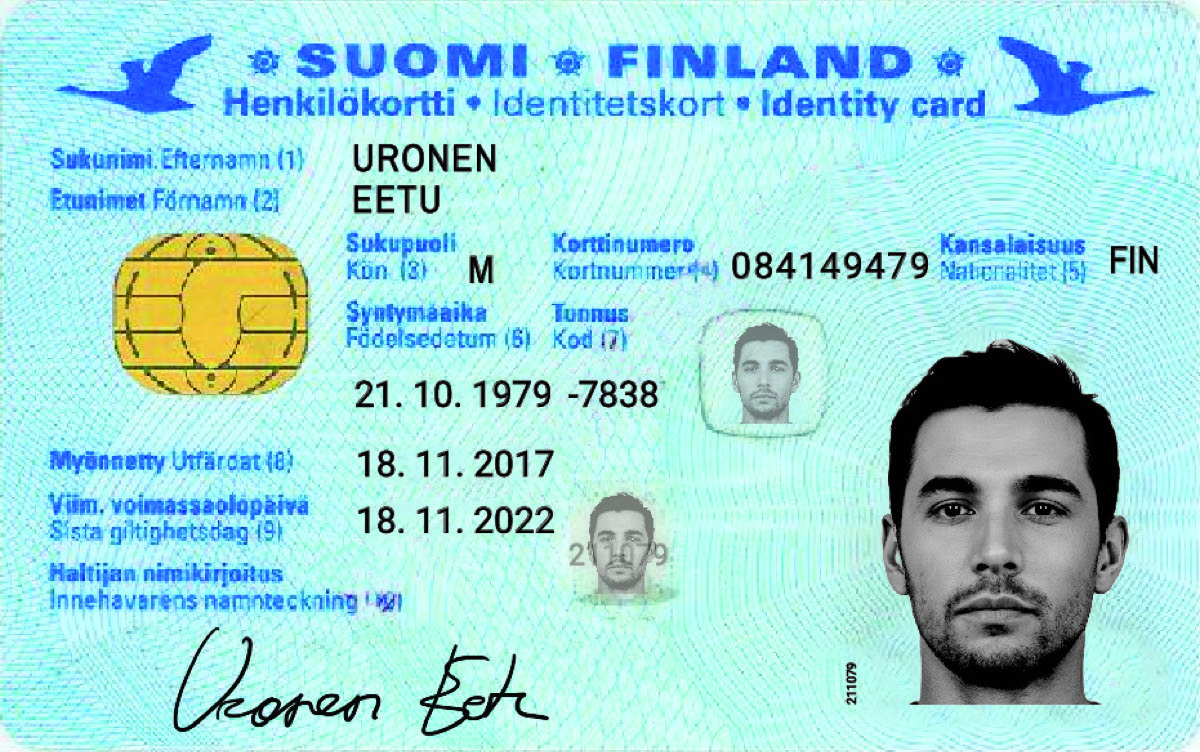}
    }
    \subfigure[\tiny Inpaint\&Rewrite fraud sample from SIDTD]{
        \includegraphics[width=0.3\textwidth]{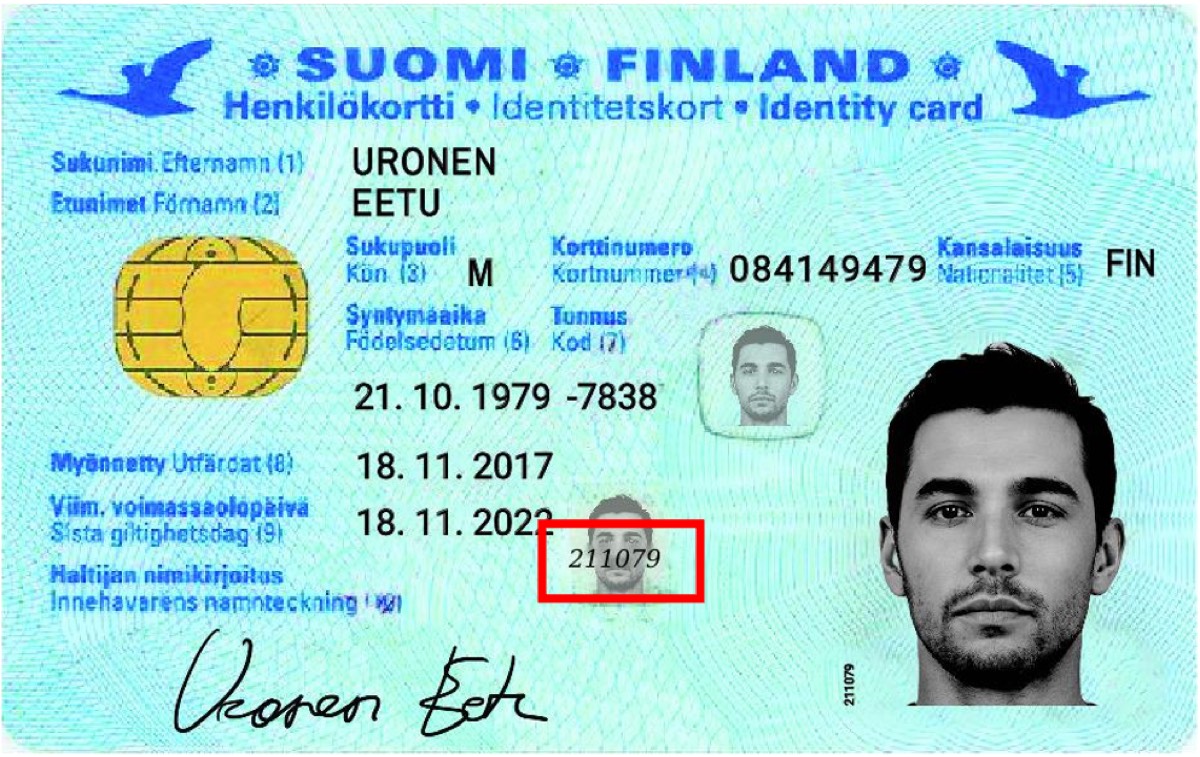}
    }
    \subfigure[\tiny Crop\&Replace fraud sample from SIDTD]{
        \includegraphics[width=0.3\textwidth]{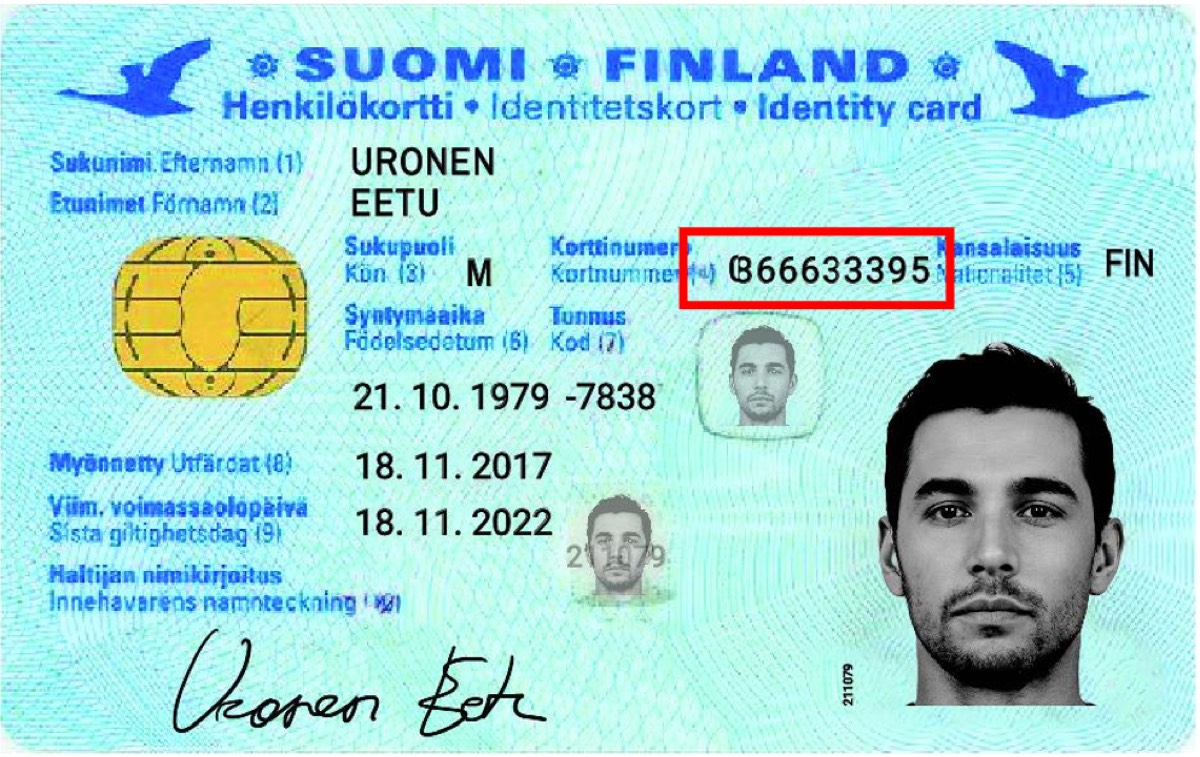}
    }

        \subfigure[\tiny Templated sample from \IDSpace]{
        \includegraphics[width=0.3\textwidth]{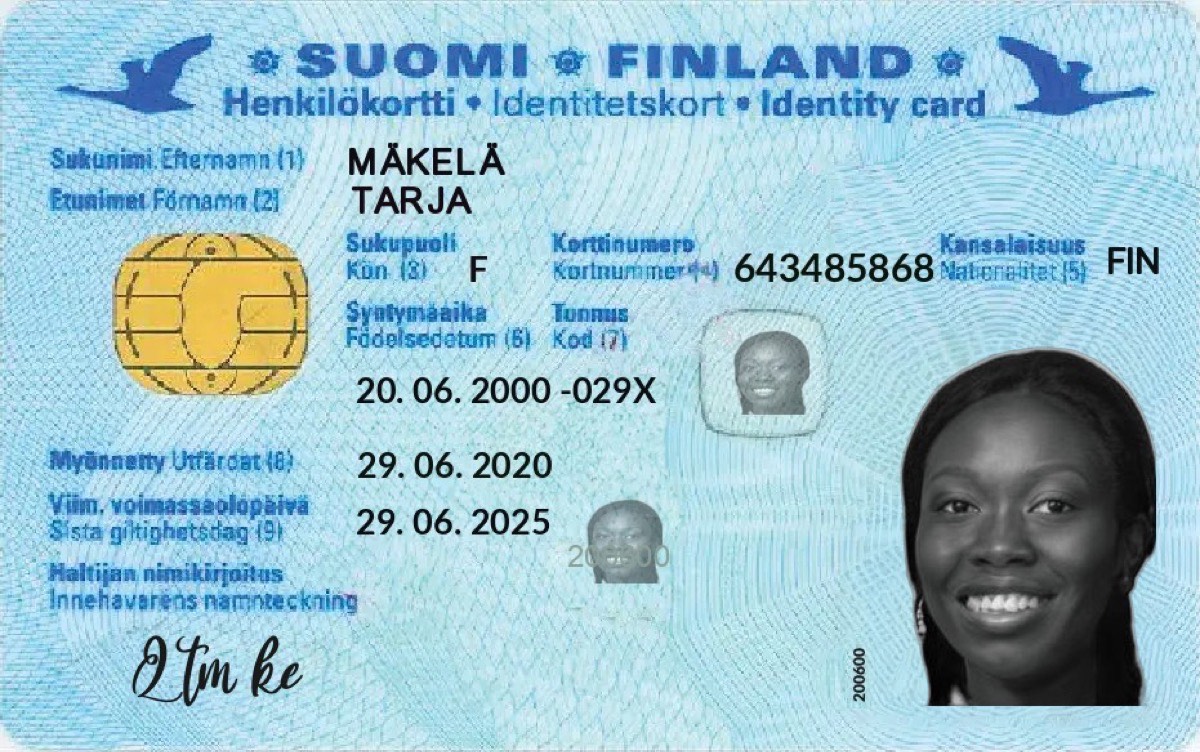}
    }
    \subfigure[\tiny Inpaint\&Rewrite fraud sample from \IDSpace]{
        \includegraphics[width=0.3\textwidth]{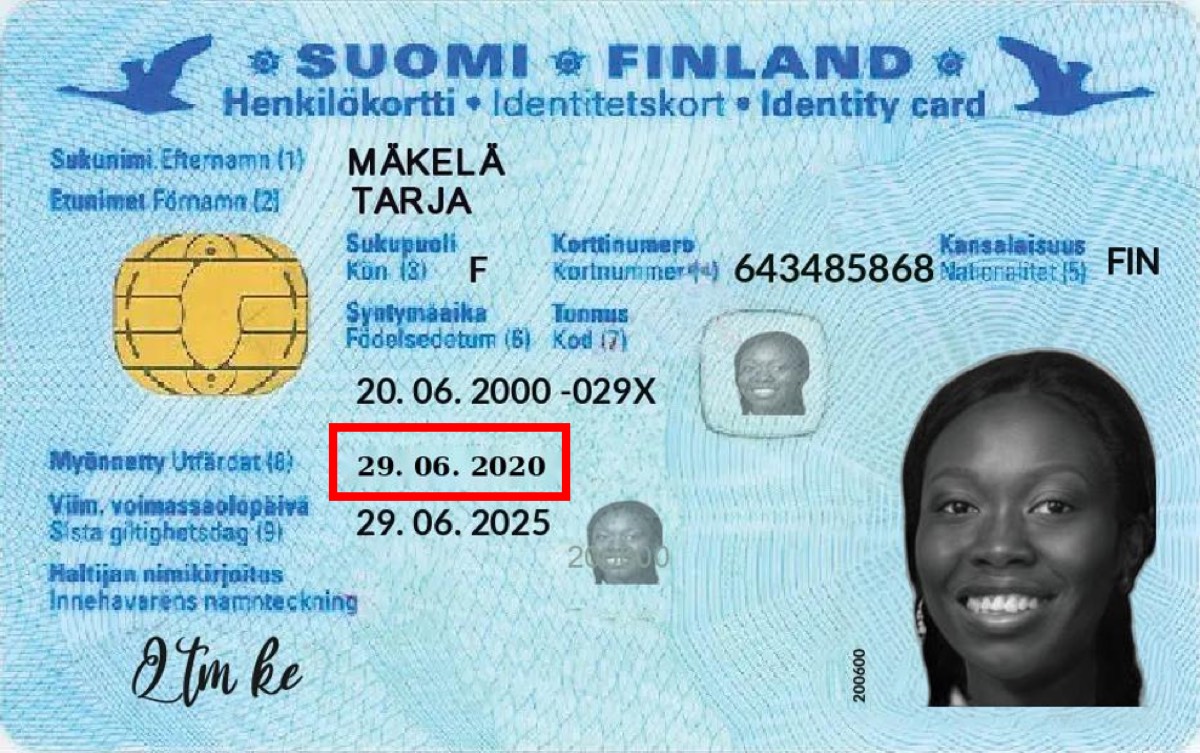}
    }
    \subfigure[\tiny Crop\&Replace fraud sample from \IDSpace]{
        \includegraphics[width=0.3\textwidth]{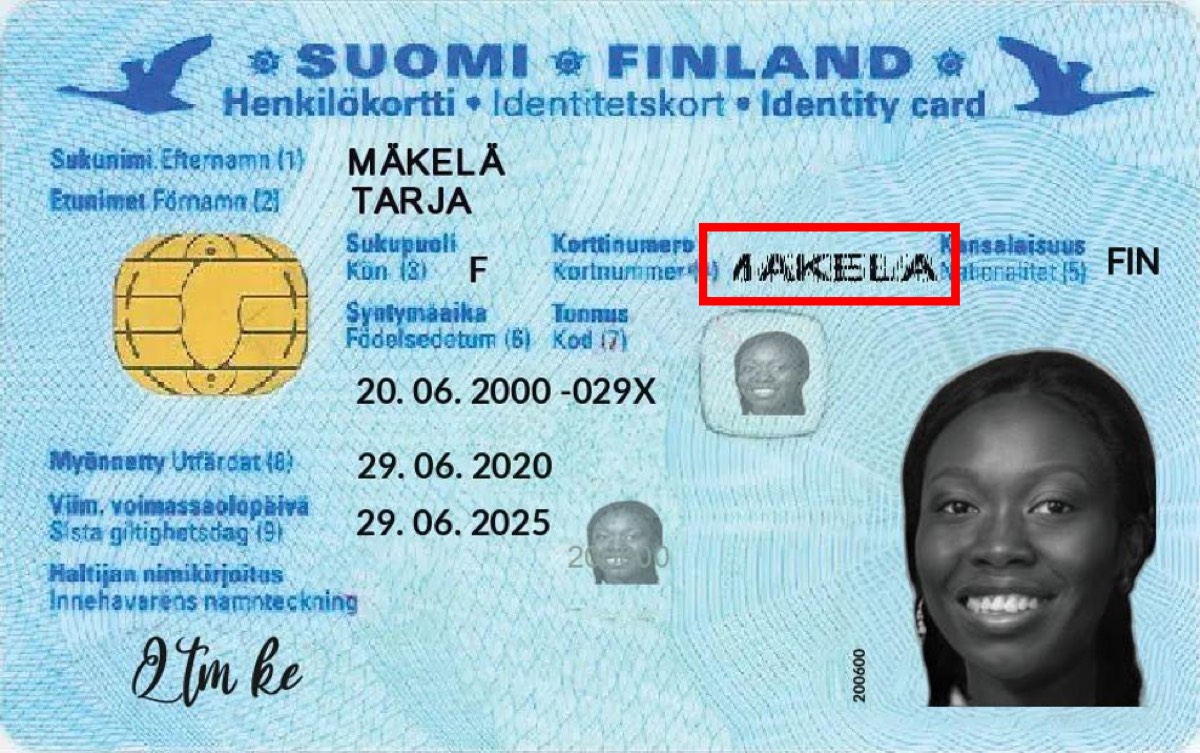}
    }

        \subfigure[\tiny Scanned sample from MIDV]{
        \includegraphics[width=0.22\textwidth]{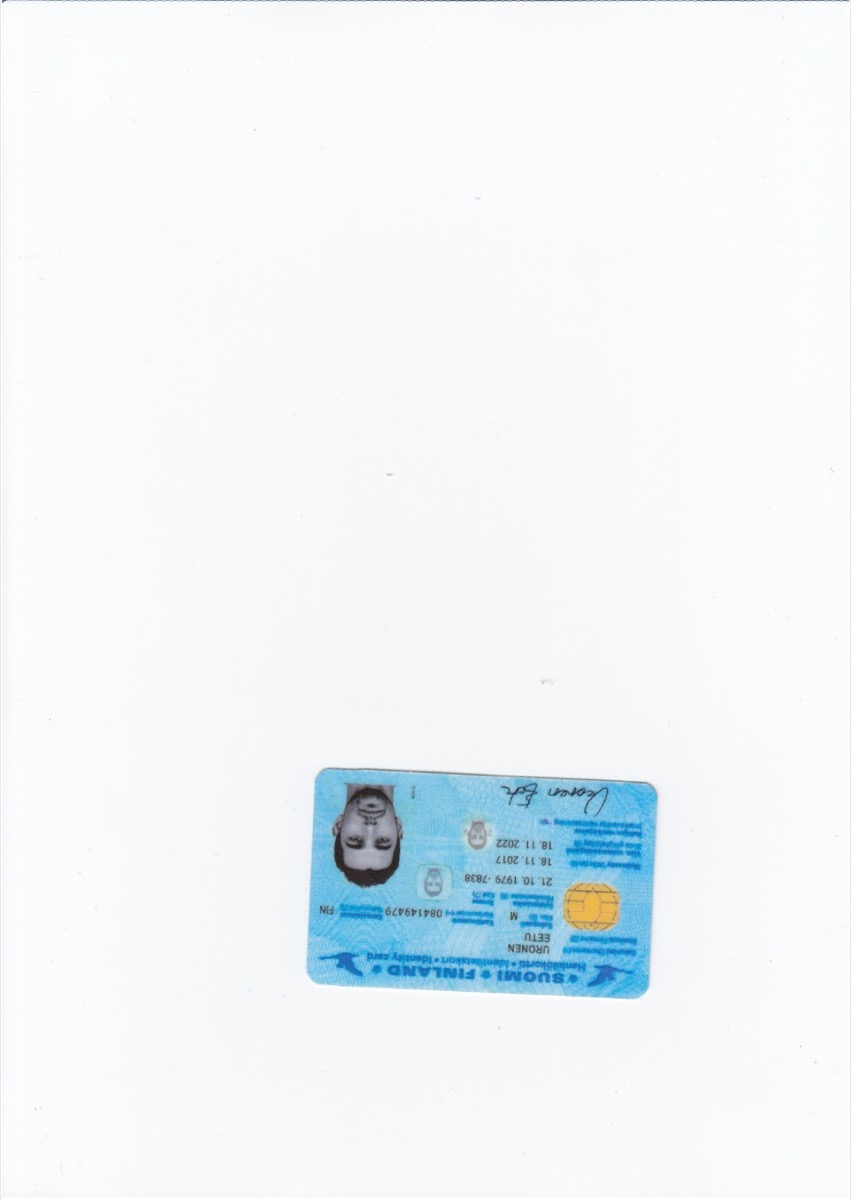}
    }
    \subfigure[\tiny Scanned sample from \IDSpace]{
        \includegraphics[width=0.22\textwidth]{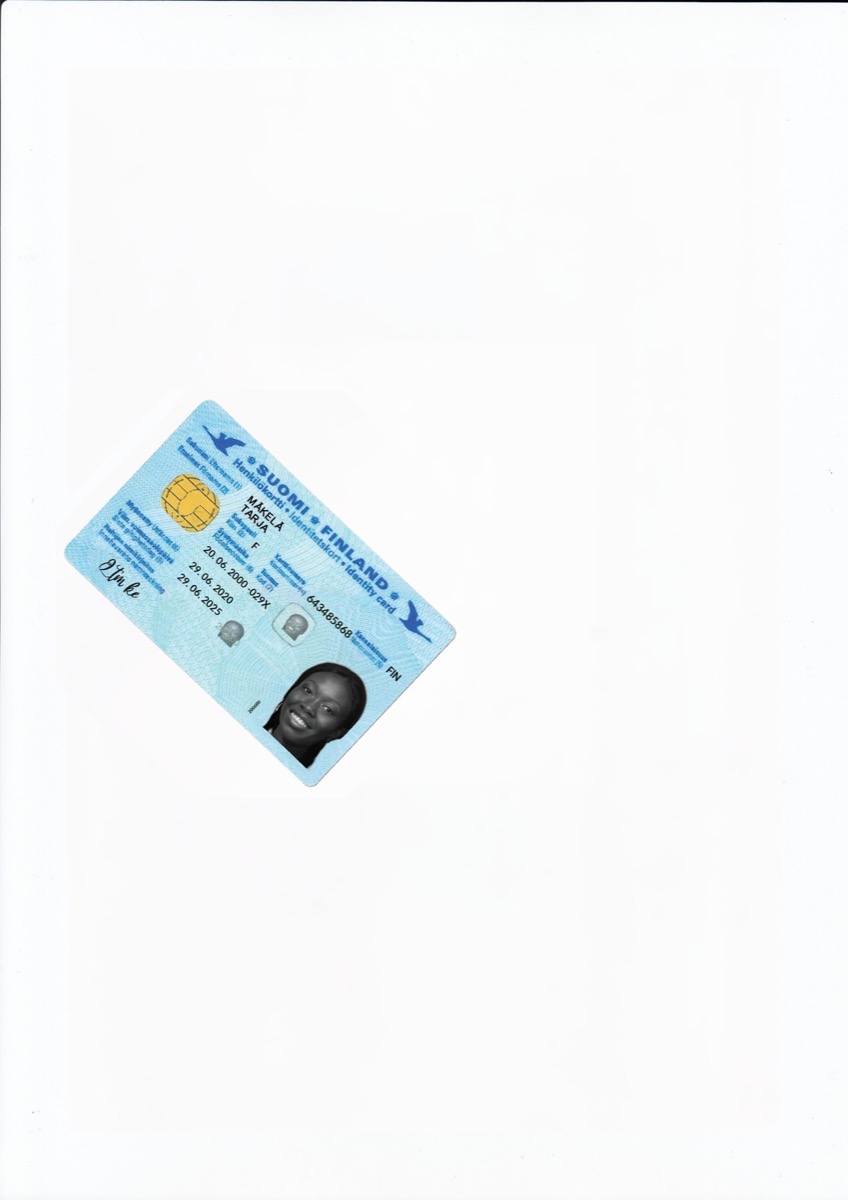}
    }
    \subfigure[\tiny Scanned sample with Inpaint\&Rewrite fraud from \IDSpace]{
        \includegraphics[width=0.22\textwidth]{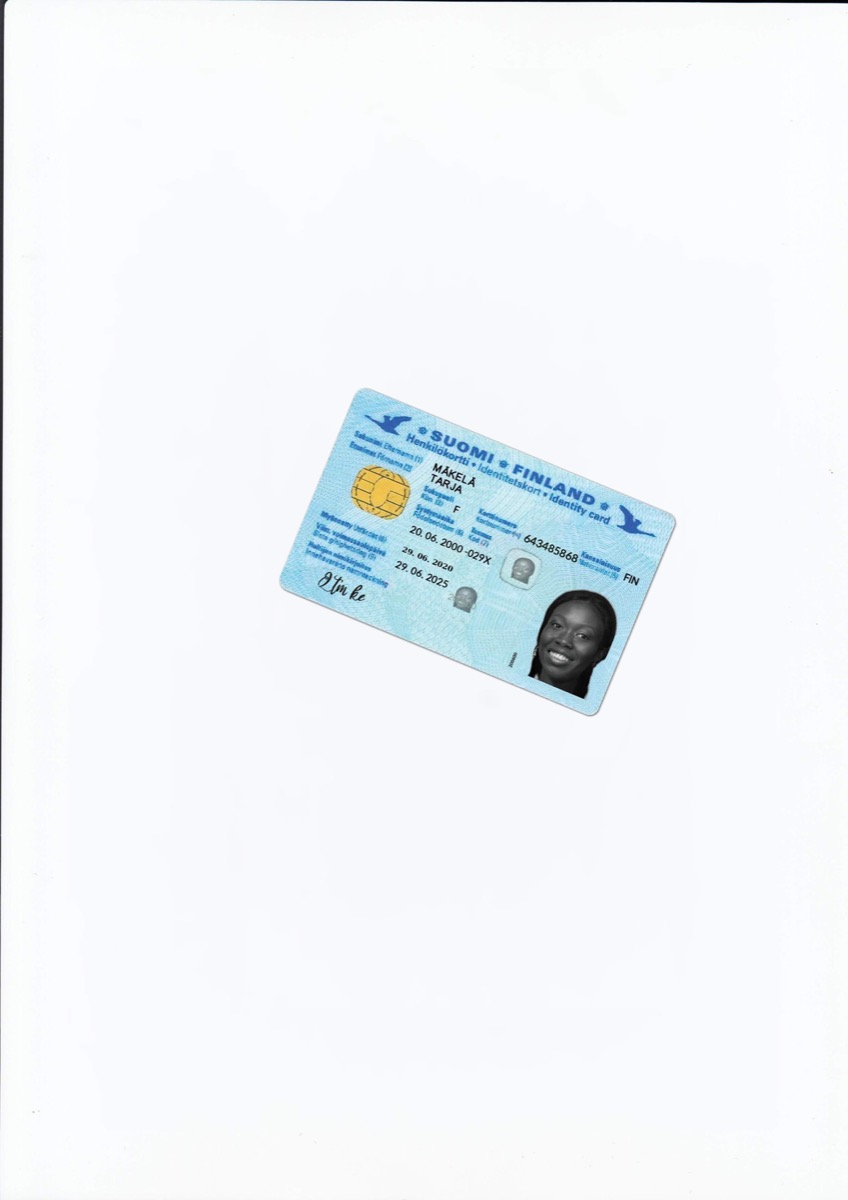}
    }
    \subfigure[\tiny Scanned sample with Crop\&Replace fraud from \IDSpace] {
        \includegraphics[width=0.22\textwidth]{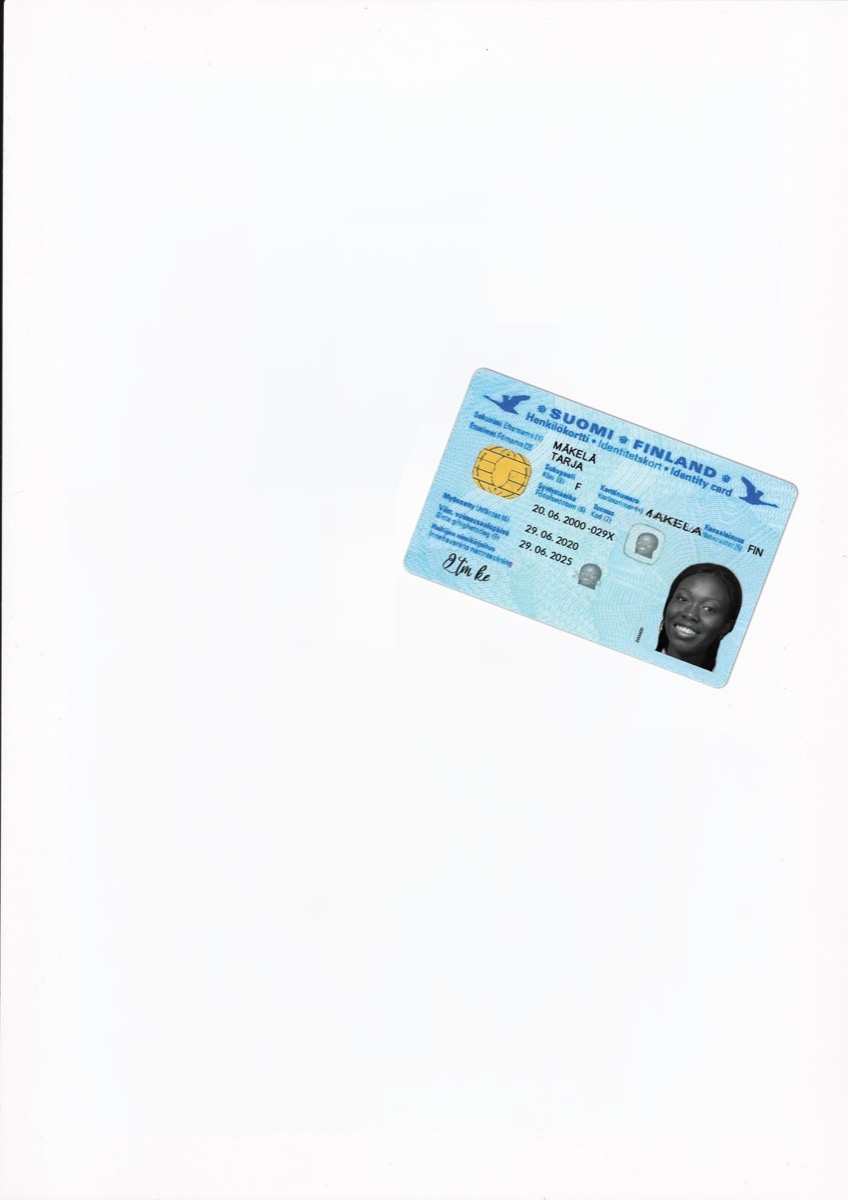}
    }
    \caption{Examples from multiple datasets containing Finnish ID card images.}
    \label{fig:fin_samples}
\end{figure*}

\begin{figure*}[h!]
    \centering
    \subfigure[\tiny Templated sample from MIDV]{
        \includegraphics[width=0.3\textwidth]{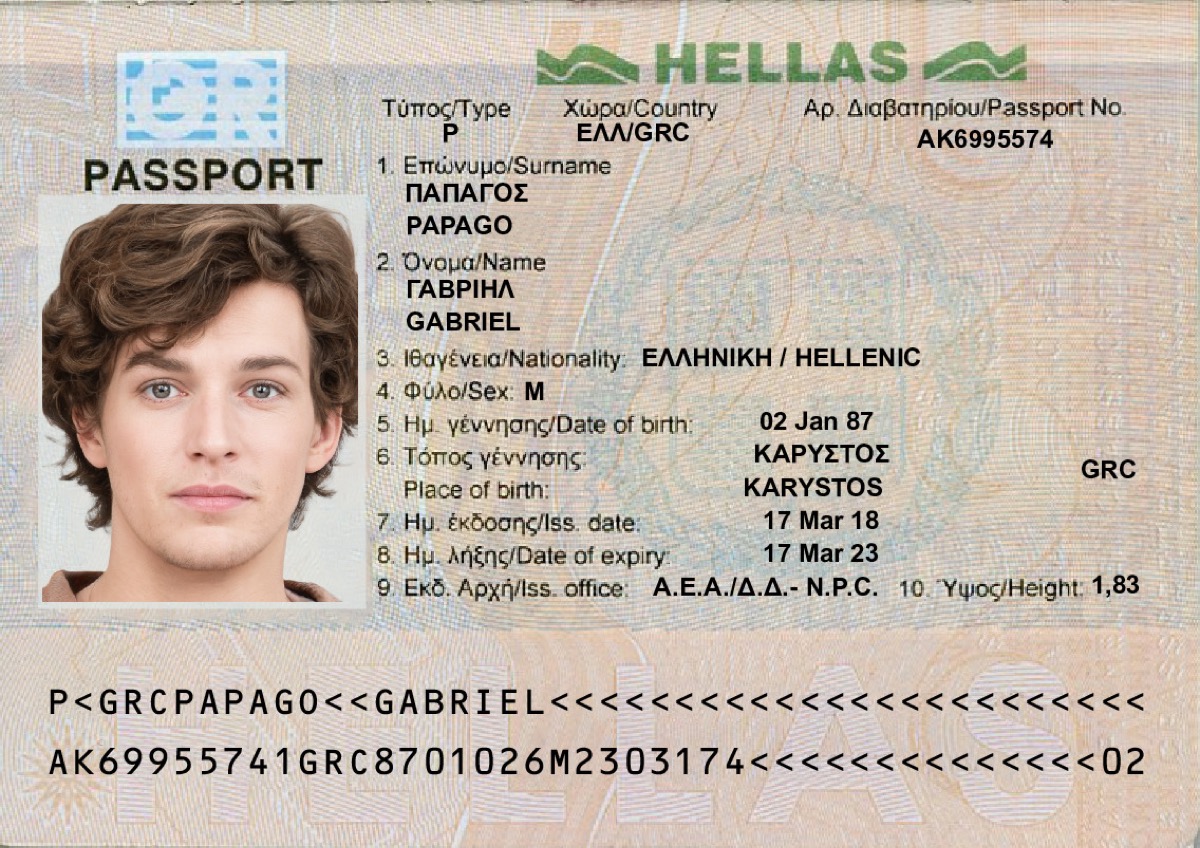}
    }
    \subfigure[\tiny Inpaint\&Rewrite fraud sample from SIDTD]{
        \includegraphics[width=0.3\textwidth]{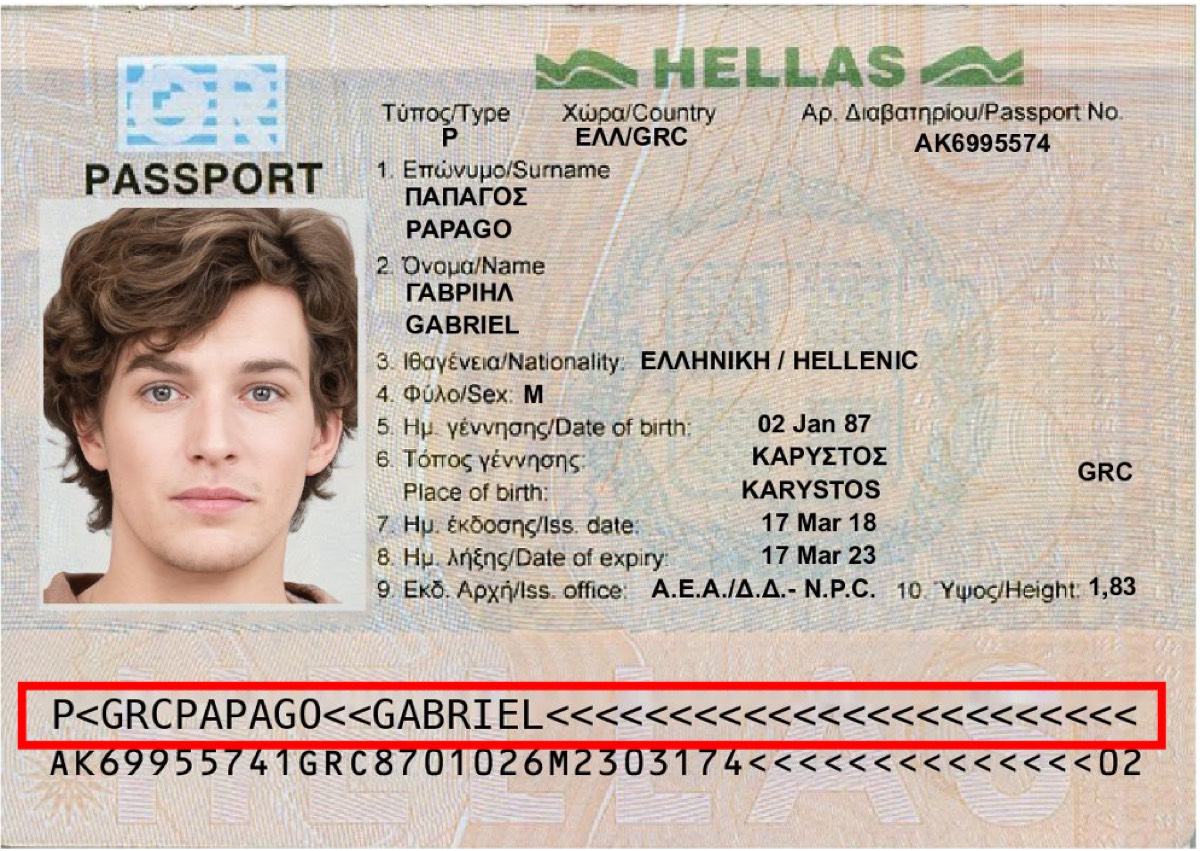}
    }
    \subfigure[\tiny Crop\&Replace fraud sample from SIDTD]{
        \includegraphics[width=0.3\textwidth]{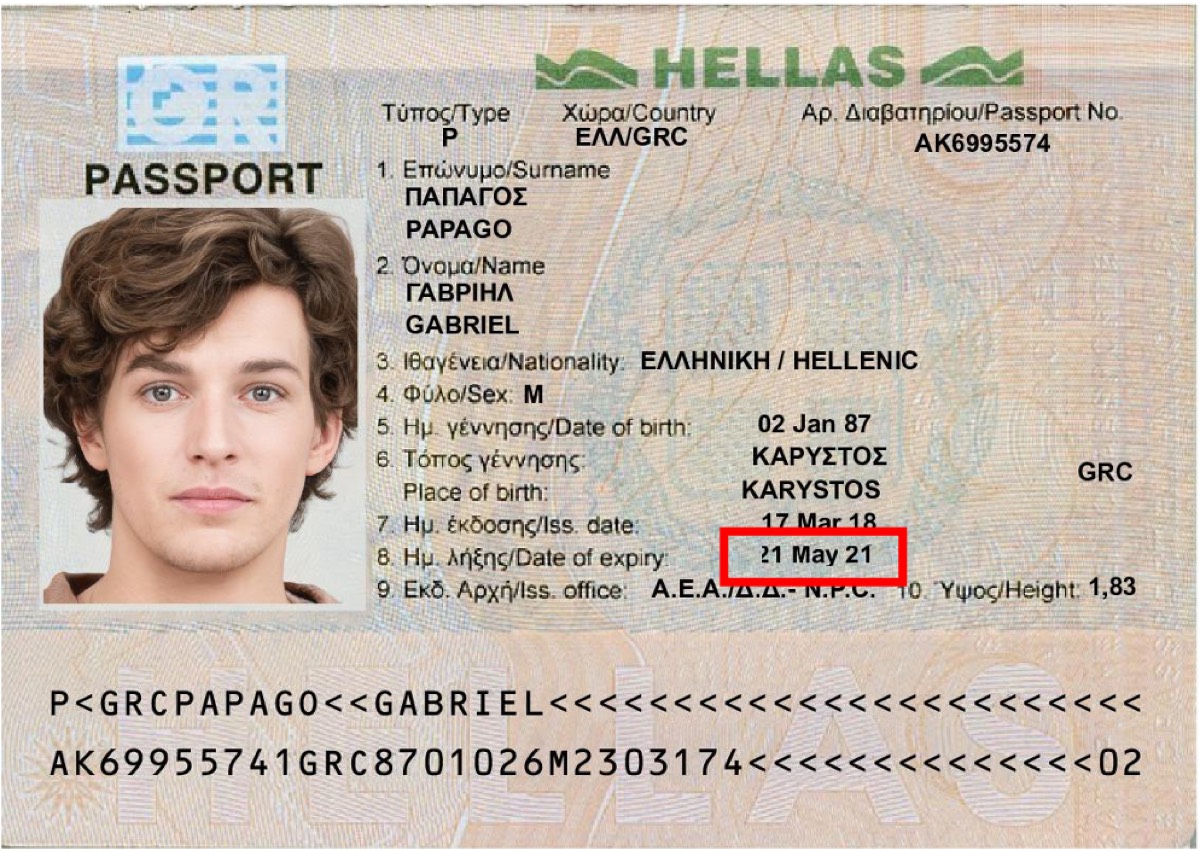}
    }

        \subfigure[\tiny Templated sample from \IDSpace]{
        \includegraphics[width=0.3\textwidth]{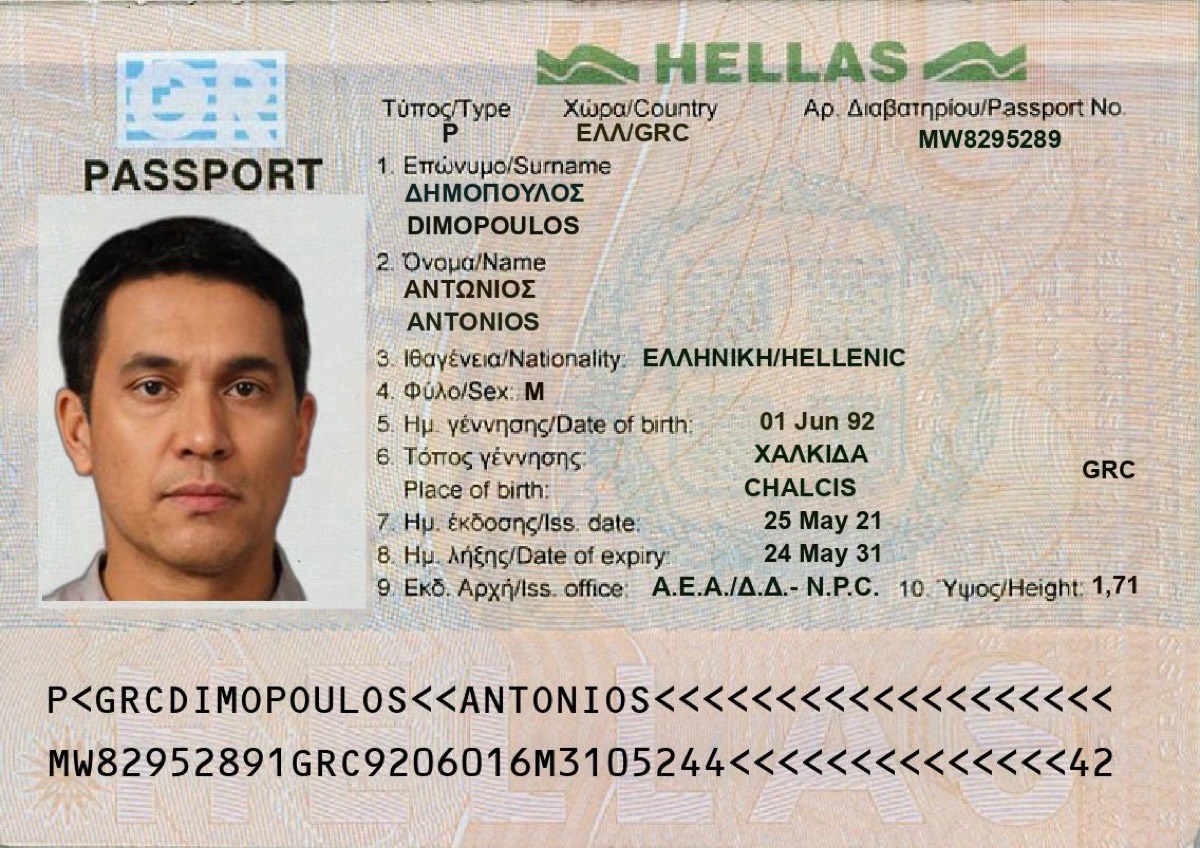}
    }
    \subfigure[\tiny Inpaint\&Rewrite fraud sample from \IDSpace]{
        \includegraphics[width=0.3\textwidth]{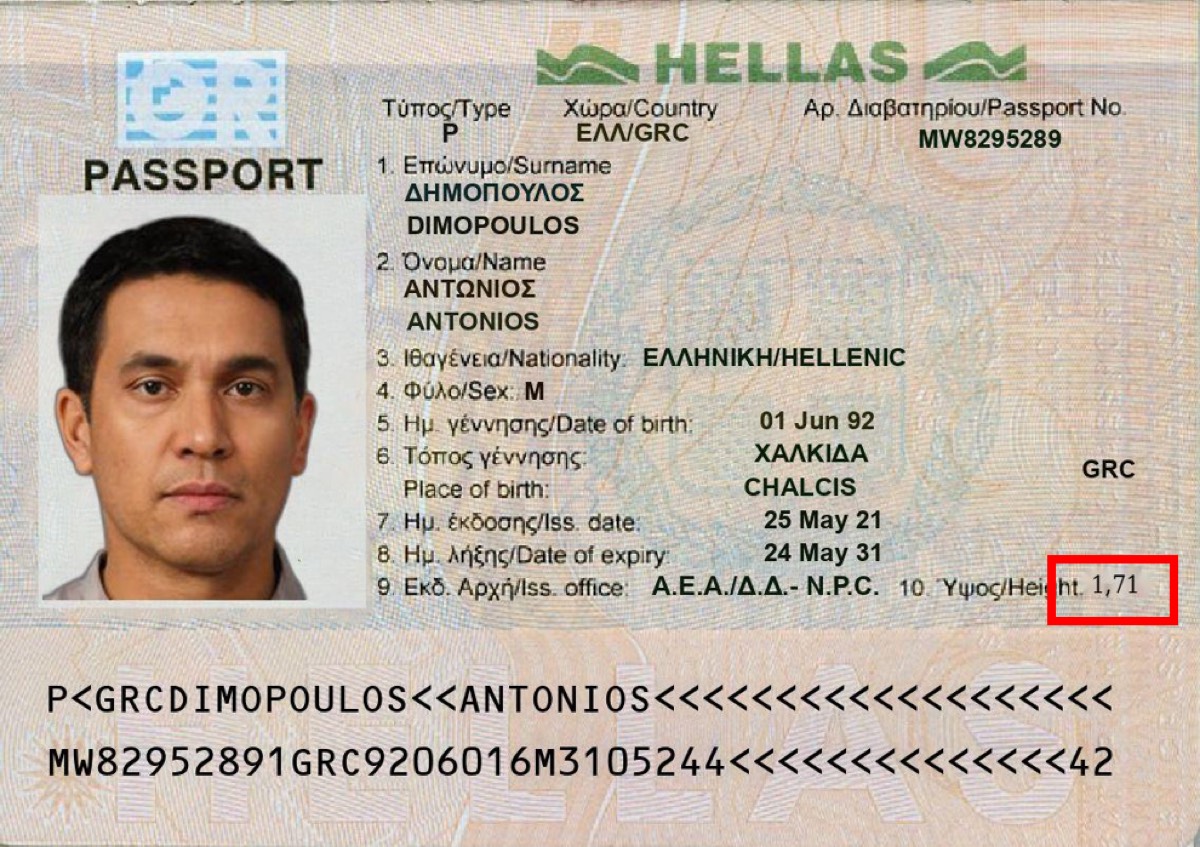}
    }
    \subfigure[\tiny Crop\&Replace fraud sample from \IDSpace]{
        \includegraphics[width=0.3\textwidth]{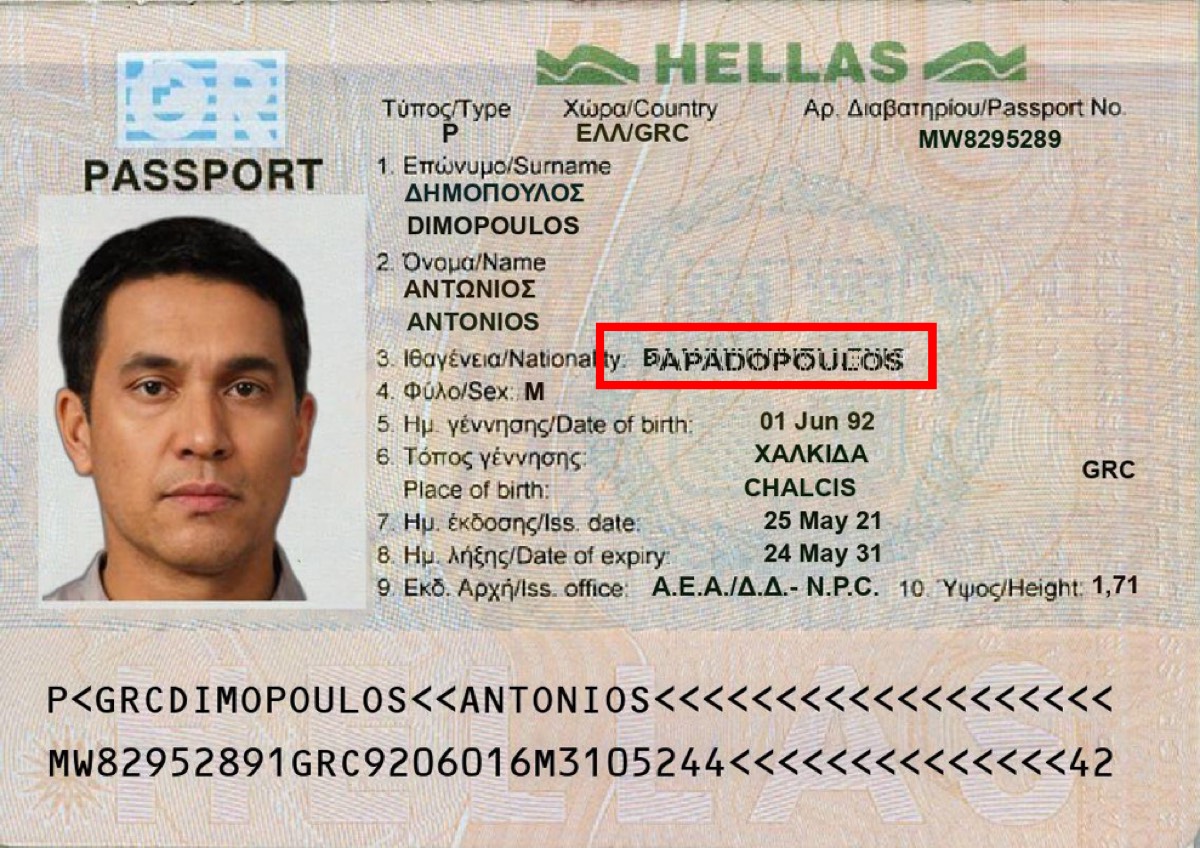}
    }

        \subfigure[\tiny Scanned sample from MIDV]{
        \includegraphics[width=0.22\textwidth]{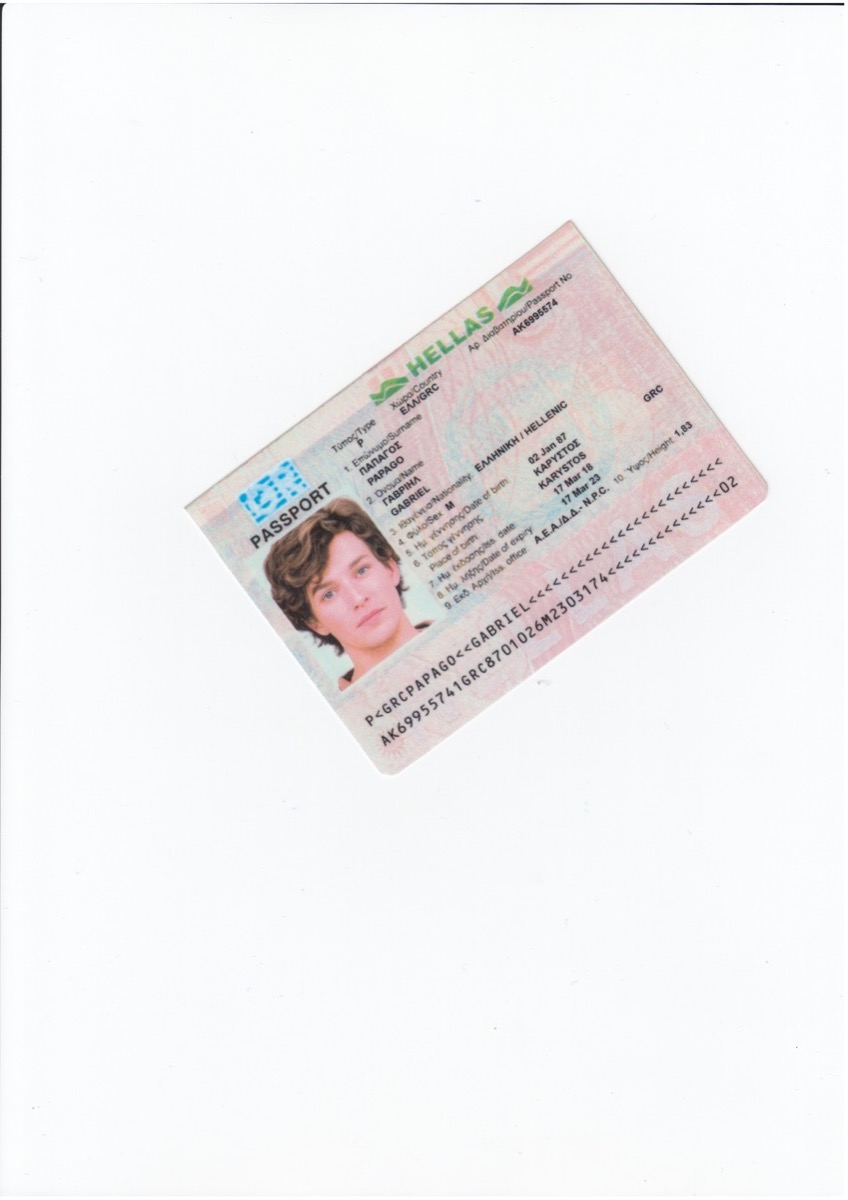}
    }
    \subfigure[\tiny Scanned sample from \IDSpace]{
        \includegraphics[width=0.22\textwidth]{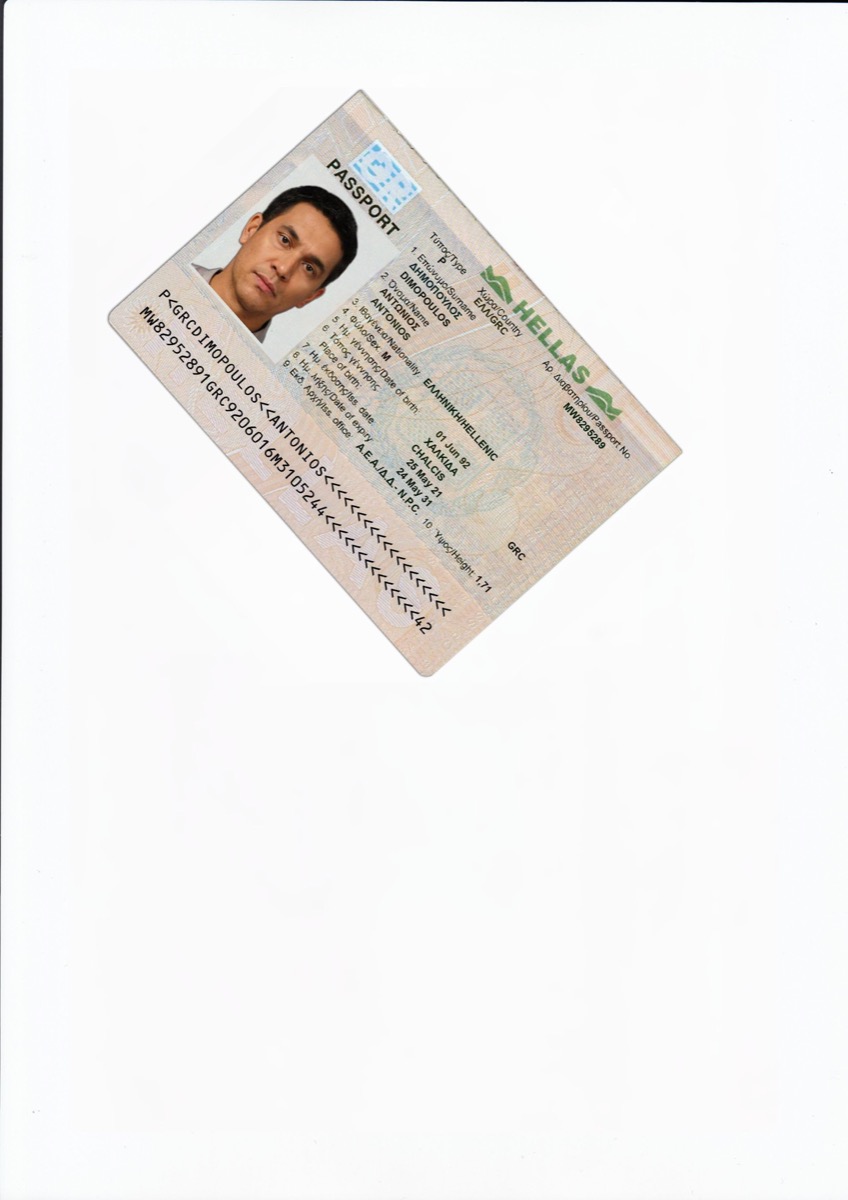}
    }
    \subfigure[\tiny Scanned sample with Inpaint\&Rewrite fraud from \IDSpace]{
        \includegraphics[width=0.22\textwidth]{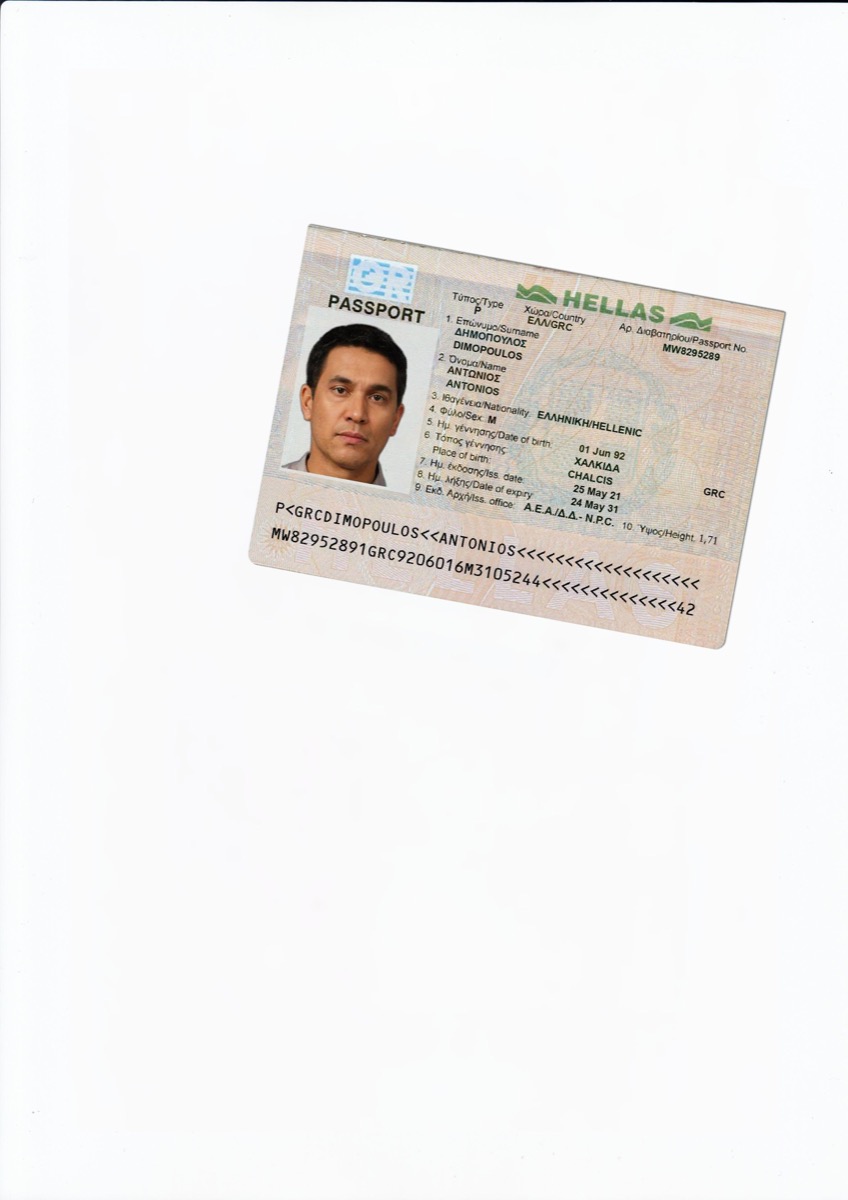}
    }
    \subfigure[\tiny Scanned sample with Crop\&Replace from fraud \IDSpace]{
        \includegraphics[width=0.22\textwidth]{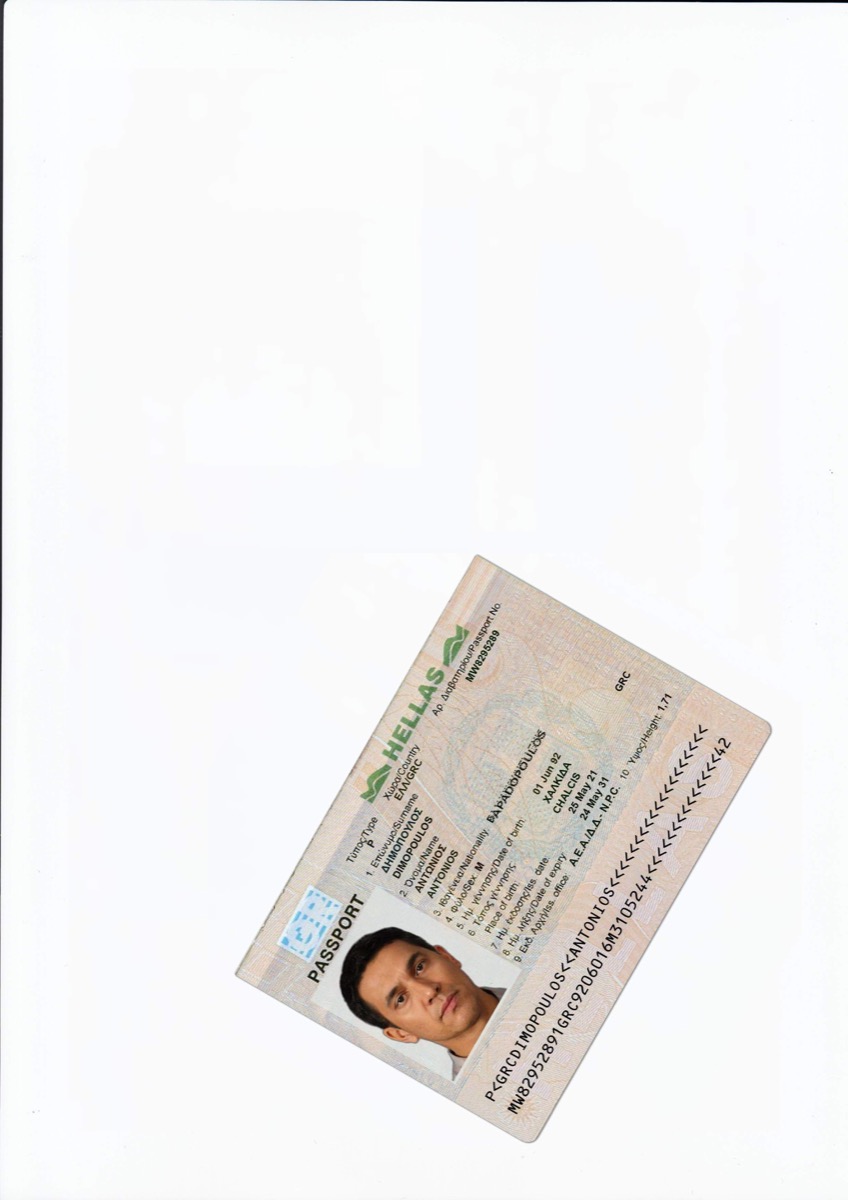}
    }
    \caption{Examples from multiple datasets containing Greek passport images.}
    \label{fig:grc_samples}
\end{figure*}

\begin{figure*}[h!]
    \centering
    \subfigure[\tiny Templated sample from MIDV]{
        \includegraphics[width=0.3\textwidth]{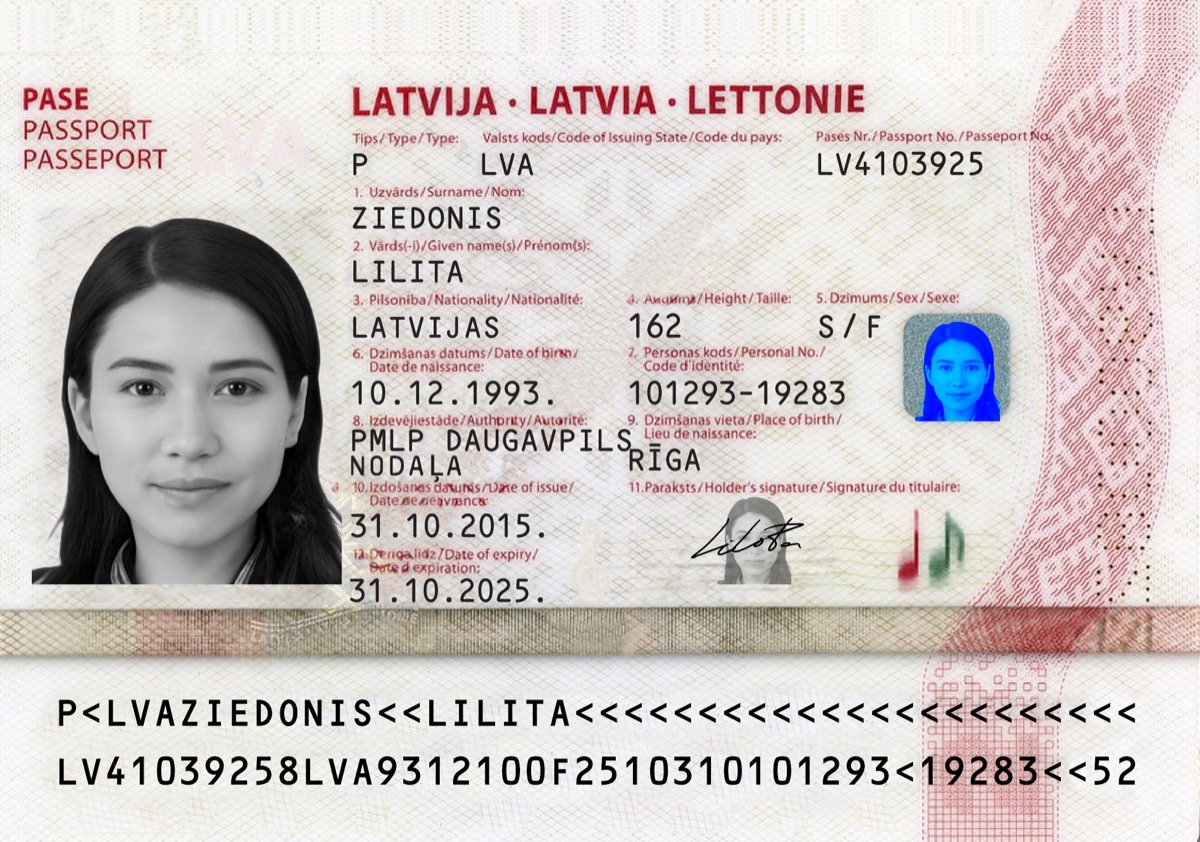}
    }
    \subfigure[\tiny Inpaint\&Rewrite fraud sample from SIDTD]{
        \includegraphics[width=0.3\textwidth]{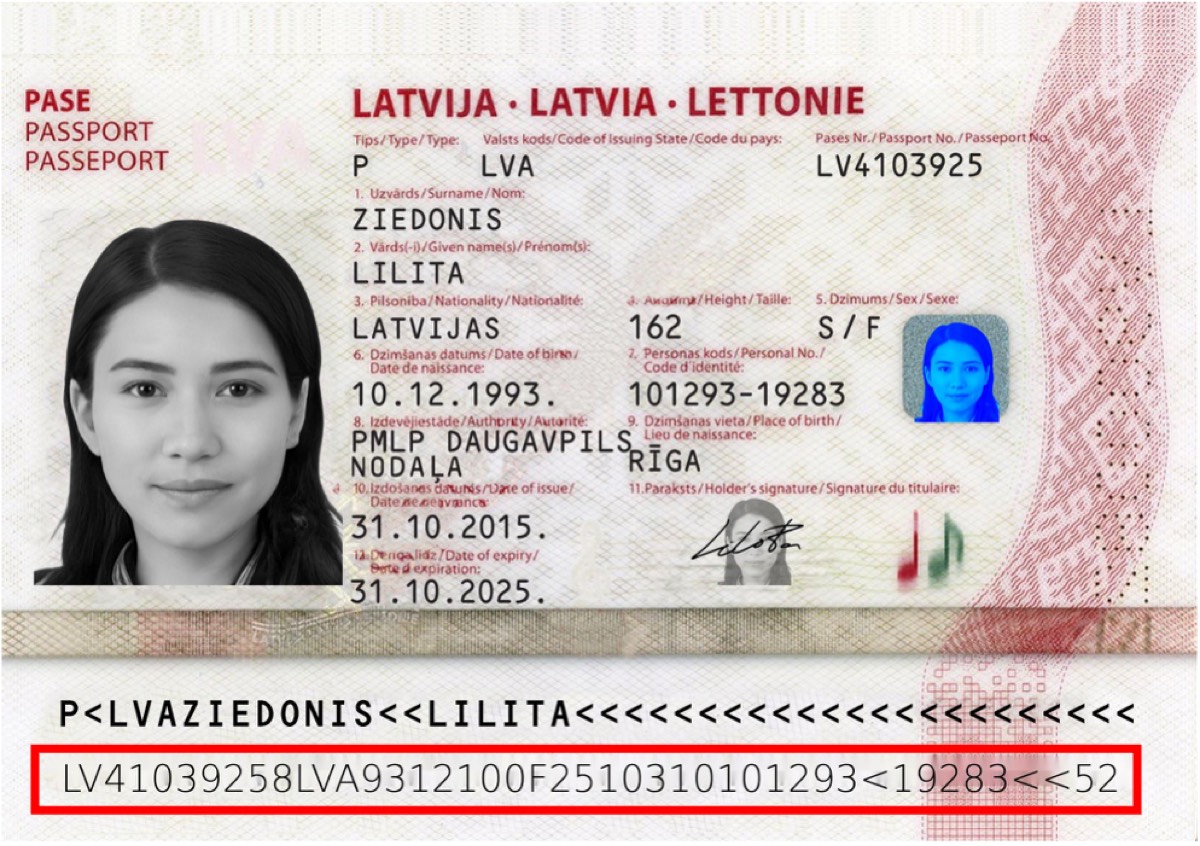}
    }
    \subfigure[\tiny Crop\&Replace fraud sample from SIDTD]{
        \includegraphics[width=0.3\textwidth]{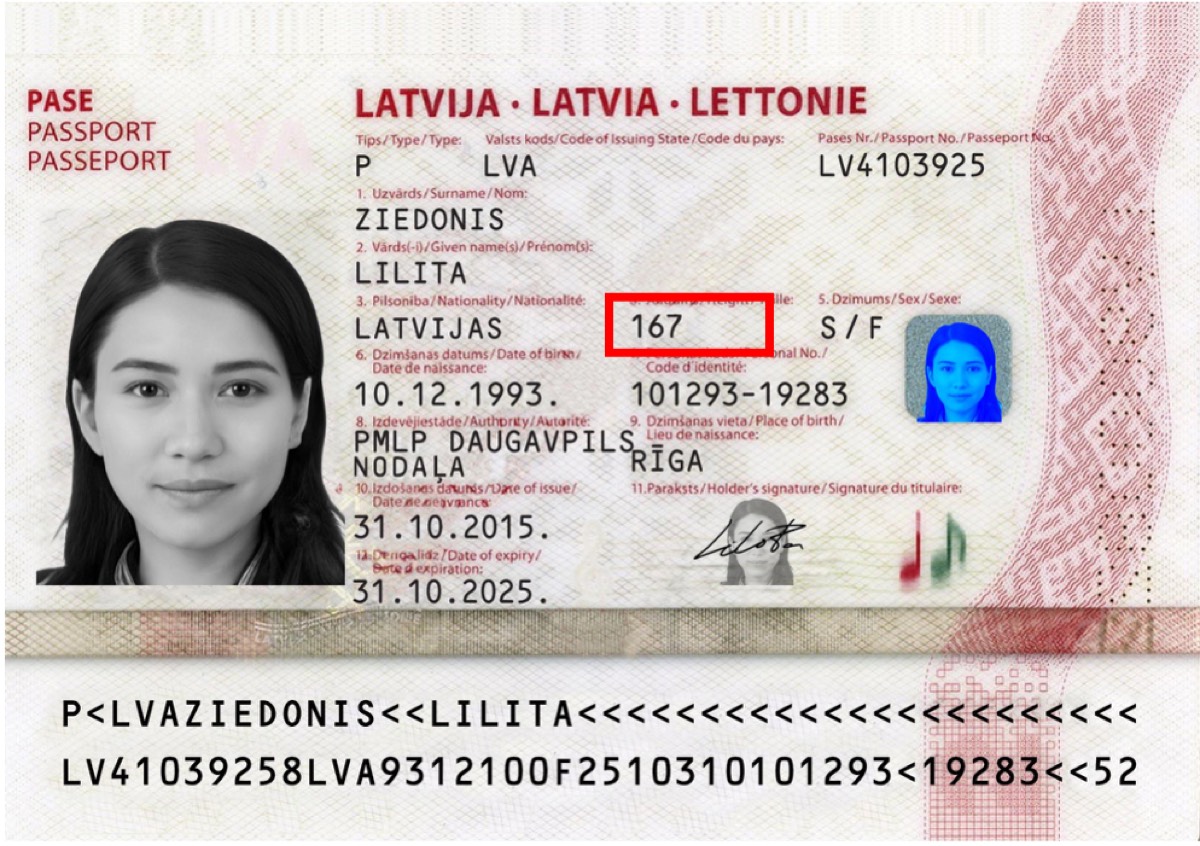}
    }
        \subfigure[\tiny Templated sample from \IDSpace]{
        \includegraphics[width=0.3\textwidth]{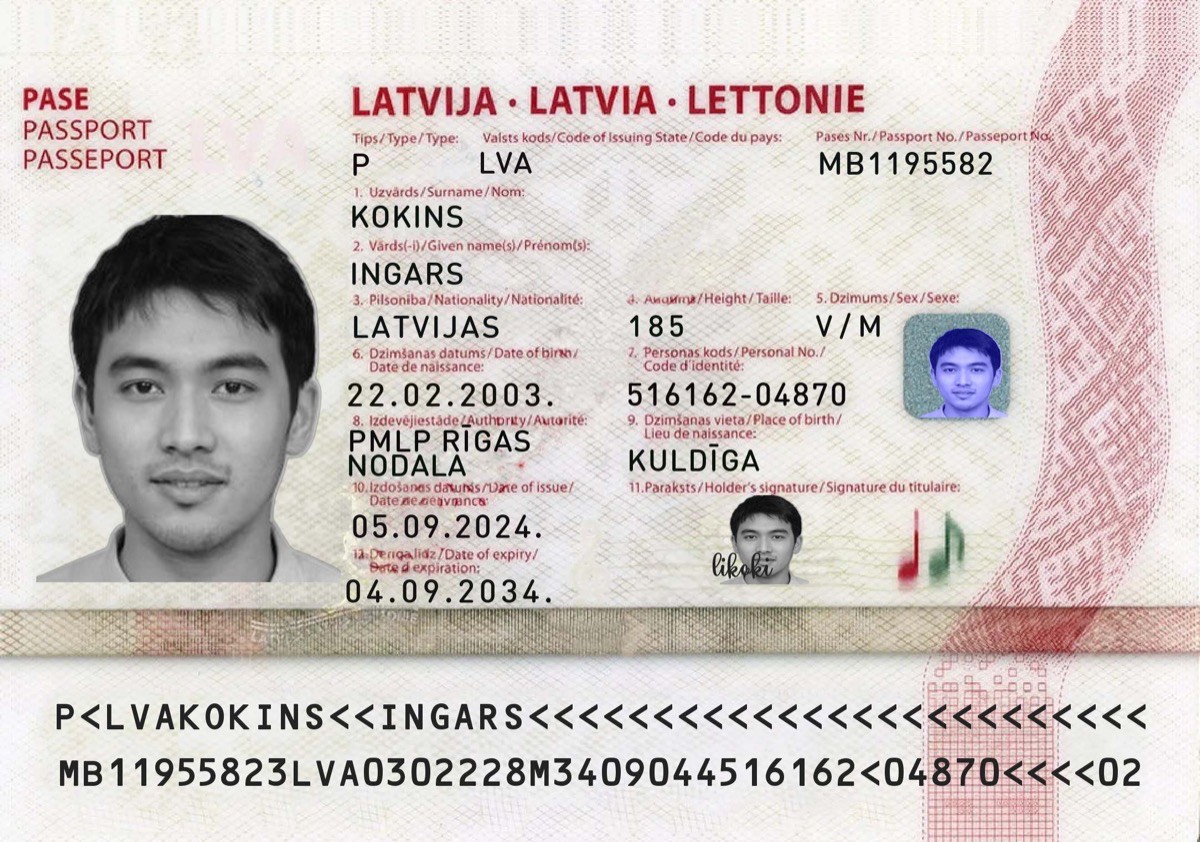}
    }
    \subfigure[\tiny Inpaint\&Rewrite fraud sample from \IDSpace]{
        \includegraphics[width=0.3\textwidth]{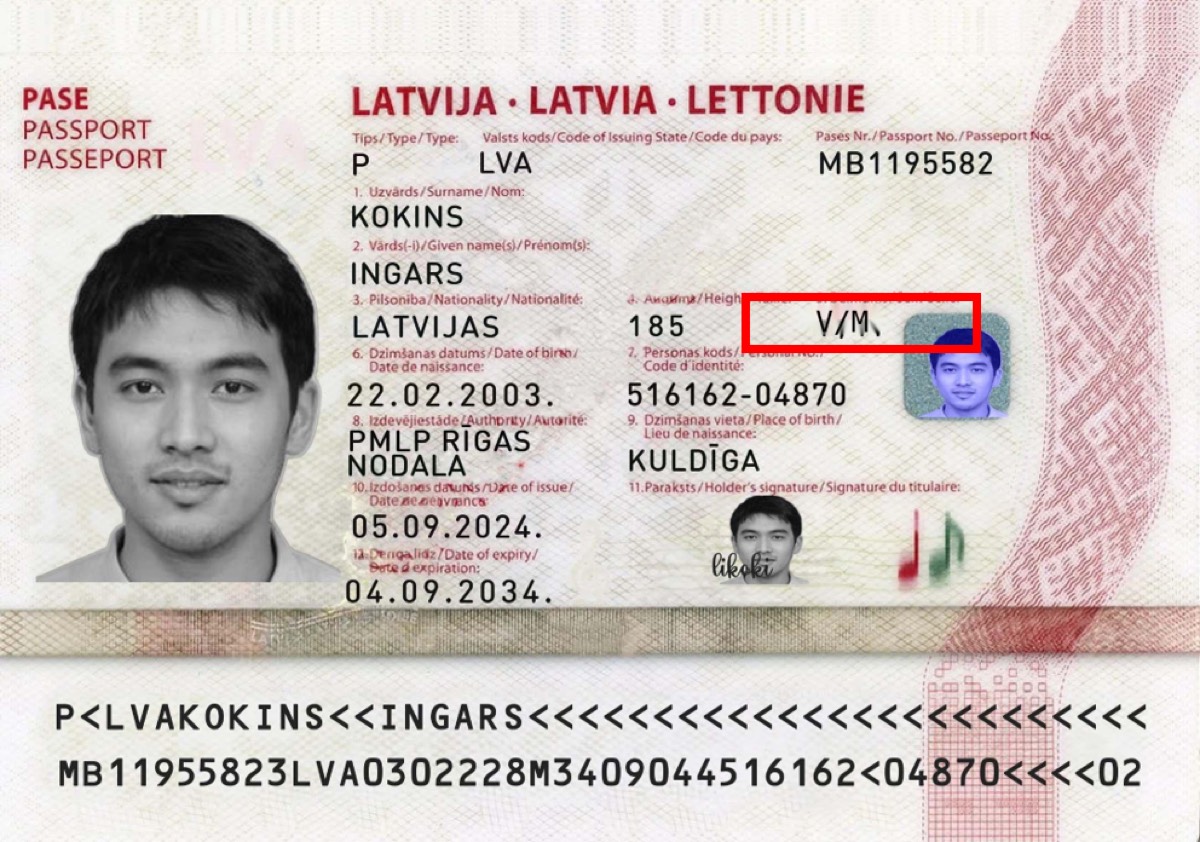}
    }
    \subfigure[\tiny Crop\&Replace fraud sample from \IDSpace]{
        \includegraphics[width=0.3\textwidth]{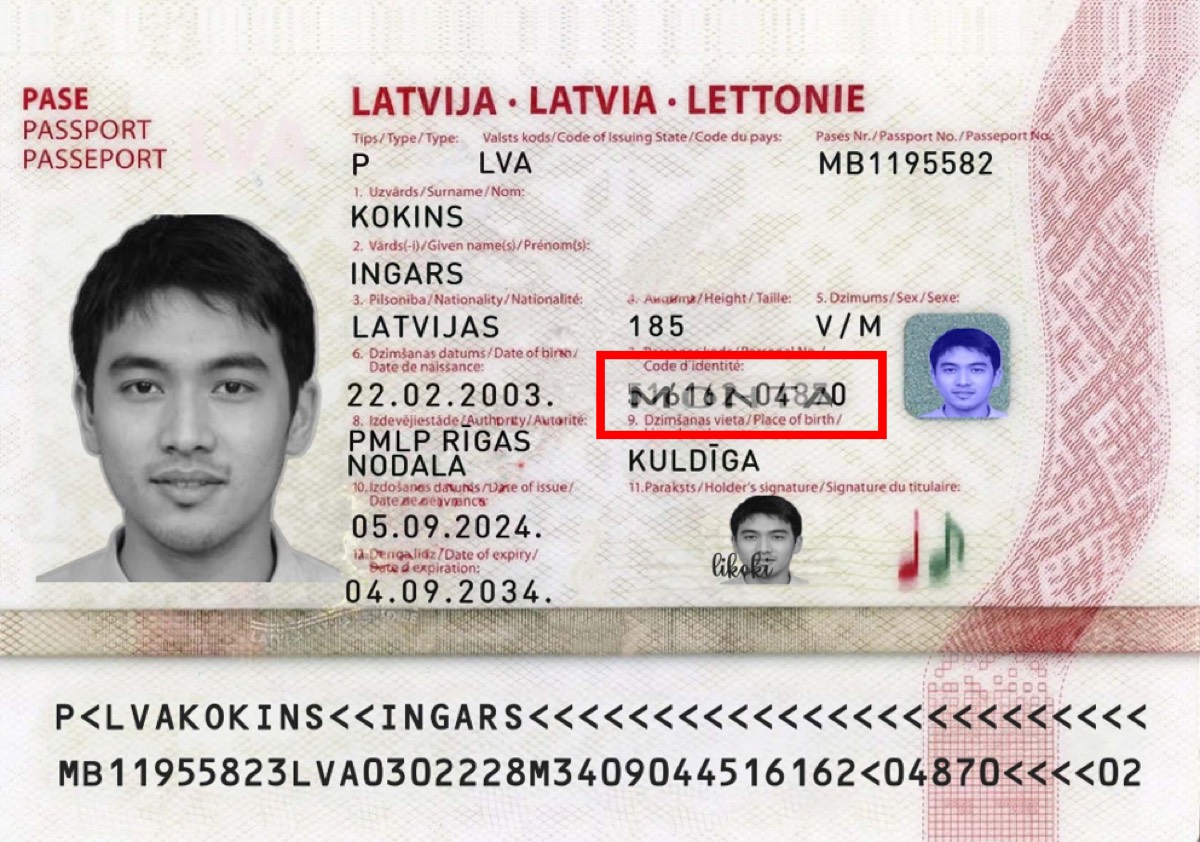}
    }

        \subfigure[\tiny Scanned sample from MIDV]{
        \includegraphics[width=0.22\textwidth]{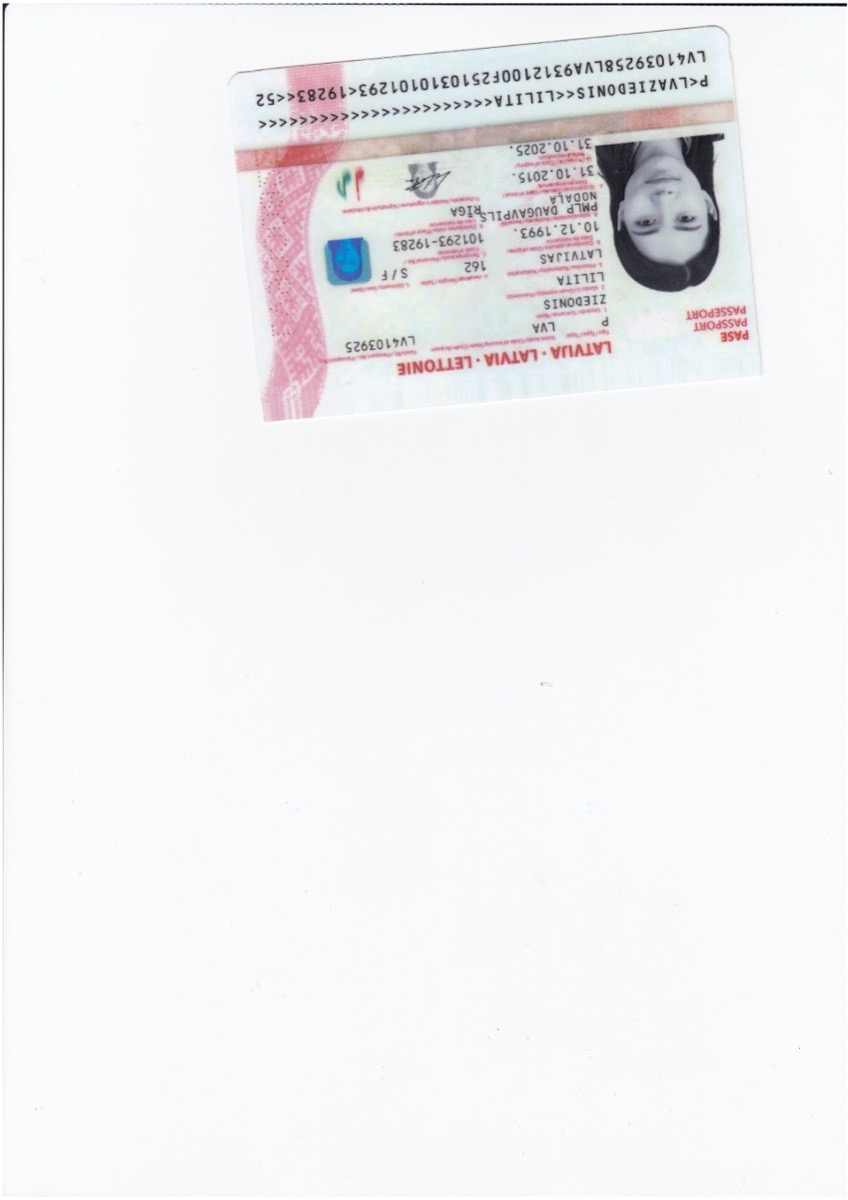}
    }
    \subfigure[\tiny Scanned sample from \IDSpace]{
        \includegraphics[width=0.22\textwidth]{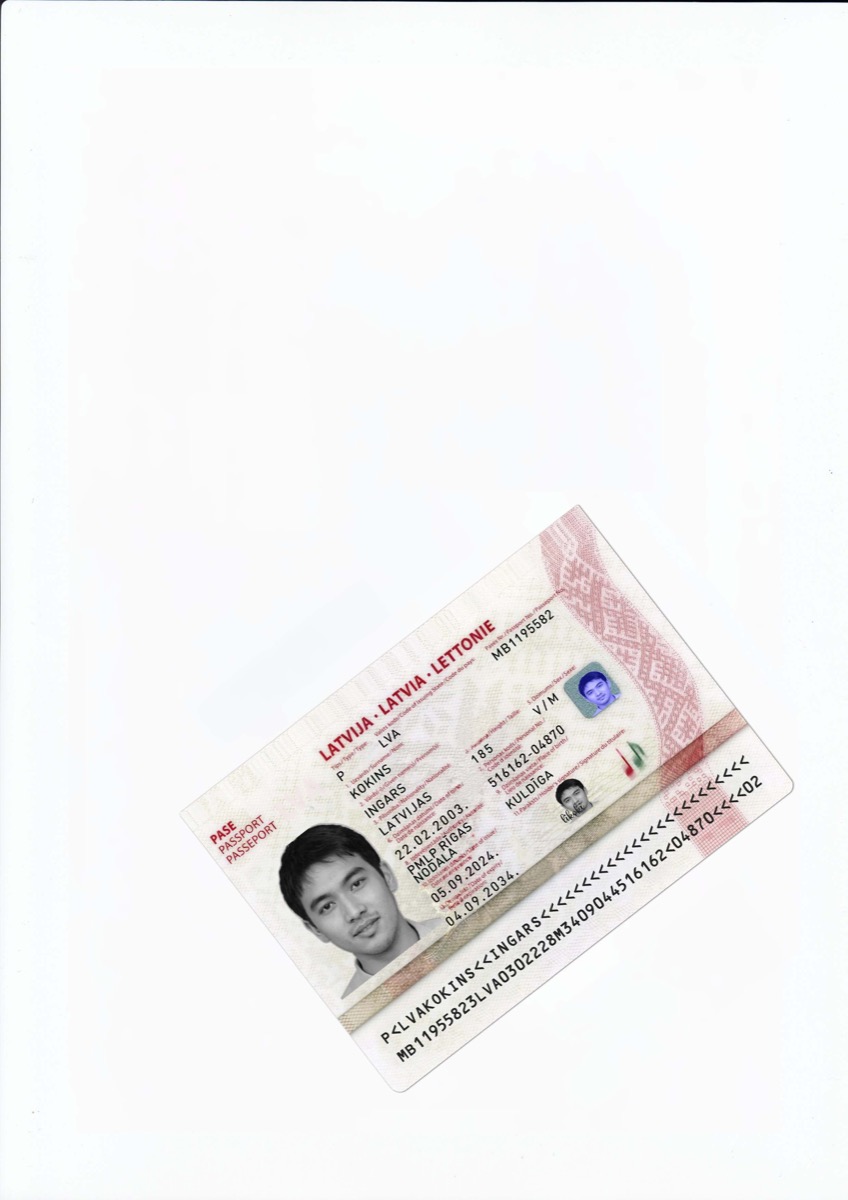}
    }
    \subfigure[\tiny Scanned sample with Inpaint\&Rewrite fraud from \IDSpace]{
        \includegraphics[width=0.22\textwidth]{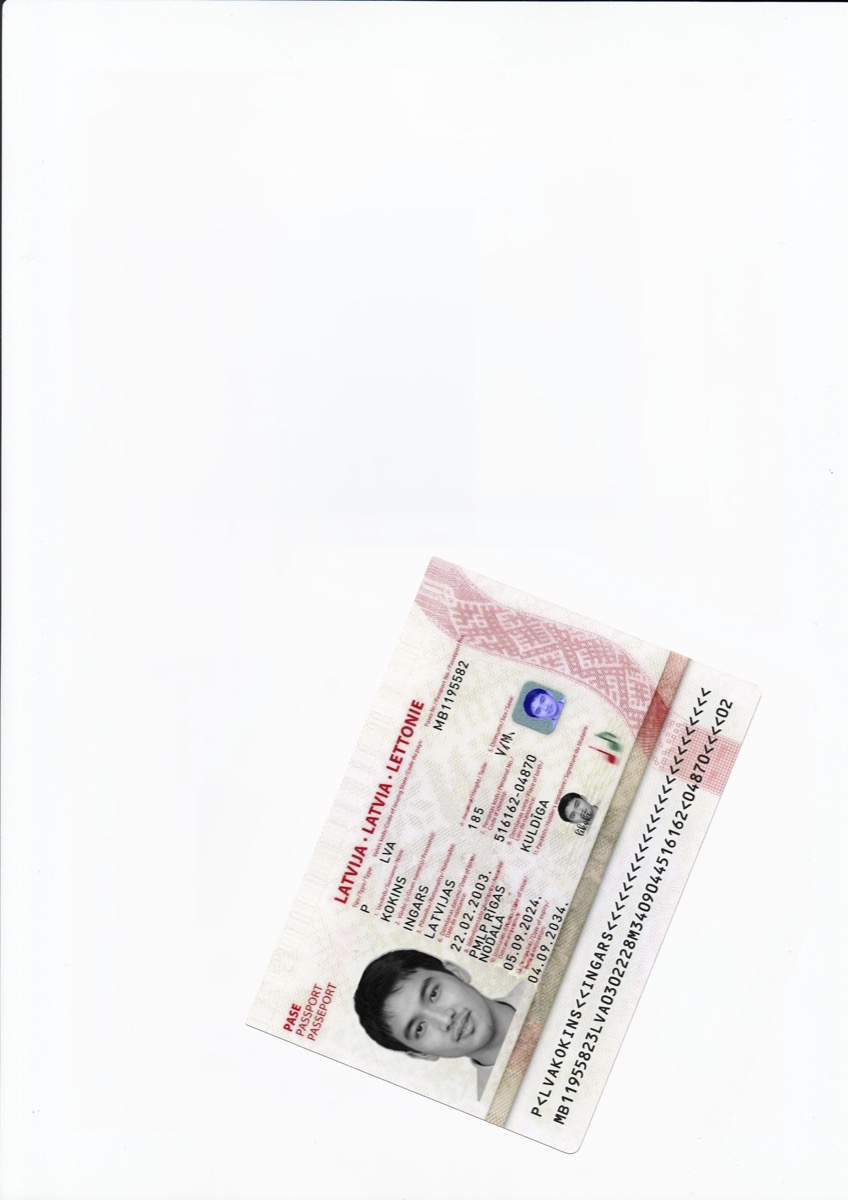}
    }
    \subfigure[\tiny Scanned sample with Crop\&Replace fraud from \IDSpace]{
        \includegraphics[width=0.22\textwidth]{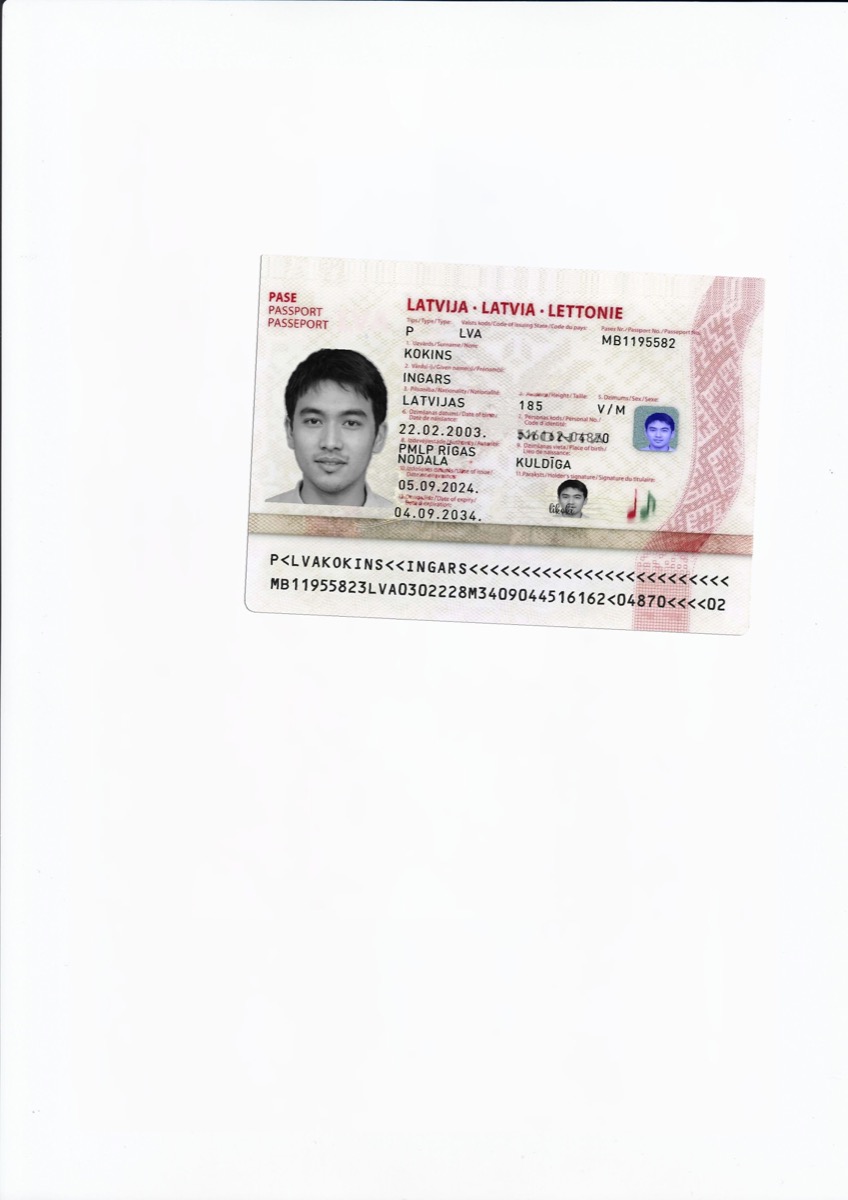}
    }
    \caption{Examples from multiple datasets containing Latvian passport images.}
    \label{fig:lva_samples}
\end{figure*}

\begin{figure*}[h!]
    \centering
    \subfigure[\tiny Templated sample from MIDV]{
        \includegraphics[width=0.3\textwidth]{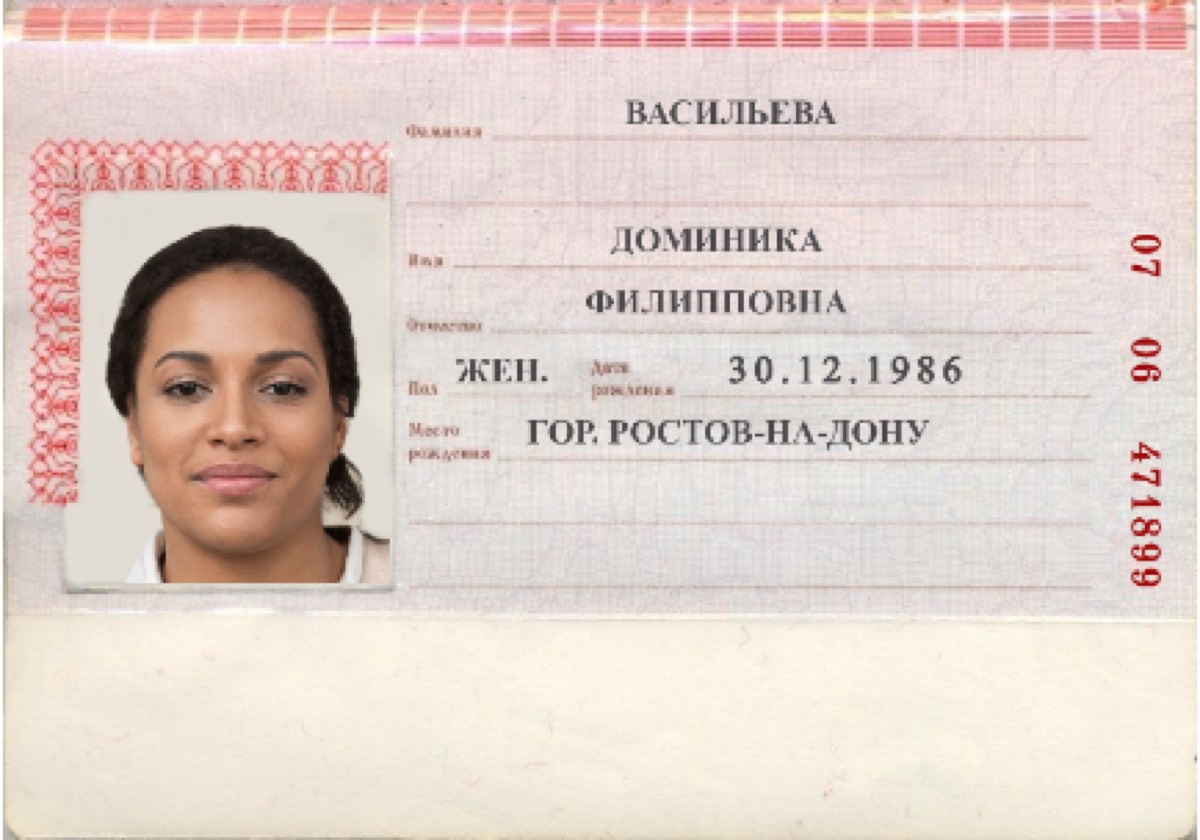}
    }
    \subfigure[\tiny Inpaint\&Rewrite fraud sample from SIDTD]{
        \includegraphics[width=0.3\textwidth]{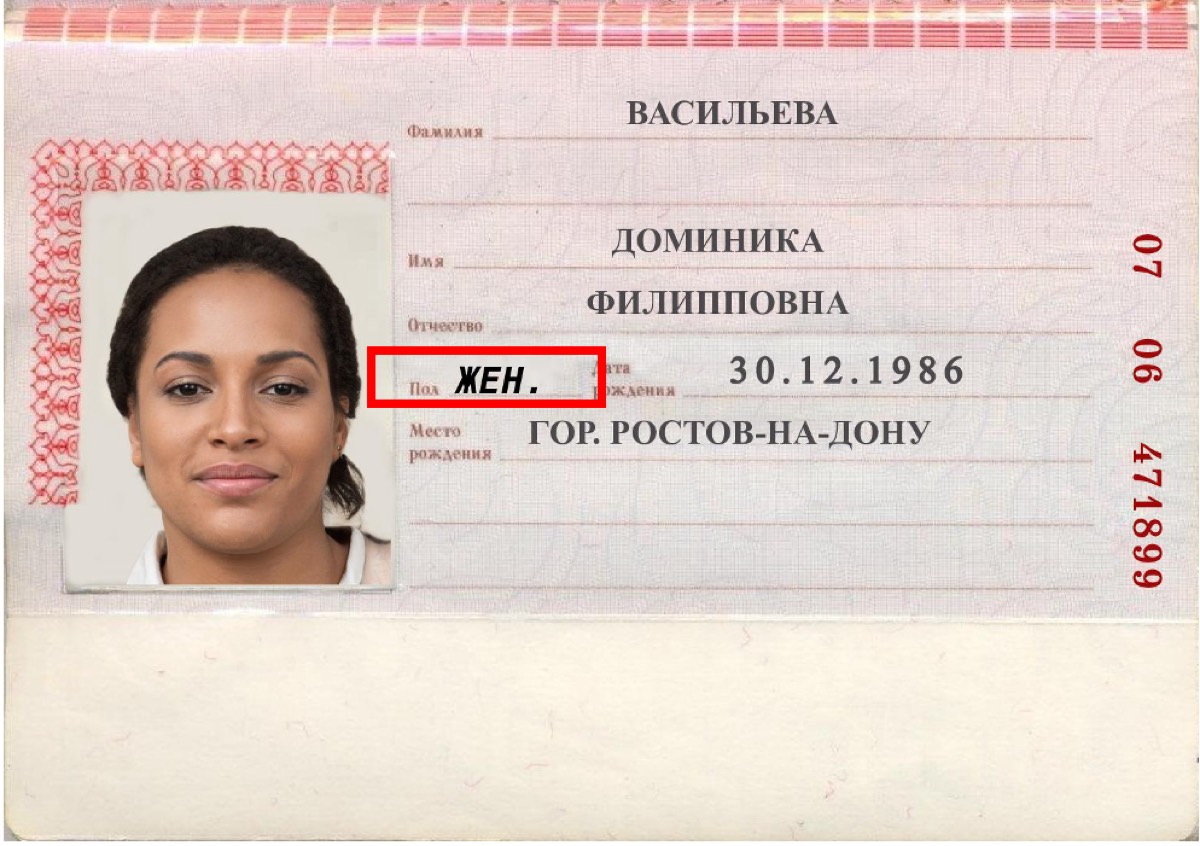}
    }
    \subfigure[\tiny Crop\&Replace fraud sample from SIDTD]{
        \includegraphics[width=0.3\textwidth]{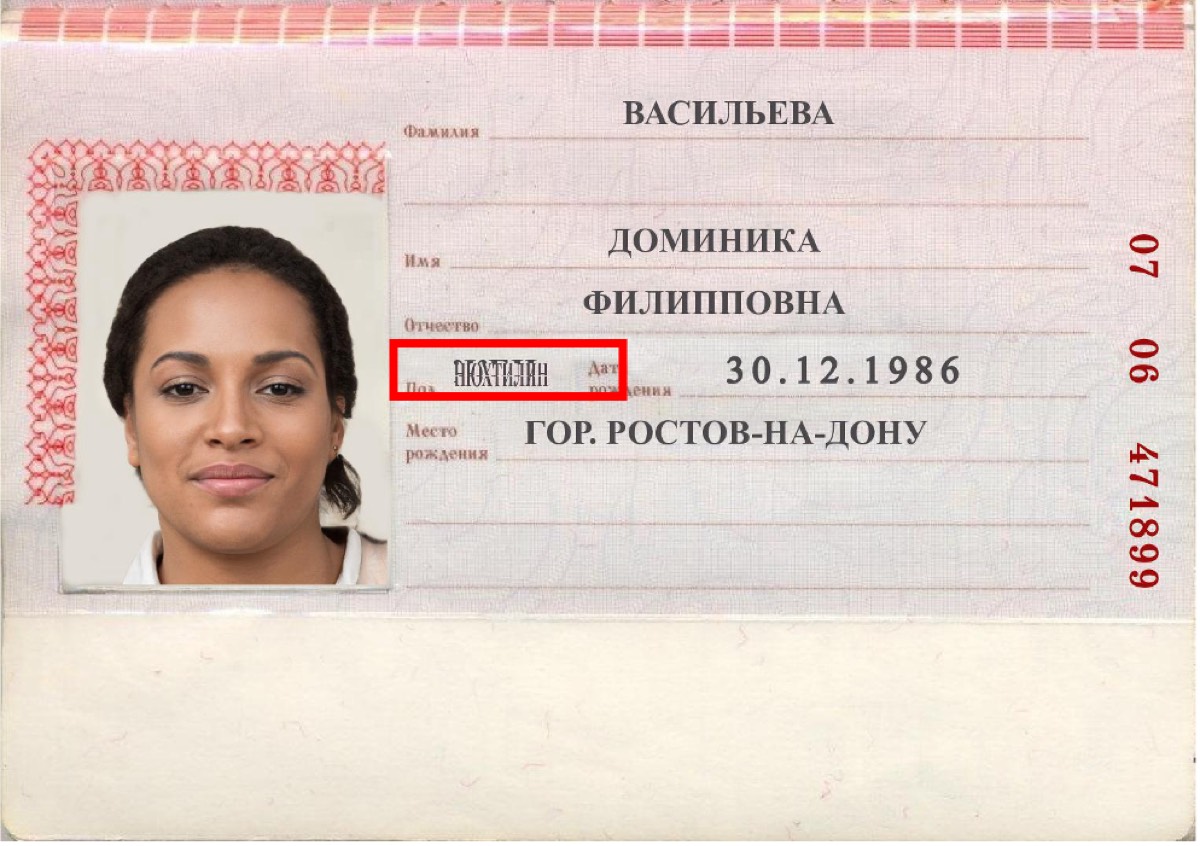}
    }
        \subfigure[\tiny Templated sample from \IDSpace]{
        \includegraphics[width=0.3\textwidth]{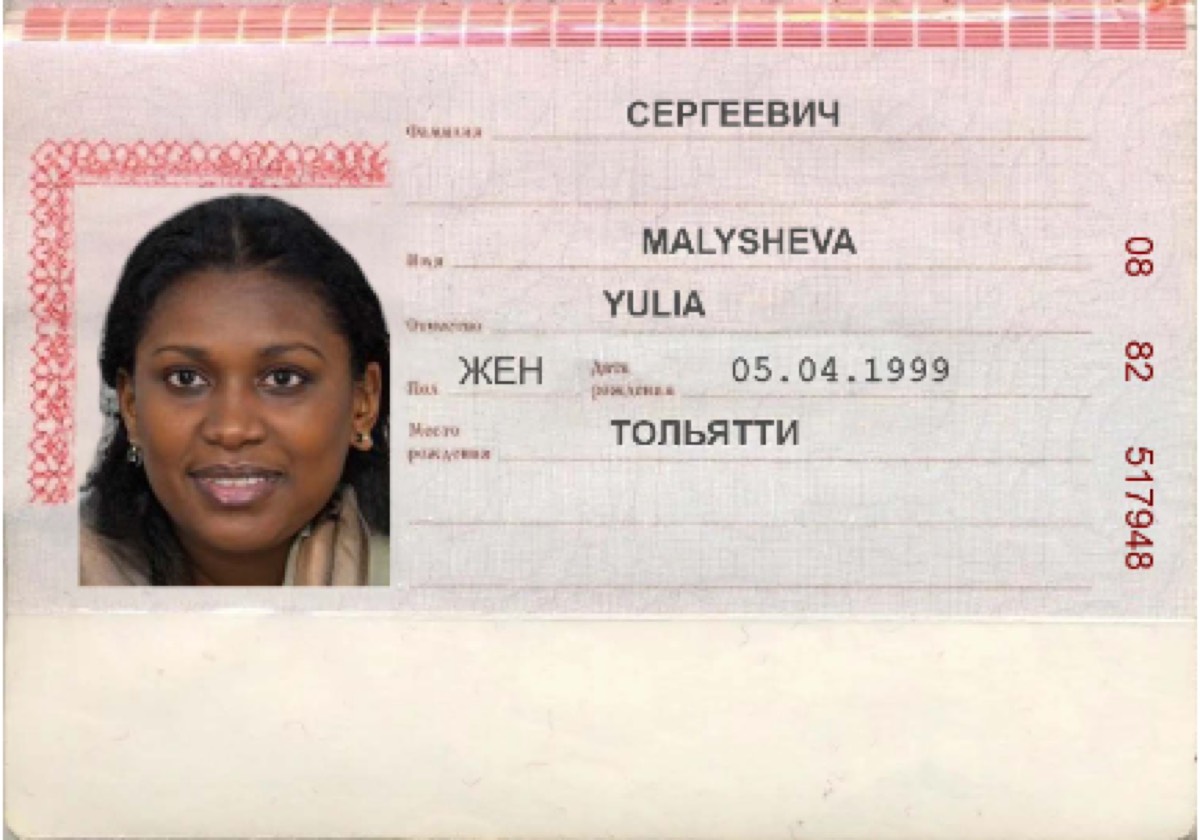}
    }
    \subfigure[\tiny Inpaint\&Rewrite fraud sample from \IDSpace]{
        \includegraphics[width=0.3\textwidth]{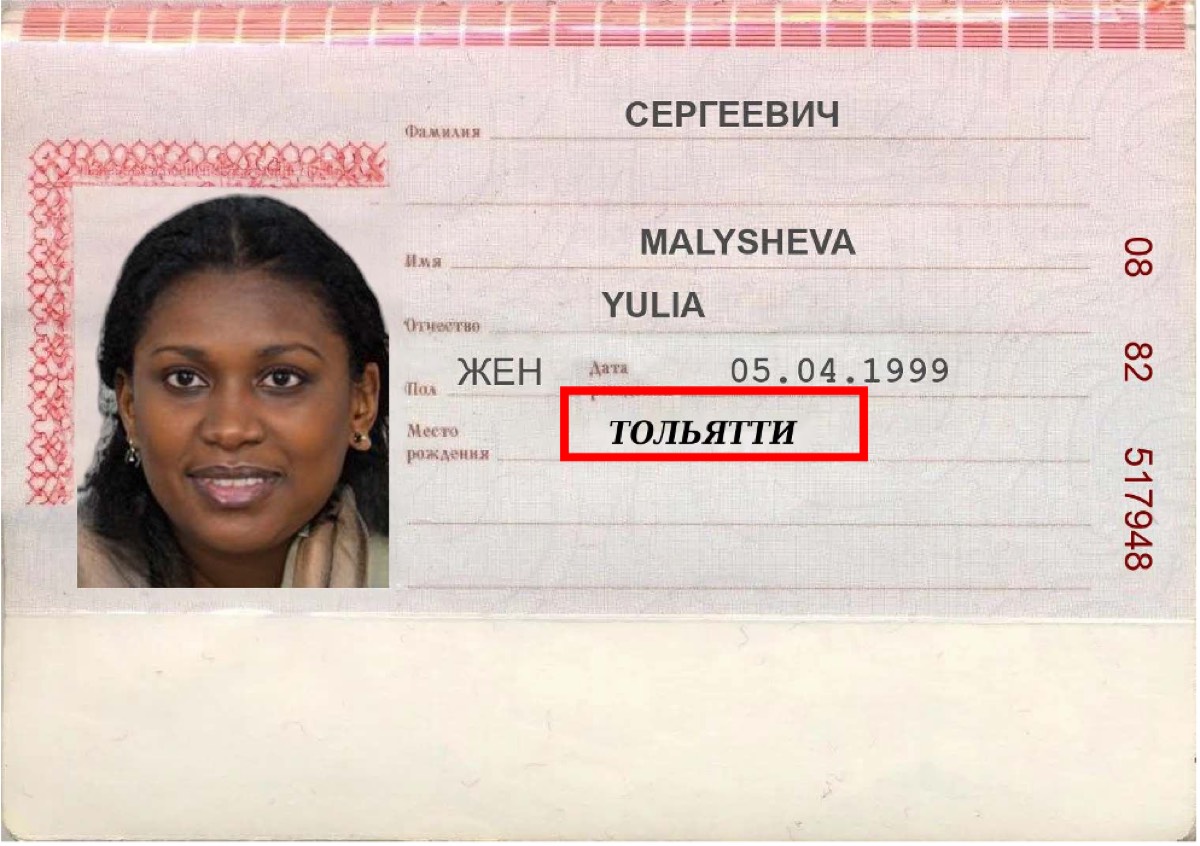}
    }
    \subfigure[\tiny Crop\&Replace fraud sample from \IDSpace]{
        \includegraphics[width=0.3\textwidth]{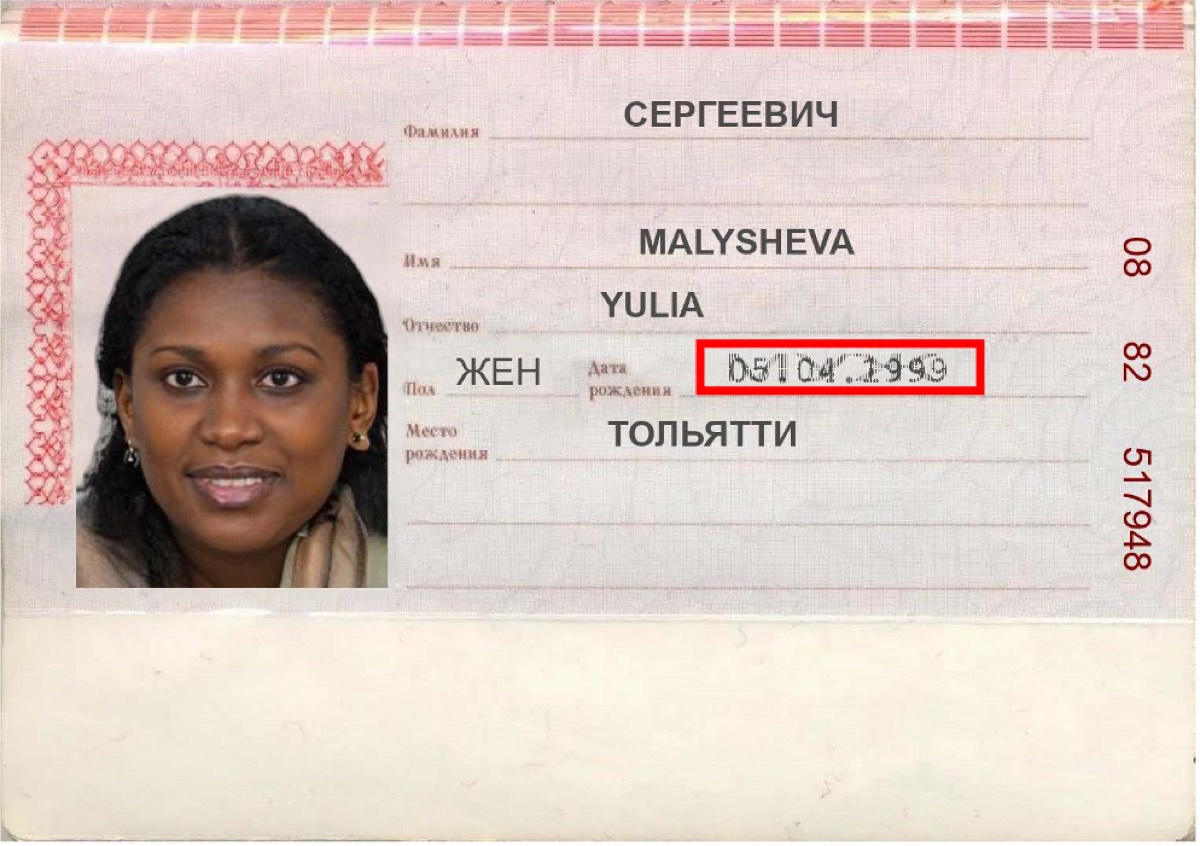}
    }

        \subfigure[\tiny Scanned sample from MIDV]{
        \includegraphics[width=0.22\textwidth]{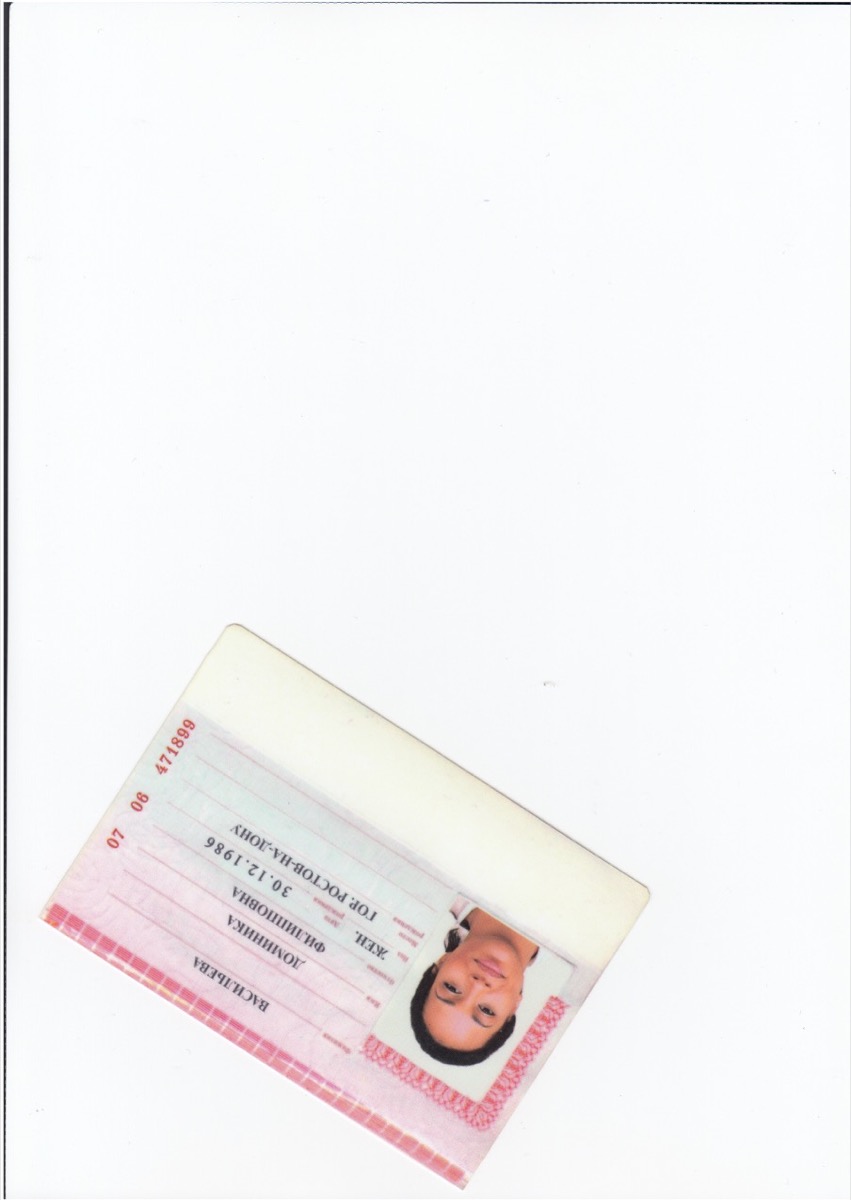}
    }
    \subfigure[\tiny Scanned sample from \IDSpace]{
        \includegraphics[width=0.22\textwidth]{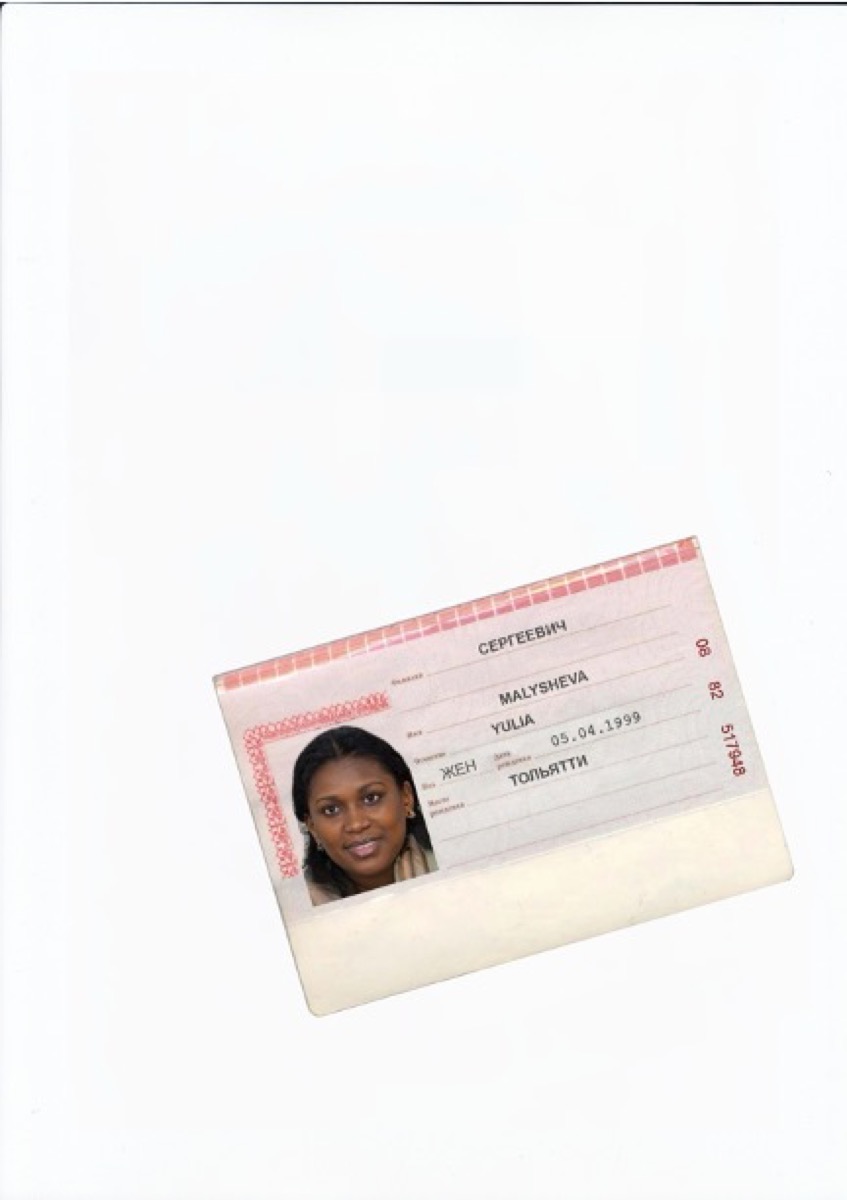}
    }
    \subfigure[\tiny Scanned sample with Inpaint\&Rewrite fraud from \IDSpace]{
        \includegraphics[width=0.22\textwidth]{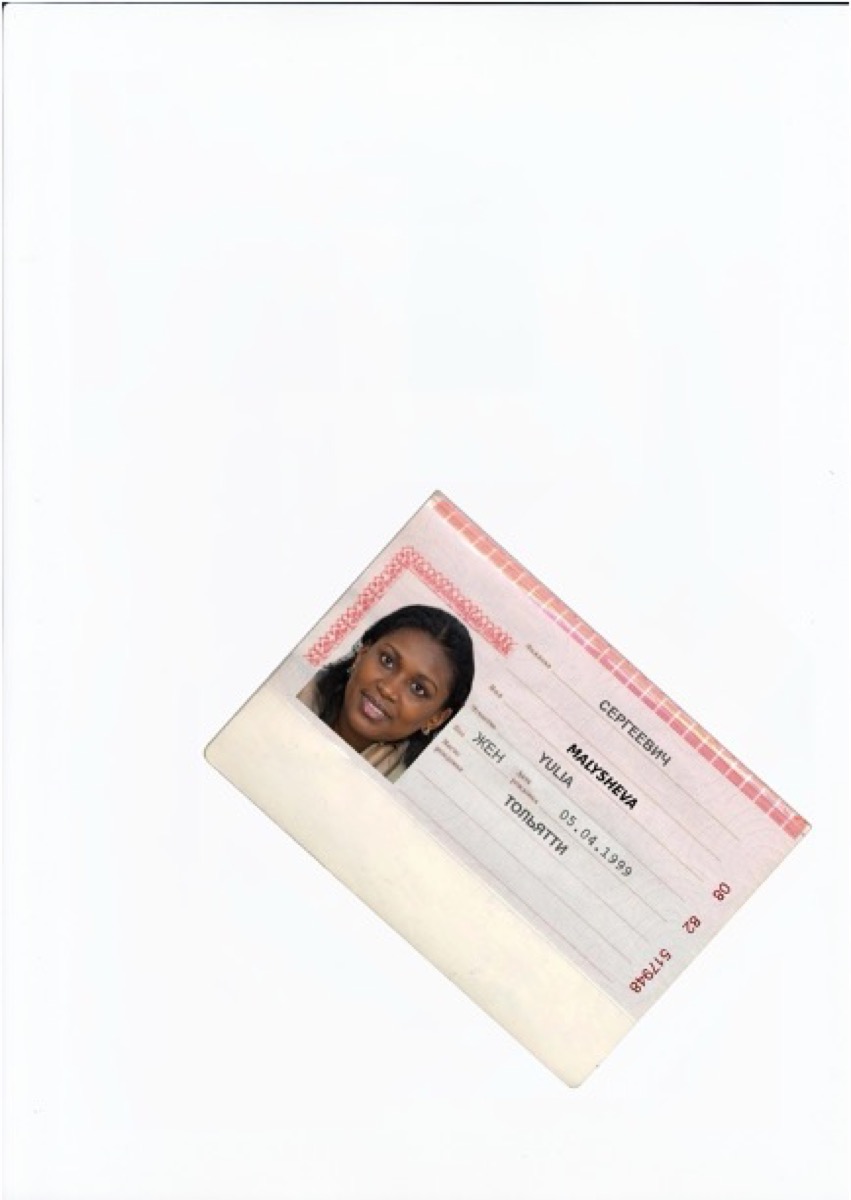}
    }
    \subfigure[\tiny Scanned sample with Crop\&Replace fraud from \IDSpace]{
        \includegraphics[width=0.22\textwidth]{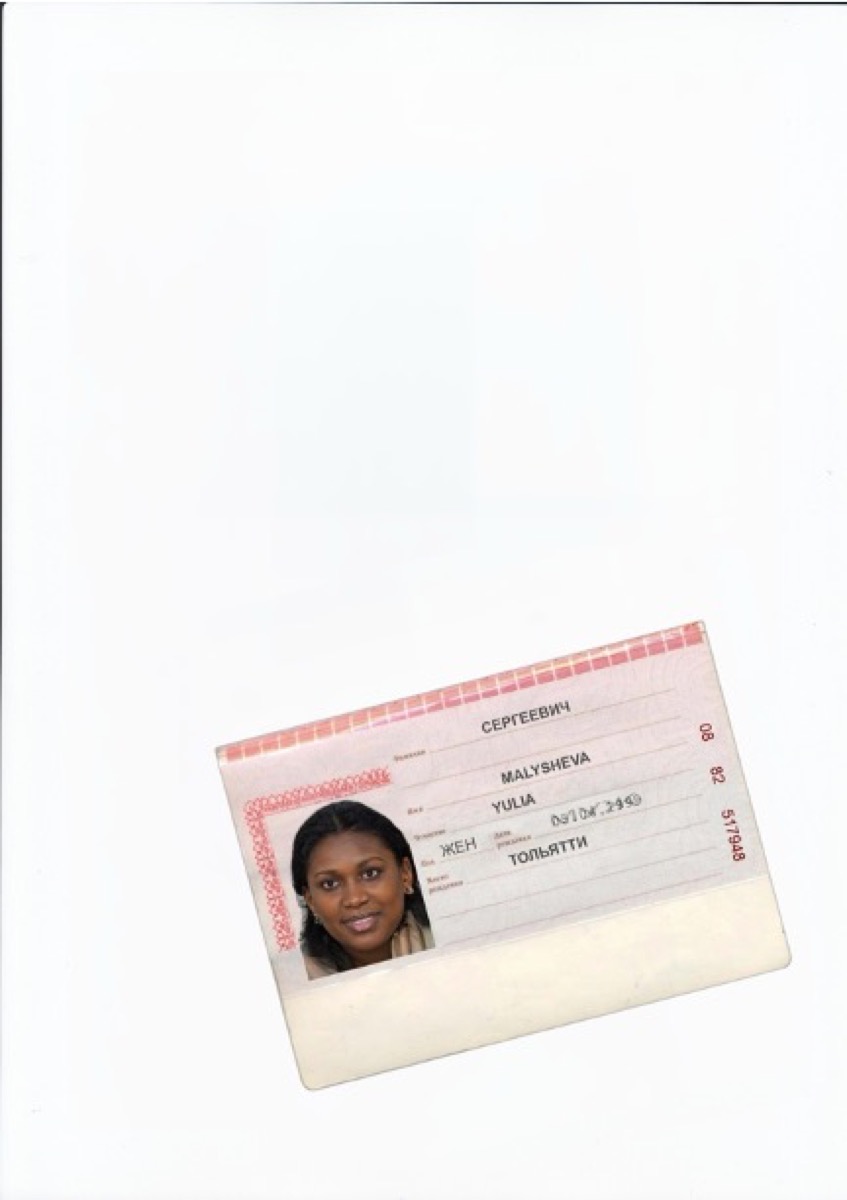}
    }
    \caption{Examples from multiple datasets containing Russian passport images.}
    \label{fig:rus_samples}
\end{figure*}

\begin{figure*}[h!]
    \centering
    \subfigure[\tiny Templated sample from MIDV]{
        \includegraphics[width=0.3\textwidth]{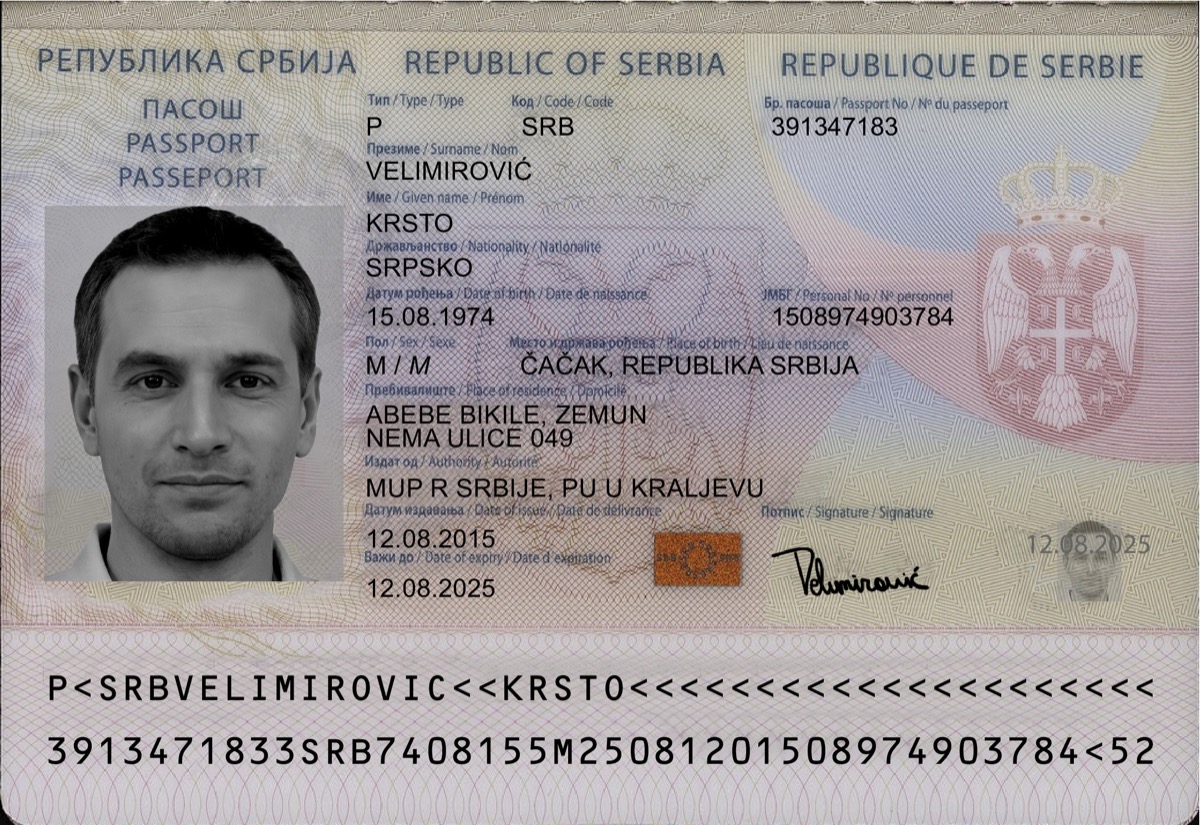}
    }
    \subfigure[\tiny Inpaint\&Rewrite fraud sample from SIDTD]{
        \includegraphics[width=0.3\textwidth]{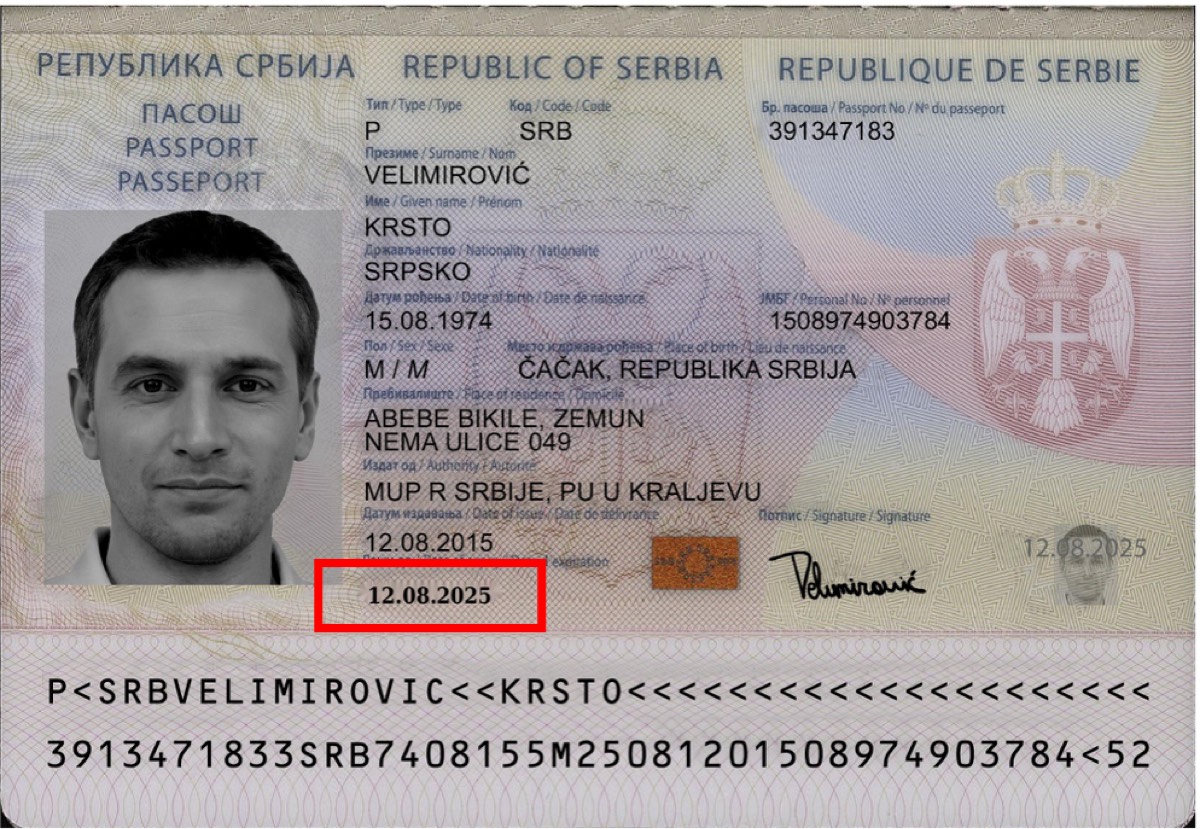}
    }
    \subfigure[\tiny Crop\&Replace fraud sample from SIDTD]{
        \includegraphics[width=0.3\textwidth]{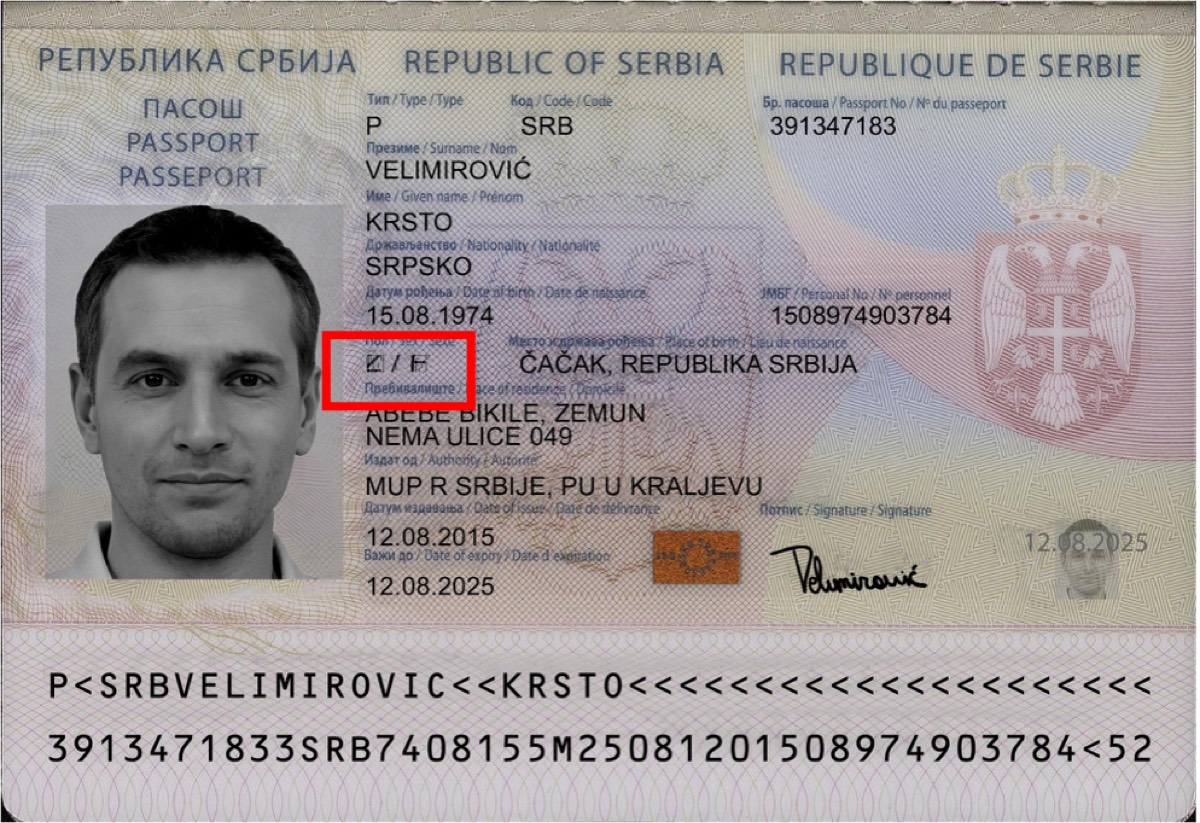}
    }
        \subfigure[\tiny Templated sample from \IDSpace]{
        \includegraphics[width=0.3\textwidth]{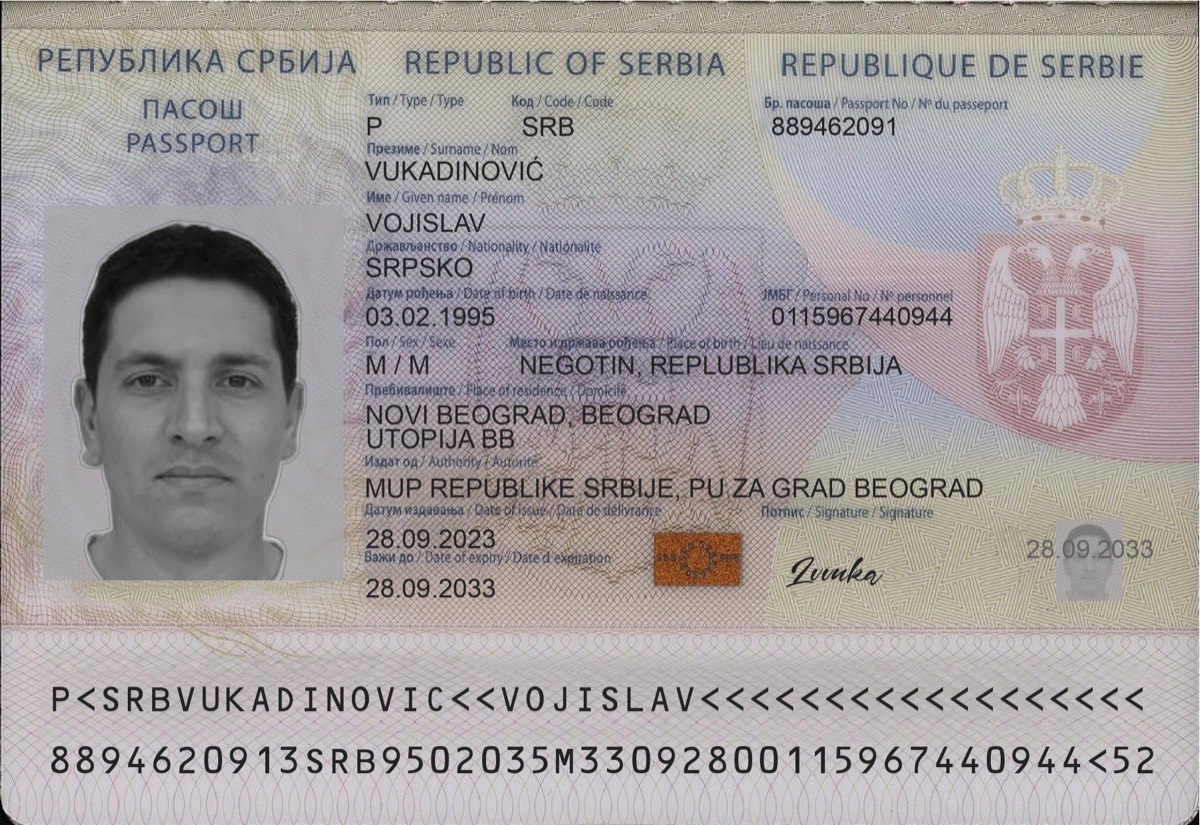}
    }
    \subfigure[\tiny Inpaint\&Rewrite fraud sample from \IDSpace]{
        \includegraphics[width=0.3\textwidth]{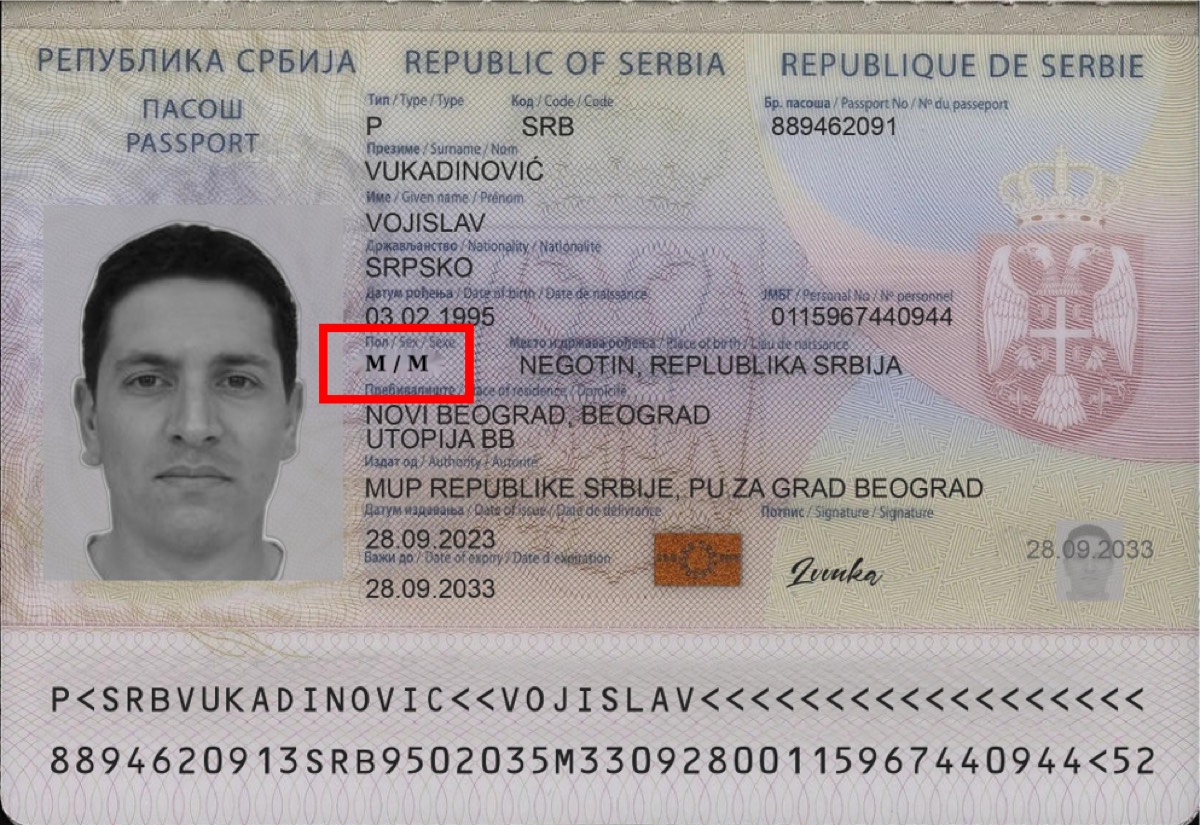}
    }
    \subfigure[\tiny Crop\&Replace fraud sample from \IDSpace]{
        \includegraphics[width=0.3\textwidth]{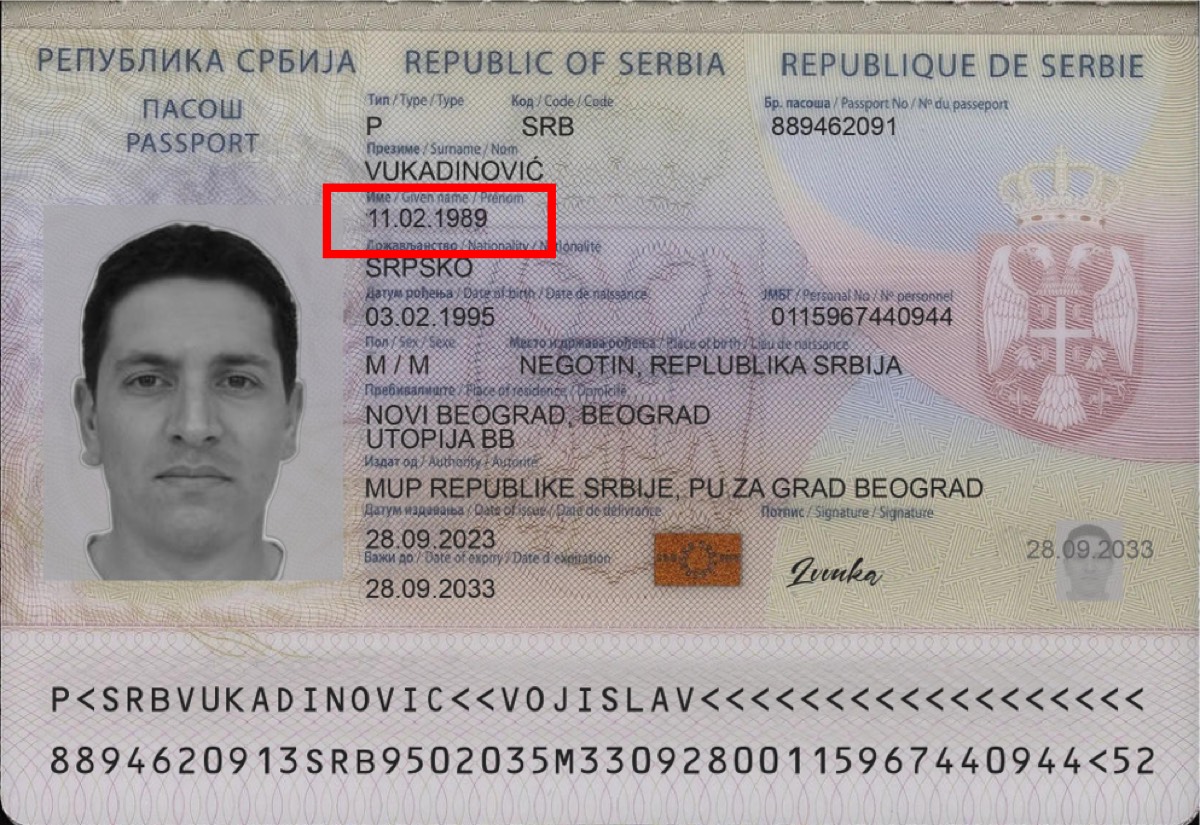}
    }

        \subfigure[\tiny Scanned sample from MIDV]{
        \includegraphics[width=0.22\textwidth]{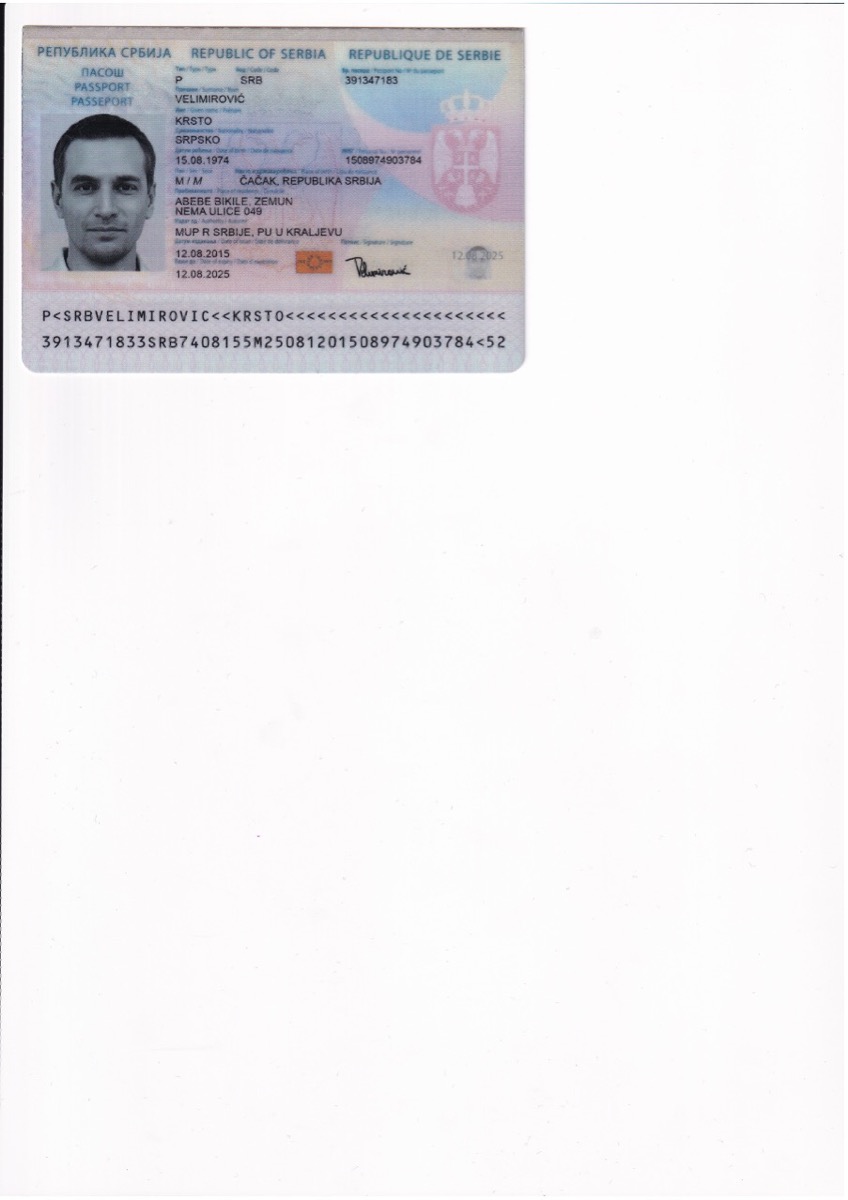}
    }
    \subfigure[\tiny Scanned sample from \IDSpace]{
        \includegraphics[width=0.22\textwidth]{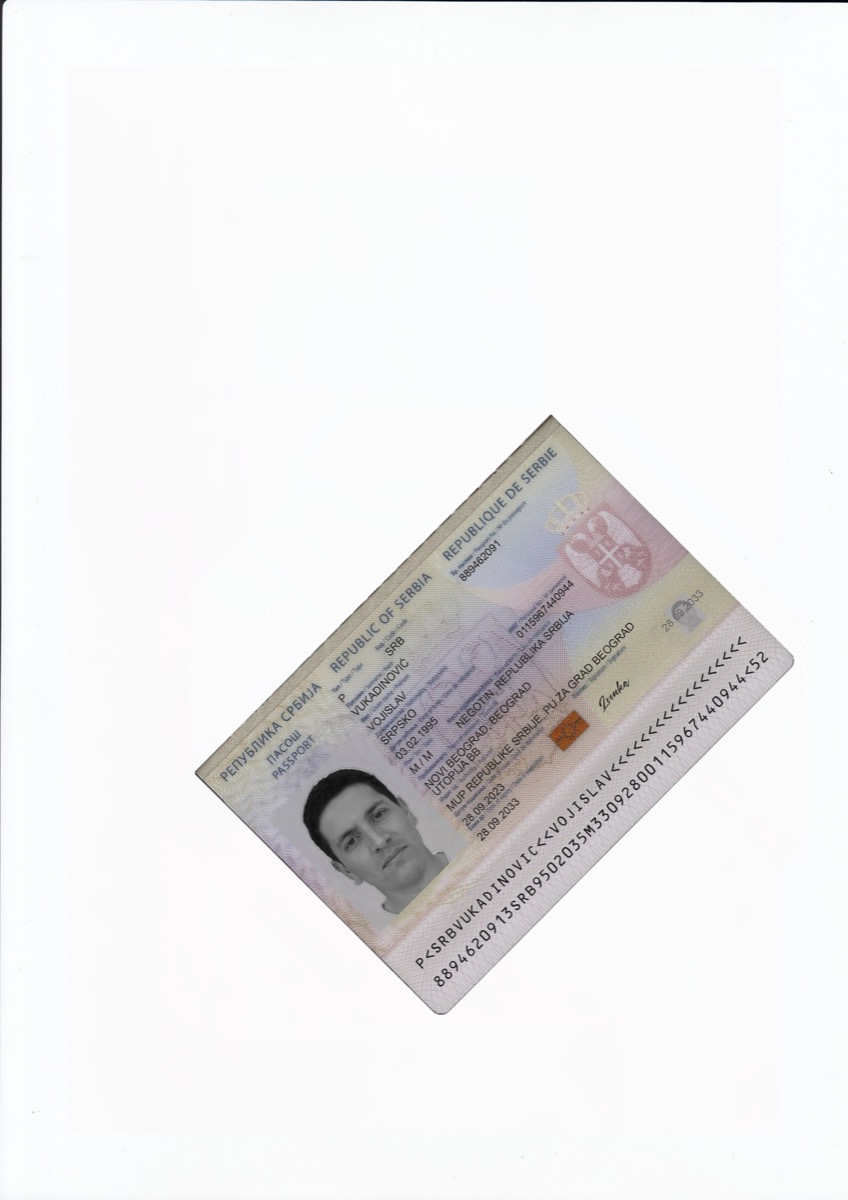}
    }
    \subfigure[\tiny Scanned sample with Inpaint\&Rewrite fraud from \IDSpace]{
        \includegraphics[width=0.22\textwidth]{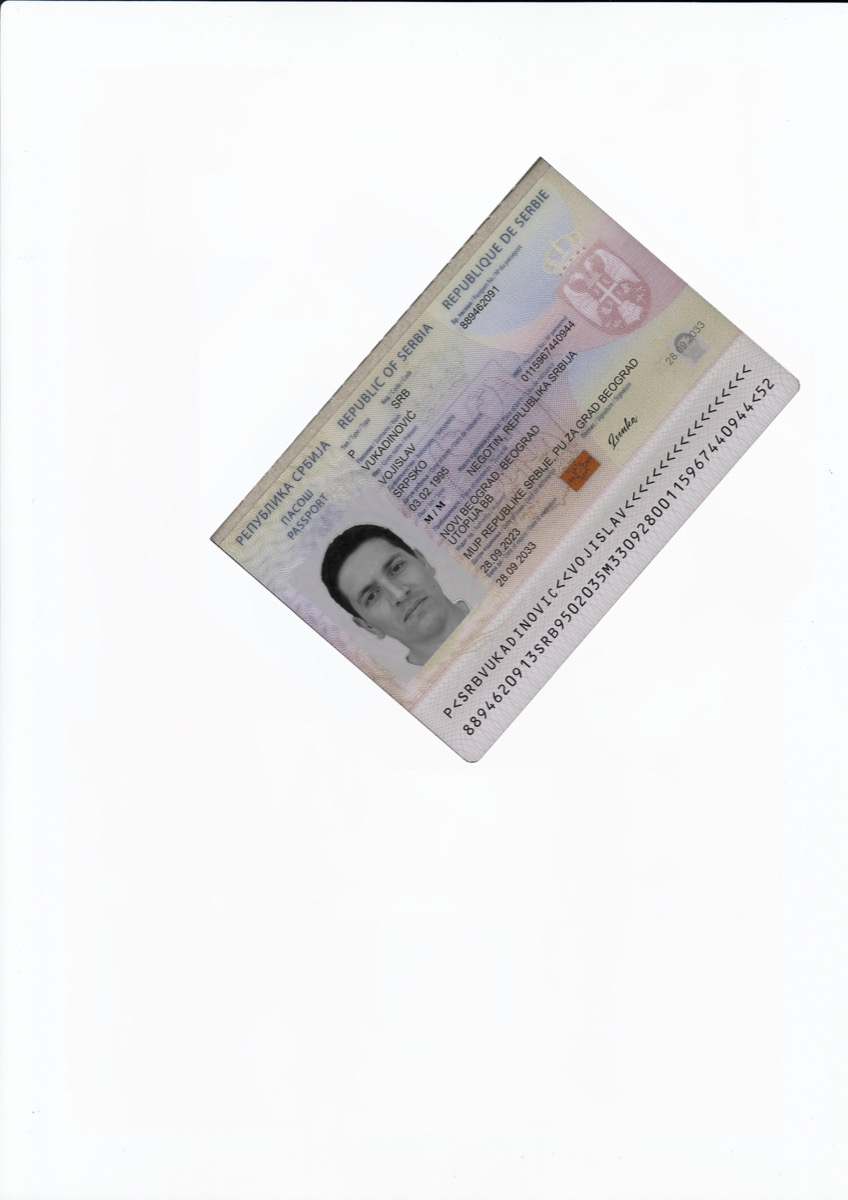}
    }
    \subfigure[\tiny Scanned sample with Crop\&Replace fraud from \IDSpace]{
        \includegraphics[width=0.22\textwidth]{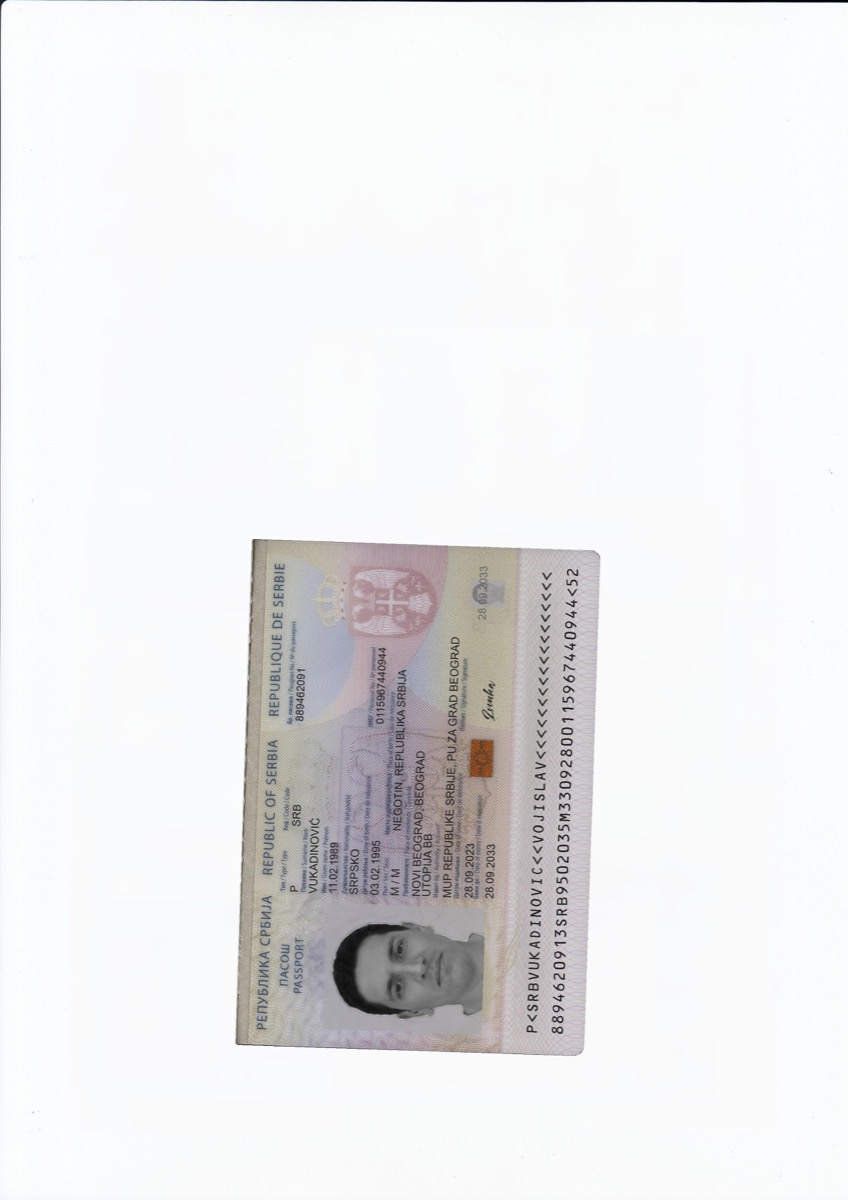}
    }
    \caption{Examples from multiple datasets containing Serbian passport images.}
    \label{fig:srb_samples}
\end{figure*}

\begin{figure*}[h!]
    \centering
    \subfigure[\tiny Templated sample from MIDV]{
        \includegraphics[width=0.3\textwidth]{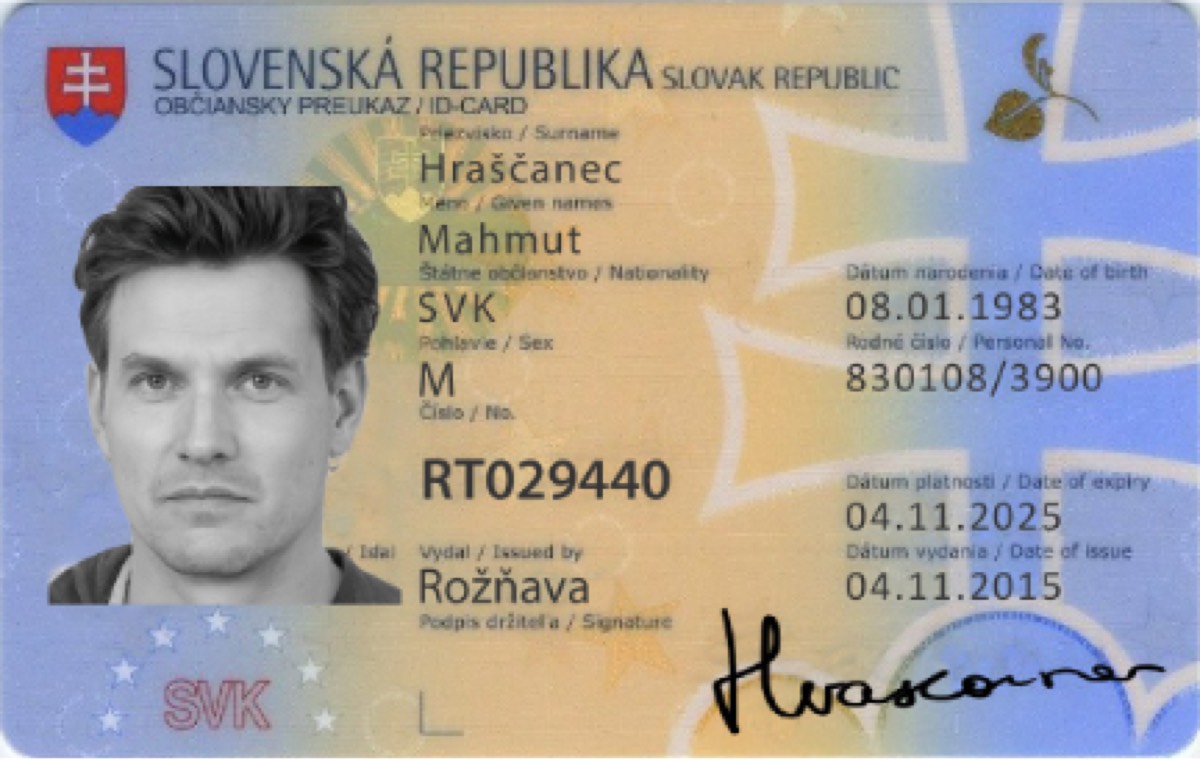}
    }
    \subfigure[\tiny Inpaint\&Rewrite fraud sample from SIDTD]{
        \includegraphics[width=0.3\textwidth]{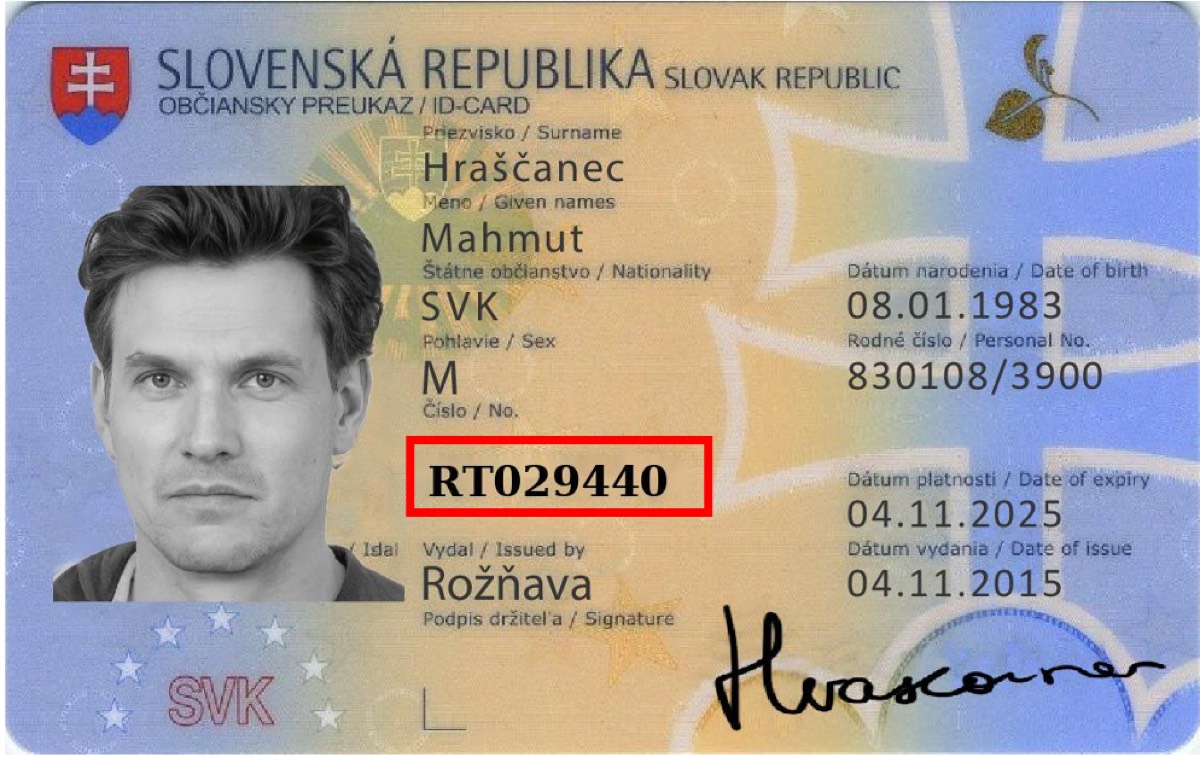}
    }
    \subfigure[\tiny Crop\&Replace fraud sample from SIDTD]{
        \includegraphics[width=0.3\textwidth]{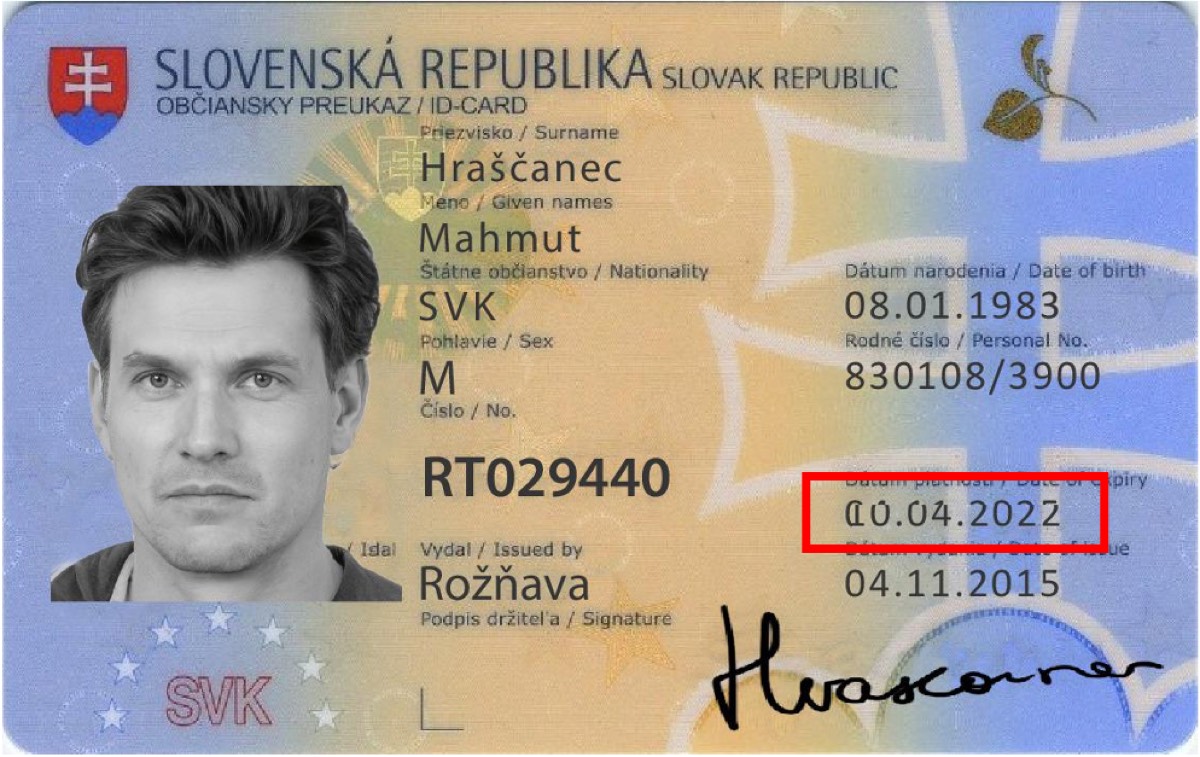}
    }
        \subfigure[\tiny Templated sample from \IDSpace]{
        \includegraphics[width=0.3\textwidth]{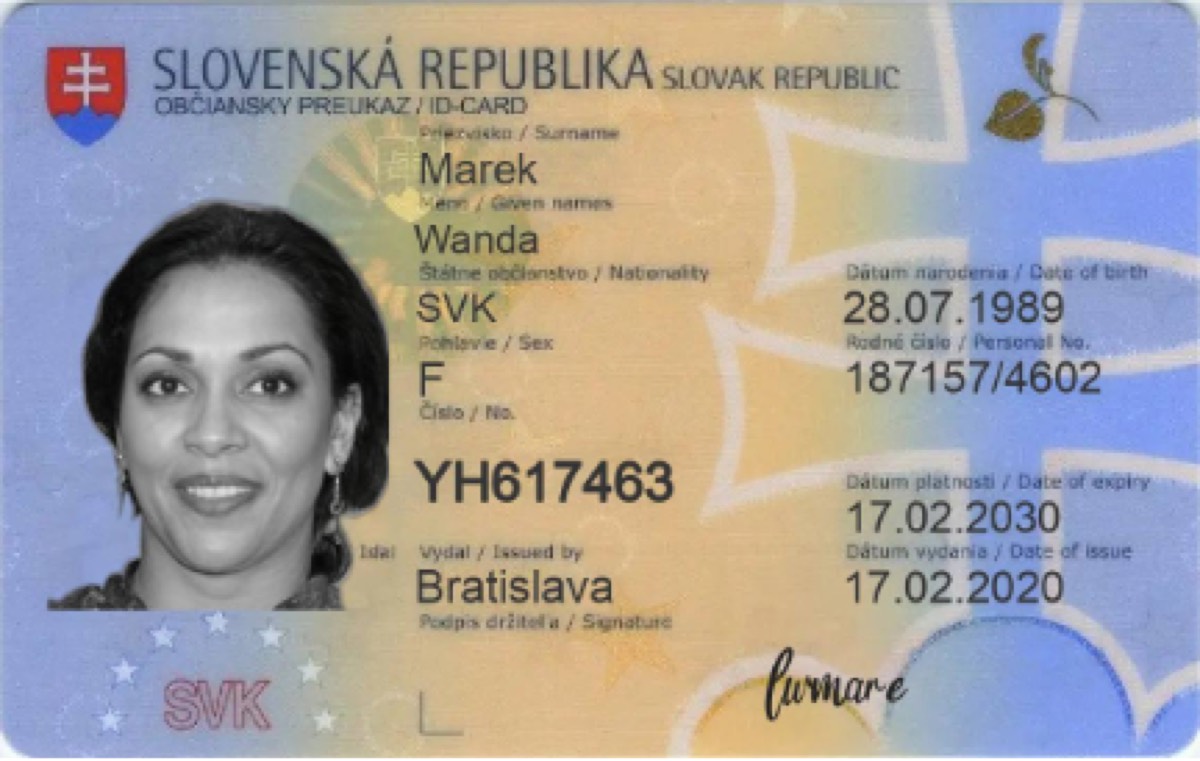}
    }
    \subfigure[\tiny Inpaint\&Rewrite fraud sample from \IDSpace]{
        \includegraphics[width=0.3\textwidth]{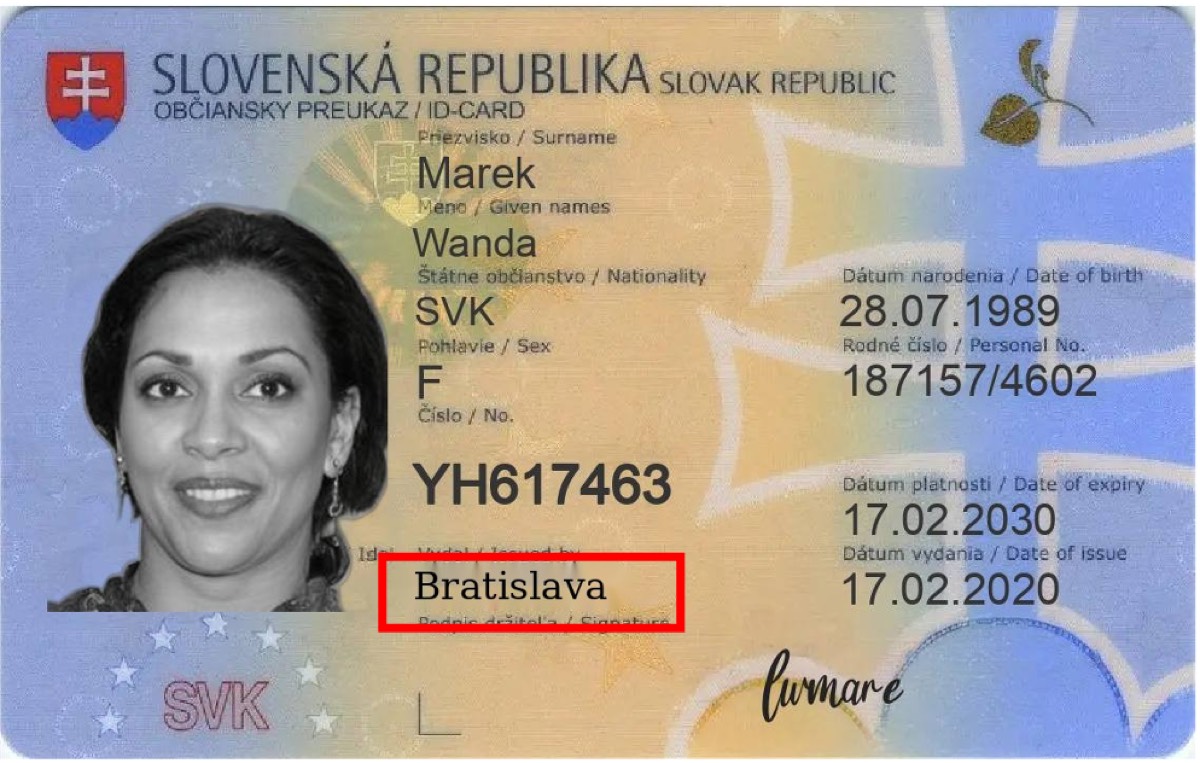}
    }
    \subfigure[\tiny Crop\&Replace fraud sample from \IDSpace]{
        \includegraphics[width=0.3\textwidth]{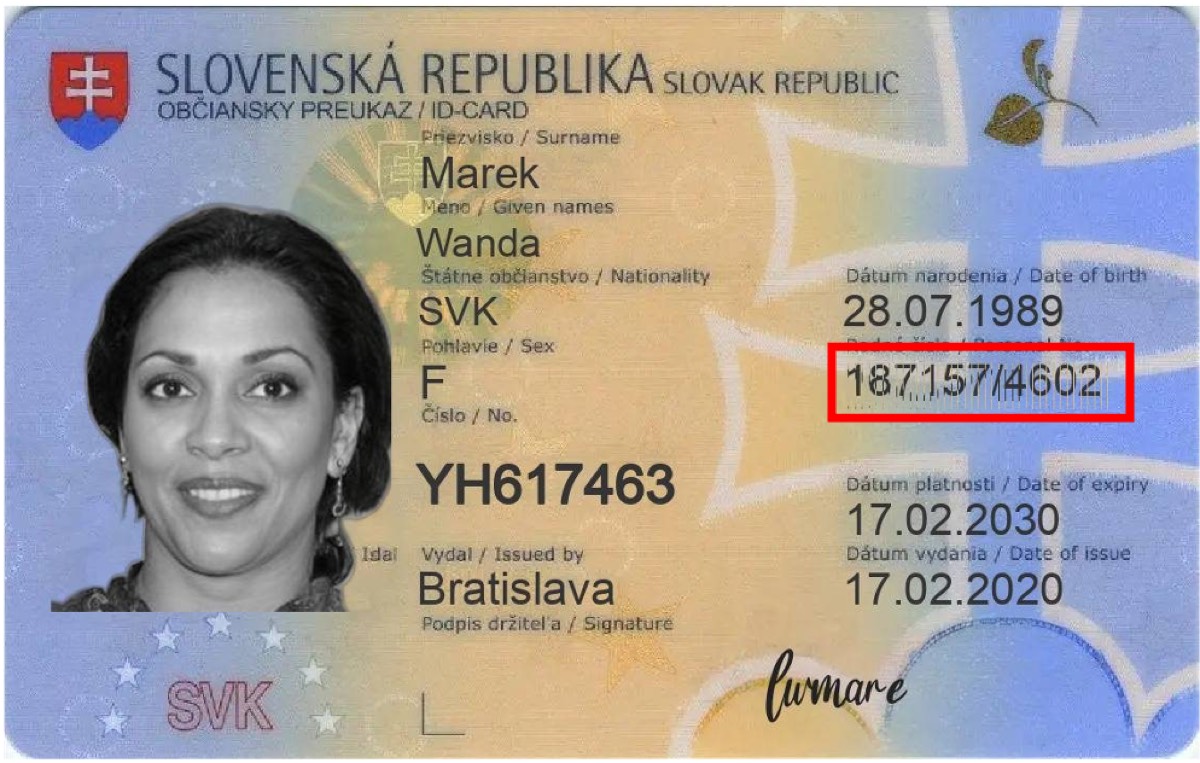}
    }

        \subfigure[\tiny Scanned sample from MIDV]{
        \includegraphics[width=0.22\textwidth]{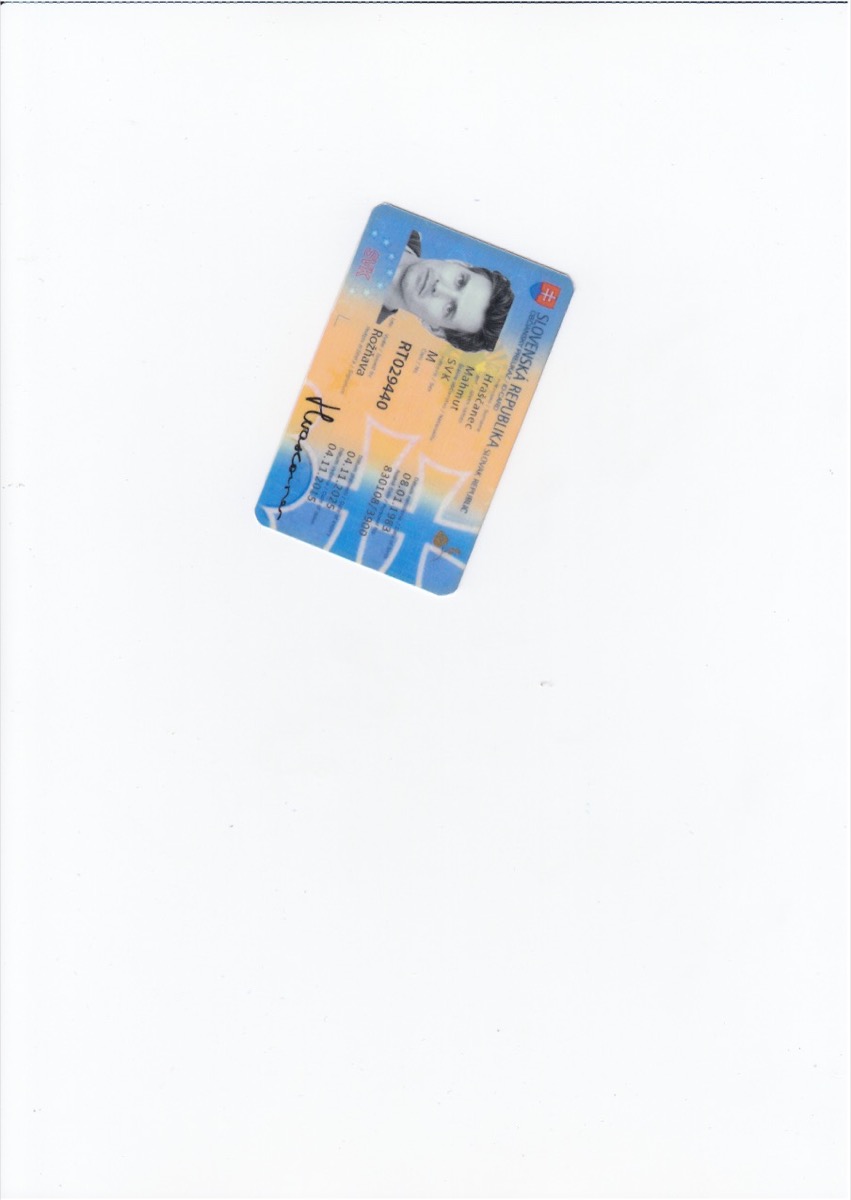}
    }
    \subfigure[\tiny Scanned sample from \IDSpace]{
        \includegraphics[width=0.22\textwidth]{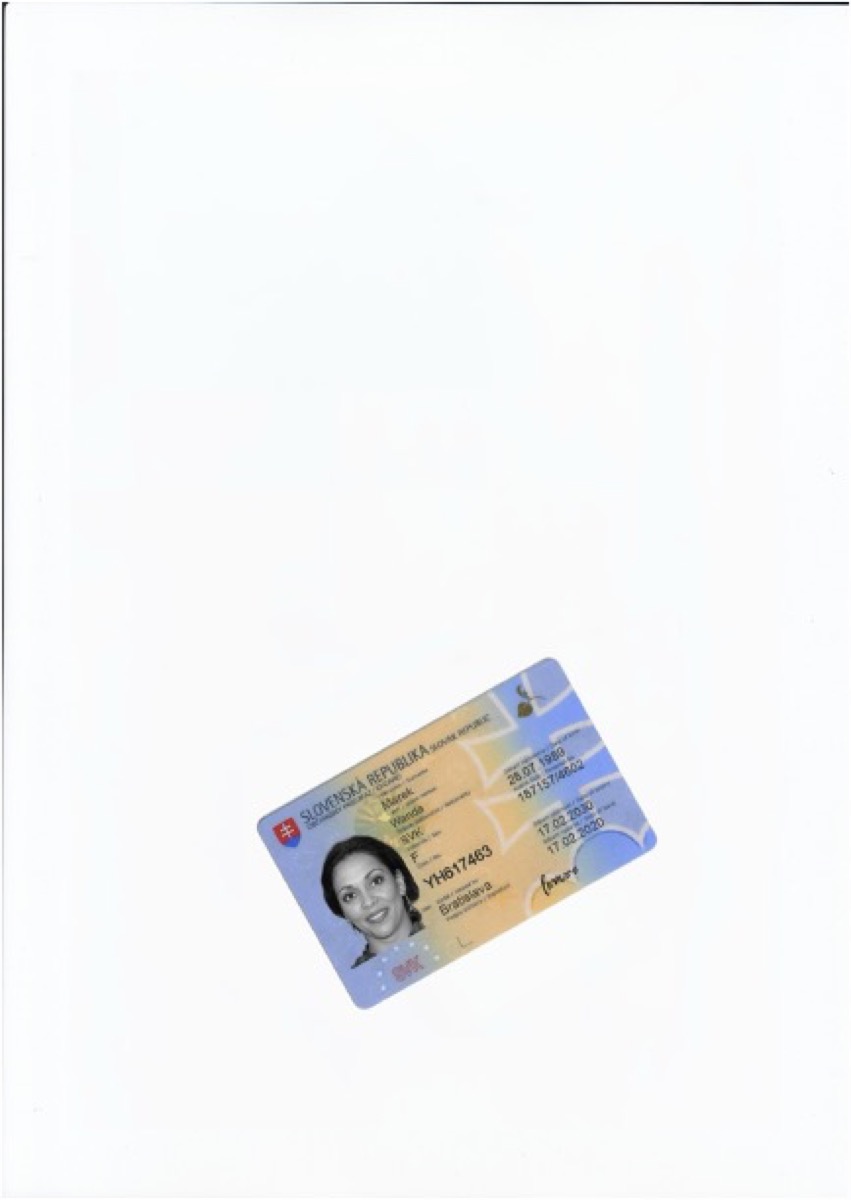}
    }
    \subfigure[\tiny Scanned sample with Inpaint\&Rewrite fraud from \IDSpace]{
        \includegraphics[width=0.22\textwidth]{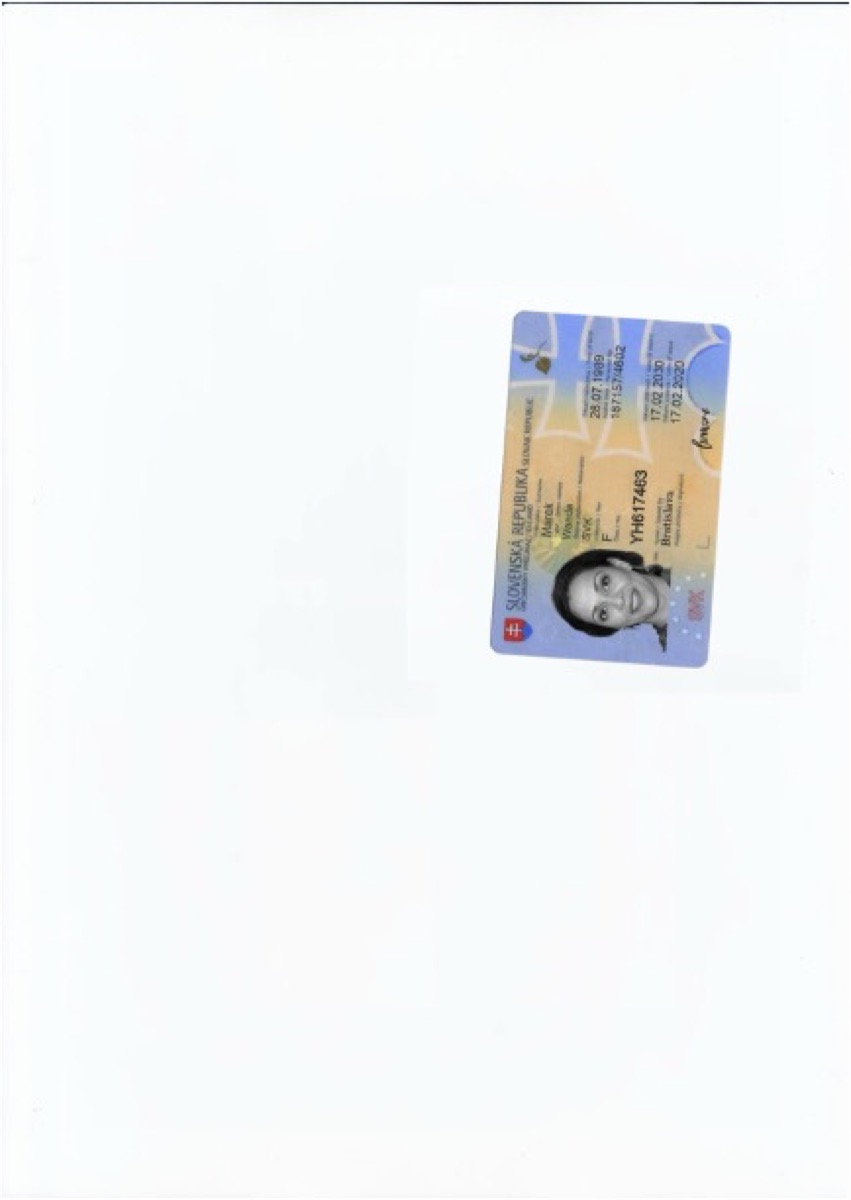}
    }
    \subfigure[\tiny Scanned sample with Crop\&Replace fraud from \IDSpace]{
        \includegraphics[width=0.22\textwidth]{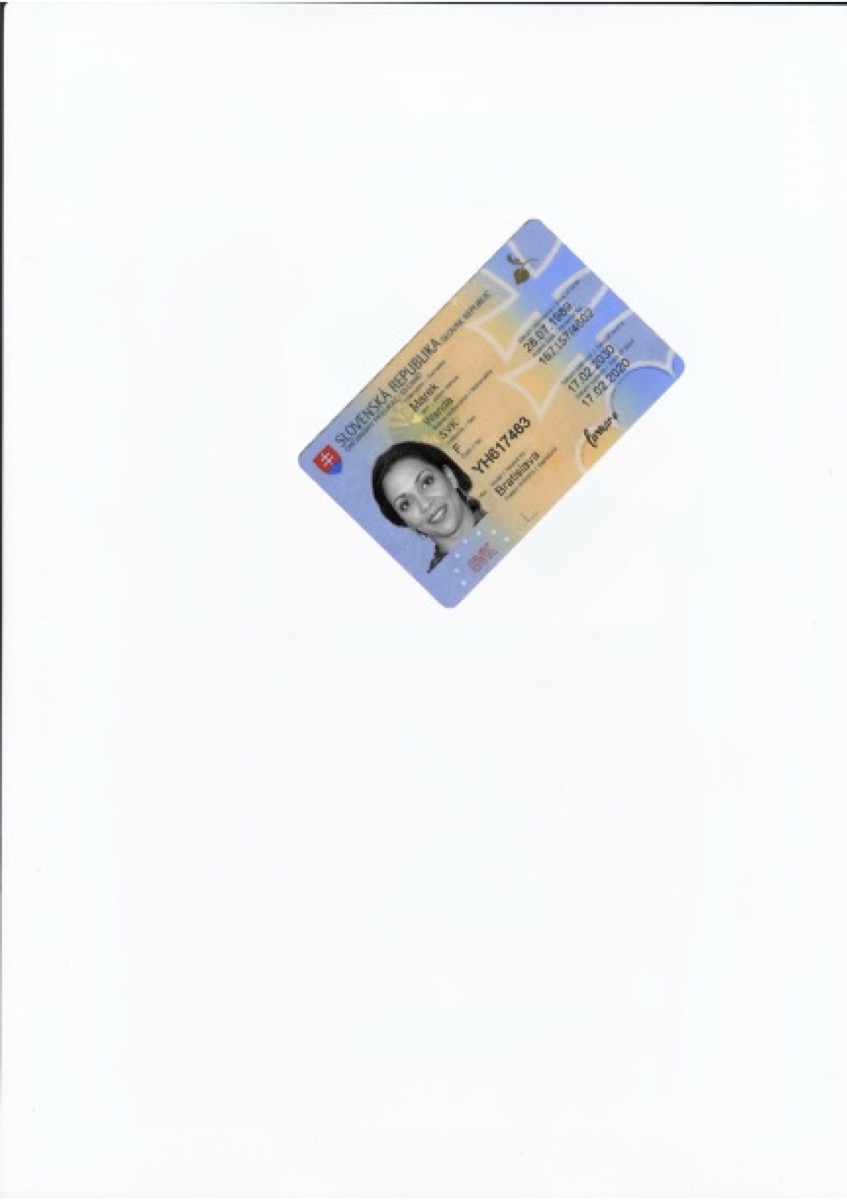}
    }
    \caption{Examples from multiple datasets containing Slovakian ID card images.}
    \label{fig:svk_samples}
\end{figure*}

Figures~\ref{fig:alb_samples} to  
\ref{fig:svk_samples} illustrate sample identity documents from datasets corresponding to 10 countries: Albania (ALB), Azerbaijan (AZE), Spain (ESP), Estonia (EST), Finland (FIN), Greece (GRC), Latvia (LVA), Russia (RUS), Serbia (SRB), and Slovakia (SVK).
The subfigures (a) through (j) in each of the three figures depict the following aspects, and the fraudulent regions are highlighted with red bounding boxes in subfigures (b), (c), (e) and (f):

\begin{itemize}[leftmargin=0.6cm]
    \item[(a)] Non-fraud template image from the MIDV dataset.
    \item[(b)] \textit{Inpaint-and-rewrite} fraud sample based on the MIDV image created by SIDTD dataset.
    \item[(c)] \textit{Crop-and-replace} fraud sample based on the MIDV image created by SIDTD dataset.
    \item[(d)] Sample from our improved dataset(\IDSpace) generated using SSIM combined with ResNet50 model guidance as the BO objective.
    \item[(e)] \textit{Inpaint-and-rewrite} fraud sample using a template generated by \IDSpace.
    \item[(f)] \textit{Crop-and-replace} fraud sample using a template generated by \IDSpace.
    \item[(g)] Scanned image from the MIDV dataset.
    \item[(h)] Simulated scanned version of the samples in MIDV dataset, replicating the scanning artifacts of (d).
    \item[(i)] Simulated scanned version of the samples in MIDV dataset, replicating the scanning artifacts of (e).
    \item[(j)] Simulated scanned version of the samples in MIDV dataset, replicating the scanning artifacts of (f).
\end{itemize}

These examples demonstrate the diversity and realism of our generated dataset.

\subsection{Consistency Evaluation Results for More Types of Customized Template Documents.}

We applied our Bayesian Optimization (BO) search-based approach using different combinations of guiding models to generate customized template documents of other countries, such as Finland, with the same experimental settings introduced in Section \ref{sec:experiment}. 
The comparison results are shown in Table~\ref{tab:cross_model_fin}. Similar to the results presented in Table~\ref{tab:cross_model}, we observe a significant improvement in consistency scores when using our proposed model-guided BO methodology, compared to the method that uses SSIM as the sole objective (first row in the table). Specifically, our model-guided BO approach achieves relative improvements in consistency scores ranging from $22.00\%$ to $34.86\%$ over the SSIM-only baseline.

\begin{table*}[t]
\centering
\caption{\small Model prediction consistency evaluation on different models (FIN)}
\label{tab:cross_model_fin}
\resizebox{\textwidth}{!}{%
\begin{tabular}{l|c|c|c|c|c|c}
\hline
\textbf{Guiding models} & \textbf{ViT-Large} & \textbf{ResNet50} & \textbf{Inception-v3} & \textbf{VGG16} & \textbf{DenseNet} & \textbf{Average} \\
\hline
SSIM-only objective & $0.5143 \pm 0.007$ & $0.5238 \pm 0.040$ & $0.5667 \pm 0.015$ & $0.7714 \pm 0.004$ & $0.5952 \pm 0.050$ & $0.5943 \pm 0.009$ \\
DenseNet & $0.8667 \pm 0.002$ & $0.8762 \pm 0.007$ & $0.8571 \pm 0.003$ & $0.8714 \pm 0.000$ & $0.9905 \pm 0.000$ & $0.8924 \pm 0.002$ \\
DenseNet + Inception-v3 & $0.9000 \pm 0.003$ & $0.8762 \pm 0.003$ & $0.8905 \pm 0.005$ & $0.8667 \pm 0.001$ & $0.9857 \pm 0.000$ & $0.9038 \pm 0.002$ \\
DenseNet + Inception-v3 + ResNet50 & $0.9333 \pm 0.001$ & $0.8905 \pm 0.003$ & $0.8714 \pm 0.001$ & $0.8857 \pm 0.000$ & $0.9952 \pm 0.000$ & $0.9152 \pm 0.002$ \\
DenseNet + Inception-v3 + VGG16 & $0.9190 \pm 0.002$ & $0.9048 \pm 0.001$ & $0.9190 \pm 0.002$ & $0.8905 \pm 0.000$ & $1.0000 \pm 0.000$ & $0.9267 \pm 0.001$ \\
DenseNet + Inception-v3 + ViT-Large & $0.9143 \pm 0.002$ & $0.9048 \pm 0.002$ & $0.9000 \pm 0.003$ & $0.8810 \pm 0.001$ & $0.9667 \pm 0.002$ & $0.9133 \pm 0.001$ \\
DenseNet + ResNet50 & $0.7905 \pm 0.002$ & $0.9190 \pm 0.002$ & $0.8238 \pm 0.003$ & $0.8810 \pm 0.001$ & $0.9952 \pm 0.000$ & $0.8819 \pm 0.005$ \\
DenseNet + ResNet50 + VGG16 & $0.8952 \pm 0.004$ & $0.8952 \pm 0.001$ & $0.8476 \pm 0.005$ & $0.8762 \pm 0.000$ & $0.9524 \pm 0.003$ & $0.8933 \pm 0.001$ \\
DenseNet + ResNet50 + ViT-Large & $0.8476 \pm 0.006$ & $0.8905 \pm 0.004$ & $0.8095 \pm 0.011$ & $0.8905 \pm 0.001$ & $0.9190 \pm 0.003$ & $0.8714 \pm 0.001$ \\
DenseNet + VGG16 & $0.8381 \pm 0.002$ & $0.8714 \pm 0.002$ & $0.8000 \pm 0.004$ & $0.8619 \pm 0.001$ & $0.9714 \pm 0.001$ & $0.8686 \pm 0.003$ \\
DenseNet + VGG16 + ViT-Large & \textcolor{black}{$0.9667 \pm 0.000$} & $0.9238 \pm 0.000$ & $0.9143 \pm 0.001$ & $0.8952 \pm 0.000$ & $0.9143 \pm 0.000$ & $0.9229 \pm 0.001$ \\
DenseNet + ViT-Large & $0.8762 \pm 0.003$ & $0.8952 \pm 0.002$ & $0.7952 \pm 0.008$ & $0.8476 \pm 0.001$ & $0.9571 \pm 0.001$ & $0.8743 \pm 0.003$ \\
Inception-v3 & $0.9048 \pm 0.002$ & $0.8619 \pm 0.004$ & $0.8619 \pm 0.006$ & $0.8667 \pm 0.000$ & $0.9810 \pm 0.000$ & $0.8952 \pm 0.002$ \\
Inception-v3 + ResNet50 & $0.9571 \pm 0.000$ & $0.9095 \pm 0.000$ & \textcolor{black}{$0.9571 \pm 0.001$} & $0.8905 \pm 0.000$ & \textcolor{black}{$1.0000 \pm 0.000$} & \textcolor{black}{$0.9429 \pm 0.002$} \\
Inception-v3 + ResNet50 + VGG16 & $0.8095 \pm 0.006$ & $0.9190 \pm 0.001$ & $0.8238 \pm 0.002$ & $0.8714 \pm 0.000$ & $0.8762 \pm 0.008$ & $0.8600 \pm 0.002$ \\
Inception-v3 + ResNet50 + ViT-Large & $0.9000 \pm 0.004$ & $0.8714 \pm 0.001$ & $0.8810 \pm 0.001$ & $0.8667 \pm 0.001$ & $0.9762 \pm 0.001$ & $0.8990 \pm 0.002$ \\
Inception-v3 + VGG16 & $0.9571 \pm 0.000$ & \textcolor{black}{$0.9667 \pm 0.000$} & $0.9381 \pm 0.000$ & \textcolor{black}{$0.9095 \pm 0.000$} & $0.8952 \pm 0.000$ & $0.9333 \pm 0.001$ \\
Inception-v3 + VGG16 + ViT-Large & $0.9286 \pm 0.003$ & $0.9190 \pm 0.001$ & $0.9238 \pm 0.002$ & $0.8952 \pm 0.000$ & $0.9238 \pm 0.003$ & $0.9181 \pm 0.000$ \\
Inception-v3 + ViT-Large & $0.9333 \pm 0.001$ & $0.9238 \pm 0.000$ & $0.8571 \pm 0.005$ & $0.8905 \pm 0.000$ & $0.9333 \pm 0.001$ & $0.9076 \pm 0.001$ \\
ResNet50 & $0.8190 \pm 0.002$ & $0.8571 \pm 0.001$ & $0.7048 \pm 0.004$ & $0.8476 \pm 0.000$ & $0.8429 \pm 0.000$ & $0.8143 \pm 0.003$ \\
ResNet50 + VGG16 & $0.8667 \pm 0.001$ & $0.8905 \pm 0.001$ & $0.8286 \pm 0.000$ & $0.8762 \pm 0.000$ & $0.9667 \pm 0.001$ & $0.8857 \pm 0.002$ \\
ResNet50 + VGG16 + ViT-Large & $0.9143 \pm 0.002$ & $0.8619 \pm 0.003$ & $0.8857 \pm 0.005$ & $0.8857 \pm 0.001$ & $0.9857 \pm 0.000$ & $0.9067 \pm 0.002$ \\
ResNet50 + ViT-Large & $0.8905 \pm 0.001$ & $0.9429 \pm 0.001$ & $0.8952 \pm 0.002$ & $0.8857 \pm 0.001$ & $0.8905 \pm 0.003$ & $0.9010 \pm 0.000$ \\
VGG16 & $0.8000 \pm 0.010$ & $0.8810 \pm 0.003$ & $0.7476 \pm 0.007$ & $0.8571 \pm 0.001$ & $0.8238 \pm 0.030$ & $0.8219 \pm 0.002$ \\
VGG16 + ViT-Large & $0.9524 \pm 0.000$ & $0.9238 \pm 0.001$ & $0.9048 \pm 0.003$ & $0.8905 \pm 0.000$ & $0.9571 \pm 0.004$ & $0.9257 \pm 0.001$ \\
ViT-Large & $0.9524 \pm 0.000$ & $0.9429 \pm 0.000$ & $0.9333 \pm 0.000$ & $0.9048 \pm 0.000$ & $0.9000 \pm 0.000$ & $0.9267 \pm 0.000$ \\

\hline
\end{tabular}%
}
\end{table*}

\subsection{Consistency Evaluation Results for Scanned Documents.}

We further evaluated the effectiveness of our synthetic data generation framework by applying the proposed method to scanned document images. 



We followed the same methodology as with the template images: training models on the generated dataset and performing Bayesian Optimization (BO) with and without model guidance to search over the predefined parameter set. We then evaluated the consistency scores using different combinations of guiding models for the BO process with scanned documents in Albania and Finland. The results are presented in Table~\ref{tab:cross_model_scanned_alb} and Table~\ref{tab:cross_model_scanned_fin}, respectively.

From these results, we observe that the consistency scores of the model-guided BO method are generally higher than those obtained on template images. The explanation is that the template region in scanned images typically occupies only around $20\%$ of the entire image. As a result, the SSIM-only objective is not sufficiently sensitive to capture subtle manipulations in these regions, allowing the model-guided approach to dominate in optimizing the objective function.

\begin{table*}[t]
\centering
\caption{\small Model prediction consistency evaluation for scanned documents on different models (ALB)}
\label{tab:cross_model_scanned_alb}
\resizebox{\textwidth}{!}{%
\begin{tabular}{l|c|c|c|c|c|c}
\hline
\textbf{Guiding models} & \textbf{ViT-Large} & \textbf{ResNet50} & \textbf{Inception-v3} & \textbf{VGG16} & \textbf{DenseNet} & \textbf{Average} \\
\hline
SSIM-only objective & $0.4933 \pm 0.000$ & $0.4933 \pm 0.000$ & $0.4933 \pm 0.000$ & $0.4933 \pm 0.000$ & $0.5733 \pm 0.000$ & $0.5093 \pm 0.001$ \\
DenseNet & \textcolor{black}{$1.0000 \pm 0.000$} & $0.9778 \pm 0.000$ & \textcolor{black}{$0.9867 \pm 0.000$} & \textcolor{black}{$1.0000 \pm 0.000$} & $0.9778 \pm 0.000$ & $0.9884 \pm 0.000$ \\
DenseNet + Inception-v3 & \textcolor{black}{$1.0000 \pm 0.000$} & $0.9733 \pm 0.000$ & $0.9822 \pm 0.000$ & $0.9689 \pm 0.002$ & $0.9600 \pm 0.000$ & $0.9769 \pm 0.000$ \\
DenseNet + Inception-v3 + ResNet50 & \textcolor{black}{$1.0000 \pm 0.000$} & $0.9733 \pm 0.000$ & \textcolor{black}{$0.9867 \pm 0.000$} & \textcolor{black}{$1.0000 \pm 0.000$} & $0.9556 \pm 0.001$ & $0.9831 \pm 0.000$ \\
DenseNet + Inception-v3 + VGG16 & \textcolor{black}{$1.0000 \pm 0.000$} & $0.9778 \pm 0.000$ & \textcolor{black}{$0.9867 \pm 0.000$} & \textcolor{black}{$1.0000 \pm 0.000$} & $0.9733 \pm 0.000$ & $0.9876 \pm 0.000$ \\
DenseNet + Inception-v3 + ViT-Large & \textcolor{black}{$1.0000 \pm 0.000$} & $0.9822 \pm 0.000$ & \textcolor{black}{$0.9867 \pm 0.000$} & \textcolor{black}{$1.0000 \pm 0.000$} & $0.9733 \pm 0.000$ & $0.9884 \pm 0.000$ \\
DenseNet + ResNet50 & \textcolor{black}{$1.0000 \pm 0.000$} & $0.9733 \pm 0.000$ & \textcolor{black}{$0.9867 \pm 0.000$} & \textcolor{black}{$1.0000 \pm 0.000$} & $0.9733 \pm 0.000$ & $0.9867 \pm 0.000$ \\
DenseNet + ResNet50 + VGG16 & \textcolor{black}{$1.0000 \pm 0.000$} & $0.9778 \pm 0.000$ & \textcolor{black}{$0.9867 \pm 0.000$} & \textcolor{black}{$1.0000 \pm 0.000$} & $0.9511 \pm 0.000$ & $0.9831 \pm 0.000$ \\
DenseNet + ResNet50 + ViT-Large & \textcolor{black}{$1.0000 \pm 0.000$} & $0.9778 \pm 0.000$ & \textcolor{black}{$0.9867 \pm 0.000$} & \textcolor{black}{$1.0000 \pm 0.000$} & $0.9689 \pm 0.000$ & $0.9867 \pm 0.000$ \\
DenseNet + VGG16 & \textcolor{black}{$1.0000 \pm 0.000$} & $0.9911 \pm 0.000$ & \textcolor{black}{$0.9867 \pm 0.000$} & \textcolor{black}{$1.0000 \pm 0.000$} & $0.9689 \pm 0.000$ & $0.9893 \pm 0.000$ \\
DenseNet + VGG16 + ViT-Large & \textcolor{black}{$1.0000 \pm 0.000$} & $0.9778 \pm 0.000$ & \textcolor{black}{$0.9867 \pm 0.000$} & \textcolor{black}{$1.0000 \pm 0.000$} & $0.9778 \pm 0.000$ & $0.9884 \pm 0.000$ \\
DenseNet + ViT-Large & \textcolor{black}{$1.0000 \pm 0.000$} & $0.9822 \pm 0.000$ & \textcolor{black}{$0.9867 \pm 0.000$} & \textcolor{black}{$1.0000 \pm 0.000$} & $0.9644 \pm 0.000$ & $0.9867 \pm 0.000$ \\
Inception-v3 & \textcolor{black}{$1.0000 \pm 0.000$} & $0.9867 \pm 0.000$ & \textcolor{black}{$0.9867 \pm 0.000$} & \textcolor{black}{$1.0000 \pm 0.000$} & $0.9689 \pm 0.000$ & $0.9884 \pm 0.000$ \\
Inception-v3 + ResNet50 & \textcolor{black}{$1.0000 \pm 0.000$} & $0.9778 \pm 0.000$ & \textcolor{black}{$0.9867 \pm 0.000$} & \textcolor{black}{$1.0000 \pm 0.000$} & $0.9644 \pm 0.001$ & $0.9858 \pm 0.000$ \\
Inception-v3 + ResNet50 + VGG16 & \textcolor{black}{$1.0000 \pm 0.000$} & $0.9733 \pm 0.000$ & $0.9822 \pm 0.000$ & \textcolor{black}{$1.0000 \pm 0.000$} & $0.9689 \pm 0.000$ & $0.9849 \pm 0.000$ \\
Inception-v3 + ResNet50 + ViT-Large & \textcolor{black}{$1.0000 \pm 0.000$} & \textcolor{black}{$0.9956 \pm 0.000$} & \textcolor{black}{$0.9867 \pm 0.000$} & \textcolor{black}{$1.0000 \pm 0.000$} & $0.9644 \pm 0.000$ & $0.9893 \pm 0.000$ \\
Inception-v3 + VGG16 & \textcolor{black}{$1.0000 \pm 0.000$} & $0.9733 \pm 0.000$ & \textcolor{black}{$0.9867 \pm 0.000$} & \textcolor{black}{$1.0000 \pm 0.000$} & $0.9644 \pm 0.000$ & $0.9849 \pm 0.000$ \\
Inception-v3 + VGG16 + ViT-Large & \textcolor{black}{$1.0000 \pm 0.000$} & $0.9778 \pm 0.000$ & \textcolor{black}{$0.9867 \pm 0.000$} & \textcolor{black}{$1.0000 \pm 0.000$} & $0.9689 \pm 0.000$ & $0.9867 \pm 0.000$ \\
Inception-v3 + ViT-Large & \textcolor{black}{$1.0000 \pm 0.000$} & $0.9733 \pm 0.000$ & \textcolor{black}{$0.9867 \pm 0.000$} & \textcolor{black}{$1.0000 \pm 0.000$} & $0.9600 \pm 0.000$ & $0.9840 \pm 0.000$ \\
ResNet50 & \textcolor{black}{$1.0000 \pm 0.000$} & $0.9867 \pm 0.000$ & $0.9333 \pm 0.006$ & $0.9689 \pm 0.002$ & $0.8978 \pm 0.006$ & $0.9573 \pm 0.001$ \\
ResNet50 + VGG16 & \textcolor{black}{$1.0000 \pm 0.000$} & $0.9911 \pm 0.000$ & \textcolor{black}{$0.9867 \pm 0.000$} & \textcolor{black}{$1.0000 \pm 0.000$} & $0.9822 \pm 0.000$ & \textcolor{black}{$0.9920 \pm 0.000$} \\
ResNet50 + VGG16 + ViT-Large & \textcolor{black}{$1.0000 \pm 0.000$} & $0.9822 \pm 0.000$ & \textcolor{black}{$0.9867 \pm 0.000$} & \textcolor{black}{$1.0000 \pm 0.000$} & $0.9511 \pm 0.000$ & $0.9840 \pm 0.000$ \\
ResNet50 + ViT-Large & \textcolor{black}{$1.0000 \pm 0.000$} & $0.9778 \pm 0.000$ & \textcolor{black}{$0.9867 \pm 0.000$} & \textcolor{black}{$1.0000 \pm 0.000$} & $0.9556 \pm 0.000$ & $0.9840 \pm 0.000$ \\
VGG16 & \textcolor{black}{$1.0000 \pm 0.000$} & $0.9778 \pm 0.000$ & \textcolor{black}{$0.9867 \pm 0.000$} & \textcolor{black}{$1.0000 \pm 0.000$} & \textcolor{black}{$0.9867 \pm 0.000$} & $0.9902 \pm 0.000$ \\
VGG16 + ViT-Large & \textcolor{black}{$1.0000 \pm 0.000$} & $0.9733 \pm 0.000$ & $0.9822 \pm 0.000$ & \textcolor{black}{$1.0000 \pm 0.000$} & $0.9822 \pm 0.000$ & $0.9876 \pm 0.000$ \\
ViT-Large & \textcolor{black}{$1.0000 \pm 0.000$} & $0.9733 \pm 0.000$ & $0.9244 \pm 0.006$ & $0.9600 \pm 0.003$ & $0.9200 \pm 0.009$ & $0.9556 \pm 0.001$ \\

\hline
\end{tabular}%
}
\end{table*}

\begin{table*}[t]
\centering
\caption{\small Model prediction consistency evaluation for scanned documents on different models (FIN)}
\label{tab:cross_model_scanned_fin}
\resizebox{\textwidth}{!}{%
\begin{tabular}{l|c|c|c|c|c|c}
\hline
\textbf{Guiding models} & \textbf{ViT-Large} & \textbf{ResNet50} & \textbf{Inception-v3} & \textbf{VGG16} & \textbf{DenseNet} & \textbf{Average} \\
\hline
SSIM-only objective & $0.4949 \pm 0.001$ & $0.5505 \pm 0.002$ & $0.5960 \pm 0.013$ & $0.5000 \pm 0.001$ & $0.5505 \pm 0.011$ & $0.5384 \pm 0.001$ \\
DenseNet & \textcolor{black}{$0.9848 \pm 0.000$} & $0.9747 \pm 0.000$ & $0.9747 \pm 0.000$ & $0.9646 \pm 0.000$ & $0.9747 \pm 0.000$ & $0.9747 \pm 0.000$ \\
DenseNet + Inception-v3 & \textcolor{black}{$0.9848 \pm 0.000$} & $0.9798 \pm 0.000$ & \textcolor{black}{$0.9848 \pm 0.000$} & $0.9747 \pm 0.000$ & \textcolor{black}{$1.0000 \pm 0.000$} & \textcolor{black}{$0.9848 \pm 0.000$} \\
DenseNet + Inception-v3 + ResNet50 & \textcolor{black}{$0.9848 \pm 0.000$} & $0.9798 \pm 0.000$ & $0.9747 \pm 0.000$ & $0.9697 \pm 0.000$ & $0.9949 \pm 0.000$ & $0.9808 \pm 0.000$ \\
DenseNet + Inception-v3 + VGG16 & \textcolor{black}{$0.9848 \pm 0.000$} & $0.9646 \pm 0.000$ & $0.9697 \pm 0.000$ & \textcolor{black}{$0.9899 \pm 0.000$} & \textcolor{black}{$1.0000 \pm 0.000$} & $0.9818 \pm 0.000$ \\
DenseNet + Inception-v3 + ViT-Large & \textcolor{black}{$0.9848 \pm 0.000$} & $0.9697 \pm 0.000$ & $0.9798 \pm 0.000$ & $0.9697 \pm 0.000$ & $0.9949 \pm 0.000$ & $0.9798 \pm 0.000$ \\
DenseNet + ResNet50 & \textcolor{black}{$0.9848 \pm 0.000$} & $0.9798 \pm 0.000$ & $0.9798 \pm 0.000$ & $0.9747 \pm 0.000$ & $0.9545 \pm 0.001$ & $0.9747 \pm 0.000$ \\
DenseNet + ResNet50 + VGG16 & \textcolor{black}{$0.9848 \pm 0.000$} & $0.9798 \pm 0.000$ & $0.9747 \pm 0.000$ & $0.9747 \pm 0.000$ & \textcolor{black}{$1.0000 \pm 0.000$} & $0.9828 \pm 0.000$ \\
DenseNet + ResNet50 + ViT-Large & \textcolor{black}{$0.9848 \pm 0.000$} & $0.9798 \pm 0.000$ & $0.9798 \pm 0.000$ & $0.9646 \pm 0.000$ & \textcolor{black}{$1.0000 \pm 0.000$} & $0.9818 \pm 0.000$ \\
DenseNet + VGG16 & \textcolor{black}{$0.9848 \pm 0.000$} & $0.9697 \pm 0.000$ & $0.9747 \pm 0.000$ & $0.9646 \pm 0.000$ & \textcolor{black}{$1.0000 \pm 0.000$} & $0.9788 \pm 0.000$ \\
DenseNet + VGG16 + ViT-Large & \textcolor{black}{$0.9848 \pm 0.000$} & $0.9646 \pm 0.000$ & $0.9747 \pm 0.000$ & $0.9798 \pm 0.000$ & \textcolor{black}{$1.0000 \pm 0.000$} & $0.9808 \pm 0.000$ \\
DenseNet + ViT-Large & \textcolor{black}{$0.9848 \pm 0.000$} & $0.9747 \pm 0.000$ & $0.9747 \pm 0.000$ & $0.9697 \pm 0.000$ & $0.9697 \pm 0.001$ & $0.9747 \pm 0.000$ \\
Inception-v3 & \textcolor{black}{$0.9848 \pm 0.000$} & $0.9747 \pm 0.000$ & $0.9747 \pm 0.000$ & $0.9596 \pm 0.000$ & $0.9848 \pm 0.000$ & $0.9758 \pm 0.000$ \\
Inception-v3 + ResNet50 & \textcolor{black}{$0.9848 \pm 0.000$} & $0.9848 \pm 0.000$ & $0.9747 \pm 0.000$ & $0.9596 \pm 0.000$ & $0.9949 \pm 0.000$ & $0.9798 \pm 0.000$ \\
Inception-v3 + ResNet50 + VGG16 & \textcolor{black}{$0.9848 \pm 0.000$} & $0.9798 \pm 0.000$ & $0.9747 \pm 0.000$ & $0.9697 \pm 0.000$ & \textcolor{black}{$1.0000 \pm 0.000$} & $0.9818 \pm 0.000$ \\
Inception-v3 + ResNet50 + ViT-Large & \textcolor{black}{$0.9848 \pm 0.000$} & \textcolor{black}{$0.9899 \pm 0.000$} & $0.9747 \pm 0.000$ & $0.9697 \pm 0.000$ & \textcolor{black}{$1.0000 \pm 0.000$} & $0.9838 \pm 0.000$ \\
Inception-v3 + VGG16 & \textcolor{black}{$0.9848 \pm 0.000$} & $0.9697 \pm 0.000$ & $0.9697 \pm 0.000$ & $0.9697 \pm 0.000$ & \textcolor{black}{$1.0000 \pm 0.000$} & $0.9788 \pm 0.000$ \\
Inception-v3 + VGG16 + ViT-Large & \textcolor{black}{$0.9848 \pm 0.000$} & $0.9848 \pm 0.000$ & $0.9747 \pm 0.000$ & $0.9596 \pm 0.000$ & $0.9848 \pm 0.000$ & $0.9778 \pm 0.000$ \\
Inception-v3 + ViT-Large & \textcolor{black}{$0.9848 \pm 0.000$} & $0.9697 \pm 0.000$ & $0.9798 \pm 0.000$ & $0.9545 \pm 0.000$ & \textcolor{black}{$1.0000 \pm 0.000$} & $0.9778 \pm 0.000$ \\
ResNet50 & \textcolor{black}{$0.9848 \pm 0.000$} & $0.9697 \pm 0.000$ & $0.9242 \pm 0.004$ & $0.9697 \pm 0.000$ & $0.9899 \pm 0.000$ & $0.9677 \pm 0.001$ \\
ResNet50 + VGG16 & \textcolor{black}{$0.9848 \pm 0.000$} & \textcolor{black}{$0.9899 \pm 0.000$} & $0.9242 \pm 0.004$ & $0.9545 \pm 0.000$ & $0.9848 \pm 0.000$ & $0.9677 \pm 0.001$ \\
ResNet50 + VGG16 + ViT-Large & \textcolor{black}{$0.9848 \pm 0.000$} & $0.9697 \pm 0.000$ & $0.9293 \pm 0.003$ & $0.9697 \pm 0.000$ & $0.9848 \pm 0.000$ & $0.9677 \pm 0.000$ \\
ResNet50 + ViT-Large & \textcolor{black}{$0.9848 \pm 0.000$} & $0.9646 \pm 0.000$ & $0.8434 \pm 0.010$ & $0.8586 \pm 0.018$ & $0.9646 \pm 0.001$ & $0.9232 \pm 0.004$ \\
VGG16 & \textcolor{black}{$0.9848 \pm 0.000$} & $0.9747 \pm 0.000$ & $0.9697 \pm 0.000$ & $0.9798 \pm 0.000$ & $0.9899 \pm 0.000$ & $0.9798 \pm 0.000$ \\
VGG16 + ViT-Large & $0.9798 \pm 0.000$ & $0.9394 \pm 0.001$ & $0.8838 \pm 0.015$ & $0.9545 \pm 0.001$ & $0.9141 \pm 0.015$ & $0.9343 \pm 0.001$ \\
ViT-Large & \textcolor{black}{$0.9848 \pm 0.000$} & $0.9747 \pm 0.000$ & \textcolor{black}{$0.9848 \pm 0.000$} & $0.9747 \pm 0.000$ & $0.9697 \pm 0.001$ & $0.9778 \pm 0.000$ \\

\hline
\end{tabular}%
}
\end{table*}

\subsection{Scaling to New Document Types.}
\label{sec:scaling}
We released documents from only ten European countries because each of these document types has around 100 high-quality identity documents for non-fraud and fraud classes, respectively, enabling us to train and drive our model-guided approach to generate meaningful synthetic data. These ten types of documents, including five types of passports and five types of ID cards, encompass most of the security features and schema elements found in general identity documents. They also cover different languages.

To demonstrate that our technique could apply to other types, we used the synthetic West Virginia driver's licenses dataset from prior work~\cite{xie2024idnet, guan2024idnet}, which consists of 5,979 synthetic non-fraud documents and 5,979 synthetic fraud documents for each of the fraud patterns.

We trained the EfficientNet-b3, ResNet50, and ViT-Large models using 1000 non-fraud documents and 1000 fraud documents, which are split into training, validation, and testing sets by 5:2:3. Similar to the experimental setting used for Table 2 in our submitted paper, we used 20 non-fraud documents and 20 fraud documents for Bayesian Optimization search with and without guiding models for finetuning the hyper-parameters.

The model consistency evaluation results (similar to Tab.~\ref{tab:cross_model}), obtained on the testing sets (including 300 non-fraud documents and 300 fraud documents), are illustrated in Tab.~\ref{tab:west-virgina-consistency1} and Tab.~\ref{tab:west-virgina-consistency2}, highlighting the good generalization capability of our \IDSpace synthetic data generation approach proposed in this work.

 \begin{table*}
  \caption{\small Consistency Score (Mean \ensuremath{\pm} Std) for Various Target Models W/O Guiding Models (Baseline) 
  }
  \label{tab:west-virgina-consistency1}
  \scriptsize
  \centering
  \begin{tabular}{c|c|c|c|c}
    \toprule
         & EfficientNet-b3	& ResNet50	& ViT-Large	& Average \\
    \midrule
BO w/ SSIM-only objective &	 0.9566 \ensuremath{\pm} 0.028	 &0.6728 \ensuremath{\pm} 0.131&	 0.6465 \ensuremath{\pm} 0.193&	 0.7586 \ensuremath{\pm} 0.172
    \\
    \bottomrule
  \end{tabular}
\end{table*}

 \begin{table*}
  \caption{\small Consistency Score (Mean \ensuremath{\pm} Std) for Various Target Models W/ Guiding Models (Our proposed approach)
  }
  \label{tab:west-virgina-consistency2}
  \scriptsize
  \centering
  \begin{tabular}{c|c|c|c|c}
    \toprule
Guiding Models	& EfficientNet-b3	& ResNet50	& ViT-Large	& Average \\
    \midrule
EfficientNet-b3	& 0.9860 \ensuremath{\pm} 0.004	& 0.7655 \ensuremath{\pm} 0.166	& 0.8531 \ensuremath{\pm} 0.169&	 0.8682 \ensuremath{\pm} 0.091\\\hline
EfficientNet-b3 + ResNet50	& 0.9851 \ensuremath{\pm} 0.004&	 0.9204 \ensuremath{\pm} 0.023	& 0.9078 \ensuremath{\pm} 0.048&	 0.9378 \ensuremath{\pm} 0.041\\\hline
EfficientNet-b3 + ViT-Large	& 0.9817 \ensuremath{\pm} 0.002&	 0.8310 \ensuremath{\pm} 0.118	& 0.9301 \ensuremath{\pm} 0.011	& 0.9143 \ensuremath{\pm} 0.077\\\hline
ResNet50	& 0.9779 \ensuremath{\pm} 0.012&	 0.9356 \ensuremath{\pm} 0.004	& 0.9301 \ensuremath{\pm} 0.018&	 0.9479 \ensuremath{\pm} 0.026\\\hline
ResNet50 + ViT-Large&	 0.9782 \ensuremath{\pm} 0.006	 &0.9135 \ensuremath{\pm} 0.026	& 0.9247 \ensuremath{\pm} 0.026&	 0.9388 \ensuremath{\pm} 0.035\\\hline
ViT-Large&	 0.9802 \ensuremath{\pm} 0.008&	 0.8597 \ensuremath{\pm} 0.038&	 0.9161 \ensuremath{\pm} 0.037&	 0.9187 \ensuremath{\pm} 0.060
    \\
    \bottomrule
  \end{tabular}
\end{table*}

Adding a new document template does require an engineering pass to identify its segments and the corresponding parameter search spaces. However, this effort is a one-time specification of field settings for each template. In practice, the required work is modest and quickly amortized by the scalability of the generation process. Importantly, this cost is orders of magnitude lower than that of collecting and annotating new real-world data.


Tab.~\ref{tab:manual-effort-breakdown} illustrates a detailed breakdown of the time spent preparing a new template for the above experiment on the West Virginia Driver's License dataset by a Ph.D student with one year of experience in the identity document design domain.

\begin{table*}
  \caption{\small Breakdown of the manual effort required to introduce a new type of identity document
  }
  \label{tab:manual-effort-breakdown}
  \scriptsize
  \centering
  \begin{tabular}{p{6cm}|p{3cm}}
    \toprule
Steps	&  Time Required (seconds)\\\hline
Identify meta data	 & 148 ( for all 21 fields)\\\hline
Generate template(prompt development and tuning)	 & 246 \\\hline
Predefine hyperparameter for each field	&  1356\\\hline
Configure the scripts for filling metadata to template	&  1210
    \\
    \bottomrule
  \end{tabular}
\end{table*}

The automatic processing times are illustrated in Tab.~\ref{tab:automatic-effort-breakdown}, which are tested on an Ubuntu Linux server equipped with 48 CPU cores (Intel Xeon Silver 4310, 2.10 GHz), 125 GB of RAM, and two NVIDIA A10 GPUs (24GB VRAM each).

\begin{table*}
  \caption{\small Breakdown of the automatic processing (i.e., our scripts) required to generate documents for the new type of identity document
  }
  \label{tab:automatic-effort-breakdown}
  \scriptsize
  \centering
  \begin{tabular}{p{3.5cm}|p{5.7cm}}
    \toprule
Steps	&  Computing Time (seconds)\\\hline
Stable diffusion Generate template	&  16.9\\\hline
Generate synthetic metadata	 & 39.53 (for automatically generating 1000 images)\\\hline
Hyperparameter search	&  42,264\\\hline
Fill metadata to template	&  1413 (for 1000 images)\\
    \bottomrule
  \end{tabular}
\end{table*}


\subsection{StyleGAN Results.}
\label{sec:stylegan}
To evaluate the identity generation capability of GANs, we fine-tuned a pre-trained StyleGAN model using $40$ Albanian 
 identity documents in the MIDV/SIDTD dataset. 
We adopted the official StyleGAN3 training configuration with the following parameters: \texttt{cfg=stylegan2}, \texttt{snap=10}, \texttt{mirror=1}, \texttt{batch=32}, and \texttt{gamma=8.2}. Training was conducted using the pretrained model \textit{stylegan2-ffhq-1024x1024.pkl}. Since StyleGAN3 requires square inputs, we resized the original image dimensions from $(2167, 1360)$ to $(1024, 1024)$, preserving the aspect ratio as much as possible. 
Model performance was assessed using the FID50K\_full metric (Fr{\'e}chet Inception Distance) against the full dataset. The best FID score of 57.52 was achieved using the pretrained model, and the corresponding checkpoint was selected for evaluation. 

\begin{figure*}[h!]
\centering
\includegraphics[width=0.4\textwidth]{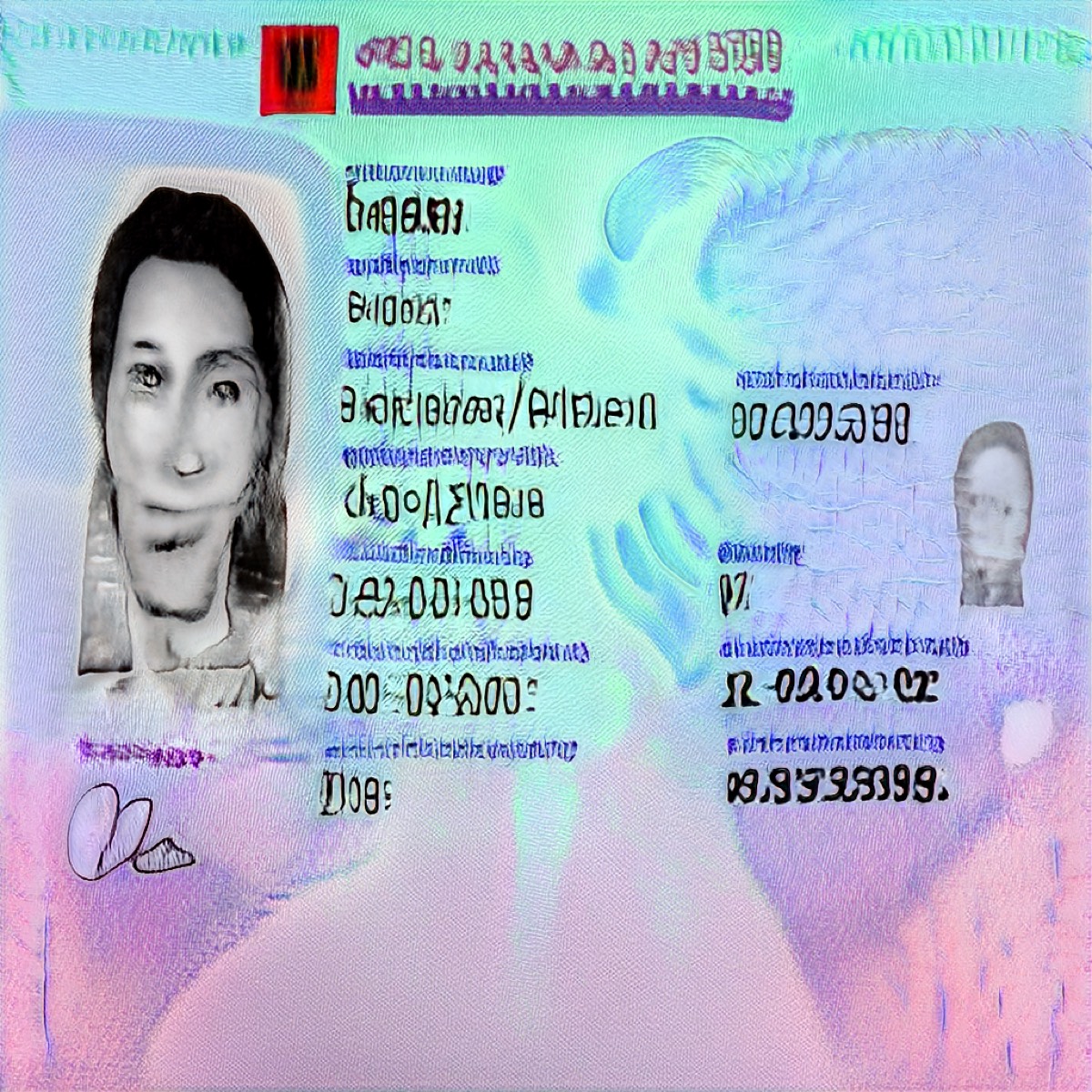} 
\caption{Images generated by StyleGAN3 using 40 identity samples from the MIDV/SIDTD dataset.}
\label{fig:stylegan}
\end{figure*}

Figure~\ref{fig:stylegan} presents example output. As illustrated, the model struggles to generate semantically meaningful or identity-consistent images mainly due to the limited size of the training dataset. GANs generally require hundreds to thousands of high-quality images to learn robust and high-fidelity representations.


\subsection{Diffusion-Based Image Inpainting.}
\label{sec:img2img}

We evaluate a diffusion-based image inpainting baseline using Stable Diffusion on a subset of the MIDV/SIDTD dataset containing (2, 20, and 40) Albanian ID cards, where the model is conditioned on an input identity document image, a localized mask indicating the target segment region, and a text prompt specifying the desired replacement. To construct training and test pairs without collecting additional annotations, we adopt a controlled region-replacement strategy. We randomly select a base identity document and extract its target segment region using the corresponding annotation. For every other image in the dataset, we replace its original target segment region with the base target segment region while keeping all other document content unchanged. This process yields paired samples consisting of an input image with a mismatched target segment region and a target image with the original correct target segment, enabling evaluation of whether the model can accurately recover the target segment by editing only the localized region while preserving the surrounding document structure. During inference, we explore a wide range of prompt formulations and diffusion parameters, including variations in noise strength, the classifier-free guidance scale, and the number of denoising steps. While sufficiently high noise strength consistently induces visible changes in the target segment region, the generated text often deviates from the target string, including incorrect characters and inconsistent spelling. Increasing guidance or denoising steps does not reliably improve textual correctness and often introduces additional artifacts, indicating that diffusion-based inpainting remains unreliable for exact text reproduction in structured identity documents, even under extensive prompt engineering and parameter tuning.

\begin{figure*}[h!]
    \centering
    \begin{tabular}{cc}
        \includegraphics[width=0.45\linewidth]{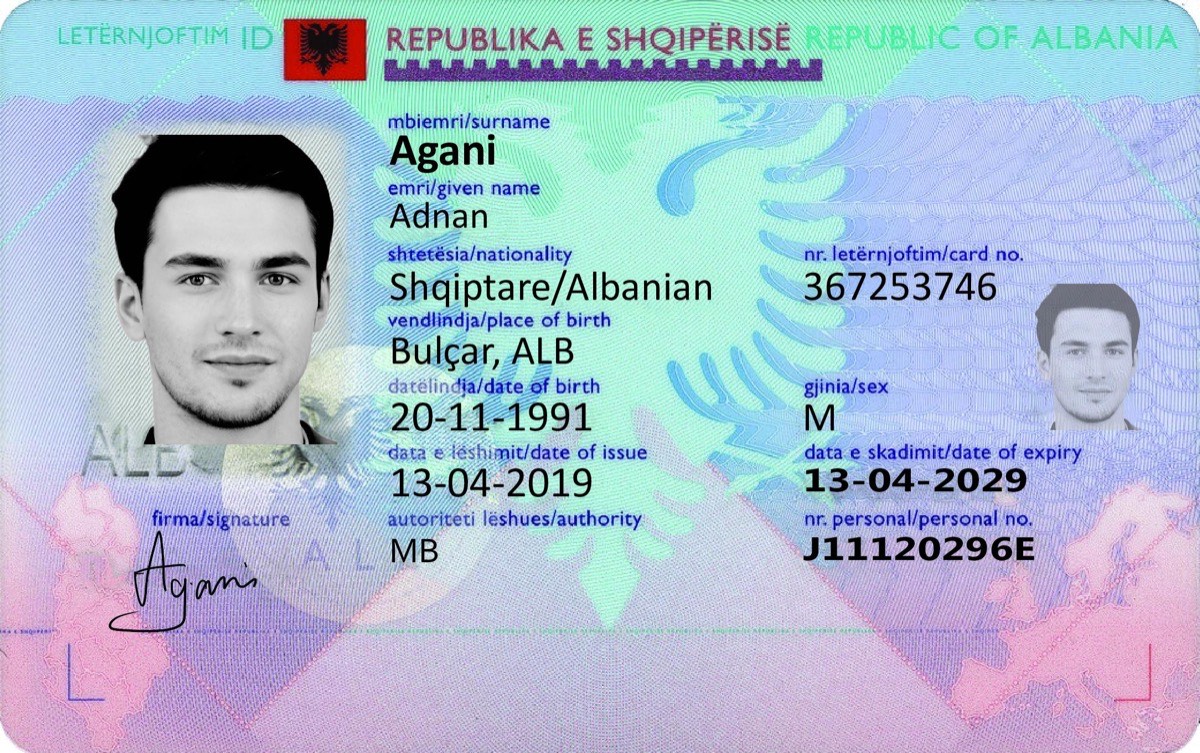} &
        \includegraphics[width=0.45\linewidth]{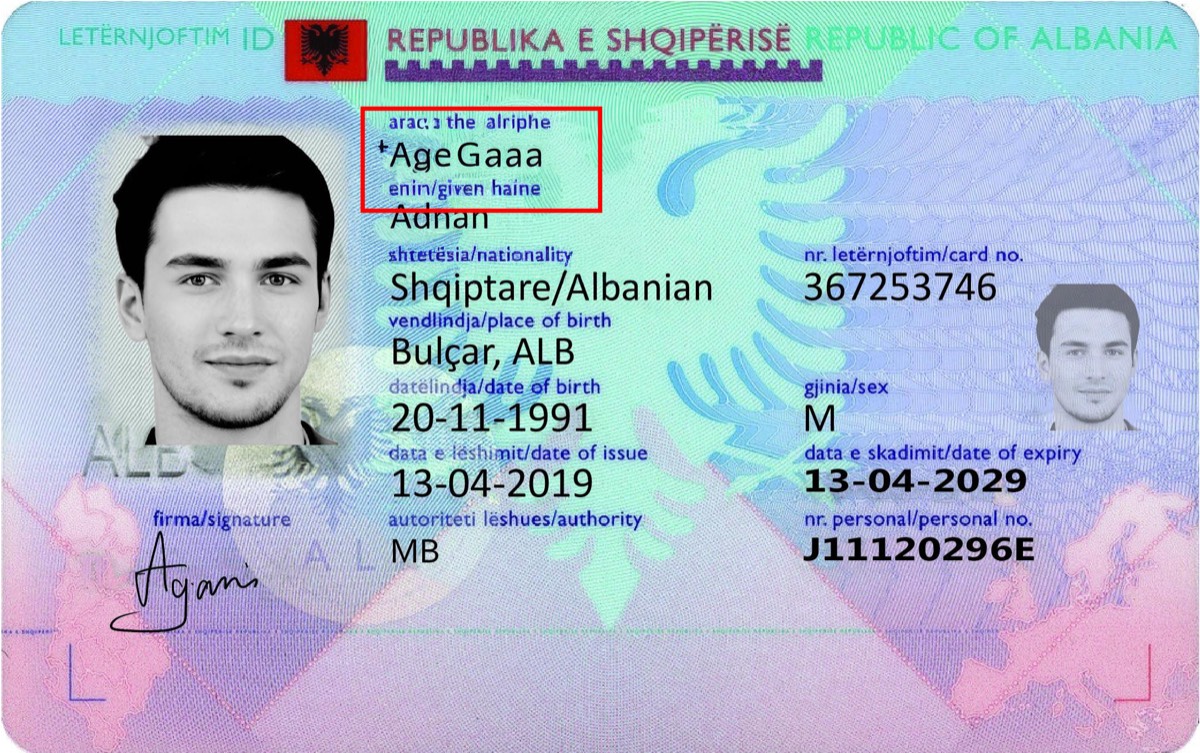} \\
        \small Target Image & \small Inpainted Output \\

    \end{tabular}
    \caption{Qualitative results of diffusion-based image inpainting.
The left image shows the target identity document samples, while the right image shows the corresponding images generated by the diffusion inpainting. The surname region is highlighted to emphasize the localized edit. Although inpainting preserves the overall document structure more effectively than text-to-image generation, the generated surnames often deviate from the target text, illustrating the limitations of diffusion models for precise text editing in structured identity documents.}
    \label{fig:diffusion_inpainting}
\end{figure*}

Representative qualitative results are shown in Figure~\ref{fig:diffusion_inpainting} with the target segment set as surname. Compared to text-to-image generation, diffusion-based inpainting substantially improves structural fidelity by explicitly conditioning on the original document layout. Nevertheless, residual artifacts and localized background drift remain, indicating that even image-conditioned diffusion models struggle to faithfully preserve the precise geometric structure and layout constraints required by identity documents.

\subsection{Text-to-Image Diffusion Generation.}
\label{sec:text2image}

We include a text-to-image diffusion baseline to evaluate whether large pretrained diffusion models can synthesize realistic identity documents from textual descriptions alone. We fine-tune a pretrained Stable Diffusion model (\texttt{stable-diffusion-v1-5}) using a LoRA-based adaptation strategy on a subset of the MIDV/SIDTD dataset containing 40 Albanian ID cards. Original images are resized from $2167 \times 1360$ to $1024 \times 768$ to match the model's maximum supported resolution while approximately preserving aspect ratio. The model is trained for 5000 epochs, with checkpoints saved every 500 epochs, and we experiment with LoRA ranks of 64 and 128.

Due to the 77-token limit of the CLIP text encoder, each ground-truth OCR transcription is compressed into a single structured prompt. An example prompt is shown below:
\begin{quote}
\small
\texttt{Albanianid template; portrait female; Cobaj Elona; 26091957; Mgull, ALB; MB; F75926997V; 499949517; 19022018; 19022028.}
\end{quote}

\begin{figure*}[h!]
\centering
\includegraphics[width=0.4\textwidth]{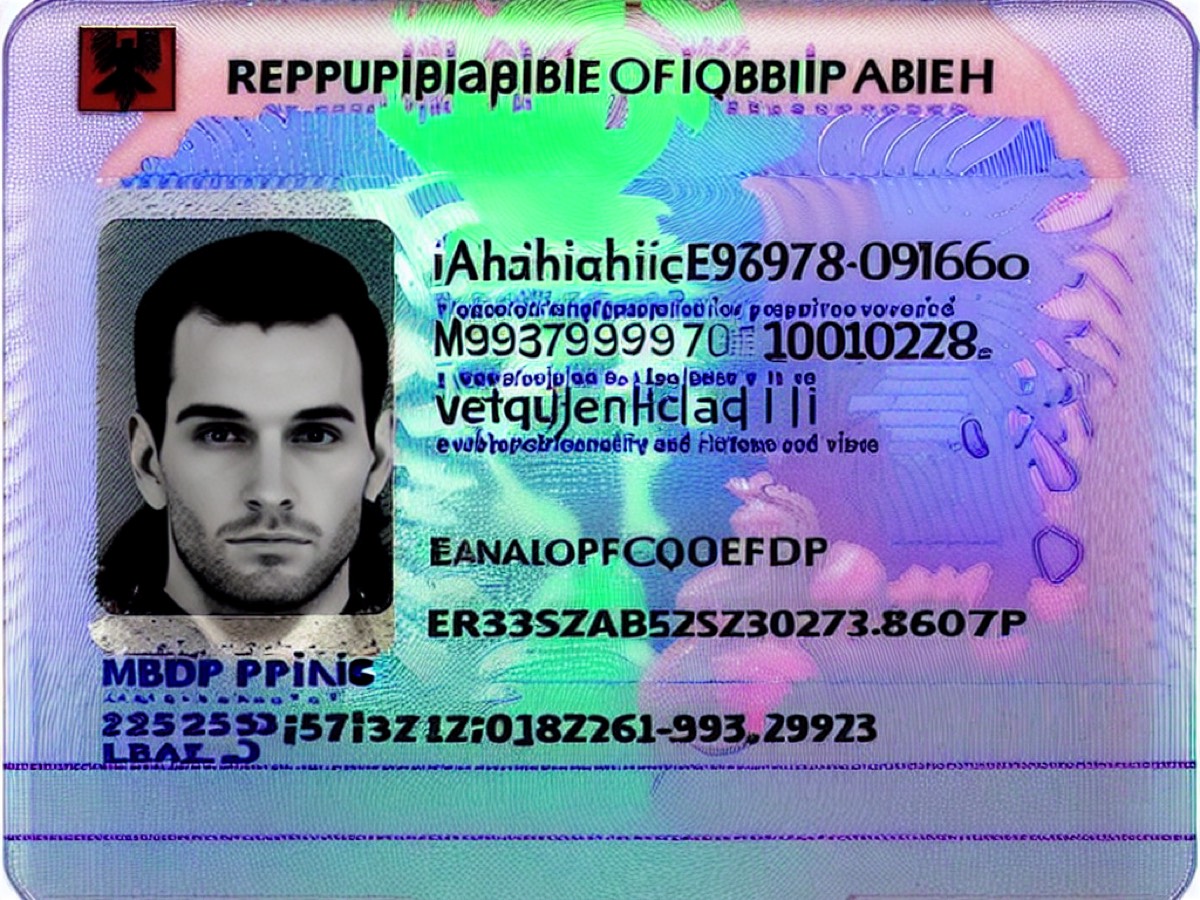} 
\caption{\textcolor{black}{Images generated by diffusion text-to-image model using 40 identity samples from the MIDV/SIDTD dataset.}}
\label{fig:diffusion_text2image}
\end{figure*}

For each checkpoint, we generate five samples using this fixed prompt. The most visually plausible result is shown in Figure~\ref{fig:diffusion_text2image}. Despite fine-tuning, the model fails to reproduce the rigid layout, typographic consistency, and fine-grained security features of authentic identity documents, highlighting the limitations of text-to-image diffusion for highly structured document synthesis.

\subsection{Generative Results.}
\label{sec:generative}

We also experimented with using GPT-4o and GPT-image-1 for ID image generation. The following prompt was provided for both of the models: \textit{``Using the provided sample as a reference, generate a realistic-looking ID card that closely mimics the layout, design, and visual elements. Replace all personal information (name, date of birth, ID number, place of birth, etc.) with clearly fictional data. Ensure that all formatting, fonts, and security features (such as watermarks, holograms, and layout positioning) remain as similar to the original as possible.''}

\begin{figure*}[h!]
    \centering
    \subfigure[Sample Image]{
        \includegraphics[width=0.45\textwidth]{figures/ALB_sample.jpg} 
    }
    \subfigure[Generated Image from GPT-4o]{
        \includegraphics[width=0.42\textwidth]{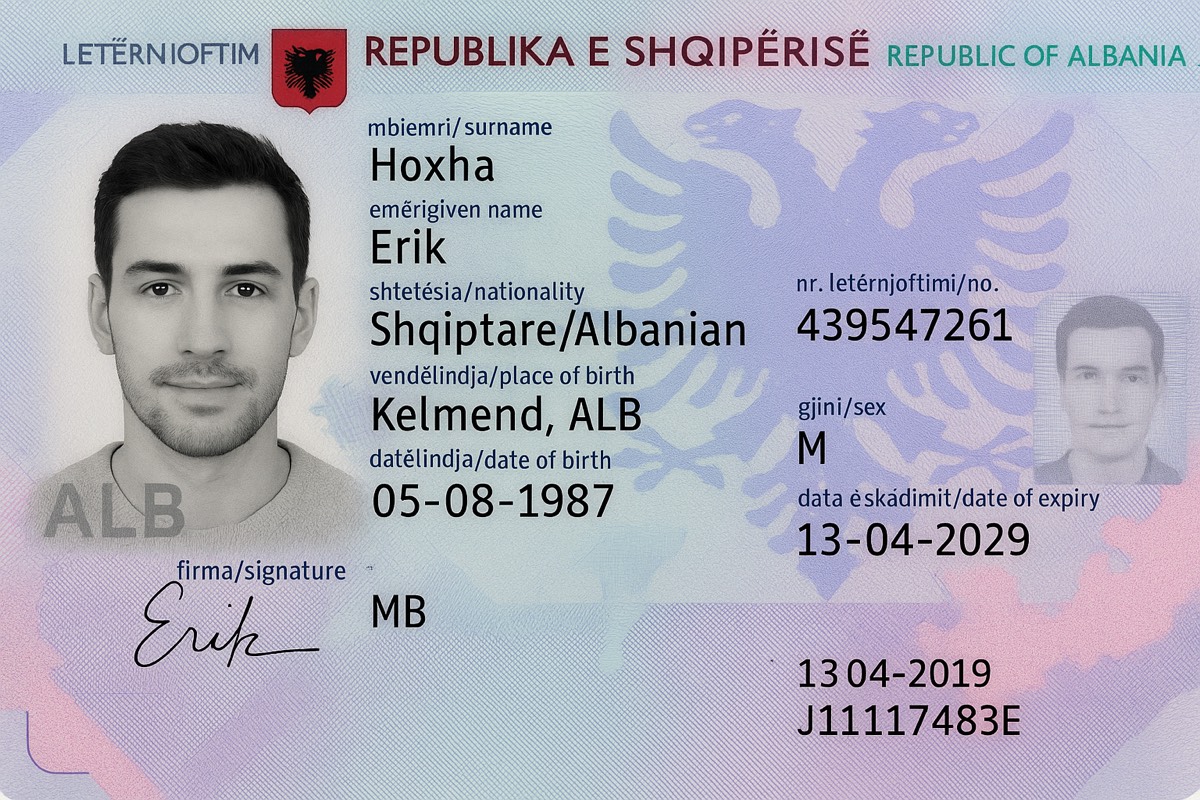} 
    }
    \subfigure[Generated Image from \IDSpace]{
        \includegraphics[width=0.45\textwidth]{figures/samples/alb/idspace_1.jpg} 
    }
    \subfigure[Generated Image from GPT-Image-1]{
        \includegraphics[width=0.42\textwidth]{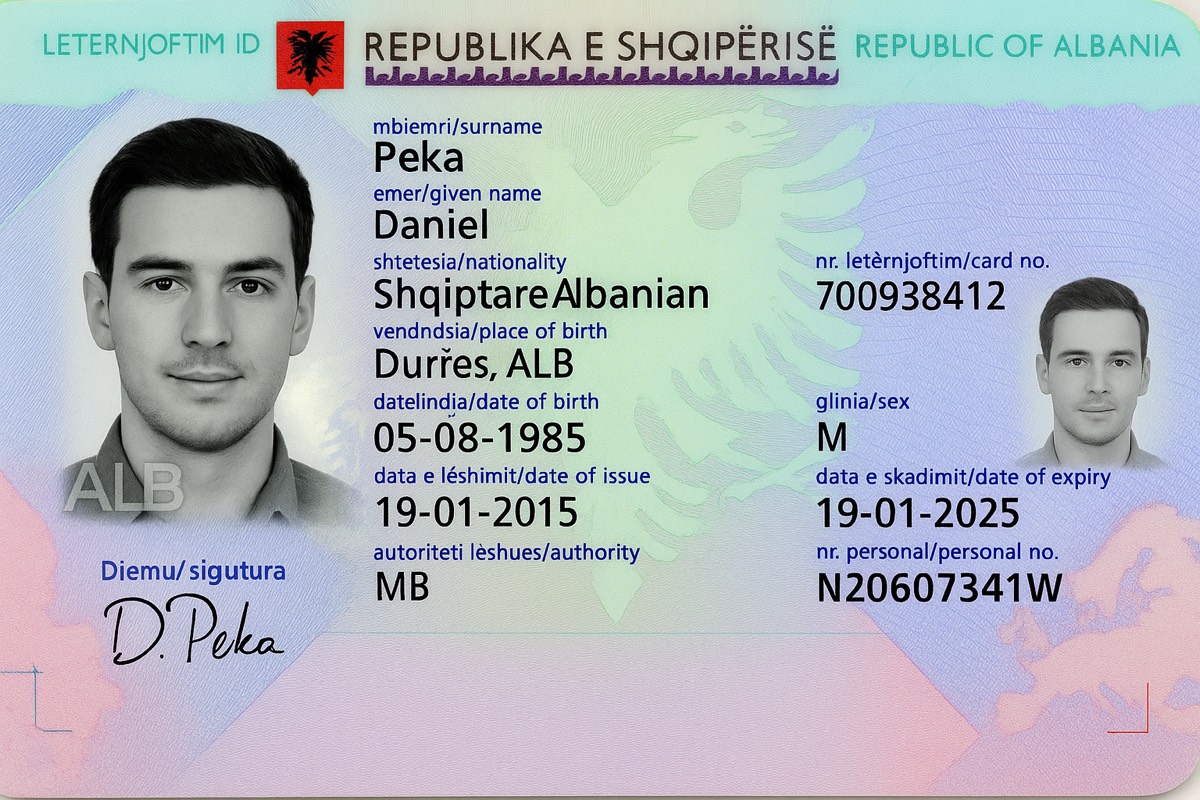} 
    }
    \caption{Comparison of Sample and Generated Image}
    \label{fig:comparison}
\end{figure*}

Figure~\ref{fig:comparison} presents four images: the original sample (a), the image generated by GPT-4o (b), the image generated by \IDSpace (c), and the image generated by GPT-Image-1 (d). As shown, the image generated by GPT-4o replicates certain background elements; however, it fails to fully preserve the original template structure. While some personal details were modified, others---such as the expiry date---remained unchanged. Additionally, the layout was inconsistently altered, affecting fields such as the issuing authority, personal number, and date of issue.

The image generated by GPT-Image-1 demonstrates a clear improvement over GPT-4o by successfully modifying all personal information as instructed. However, it changes the background design of the ID, resulting in a noticeable difference that can be easily identified by a human observer. Furthermore, both models fail to accurately replicate the font size and style, which are critical for maintaining the authenticity of identity documents. The output image dimensions from both GPT-4o and GPT-Image-1 also differ from those of the original input.

These inconsistencies suggest that although GPT-4o and GPT-Image-1 exhibit some capacity for layout replication, they lack the precision necessary to maintain the structural and semantic fidelity required for realistic ID template generation.

\subsection{Ablation Study: Tuning of $\lambda_i$.}

To investigate the impact of the $\lambda$ coefficients on consistency in the objective function, we conducted additional experiments on the model-guided BO method using the SIDTD dataset for the Albanian (ALB) region. We fixed the guiding model to ResNet50 and set the SSIM weight ($\lambda_0$) to 1. We then varied the consistency weight ($\lambda_1$) across a range of values: 0, 0.2, 0.5, 1, 1.5, 2, 5, and 10. The results are presented in Table~\ref{tab:lambda1}.

From the results, we observe that the consistency score increases as the consistency weight $\lambda_1$ increases, up to a value of 2. Beyond this point, the consistency score begins to slightly decline. This trend suggests that both SSIM and the consistency score play important roles in guiding the generation process. SSIM, which measures perceptual similarity to the target image, remains fundamental for maintaining visual quality, while the consistency term helps ensure semantic alignment. Therefore, a balanced combination of these two objectives is essential for optimal performance.

\begin{table*}
\centering
\caption{Consistency for different $\lambda_1$(Mean \ensuremath{\pm} Variance)}
\label{tab:lambda1}
\resizebox{\textwidth}{!}{
\begin{tabular}{l|c|c|c|c|c|c|c|c}
\hline
\textbf{$\lambda_1$} & 0 & 0.2 & 0.5 & 1 & 1.5 & 2 & 5 & 10 \\
\hline
Consistency & $0.4889 \pm 0.061$ &$0.9315 \pm 0.000$ &$0.9167 \pm 0.001$ &$0.9407 \pm 0.001$ &$0.9481 \pm 0.000$ & $0.9537 \pm 0.000$ & $0.9509 \pm 0.000$ & $0.9306 \pm 0.001$\\
\hline
\end{tabular}
}

\end{table*}

\subsection{Identity Document Fraud Detection Models used in Academic and Industry}
\label{sec:model-survey}

First, this work focuses on fraud detection in documents digitally captured under white light conditions, rather than using multi-spectral imaging techniques such as near-infrared and ultraviolet light. It also focuses on a binary classification task for specific fraud patterns following a broad class of academic works in this area~\cite{boned2024synthetic}. We utilized six commonly used, open-source vision architectures (ViT, ResNet, Inception, DenseNet, VGG16, and EfficientNet), each widely employed in recent academic research~\cite{boned2024synthetic, park2023kid34k, khare2024predictive}
 and industrial fraud detection pipelines~\cite{onfido, bruveris2020reducing, gietema2024method, mahadevan2023generalized, bayer2025authentication} for remote identity verification, as shown in Tab.~\ref{tab:academic-models} and Tab.~\ref{tab:industry-models}. Our selected fraud detection models, fine-tuned on real data, serve as strong surrogates for generalizable fraud detection behavior.
Our synthetic data generation method can be easily adopted in commercial black box platforms that are typically inaccessible due to IP restrictions. These platforms could use their models as guiding models to apply our approach to generate documents for model evaluation.

\begin{table*}
  \caption{\small Vision Models Used in Recent Academic Research 
  }
  \label{tab:academic-models}
  \scriptsize
  \centering
  \begin{tabular}{p{2cm}|p{4.5cm}}
    \toprule
         & Models Used for Fraud Detection  \\
    \midrule
    SIDTD~\cite{boned2024synthetic} & EfficientNet-B3, ResNet50, ViT-large, etc.\\\hline
    Kid34k~\cite{park2023kid34k} & ResNet18, ResNet34, EfficientNet, DenseNet, etc.\\\hline 
    ~\cite{khare2024predictive} & CNN, EfficientNet, etc.
    \\
    \bottomrule
  \end{tabular}
\end{table*}

\begin{table*}
  \caption{\small Vision Models in Industrial Fraud Detection Pipelines 
  }
  \label{tab:industry-models}
  \scriptsize
  \centering
  \begin{tabular}{p{2.5cm}|p{4.5cm}}
    \toprule
         & Models Used for Fraud Detection  \\
    \midrule
    Onfido (Now Entrust) & Onfido's Atlas AI platform supports micro-model ensembles (~10k ML models~\cite{onfido}), including convolutional models~\cite{gietema2024method} such as ResNet~\cite{bruveris2020reducing}, VGG16~\cite{mahadevan2023generalized}, and ViT~\cite{mahadevan2023generalized} backbones.\\\hline
   MicroBlink & MicroBlink's Know Your Customer (KYC) platform leverages ViT models for core platform and EfficientNet models for edge~\cite{microblink2025blog}.\\\hline 
 Jumio & CNN-based document neural networks~\cite{bayer2025authentication}.\\
    \bottomrule
  \end{tabular}
\end{table*}
 
\subsection{Mobile Document Generation}
\label{sec:mobile-document}
A key application of the \IDSpace is the generation of realistic mobile document images, including photographs of identity documents captured under diverse real-world backgrounds. In this section, we present a pipeline for generating such data by replacing a document in an existing background image with a document generated by \IDSpace. The challenge in this process is to ensure that the inserted document appears natural and visually consistent within the context of the original image. This involves solving several technical problems, including accurately detecting and localizing the original document, segmenting it from the background, aligning the new document to match the original perspective, and blending it seamlessly into the scene.

To address these challenges, we employ a combination of advanced computer vision models and image processing techniques. In particular, we use Grounding DINO~\cite{liu2023grounding}, a state-of-the-art model that integrates object detection and language grounding, to detect and localize the document in the original background image. Once localized, the Segment Anything Model (SAM)~\cite{kirillov2023segment} is applied to obtain an accurate segmentation mask of the original document. The prompt-based interface of SAM allows for precise and flexible segmentation, which is critical for accurate geometric alignment.

For the blending stage, we adopt the Deep Image Blending (DIB) framework~\cite{Zhang2019DeepIB}, which synthesizes high-quality images by optimizing a combination of loss functions including Poisson gradient loss, content loss, style loss, histogram loss, and total variation loss. To enhance structural fidelity, we extend the DIB loss with an additional differentiable Structural Similarity Index (SSIM) loss. This augmentation improves both local and global consistency between the blended image and the background.
The SSIM loss is formally defined as
$L_{\text{SSIM}} = 1 - \text{SSIM}(I_{\text{blend}}, I_{\text{background}})$,
where $\text{SSIM}(I_{\text{blend}}, I_{\text{background}})$ measures the structural similarity between the blended image $I_{\text{blend}}$ and the original background $I_{\text{background}}$ in terms of luminance, contrast, and structural features. The SSIM index for image patches $x$ and $y$ is computed as
$\text{SSIM}(x, y) = \frac{(2\mu_x \mu_y + C_1)(2\sigma_{xy} + C_2)}{(\mu_x^2 + \mu_y^2 + C_1)(\sigma_x^2 + \sigma_y^2 + C_2)}$,
where $\mu_x$ and $\mu_y$ are the means, $\sigma_x^2$ and $\sigma_y^2$ are the variances, and $\sigma_{xy}$ is the covariance between $x$ and $y$. $C_1$ and $C_2$ are constants to ensure numerical stability.

The complete loss function for the blending process becomes $L_{\text{total}} = L_{\text{DIB}} + \lambda L_{\text{SSIM}}$,
where $L_{\text{DIB}}$ represents the original DIB loss and $\lambda$ is a balancing coefficient. The inclusion of the SSIM term helps reduce visual artifacts such as inconsistent lighting and sharp edges at the insertion boundary. Consequently, the synthesized images exhibit improved structural coherence and perceptual realism. This enhanced blending quality contributes to the usability of the \IDSpace in real-world scenarios, facilitating the evaluation under realistic conditions.

\textbf{Parameterization.} Once we have the enhanced DIB model trained, for each generation, a user can flexibly specify to use the background of a certain existing mobile ID image (e.g., a picture of A's driver's license (DL) placed on top of the keyboard of a computer), and a template image, e.g., B's DL, to be blended by replacing A's DL in the image by B's DL. When generating a batch of documents, the users can flexibly specify the distribution of background images. For example, given a database of existing mobile documents that are annotated with labels describing the objects in the background, indoor or outdoor, lighting condition, the type of mobile phone used to capture the image, etc., e.g., MIDV's collection of mobile documents,  users can specify whether the new mobile dataset to generate will focus on indoor settings or outdoor settings, or it should involve $50\%$ of indoor images and $50\%$ of outdoor images. Users can also specify the distribution of age, gender, ethnicity groups, and fraud patterns of the entities involved in the new template images to be blended with existing documents. For example, using our tool and a database of existing mobile documents, a user can easily generate a batch of mobile documents featuring $100$ Spanish IDs for Asian Females with ages uniformly distributed from $5$ to $95$, with an indoor background captured by a discontinued Samsung Galaxy C5 mobile phone, for testing their newly trained models. The control parameters to be automatically tuned are the same with scanned images, as shown in Tab.~\ref{tab:parameters}.

\begin{figure*}[h!]
    \centering
    \subfigure[Background for Fig.~\ref{fig:mobile_docs}(a)]{
        \includegraphics[width=0.28\textwidth]{figures/mobile_backgrounds/a.jpg}
    }
    \subfigure[Background for Fig.~\ref{fig:mobile_docs}(b)]{
        \includegraphics[width=0.28\textwidth]{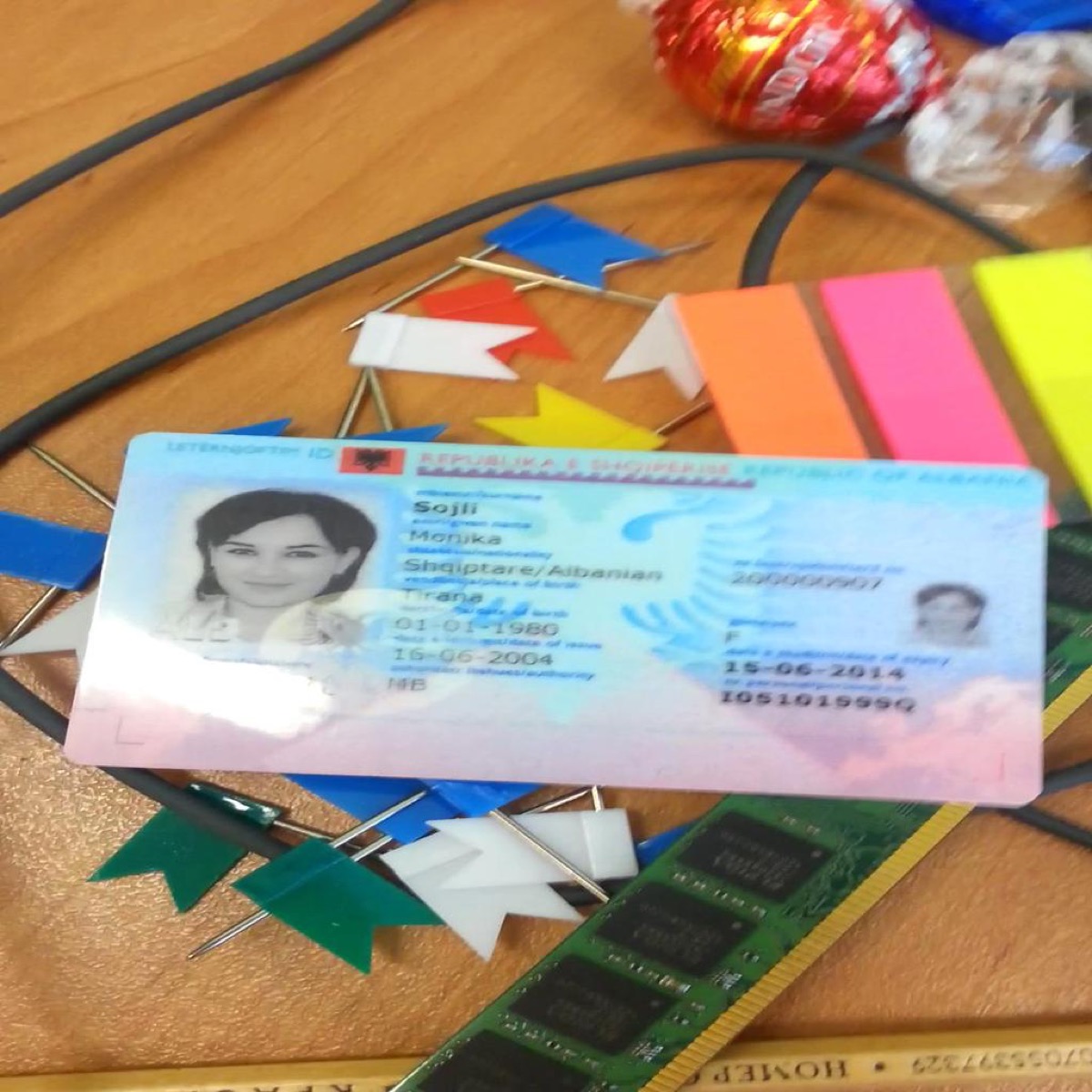}
    }
    \subfigure[Background for Fig.~\ref{fig:mobile_docs}(c)]{
        \includegraphics[width=0.28\textwidth]{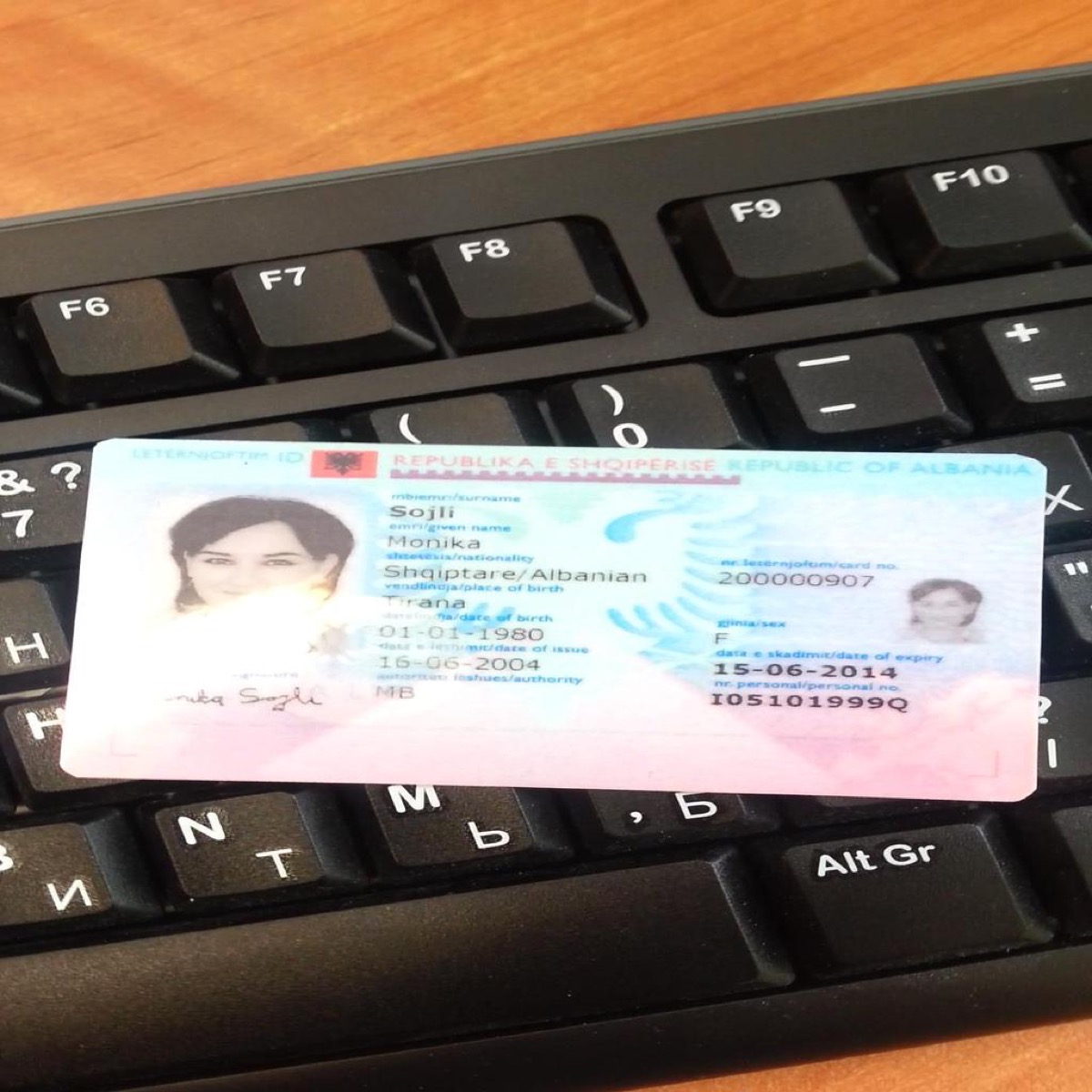}
    }
    \subfigure[Background for Fig.~\ref{fig:mobile_docs}(d)]{
        \includegraphics[width=0.28\textwidth]{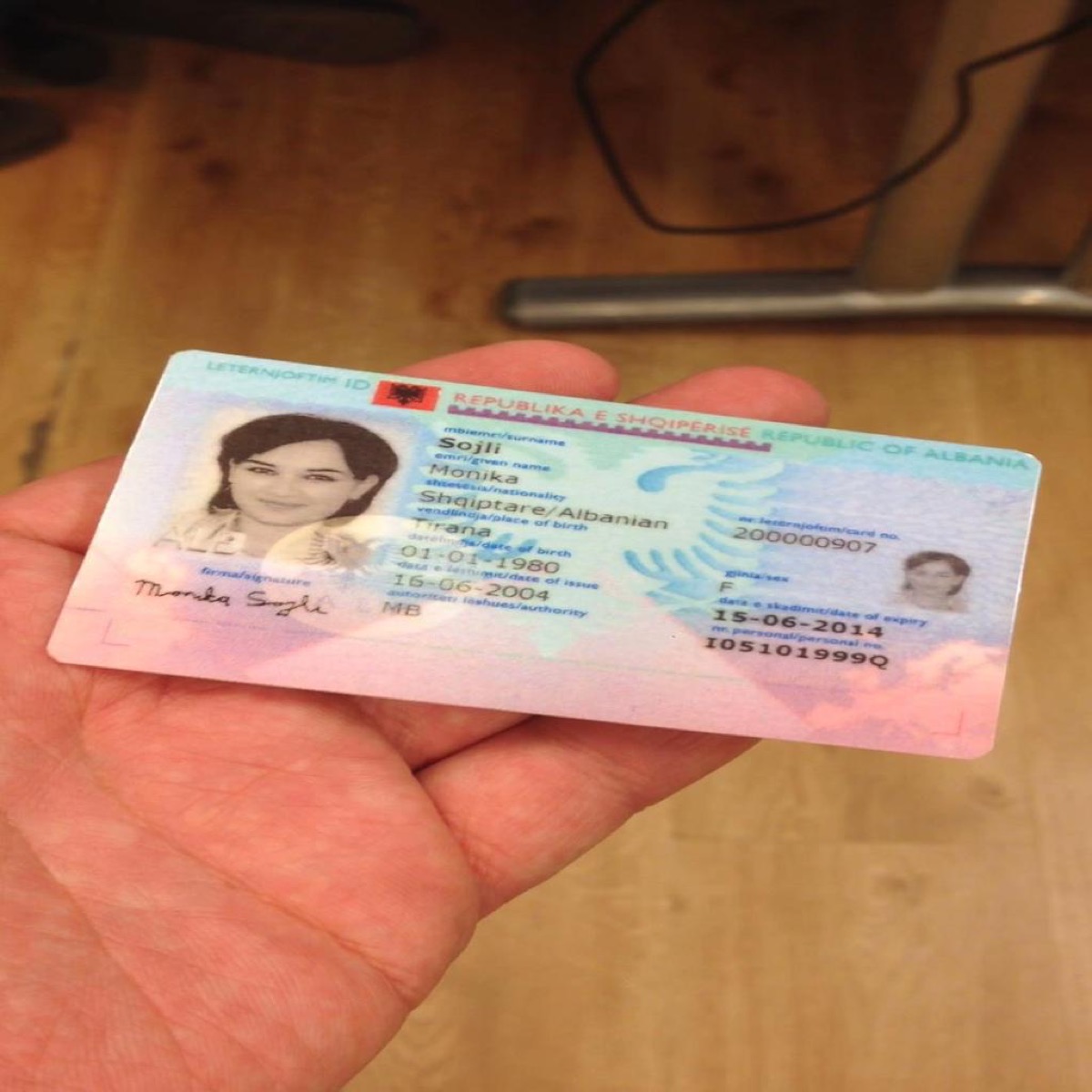}
    }
    \subfigure[Background for Fig.~\ref{fig:mobile_docs}(e)]{
        \includegraphics[width=0.28\textwidth]{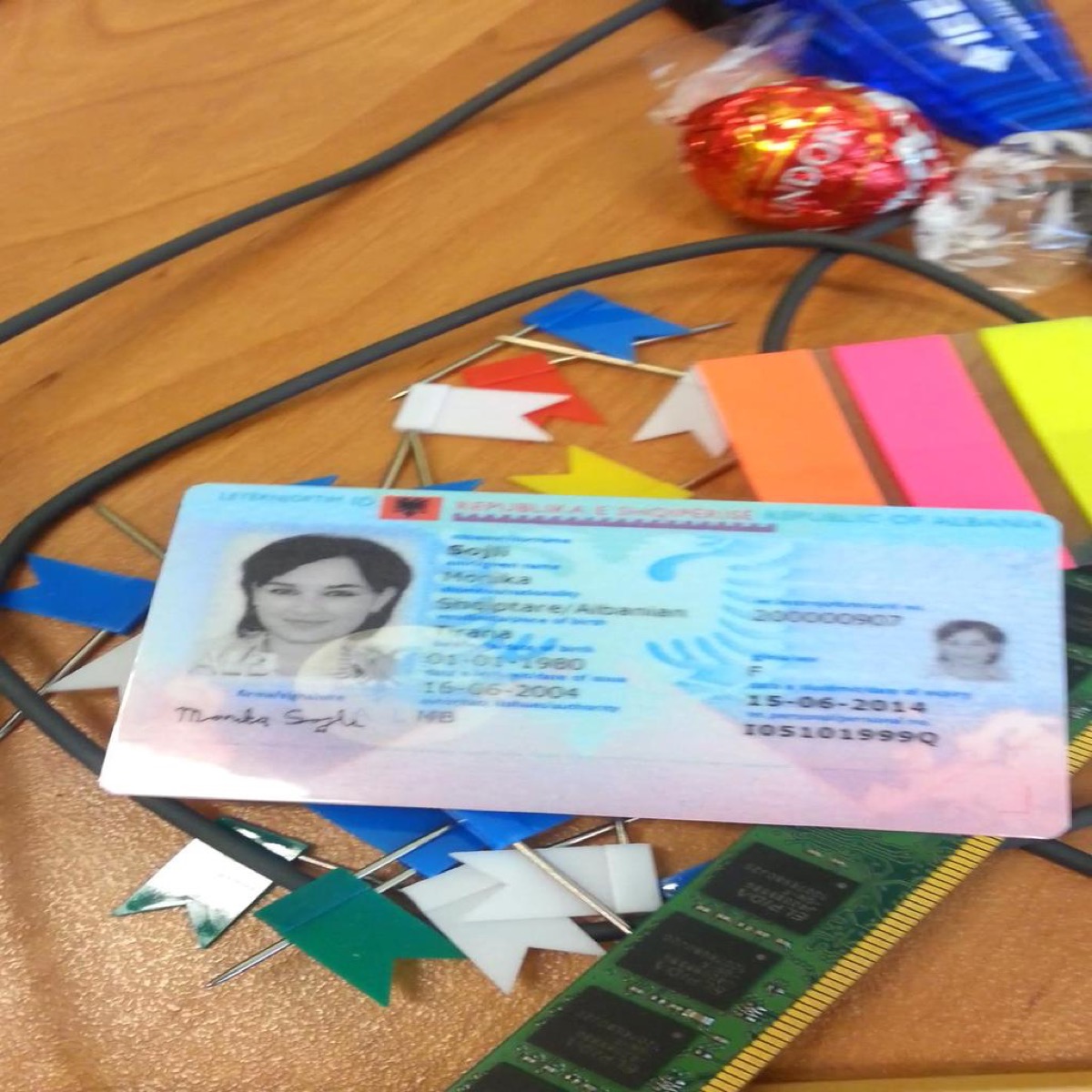}
    }
    \subfigure[Background for Fig.~\ref{fig:mobile_docs}(f)]{
        \includegraphics[width=0.28\textwidth]{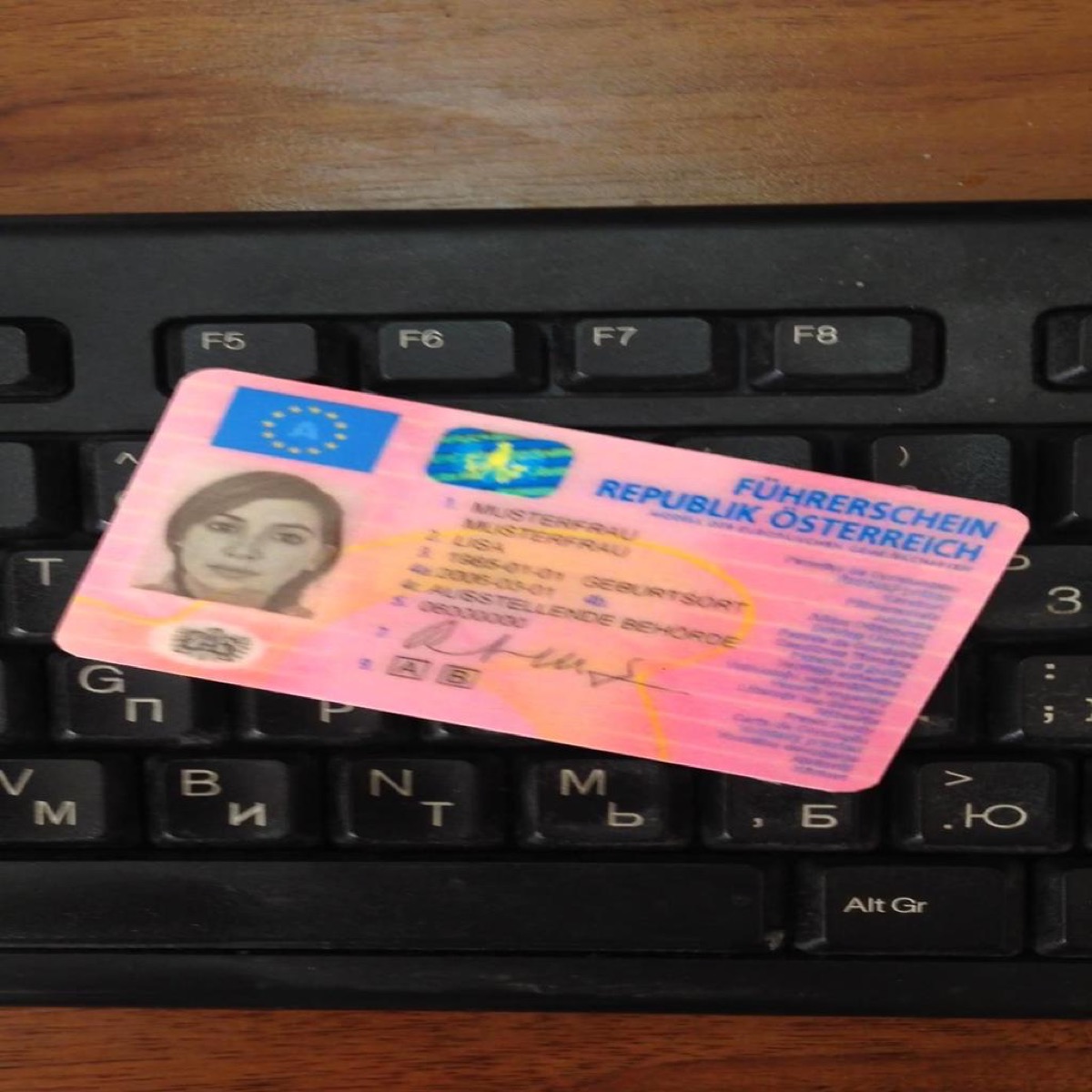}
    }
    \subfigure[Background for Fig.~\ref{fig:mobile_docs}(g)]{
        \includegraphics[width=0.28\textwidth]{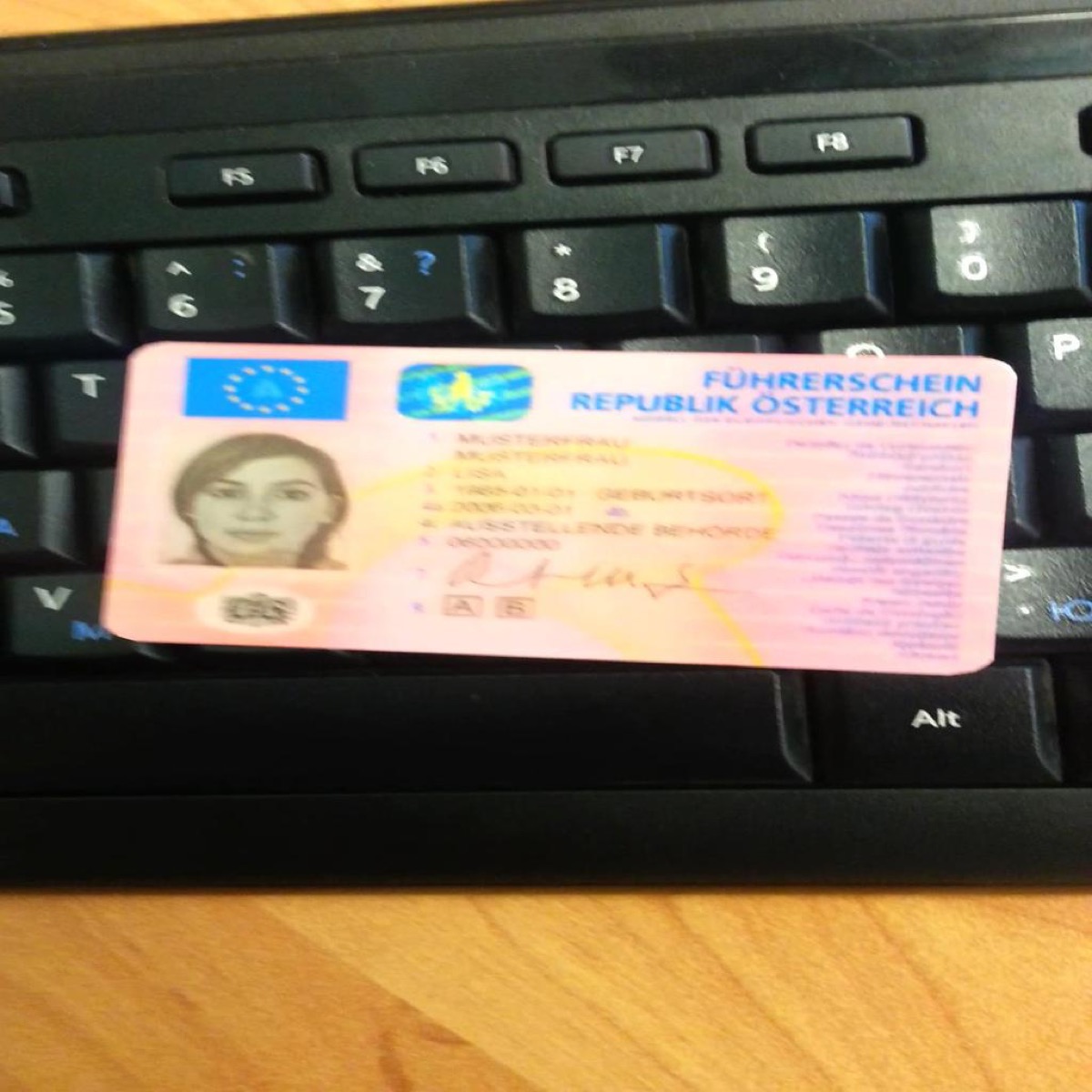}
    }
    \subfigure[Background for Fig.~\ref{fig:mobile_docs}(h)]{
        \includegraphics[width=0.28\textwidth]{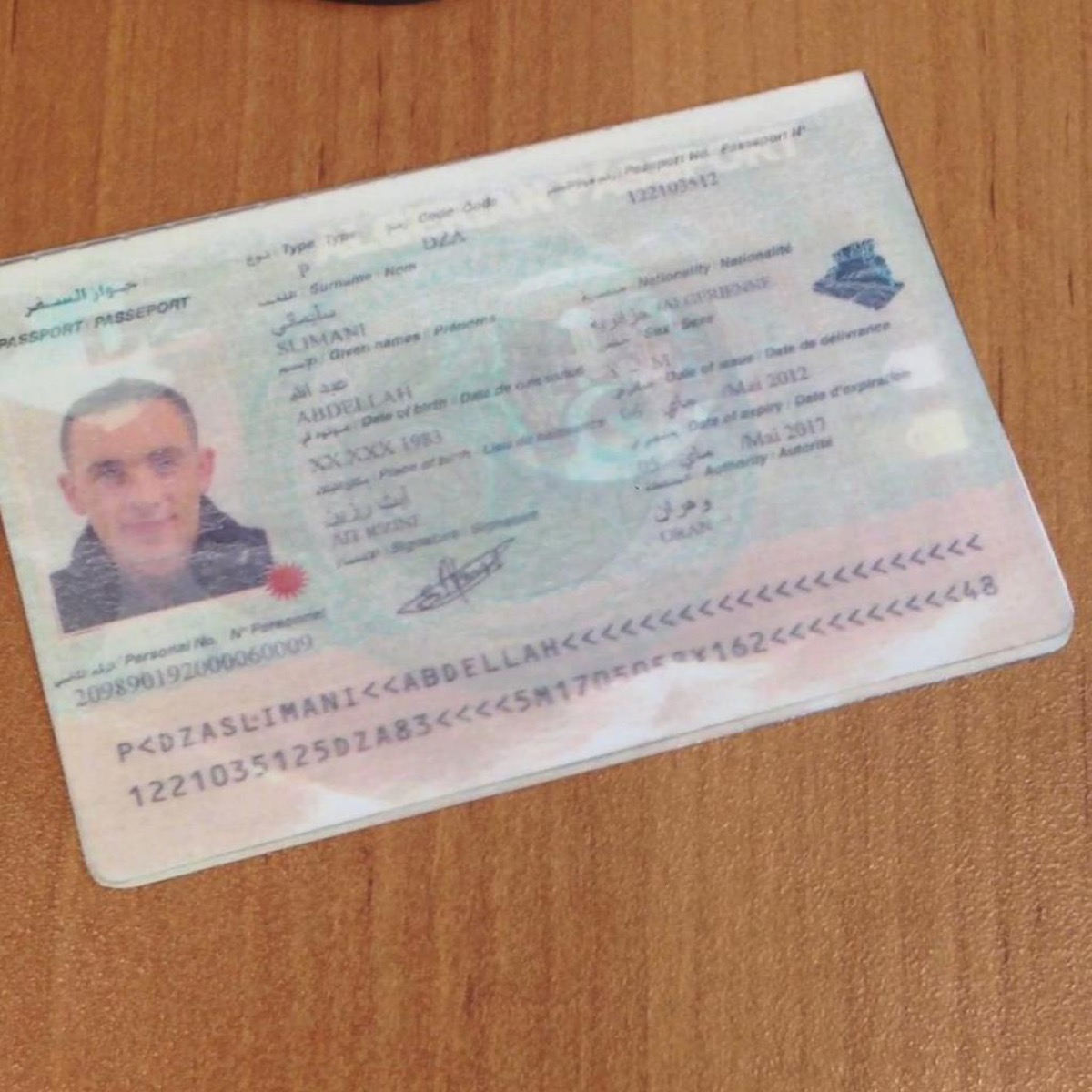}
    }
    \subfigure[Background for Fig.~\ref{fig:mobile_docs}(i)]{
        \includegraphics[width=0.28\textwidth]{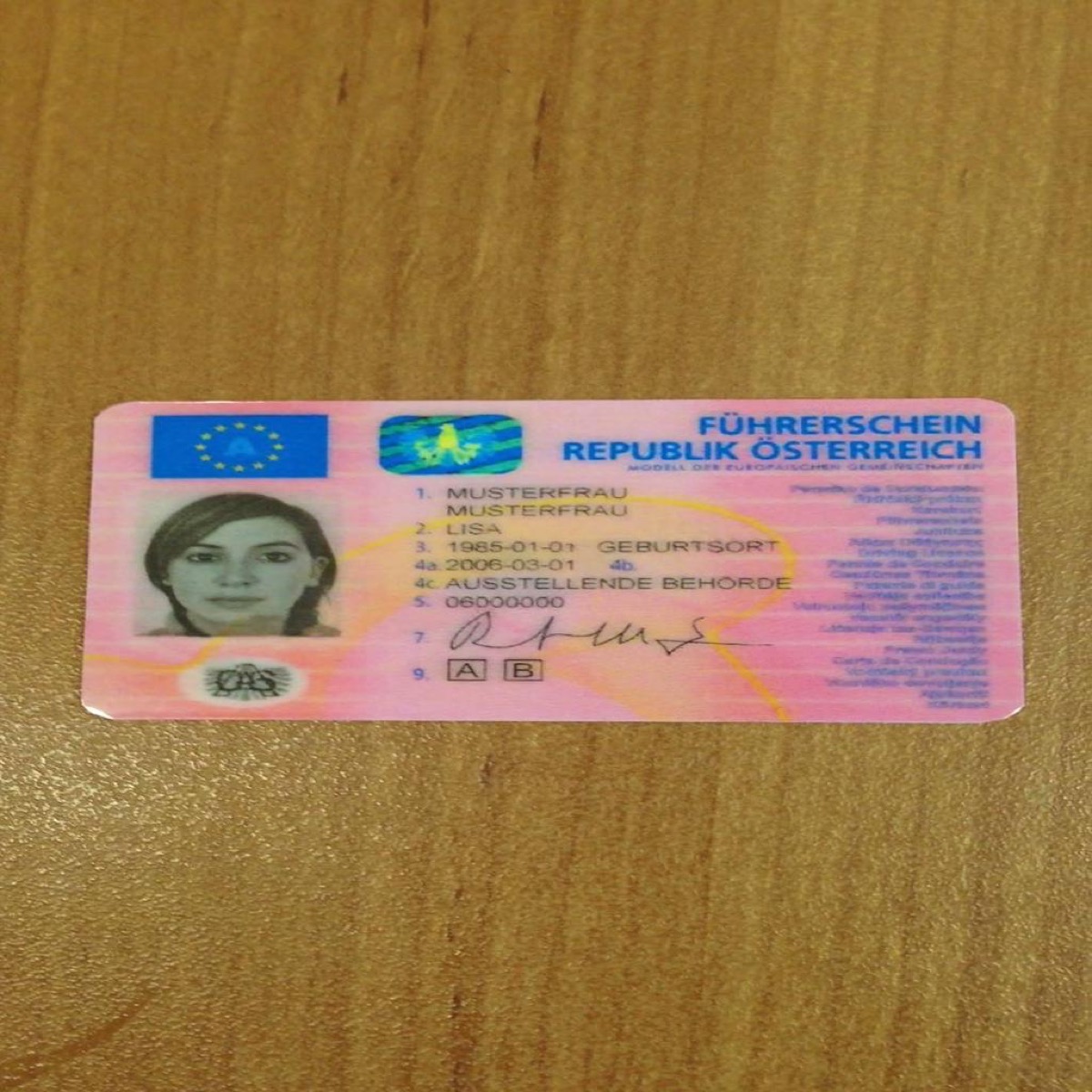}
    }
    \subfigure[ Background for Fig.~\ref{fig:mobile_docs}(j)]{
        \includegraphics[width=0.28\textwidth]{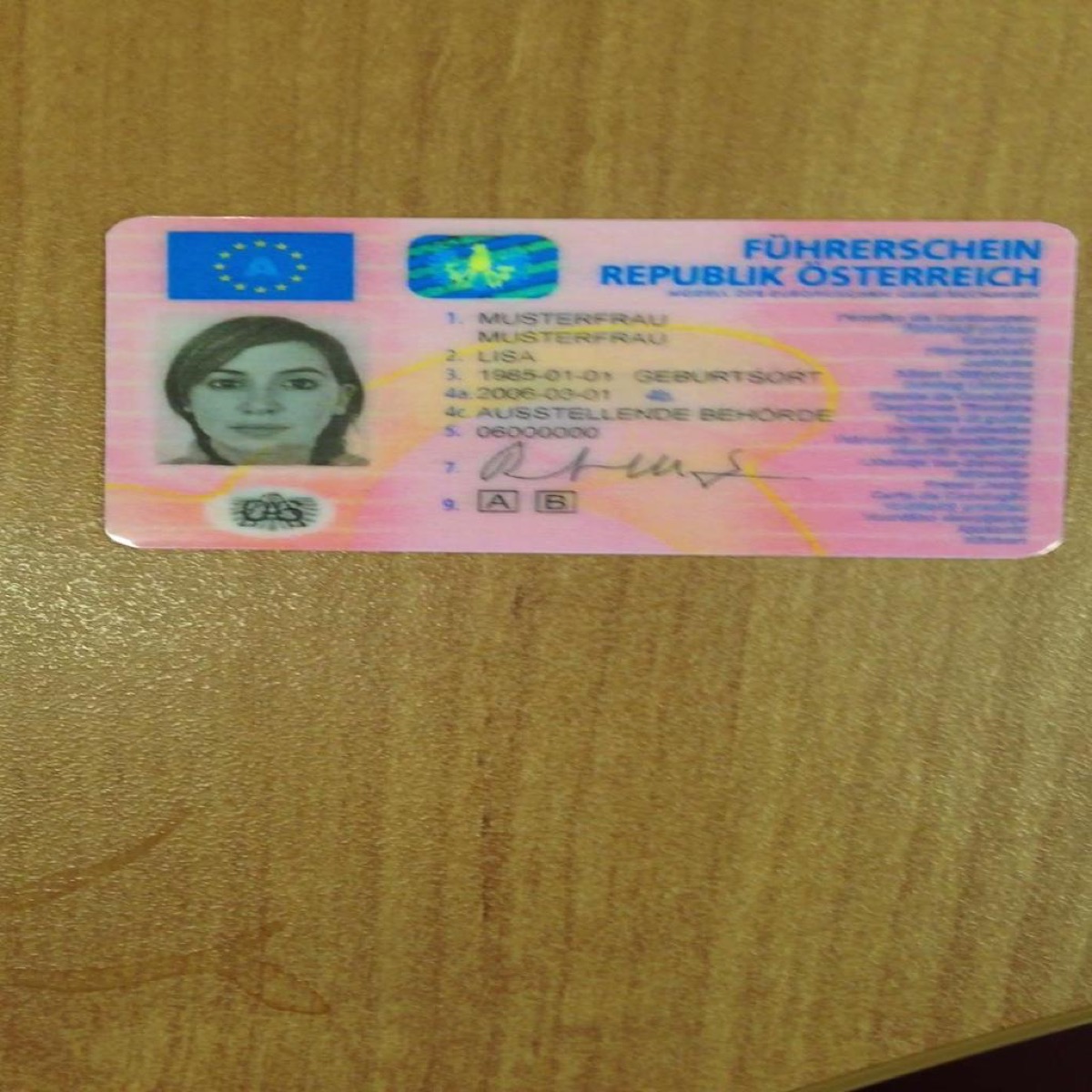}
    }
    \caption{Background images used in Fig.~\ref{fig:mobile_docs}.}
    \label{fig:background_docs}
\end{figure*}

Using the above methodology, we have generated $50$ mobile documents for each of the ten European identity document types included in the \IDSpace dataset to illustrate the robust use case of our \IDSpace framework. Samples of the mobile documents generated for each of ten European identity document types are described in Figure~\ref{fig:mobile_docs}, with background images in Figure~\ref{fig:background_docs}. While some small issues remain to be improved, such as collecting a diverse set of background images with detailed annotations to facilitate semantic search of backgrounds to match user requirements, and addressing inconsistent sizes and lighting conditions between the document in the background image and user requirements, we believe the solutions to these problems are orthogonal with the parameterized and model-guided framework we proposed in this work, and can be addressed in our future works.

\begin{figure*}[h!]
    \centering
    \subfigure[ Albania]{
        \includegraphics[width=0.28\textwidth]{figures/mobile_images/alb.jpg}
    }
    \subfigure[ Azerbaijan]{
        \includegraphics[width=0.28\textwidth]{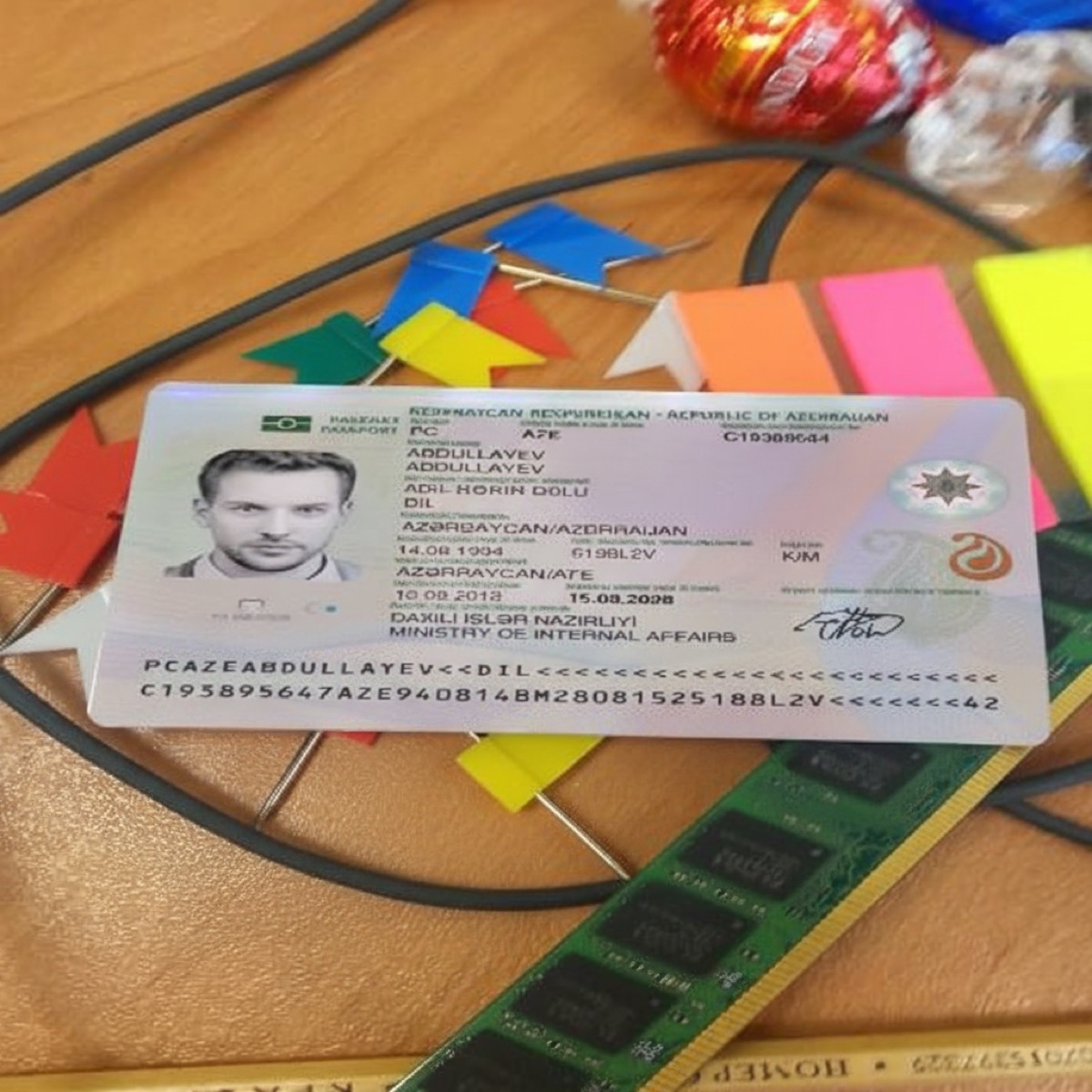}
    }
    \subfigure[ Spain]{
        \includegraphics[width=0.28\textwidth]{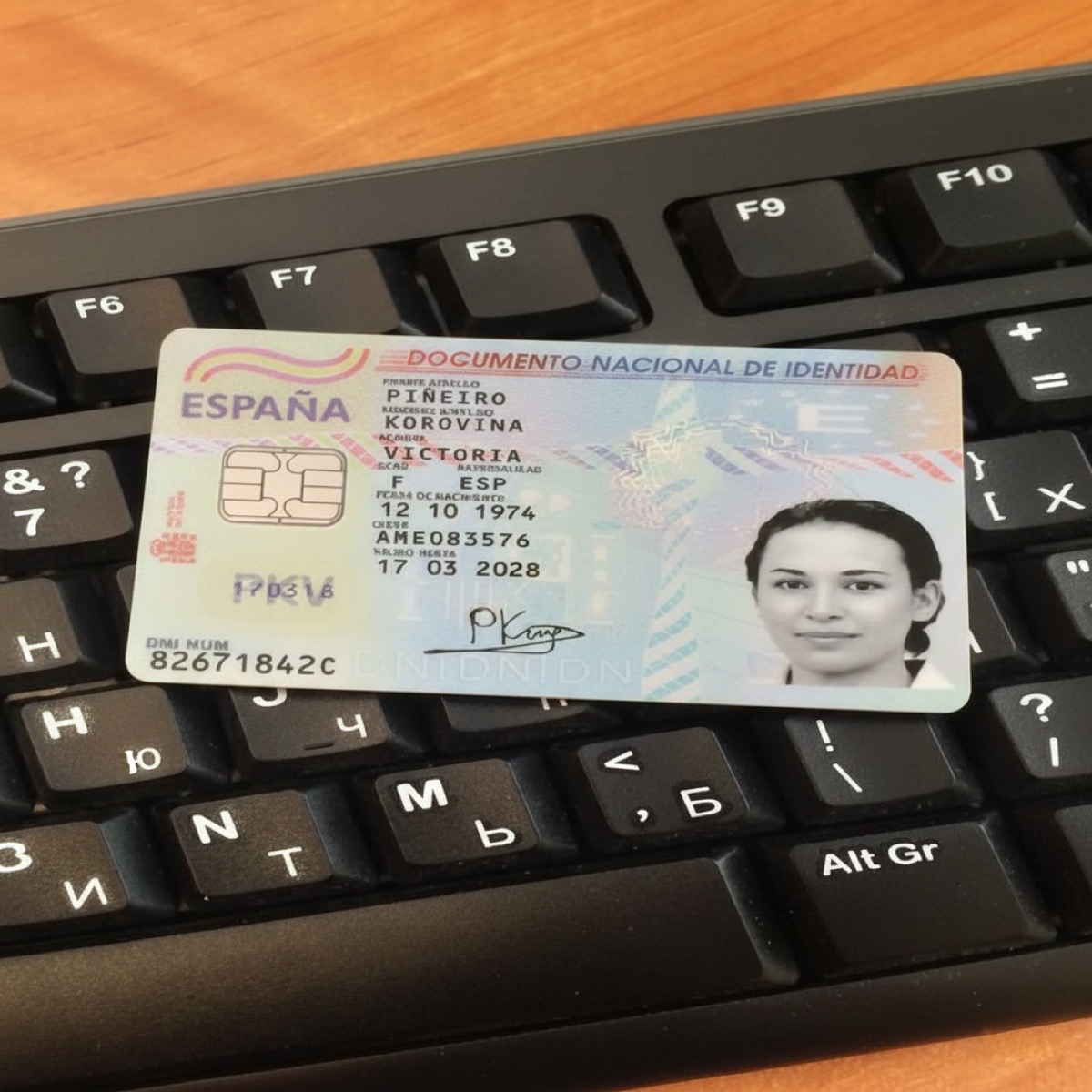}
    }
    \subfigure[ Estonia]{
        \includegraphics[width=0.28\textwidth]{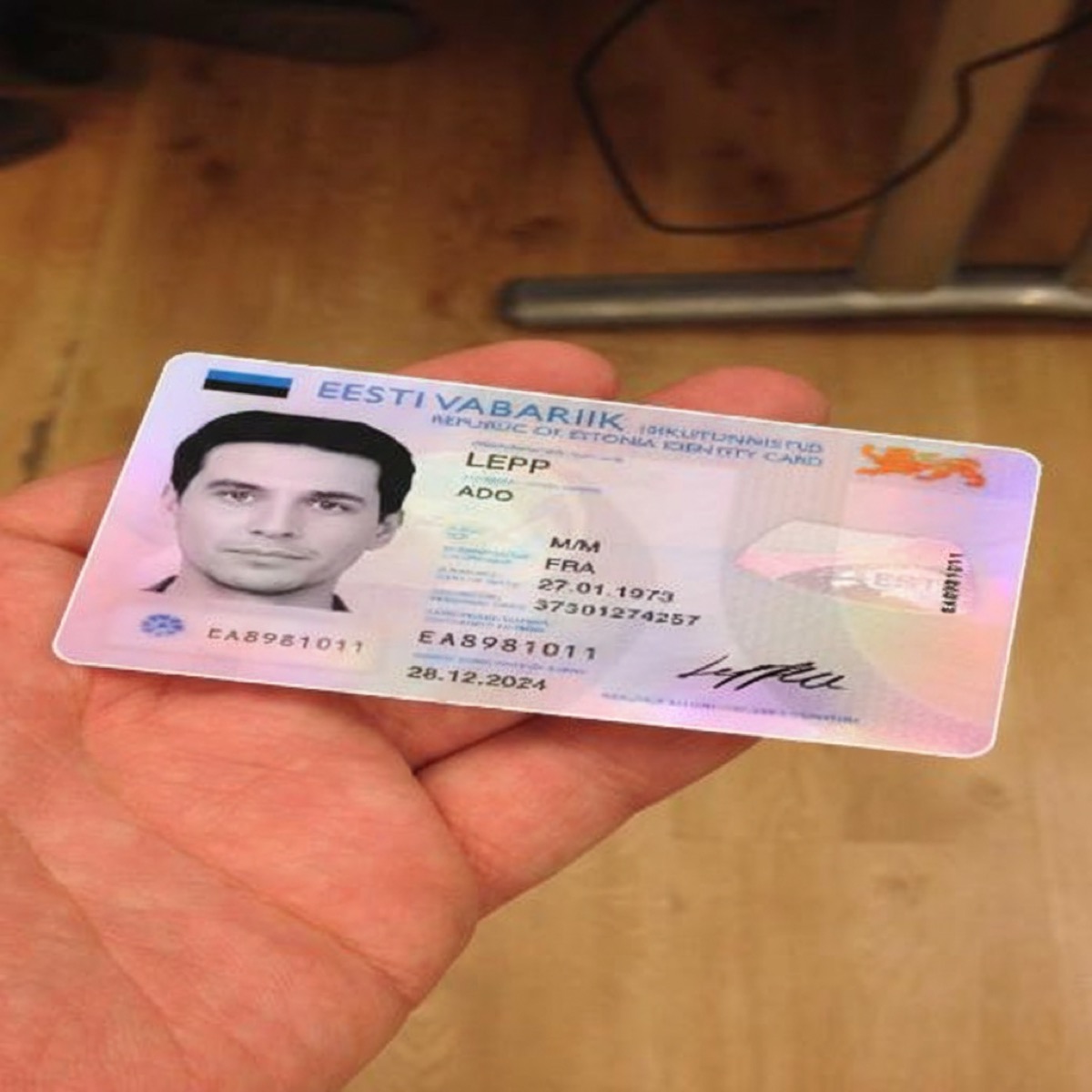}
    }
    \subfigure[ Finland]{
        \includegraphics[width=0.28\textwidth]{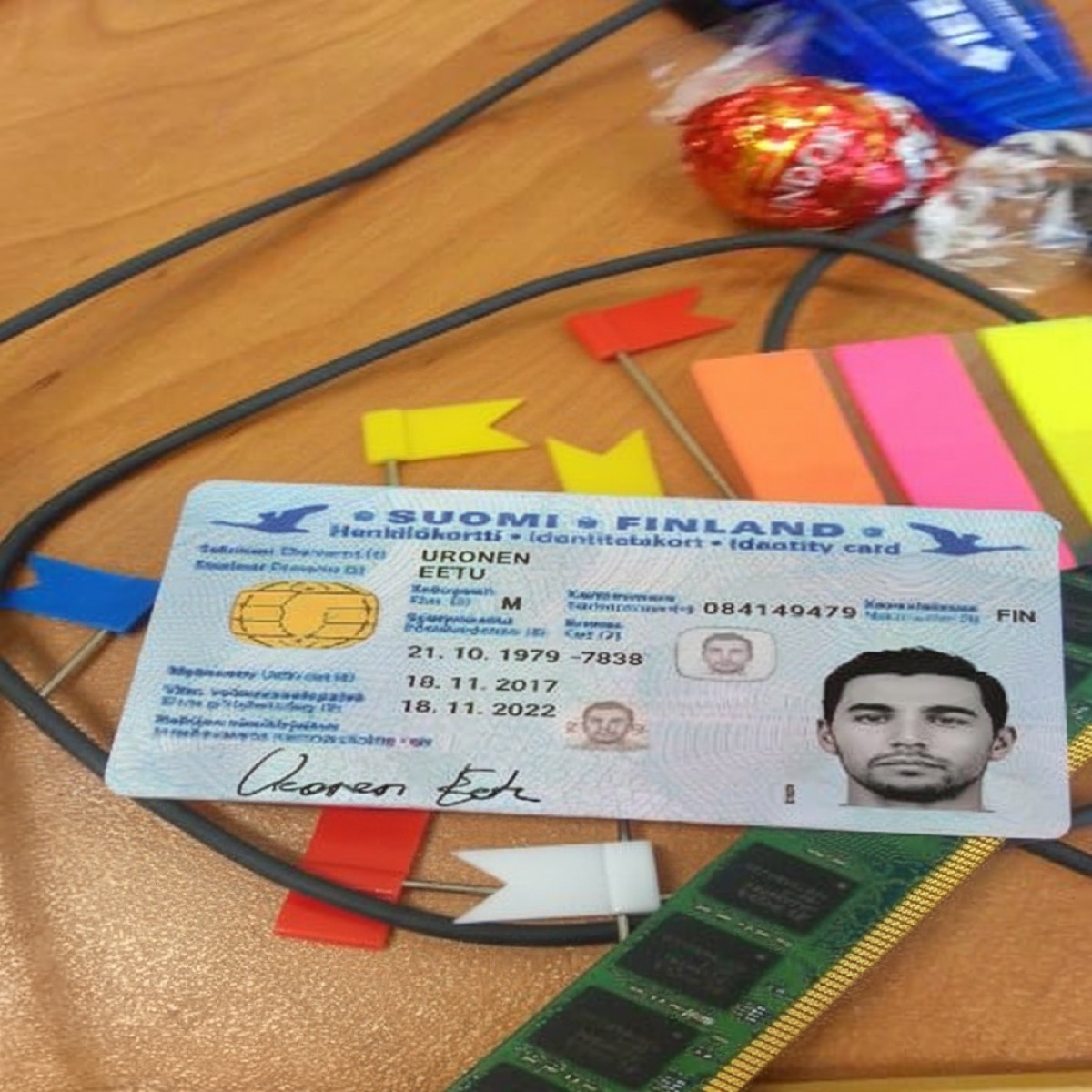}
    }
    \subfigure[ Greece]{
        \includegraphics[width=0.28\textwidth]{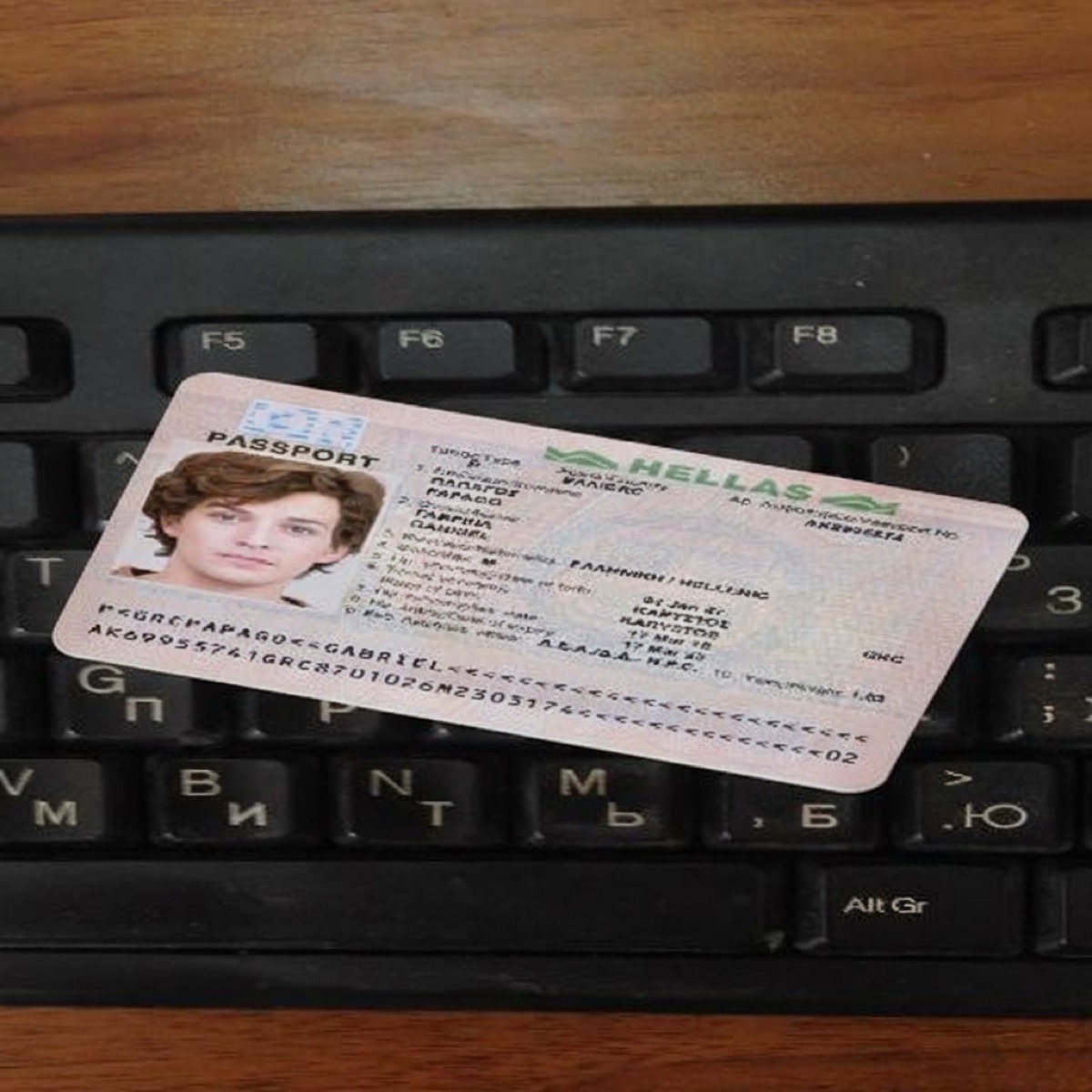}
    }
    \subfigure[ Latvia]{
        \includegraphics[width=0.28\textwidth]{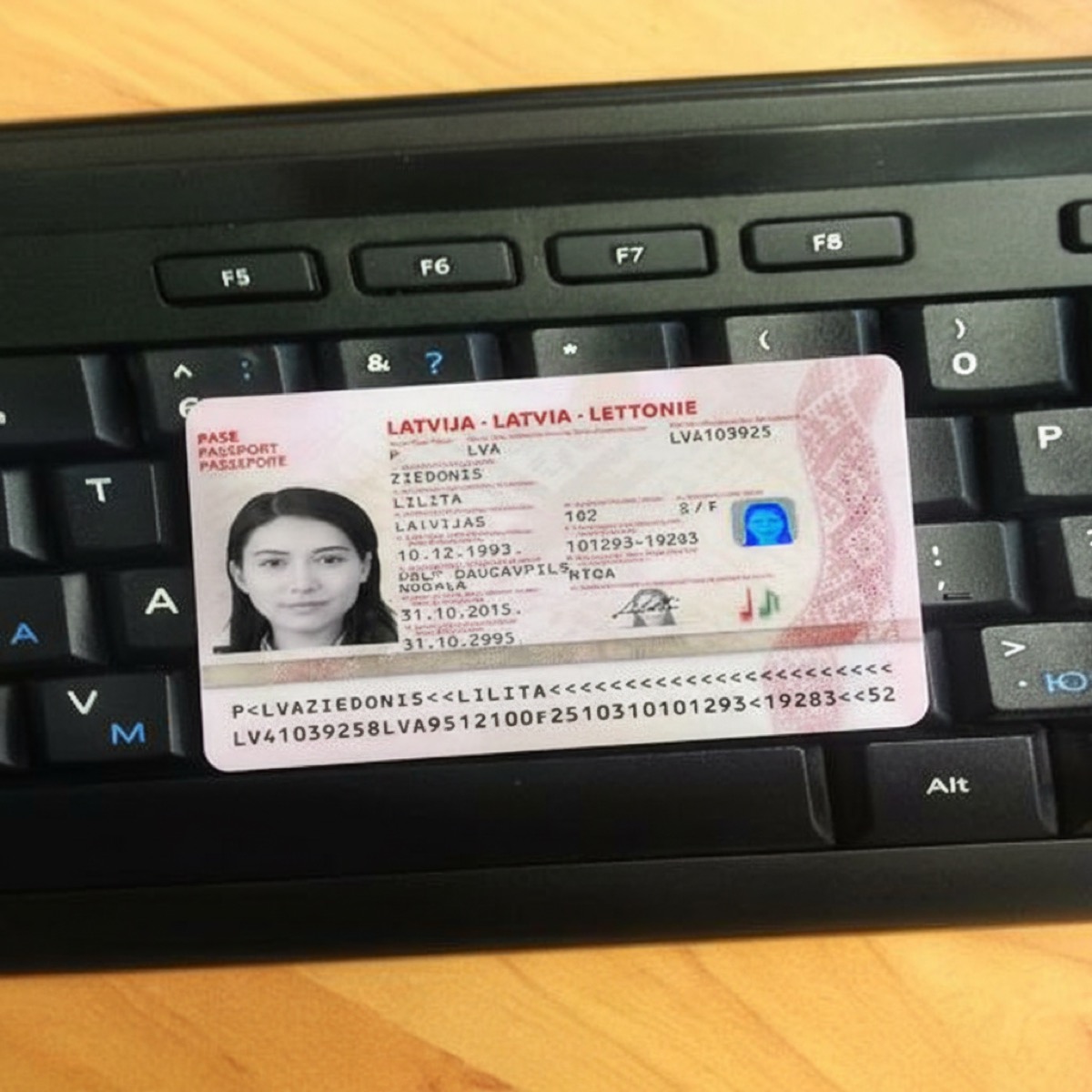}
    }
    \subfigure[ Russia]{
        \includegraphics[width=0.28\textwidth]{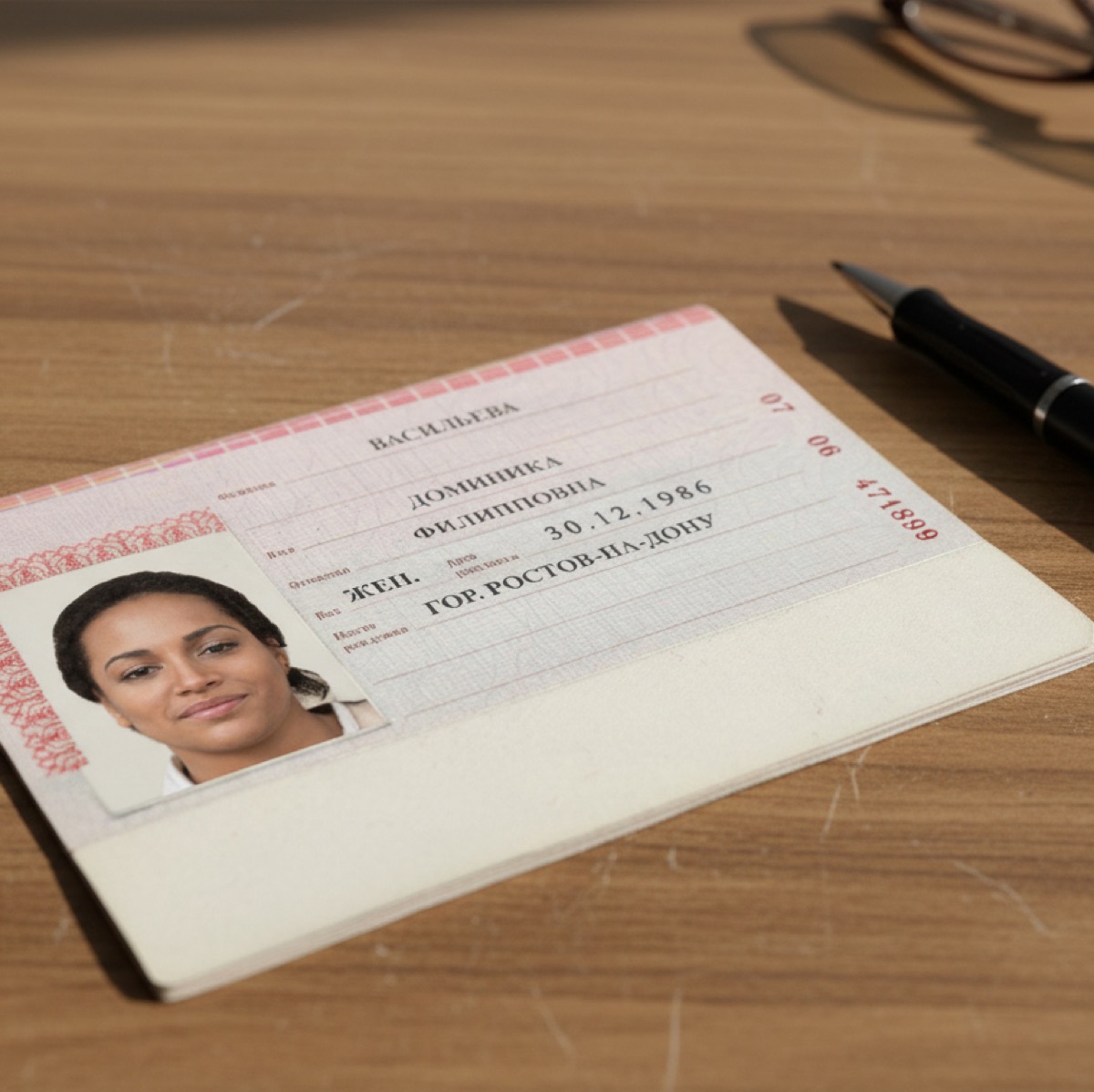}
    }
    \subfigure[ Serbia]{
        \includegraphics[width=0.28\textwidth]{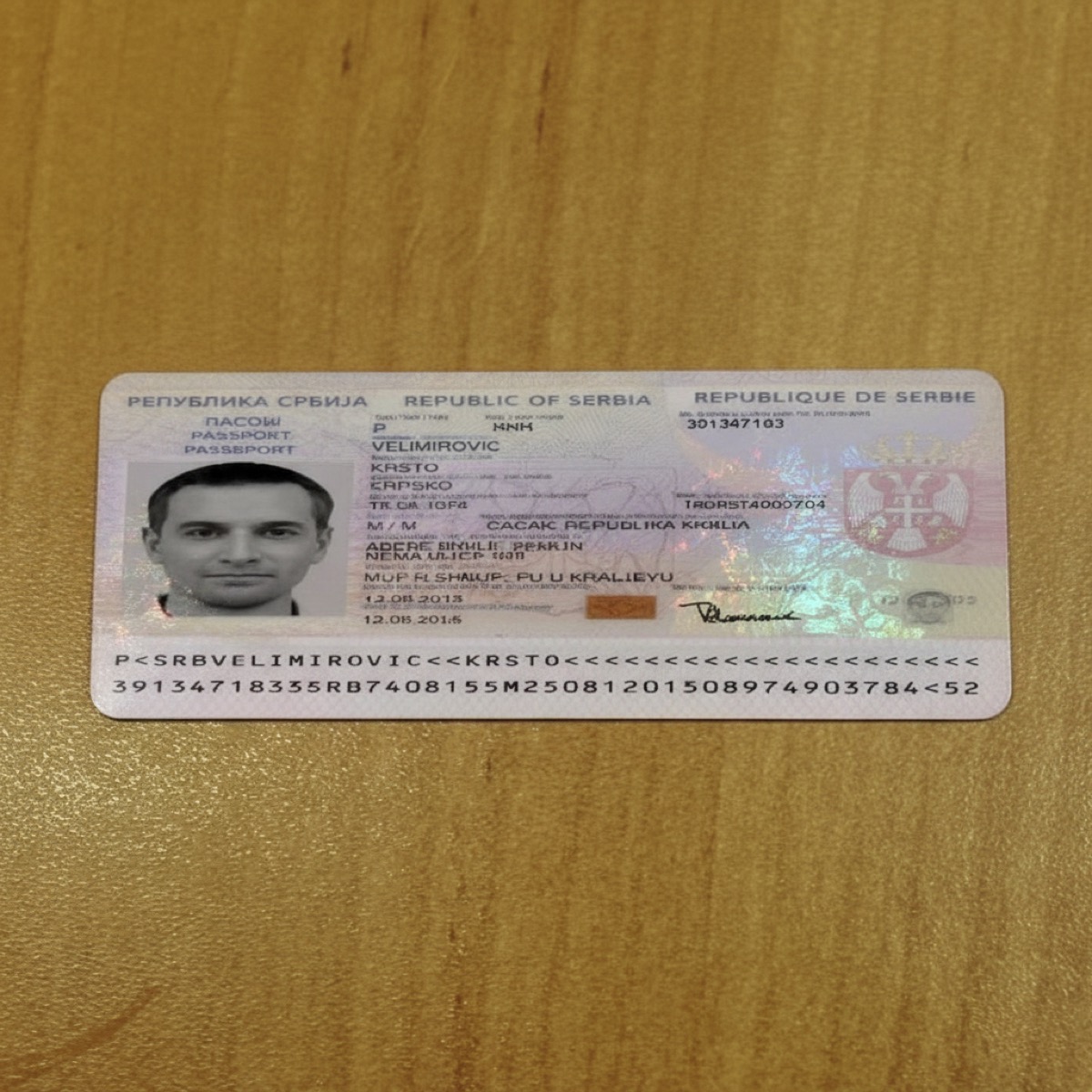}
    }
    \subfigure[ Slovakia]{
        \includegraphics[width=0.28\textwidth]{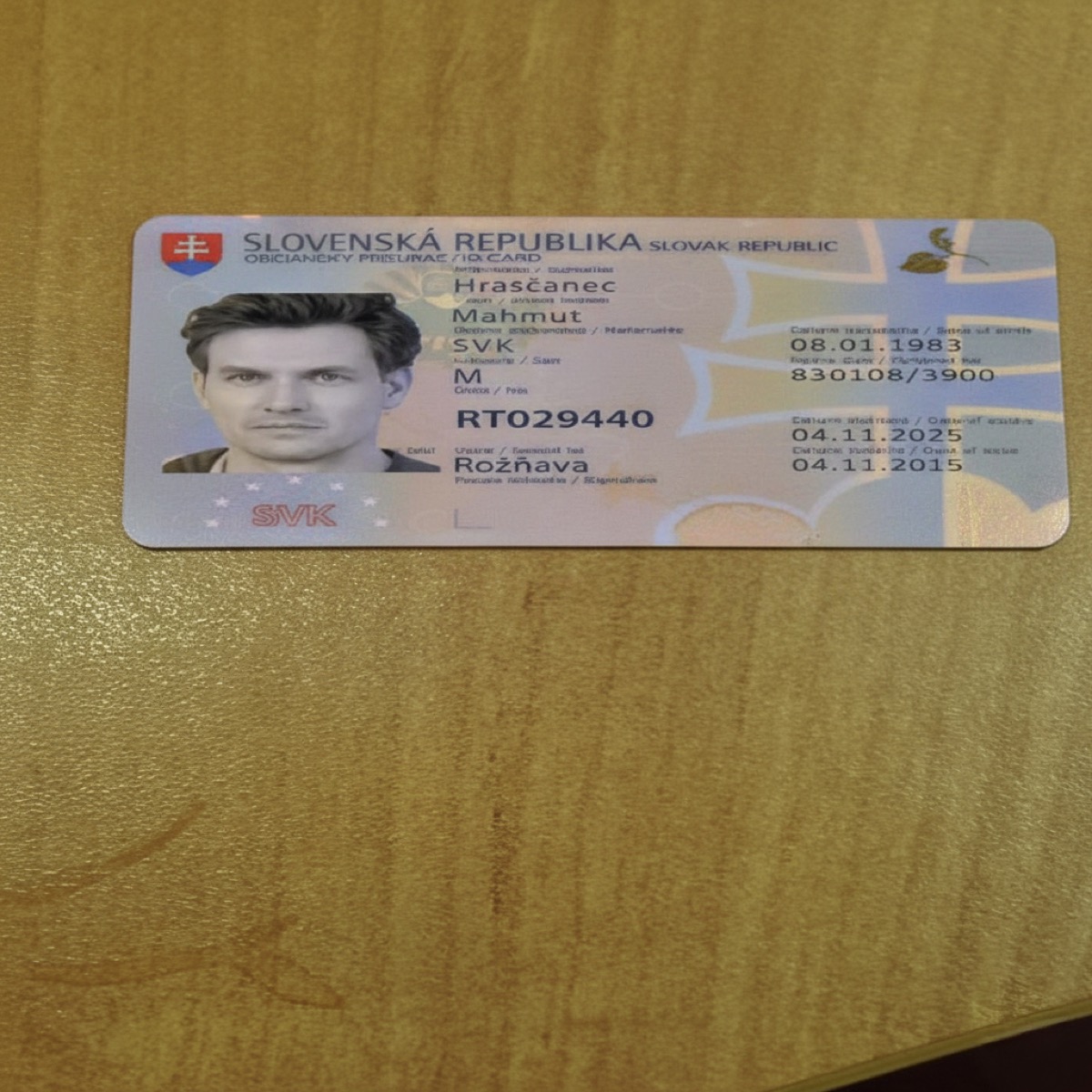}
    }
    \caption{Example Mobile document images generated for the $10$ different templates.}
    \label{fig:mobile_docs}
\end{figure*}

\end{document}